\documentclass[12pt]{report}

\usepackage{tamuconfig}

\usepackage{times}
\usepackage[T1]{fontenc}

\usepackage{graphicx}

\DeclareGraphicsExtensions{.png, .jpg, .pdf}

\graphicspath{ {./graphic/} }

\usepackage[hidelinks]{hyperref}

\usepackage{mymacros}

\begin{document}

\renewcommand{\tamumanuscripttitle}{Human-in-the-loop Methods for Data-driven and Reinforcement Learning Systems}

\renewcommand{\tamupapertype}{Dissertation}

\renewcommand{\tamufullname}{Vinicius Guimaraes Goecks}

\renewcommand{\tamudegree}{Doctor of Philosophy}
\renewcommand{\tamuchairone}{John Valasek}

\renewcommand{\tamumemberone}{Gregory Chamitoff}
\newcommand{\tamumembertwo}{Daniel Selva}
\newcommand{\tamumemberthree}{Dylan Shell}
\renewcommand{\tamudepthead}{Rodney D. W. Bowersox}

\renewcommand{\tamugradmonth}{May}
\renewcommand{\tamugradyear}{2020}
\renewcommand{\tamudepartment}{Aerospace Engineering}

%
%
%
%


\providecommand{\tabularnewline}{\\}

\begin{titlepage}
\begin{center}
\MakeUppercase{\tamumanuscripttitle}
\vspace{4em}

A \tamupapertype

by

\MakeUppercase{\tamufullname}

\vspace{4em}

\begin{singlespace}

Submitted to the Office of Graduate and Professional Studies of \\
Texas A\&M University \\

in partial fulfillment of the requirements for the degree of \\
\end{singlespace}

\MakeUppercase{\tamudegree}
\par\end{center}
\vspace{2em}
\begin{singlespace}
\begin{tabular}{ll}
 & \tabularnewline
& \cr
Chair of Committee, & \tamuchairone\tabularnewline
Committee Members, & \tamumemberone\tabularnewline
 & \tamumembertwo\tabularnewline
 & \tamumemberthree\tabularnewline
Head of Department, & \tamudepthead\tabularnewline

\end{tabular}
\end{singlespace}
\vspace{3em}

\begin{center}
\tamugradmonth \hspace{2pt} \tamugradyear

\vspace{3em}

Major Subject: \tamudepartment \par
\vspace{3em}
Copyright \tamugradyear \hspace{.5em}\tamufullname 
\par\end{center}
\end{titlepage}
\pagebreak{}

%
%
%
%

\chapter*{ABSTRACT}
\addcontentsline{toc}{chapter}{ABSTRACT} 

\pagestyle{plain} 
\pagenumbering{roman} 
\setcounter{page}{2}





Recent successes combine reinforcement learning algorithms and deep neural networks, despite reinforcement learning not being widely applied to robotics and real world scenarios.
This can be attributed to the fact that current state-of-the-art, end-to-end reinforcement learning approaches still requires thousands or millions of data samples to converge to a satisfactory policy and are subject to catastrophic failures during training.
Conversely, in real world scenarios and after just a few data samples, humans are able to either provide demonstrations of the task, intervene to prevent catastrophic actions, or simply evaluate if the policy is performing correctly.
This research investigates how to integrate these human interaction modalities to the reinforcement learning loop, increasing sample efficiency and enabling real-time reinforcement learning in robotics and real world scenarios.
The theoretical foundation of this research work builds upon the actor-critic reinforcement learning architecture, the use of function approximation to represent action- and value-based functions, and the integration of different human interaction modalities; namely, task demonstration, intervention, and evaluation, to these functions and to reward signals.
This novel theoretical foundation is called \emph{Cycle-of-Learning}, a reference to how different human interaction modalities are cycled and combined to reinforcement learning algorithms.
This approach is validated on an Unmanned Air System (UAS) collision avoidance and landing scenario using a high-fidelity simulated environment and several continuous control tasks standardized to benchmark reinforcement learning algorithms.
Results presented in this work show that the reward signal that is learned based upon human interaction accelerates the rate of learning of reinforcement learning algorithms, when compared to traditional handcrafted or binary reward signals returned by the environment.
Results also show that learning from a combination of human demonstrations and interventions is faster and more sample efficient when compared to traditional supervised learning algorithms.
Finally, \emph{Cycle-of-Learning} develops an effective transition between policies learned using human demonstrations and interventions to reinforcement learning. It learns faster and uses fewer interactions with the environment when compared to state-of-the-art algorithms.
The theoretical foundation developed by this research opens new research paths to human-agent teaming scenarios where autonomous agents are able to learn from human teammates and adapt to mission performance metrics in real-time and in real world scenarios.

\pagebreak{}

%
%
%
%

\chapter*{DEDICATION}
\addcontentsline{toc}{chapter}{DEDICATION}  

\begin{center}
\vspace*{\fill}
To my wife, family, and friends who have always supported me\\ throughout my academic journey.
\vspace*{\fill}
\end{center}

\pagebreak{}

%
%
%
%

\chapter*{ACKNOWLEDGMENTS}
\addcontentsline{toc}{chapter}{ACKNOWLEDGMENTS}  

First of all, I would not be able to complete this doctorate degree without the support of my loving wife, Lucieni, and my parents, Elizabeth and Claudio, and brother, Thiago. My wife has been always by my side when facing the challenges of graduate school and also sharing all the happy moments that came with it. My parents, even though in a another country, were always supportive of me when pursuing this degree. I would not be able to complete this dissertation without their emotional support.

I would like to acknowledge and thank who contributed to the development of this research and the successful completion of this dissertation.
Dr. John Valasek, who has been an exceptional advisor and mentor not only on academic matters, but also on leadership and management, which I see as invaluable skills for my future career. My committee members, Dr. Gregory Chamitoff, Dr. Dylan Shell, and Dr. Daniel Selva for their guidance when grounding the research fundamentals and support when defining the research direction.
I would like to specially thank Dr. Nicholas Waytowich, Dr. Gregory Gremillion, and Dr. Vernon Lawhern, from the U.S. Army Research Laboratory, for being an integral part of the development process of this research, for working together when defining the research direction and research questions to be answered, and for all support when presenting and publishing the research results. This research would not be successful without their technical contributions, guidance, and support.


Finally, I would like to also thank my graduate student peers and personal friends that directly contributed to my formation and supported me during my doctorate journey: Humberto Ramos, Niladri Das, Hakjoo Kim, Jack Lu, Bochan Lee, Josh Harris, Robyn Woollands, Austin Probe, Clark Moody, and more recently, Ritwik Bera and Morgan Wood. Thank you for everything you taught me throughout this graduate school years and for being part of this journey.


\pagebreak{}
%
%
%
%

\chapter*{CONTRIBUTORS AND FUNDING SOURCES}
\addcontentsline{toc}{chapter}{CONTRIBUTORS AND FUNDING SOURCES}  

\subsection*{Contributors}
This work was supported by a dissertation committee consisting of Professor John Valasek, Gregory Chamitoff, and Daniel Selva of the Department of Aerospace Engineering and Professor Dylan Shell of the Department of Computer Science.

The data analyses for chapters \ref{ch:cybersteer} through \ref{ch:col_loss} were conducted in collaboration with Dr. Nicholas R. Waytowich, Dr. Vernon J. Lawhern, and Dr. Gregory M. Gremillion from the U.S. Army Research Laboratory and were published in 2018 and 2019 in articles listed in the Biographical Sketch.

All other work conducted for the dissertation was completed by the student independently.

\subsection*{Funding Sources}
Graduate study was supported by a fellowship from Coordenação de Aperfeiçoamento de Pessoal de Nível Superior, by a fellowship from the U.S. Army Research Laboratory through the Oak Ridge Associated Universities, and by a fellowship from the Department of Aerospace Engineering at Texas A\&M University.

Research was sponsored by the U.S. Army Research Laboratory and was accomplished under Cooperative Agreement Number W911NF-18-2-0134 and W911NF-17-2-0078. The views and conclusions contained in this document are those of the authors and should not be interpreted as representing the official policies, either expressed or implied, of the Army Research Laboratory or the U.S. Government. The U.S. Government is authorized to reproduce and distribute reprints for Government purposes notwithstanding any copyright notation herein.

\pagebreak{}
%
%
%
%


\chapter*{NOMENCLATURE}
\addcontentsline{toc}{chapter}{NOMENCLATURE}  


\hspace*{-1.25in}
\vspace{12pt}
\begin{spacing}{1.0}
	\begin{longtable}[htbp]{@{}p{0.35\textwidth} p{0.62\textwidth}@{}}
		
		A3C & Asynchronous Advantage Actor-Critic\\ [2ex]
		AAAI & Association for the Advancement of Artificial Intelligence\\ [2ex]
		$A(s,a)$ & Advantage function\\ [2ex]
		$a_t$ & Action vector at time $t$\\ [2ex]
		AGL & Altitude Above Ground Level\\ [2ex]
		AIAA & American Institute of Aeronautics and Astronautics\\ [2ex]
		ANN & Artificial Neural Network\\ [2ex]
		A-RLC & Adaptive-Reinforcement Learning Control\\ [2ex]
		BCE & Binary Cross-Entropy\\ [2ex]
		CE & Cross-Entropy\\ [2ex]
		CNN & Convolutional Neural Network\\ [2ex]
		CoL & Cycle-of-Learning for Autonomous Systems\\ [2ex]
		BC & Behavior Cloning\\ [2ex]
		CEM & Cross-Entropy Method\\ [2ex]
		CMA-ES & Covariance-Matrix Adaptation Evolution Strategy\\ [2ex]
		COACH & Convergent Actor-Critic by Humans\\ [2ex]
		CPU & Central Processing Unit\\ [2ex]
		$D$ & Discriminator network\\ [2ex]
		DAgger & Dataset Aggregation\\ [2ex]
		DAPG & Demo-Augmented Policy Gradient\\ [2ex]
		Deep RL & Deep Reinforcement Learning\\ [2ex]
		DDPG & Deep Deterministic Policy Gradient\\ [2ex]
		DDPGfD & Deep Deterministic Policy Gradient from Demonstrations\\ [2ex]
		DPG & Deterministic Policy Gradient\\ [2ex]
		DL & Deep Learning\\ [2ex]
		DQN & Deep Q-Network\\ [2ex]
		DQfD & Deep Q-learning from Demonstrations\\ [2ex]
		DOF & Degree-of-Freedom\\ [2ex]
		ELU & Exponential Linear Unit\\ [2ex]
		$J(\cdot)$ & Objective function\\ [2ex]
		HCRL & Human-Centered Reinforcement Learning\\ [2ex]
		HER & Hindsight Experience Replay\\ [2ex]
		HRI & Human-Robot Interaction\\ [2ex]
		I2A & Imagination-Augmented Agents\\ [2ex]
		IEEE & Institute of Electrical and Electronics Engineers\\ [2ex]
		IJCNN & International Joint Conference on Neural Networks\\ [2ex]
		IL & Imitation Learning\\ [2ex]
		iLQR & Iterative Linear Quadratic Regulator\\ [2ex]
		IMPALA & Importance Weighted Actor-Learner Architecture\\ [2ex]
		IOC & Inverse Optimal Control\\ [2ex]
		IPG & Interpolated Policy Gradient\\ [2ex]
		IRL & Inverse Reinforcement Learning\\ [2ex]
		IRM & Intrinsic Reward Module\\ [2ex]
		FCN & Fully-Connected Network\\ [2ex]
		FCCN & Fully-Connected Convolutional Network\\ [2ex]
		$G$ & Generator network\\ [2ex]
		GAE & Generalized Advantage Estimator\\ [2ex]
		GAIL & Generative Adversarial Imitaiton Learning\\ [2ex]
		GAN & Generative Adversarial Network\\ [2ex]
		GCL & Guided Cost Learning\\ [2ex]
		GPS & Global Positioning System\\ [2ex]
		GPU & Graphics Processing Unit\\ [2ex]
		KL & Kullback-Leibler\\ [2ex]
		$\mathcal{L}$ & Loss function\\ [2ex]
		LfD & Learning from Demonstrations\\ [2ex]
		LfE & Learning from Evaluations\\ [2ex]
		LfI & Learning from Interventions\\ [2ex]
		LSTM & Long Short-Term Memory\\ [2ex]
		LP & Linear Programming\\ [2ex]
		M-PAC & Multi-Preference Actor Critic\\ [2ex]
		MAE & Mean Absolute Error\\ [2ex]
		MADCAT & Mission Adaptive Digital Composite Aerostructure Technologies\\ [2ex]
		MDN & Mixture Density Network\\ [2ex]
		MDP & Markov Decision Process\\ [2ex]
		MLP & Multilayer Perceptron\\ [2ex]
		MPC & Model Predictive Control\\ [2ex]
		MSE & Mean Squared Error\\ [2ex]
		MVE & Model Value Expansion\\ [2ex]
		NAF & Normalized Advantage Function\\ [2ex]
		$o_t$ & Observation vector at time $t$\\ [2ex]
		OML & Outer Mold Line\\ [2ex]
		$\pi$ & Policy\\ [2ex]
		$\pi_0$ & Initial Policy\\ [2ex]
		PDMS & Polydimethylsiloxane\\ [2ex]
		PER & Prioritized Experience Replay\\ [2ex]
		PG & Policy Gradient\\ [2ex]
		POfD & Policy Optimization with Demonstrations\\ [2ex]
		POMDP & Partially Observable Markov Decision Process\\ [2ex]
		PPO & Proximal Policy Optimization\\ [2ex]
		$Q(s,a)$ & State-action value function\\ [2ex]
		$\mathcal{R}$ & Replay Buffer\\ [2ex]
		$R_1$ & $1$-step return\\ [2ex]
		$R_n$ & $n$-step return\\ [2ex]
		$r_t$ & Reward value at time $t$\\ [2ex]
		ReLU & Rectified Linear Unit\\ [2ex]
		RGB & Visible spectrum color code\\ [2ex]
		RL & Reinforcement Learning\\ [2ex]
		RMSProp & Root Mean Square Propagation\\ [2ex]
		RNN & Recurrent Neural Network\\ [2ex]
		$s_t$ & State vector at time $t$\\ [2ex]
		SAA & Sense-and-Avoid\\ [2ex]
		SAC & Soft Actor-Critic\\ [2ex]
		SAMI & Structured Adaptive Model Inversion\\ [2ex]
		SC2LE & StarCraft II Learning Environment\\ [2ex]
		SMA & Shape-Memory Alloy\\ [2ex]
		sUAS & Small Unmanned Air System\\ [2ex]
		SVM & Support Vector Machine\\ [2ex]
		$t$ & Time step\\ [2ex]
		$T$ & Total time steps in one episode\\ [2ex]
		$\tau$ & Trajectory\\ [2ex]
		TAMER & Training an Agent Manually via Evaluative Reinforcement\\ [2ex]
		TAMU & Texas A\&M University\\ [2ex]
		Tanh & Hyperbolic Tangent\\ [2ex]
		TD & Temporal Difference\\ [2ex]
		TD3 & Twin Delayed DDPG\\ [2ex]
		$\theta_\pi$ & Parameters of policy $\pi$ (actor)\\ [2ex]
		$\theta_Q$ & Parameters of the critic\\ [2ex]
		TNPG & Truncated Natural Policy Gradient\\ [2ex]
		TRPO & Trust Region Policy Optimization\\ [2ex]
		$\mu$ & Deterministic policy\\ [2ex]
		UAS & Unmanned Air System\\ [2ex]
		UAV & Unmanned Air Vehicle\\ [2ex]
		UNREAL & Unsupervised Reinforcement and Auxiliary Learning\\ [2ex]
		$V(s)$ & State value function\\ [2ex]
		VAE & Variational Autoencoder\\ [2ex]
		VICE-RAQ & Variational Inverse Control with Events - Reinforcement learning with Active Queries\\ [2ex]
		VPG & Vanilla Policy Gradient\\ [2ex]
		VSCL & Vehicle Systems \& Control Laboratory\\ [2ex]
		$w$ & Weight vector of a model\\ [2ex]
		$Z$ & Probability normalization term\\ [2ex]
		
	\end{longtable}
\end{spacing}

\pagebreak{}

%
%
%
%

\phantomsection
\addcontentsline{toc}{chapter}{TABLE OF CONTENTS}  

\begin{singlespace}
\renewcommand\contentsname{\normalfont} {\centerline{TABLE OF CONTENTS}}

\setcounter{tocdepth}{4} 

\setlength{\cftaftertoctitleskip}{1em}
\renewcommand{\cftaftertoctitle}{%
\hfill{\normalfont {Page}\par}}

\tableofcontents

\end{singlespace}

\pagebreak{}


\phantomsection
\addcontentsline{toc}{chapter}{LIST OF FIGURES}  

\renewcommand{\cftloftitlefont}{\center\normalfont\MakeUppercase}

\setlength{\cftbeforeloftitleskip}{-12pt} 
\renewcommand{\cftafterloftitleskip}{12pt}

\renewcommand{\cftafterloftitle}{%
\\[4em]\mbox{}\hspace{2pt}FIGURE\hfill{\normalfont Page}\vskip\baselineskip}

\begingroup

\begin{center}
\begin{singlespace}
\setlength{\cftbeforechapskip}{0.4cm}
\setlength{\cftbeforesecskip}{0.30cm}
\setlength{\cftbeforesubsecskip}{0.30cm}
\setlength{\cftbeforefigskip}{0.4cm}
\setlength{\cftbeforetabskip}{0.4cm}



\listoffigures

\end{singlespace}
\end{center}

\pagebreak{}

%
\phantomsection
\addcontentsline{toc}{chapter}{LIST OF TABLES}  

\renewcommand{\cftlottitlefont}{\center\normalfont\MakeUppercase}

\setlength{\cftbeforelottitleskip}{-12pt} 

\renewcommand{\cftafterlottitleskip}{1pt}

\renewcommand{\cftafterlottitle}{%
\\[4em]\mbox{}\hspace{2pt}TABLE\hfill{\normalfont Page}\vskip\baselineskip}

\begin{center}
\begin{singlespace}

\setlength{\cftbeforechapskip}{0.4cm}
\setlength{\cftbeforesecskip}{0.30cm}
\setlength{\cftbeforesubsecskip}{0.30cm}
\setlength{\cftbeforefigskip}{0.4cm}
\setlength{\cftbeforetabskip}{0.4cm}

\listoftables 

\end{singlespace}
\end{center}
\endgroup
\pagebreak{}  

%
%
%
%


\pagestyle{plain} 
\pagenumbering{arabic} 
\setcounter{page}{1}

\chapter{INTRODUCTION} \label{ch:introduction}

``We call ourselves \emph{Homo sapiens} --- man the wise --- because our intelligence is so important to us. For thousand of years, we have tried to understand how we think; that is, how a mere handful of matter can perceive, understand, predict, and manipulate a world far larger and more complicated than itself. The field of artificial intelligence, or AI, goes further still: it attempts not just to understand but also build intelligent entities.''

--- Stuart Russell. ``\textit{Artificial Intelligence: A Modern Approach}'', 2003 \cite{Russell2003}.

\section{Research Problem Overview}
Data-driven approaches and learning algorithms are well suited to solve high-level prediction and control problems in an information-rich world. Learning algorithms are able to learn directly from examples, to search for an underlying pattern of an apparently patternless data, and to improve a decision-making process by continuously repeating it and observing the results.

The primary goal of learning methodologies is to imbue intelligent agents with the capability to autonomously and successfully perform complex tasks, when \emph{a priori} design of the necessary behaviors is intractable. 
Instead, given an objective function for the desired behavior, learning techniques can be used to empirically discover the policy or controller required to satisfy it.
Several classes of these techniques have yielded promising results, including reinforcement learning, learning from demonstrations, interventions, and evaluations.

Reinforcement learning has been shown to work on scenarios with well-designed reward
functions and easily available interactions with the environment.
However, in real-world robotic applications, explicit reward functions are non-existent, and interactions with the hardware are expensive and susceptible to catastrophic failures.
This motivates leveraging human interaction to supply this reward function and task knowledge, to reduce the amount of high-risk interactions with the environment and to safely shape the behavior of robotic agents, thus, enabling \emph{Real-Time Human-in-the-Loop Reinforcement Learning}.

\section{Literature Review} \label{sec:lit_review}

This research focuses on two main areas: learning from human interaction and reinforcement learning.

\subsection{Learning from Human Interaction} \label{subsec:lit_human_interaction}

This section addresses the state-of-the-art and current challenges when learning from human interaction, including learning from human demonstrations, interventions, and evaluations. Other human interaction modalities considered but not included on this work are natural language, eye gaze tracking, gestures, brain electrical activity, and domain knowledge that is applied only to a single task.

\subsubsection{Learning from Demonstrations} \label{sssec:lfd}

Learning from Demonstrations (LfD) can be used to provide a more directed path to these intended behaviors by utilizing examples of humans performing the task.
This technique has the advantage of quickly converging to more stable behaviors.
However, given that it is typically performed offline, it does not provide a mechanism for corrective or preventative inputs when the learned behavior results in undesirable or catastrophic outcomes, potentially due to unseen states.
LfD also inherently requires the maximal burden on the human, requiring them to perform the task many times until the state space has been sufficiently explored, so as to generate a robust policy. 
Also, it necessarily fails when the human is incapable of performing the task successfully at all.

There are many empirical successes of using imitation learning to train a policy or controller based on human demonstrations \cite{Argall2009}.
Early research on learning from human demonstrations was primarily focused on teaching higher-level commands, as for example ``pick", ``move", and ``place" when controlling a robotic arm \cite{kang1997toward,kuniyoshi1994learning,osa2018algorithmic}, which later shifted to trajectory-level planning when the term ``Learning from Demonstrations" became popular \cite{schaal1997learning,schaal1999imitation,atkeson1997robot,osa2018algorithmic}.

For self-driving cars, the earlier Autonomous Land Vehicle In a Neural Network (ALVINN) \cite{pomerleau1989alvinn} by \citeauthor{pomerleau1989alvinn} learned from demonstrations to map from images to discrete actions using a single hidden-layer neural network.
Most recent research also successfully used human demonstrations to train a policy that mapped from front-facing camera images to steering wheel commands using around one hundred hours of human driving data \cite{Bojarski2016}. 
Similar approaches have been taken to train small unmanned air system (sUAS) to navigate through cluttered environments while avoiding obstacles \cite{Giusti2015}.
\citeauthor{Nair2017} \cite{Nair2017} combining self-supervised learning and behavior cloning used images as inputs to create a pixel-level inverse dynamics model of a robotic rope manipulation task.
Related to navigation, there is related work that uses inverse optimal control on human demonstrations to learn navigation skill in a real world robot, and detect failure states and learn recover policies using Gaussian processes \cite{del2018not}.

\citeauthor{Rahmatizadeh2016} \cite{Rahmatizadeh2016} demonstrated that humans can also aid robotic learning through demonstration of the goal task. Long Short-Term Memory (LSTM) networks \cite{hochreiter1997long} can be used to generalize human demonstration from virtual environments and have the learned policy transferred to a physical robot.
When human data are limited, \citeauthor{Deisenroth2015} \cite{Deisenroth2015} used Gaussian Processes to extract more information about each interaction between human and robot to reduce the time required to learn the robotic tasks. These approaches currently limit the robot behavior to what has been demonstrated by the human expert.

Another example of work that attempts to augment learning from demonstrations with additional human interaction is the Dataset Aggregation (DAgger) algorithm \cite{Ross2011}.
DAgger is an iterative algorithm that consists of two policies, a primary agent policy that is used for direct control of a system, and a reference policy that is used to generate additional labels to fine-tune the primary policy towards optimal behavior. Importantly, the reference policy's actions are not taken, but are instead aggregated and used as additional labels to re-train the primary policy for the next iteration. In \cite{Ross2013} DAgger was used to train a collision avoidance policy for an autonomous quadrotor using imitation learning on a set of human demonstrations to learn the primary policy and using the human observer as a reference policy. 
There are some drawbacks to this approach that are worth discussing. As noted by \citeauthor{Ross2013} \cite{Ross2013}, because the human observer is never in direct control of the policy, safety is not guaranteed, since the agent has the potential to visit previously unseen states, which could cause catastrophic failures. 
Additionally, the subsequent labeling by the human can be suboptimal both in the amount of data recorded (perhaps recording more data in suboptimal states than is needed to learn an optimal policy) as well as in capturing the intended result of the human observer's action (as in distinguishing a minor course correction from a sharp turn, or the appropriate combination of actions to perform a behavior). 
Another limitation of DAgger is that the human feedback was provided \emph{offline} after each run while viewing a slower replay of the video stream to improve the resulting label quality. 
This prevents the application to tasks where real-time interaction between humans and agents are required. 

Demonstrations can also be used to infer the cost function used by the demonstrator while performing the task, known as Inverse Optimal Control (IOC) \cite{sammut2011encyclopedia,levine2012continuous,johnson2013inverse,sanchez2017discrete} or Inverse Reinforcement Learning (IRL) \cite{Ng2000,ramachandran2007bayesian,sammut2011encyclopedia} when it is inferred the reward function, which is the negative of the cost function.
Following this principle, related research uses the maximum entropy principle to maximize the entropy of the model distribution subject to the feature constraints from demonstration data \cite{Ziebart2008,Wulfmeier2015a}, evaluated using the difference between value function for the optimal policy obtained using the learned reward model and using the ground truth reward, or expected value difference \cite{Wulfmeier2015a}. Leveraging the principle of maximum entropy, previous approaches \cite{finn2016guided} learns the cost function and optimizes a policy for it at the same time, compares this policy to the demonstrated trajectories, performs and evaluation step, and repeats the procedure until convergence or the desired performance is achieved.
Alternatively to IRL, demonstrations can be integrated directly to Reinforcement Learning (RL) algorithms to increase their sample-efficiency during training. 

\subsubsection{Learning from Interventions}

\emph{Learning from interventions} (LfI), where a human acts as an overseer while an agent is performing a task and periodically takes over control or intervenes when necessary \cite{Akgun2012, Akgun2012a}, can provide a method to improve the agent policy while preventing or mitigating catastrophic behaviors \cite{Saunders2017}.
Similar work by \citeauthor{Hilleli2018} \cite{Hilleli2018} has proposed using human interaction to train a classifier to detect unsafe states, which would then trigger the intervention by a safe policy previously trained based on human demonstration of the task.
This technique can also reduce the number of direct interactions with the agent, when compared to learning from demonstration.
However, this technique suffers from the disadvantage that desired behaviors must be discovered through more variable exploration, resulting in slower convergence and less stable behavior.

Related work includes slowing down the execution of the task controlled by RL agent during training so the human can intervene at any step to prevent catastrophic actions by replacing agent's actions by safe human actions. On \citeauthor{Saunders2017} \cite{Saunders2017}, the task is paused and a model is trained to imitate human intervention decisions. The trained intervention model replaces human and the training continues. They show that this approach works well for simple cases when the human is able to prevent catastrophic actions but it does not scale well to more complex tasks due to the amount of human intervention required. Related work uses interventions to build upon policies trained with human demonstration for robot navigation in the real world \cite{wignessline} and kinesthetic teaching of a robot via keyframes that can be connected to generate trajectories \cite{Akgun2012a}.

Several cases include the combination of demonstrations and mixed initiative control for training robotic polices \cite{Grollman2007} as well as combining imitation learning with interactive reward shaping in a simulated racing game \cite{Hilleli2018}.
Previous approaches combined human actions with robot autonomy to achieve a common goal. On \citeauthor{javdani2015shared} \cite{javdani2015shared}, the robot did not know the goal a-priori and used inverse optimal control and the maximum entropy principle to estimate the distribution over the human's goal based on previous inputs. They showed this approach enabled faster task completion using less data samples when compared to the traditional approach where the robot first predicts the goal then assist for it. Other approaches \cite{reddy2018shared} combined a pre-trained Q-learning policy with human input and showed that the combination is better than the parts by themselves. The main limitation is that it needs a Q-function trained already, which might not be realistic for real-world tasks.

\subsubsection{Learning from Evaluations}

\emph{Learning from evaluation} (LfE) is one such way to leverage human domain knowledge and intent to shape agent behavior through sparse interactions in the form of evaluative feedback, possibly allowing for the approximation of a reward function \cite{knox2009interactively,MacGlashan2017,Warnell2018}.
This technique has the advantage of minimally tasking the human evaluator and can be used when training behaviors they themselves cannot perform, only requiring an understanding of the task goal.
An example would be maneuvering a robotic arm with multiple degrees of freedom in a constrained space. Due to the number of joints and obstacles, the human might not be able to provide complete demonstrations but is able to evaluate if the maneuver was executed successfully or not.
Additionally, if the time-scale of the autonomous system is faster than human reaction time, then it can be challenging for the autonomous system to attribute which actions correspond to the provided feedback (credit assignment problem).

Similar to LfI, it can be slow to converge as the agent can only identify desired or even stable behaviors through more random exploration or indirect guidance from human negative reinforcement of unwanted actions, rather than through more explicit examples of desired behaviors.
Another disadvantage is that the human evaluation signals are generally non-stationary and policy-dependent, for example, what was classified as a good action in the past may not be classified in the same way in the present depending on the human perception of the autonomous system's policy.

\citeauthor{Thomaz2006} \cite{Thomaz2006}  addressed how humans want to reward machines guided by RL algorithms, how to design machines that learn effectively from natural human interaction, and the motivation to maintain safe exploration of new actions and states when using RL algorithms guided by humans.
\citeauthor{Knox2013} \cite{knox2009interactively, Knox2012, Knox2013, Knox2015} have developed methods for adapting human inputs to classic machine learning algorithms via the ''Training an Agent Manually via Evaluative Reinforcement" (TAMER) framework. The human trainer rewards the robot based on its past actions and the framework handles the distribution of these rewards along the state-action pairs to shape the policy of the intelligent agent.
This work was later extended to use deep neural networks as learning representation to solve ATARI games using raw images as inputs, called Deep TAMER \cite{Warnell2018}.
\citeauthor{Leon2013} \cite{Leon2013}  mixed human demonstration and direct natural language human feedback during the task execution as an additional dynamic reward shaping mechanism. So far these methods have only been applied to low-dimensional RL problems.

Differently than using the human as a reward function generator, \citeauthor{MacGlashan2017} presents the Convergent Actor-Critic by Humans (COACH) algorithm, which uses the human as the temporal difference (TD) error. COACH is based on the insight that the advantage function is a good model of human feedback and that actor-critic algorithms update a policy using the critic’s TD error, which is an unbiased estimate of the advantage function.
This work was later extended by \citeauthor{arumugam2019deep} \cite{arumugam2019deep} to use deep neural networks, called Deep COACH, to learn policies mapping raw pixel inputs to actions in Minecraft.

\citeauthor{Christiano2017} \cite{Christiano2017} provided a framework to shape the behavior of an RL agent when the task is relatively complicated to be demonstrated by a human operator. Instead of performing the task, the agent presents two different trajectories (sequence of states and actions) and queries the human the preferred option. The resultant behavior is more natural when compared to the one achieved by human-handcrafted rewards.
Similar to this approach, \citeauthor{singh2019end} \cite{singh2019end} presents VICE-RAQ (Variational Inverse Control with Events - Reinforcement learning with Active Queries) which constructs a classifier to function as a reward function based on example of successful outcomes provide by humans instead of relying on a handcrafted environment reward. After that, the agent can query the human for more labels when it encounters possible goal states.
\citeauthor{ibarz2018reward} \cite{ibarz2018reward} combines learning from expert demonstrations to learning from preferences to solve ATARI games without using the game score. They train an behavior cloning policy then learn a reward using human feedback and trajectory preferences and show that this combined approach outperforms the use of only preferences or only demonstrations. Additionally, they also show that humans can prevent reward hacking due to the constant and online feedback.

\subsection{Reinforcement Learning} \label{subsec:lit_rl}

This section addresses the state-of-the-art and current challenges relevant to the Reinforcement Learning (RL) literature, including sample-efficiency when training RL algorithms and deploying these algorithms in real-world and robotics applications.

\subsubsection{State-of-the-Art Algorithms}

Reinforcement Learning is a fast-moving field and according to \citeauthor{SchulmanMLSS2016} \cite{SchulmanMLSS2016}, there is still no consensus on which RL algorithm would perform better on different applications and robotic systems: actor-only and actor-critic methods, policy gradient methods (e.g., score function, natural or non-natural re-parameterization), value learning (e.g., Q-learning and its derived algorithms), or derivative-free optimization approaches (e.g., cross-entropy method). According to \citeauthor{grondman2012survey} \cite{grondman2012survey}, in (quasi-)stationary Markov Decision Processes (MDP), actor-critic methods should provide policy gradients with lower variance than actor-only methods. However, actor-only methods are more resilient to fast changing non-stationary environments due to the critic not being able to adapt as quick as the actor to the new environment and, consequently, providing poor information for actor updates.

It could be argued that policy gradient reinforcement learning started with \citeauthor{sutton2000policy} \cite{sutton2000policy}, work that paved the way to modern policy gradient algorithms using function approximation. Another classic work by \citeauthor{kakade2002approximately} \cite{kakade2002approximately} proposed an algorithm to estimate the lower bound of a policy performance, which can be used to constrain the policy updates and guarantee policy improvement.
Policy gradient is a sub-field of policy search \cite{deisenroth2013survey} where the gradient is used to guide the search. Alternatively, \citeauthor{levine2013guided} \cite{levine2013guided} uses iterative linear quadratic regular (iLQR \cite{li2004iterative}), initialized from demonstrations, for trajectory optimization and direct policy search. The iLQR samples trajectories from high-reward regions, which are incorporated to the policy search by using regularized importance sampling.

In modern policy gradient methods for deep reinforcement learning (Deep RL), Trust Region Policy Optimization (TRPO) and Truncated Natural Policy Gradient (TNPG) by \citeauthor{Schulman2015} \cite{Schulman2015} iteratively optimize policies with guaranteed monotonic improvement using the natural policy gradient to optimize a given model in the policy-space, constrained by a policy trust region, instead of optimize it in the parameter-space performed by gradient descent.
Similarly, Proximal Policy Optimization (PPO) \cite{Schulman2017} incorporates similar ideas to TRPO in terms of constraining policy updates in the policy-space instead of parameter-space by clipping the objective function and also penalizes the KL divergence between new and old policy in the objective function. The bias-variance trade-off for this methods can be controlled using Generalized Advantage Estimator (GAE) \cite{schulman2015high}, leading to high-performance agents in complex controls benchmarks. Multiple Deep RL methods for continuous control are benchmarked by \citeauthor{Duan2016} \cite{Duan2016} showing the efficacy of the natural policy gradient methods.

Policy gradient methods can also be parallelized. \citeauthor{heess2017emergence} \cite{heess2017emergence} implements a distributed form of PPO where data collection and gradient calculation are distributed between parallel workers. The authors show that agents trained in rich environments without an specifically handcrafted reward function can lead to the development of non-trivial locomotion skills that would be difficult to be encoded in a reward function. 
\citeauthor{Mnih2016} \cite{Mnih2016} introduces the Asynchronous Advantage Actor-Critic (A3C) that substitutes the experience replay buffer by running multiple copies of the environment.
A3C maintains a policy and a value function and uses a mix of $n$-step returns to update both. Policy is updated using the policy gradient based on advantage function, while the value function is updated based on the mean squared error with the $n$-step return. This work also uses target networks to stabilize training, shared layers between policy and value function, and an entropy regularization term with respect to the policy parameters.
The IMPALA (Importance Weighted Actor-Learner Architecture) algorithm by \citeauthor{espeholt2018impala} \cite{espeholt2018impala} generates, in parallel, multiple trajectories and communicate them to a central learner. Since the policy generating the trajectories might be different to the one being updated, they propose the new V-trace algorithm for off-policy correction using truncated importance sampling. This new approach achieves better performance than previous agents
with less data on controls and ATARI benchmarks.

For value learning in discrete action-spaces, for example, when the agent learns to predict based on the temporal differences \cite{sutton1988learning} or learns a Q-function \cite{watkins1992q} representing the expected discounted sum of rewards to be received at the end of an episode when the agent is a given state and performs a given action, much of the recent progress came after the work by \citeauthor{Mnih2013} \cite{Mnih2013,Mnih2015a}, which successfully integrated deep neural networks to reinforcement learning in the Deep Q-Network (DQN) algorithm to solve multiple ATARI games learning directly from images (pixel inputs) --- a significantly larger state-space when compared to learning from low-dimensional states.
Key insight from the DQN work is the benefit of having a replay buffer \cite{lin1992self} to storage previous interactions between agent and environment, more specifically using Prioritized Experience Replay (PER) \cite{Schaul2015}.
By using an experience replay and random sampling, the samples become closer to satisfy the i.i.d. (independent and identically distributed) assumptions used in stochastic gradient-based algorithms and liberates online learning agents from processing transitions in the exact order they are experienced \cite{Schaul2015}.
In addition to that and to replace random sampling, PER samples transition with higher TD-error, corrected with importance sampling, and diversity alleviated using stochastic prioritization, liberating agents from considering transitions with the same frequency that they are experienced \cite{Schaul2015}.
Successful transitions can also be added to this buffer to aid the learning process \cite{lipton2016efficient}.

The DQN work was extended by \citeauthor{Hasselt2015} \cite{Hasselt2010, Hasselt2015} with the Double Deep Q-Networks (DDQN) algorithm, correcting the overestimation of the Q values typical of DQN approach due to the maximization step over estimated action values. DDQN adds a second Q network (called target network) that decouples action selection and action evaluation and leading to better performance on the ATARI benchmark.
Further extending DQN, \citeauthor{wang2015dueling} \cite{wang2015dueling} proposes a neural network architecture that decouples the Q-function in values and advantages while sharing a common feature learning module. This switches the focus to finding the more valuable states and removing the need to evaluate the Q-function for each action, improving DQN performance on tasks with larger action-space.
Instead of only learning the expected value for the value functions, \citeauthor{bellemare2017distributional} \cite{bellemare2017distributional} with Distributional Q-learning proposes learning complete value function distributions. Authors argue that learning the complete distribution, when combined with function approximation, leads to a more stable policy during training.
Many of these advances and extensions to DQN were combined in a single algorithm, Rainbow DQN \cite{hessel2018rainbow}, to achieve state-of-the-art performance on ATARI games.
Rainbow DQN uses noisy networks for exploration \cite{fortunato2017noisy}, instead of the traditional $\epsilon$-greedy approach used in Q-learning which adds noise to the action-space. In the work by \citeauthor{fortunato2017noisy}, the authors add noise to the parameters of the policy (to the weights, if using a neural networks), which are also learned with gradient descent. This approach lead to better performance when compared to traditional RL approaches for exploration.

For value learning in continuous action-spaces, the Deep Deterministic Policy Gradient (DDPG) \cite{Silver2014,lillicrap2015continuous} by \citeauthor{lillicrap2015continuous} was one the first adaptations of DQN for continuous action-spaces. DDPG uses two deep neural networks to learn at the same time a deterministic policy and an Q-function, which is assumed to be differentiable with respect to the actions and used to guide the policy updates based on its gradients, substituting the $\max_a Q(s,a)$ used in DQN.
Alternatively to DDPG, \citeauthor{gu2016continuous} \cite{gu2016continuous} proposes the Normalized Advantage Function (NAF) algorithm that parameterizes the advantage term as a quadratic function so the actions are computed analytically and the agent only needs to learn the Q-values.
The authors also show that incorporating a model-based approach to their method to generate off-policy samples during the beginning of training leads to better performance when compared to just using a model-free approach. Interestingly, the model-based part needs to be switched off a later stages of training because Q-learning performs better with on-policy samples when the policy is already better developed.
Similar to DQN, the main problem with DDPG is overestimating the Q-function and having the policy to exploit this error, leading to policy breaking during training. \citeauthor{fujimoto2018addressing} \cite{fujimoto2018addressing} presents the Twin Delayed DDPG (TD3) algorithm that solves this issue by using two clipped Q-networks, learning the policy slower than the critic, and adding noise and smoothing actions to prevent exploitation of overestimated Q-values. This approach leads to better performance when compared to DDPG.
As an alternative to the ``deterministic policy" of DDPG, \citeauthor{haarnoja2018soft} \cite{haarnoja2018soft, haarnoja2018softapp} introduces the Soft Actor-Critic (SAC) algorithm combining off-policy actor-critic training with a stochastic actor, and further aims to maximize the entropy of this actor with an entropy maximization objective. The entropy term on the objective functions allow to control for the exploration-exploitation trade-off in RL while learning a more robust policy to noisy observations.

Hybrid approaches as the Interpolated Policy Gradient (IPG) algorithm \cite{gu2017interpolated} and Q-Prop \cite{gu2016q} integrate the sample-efficiency of off-policy RL algorithms with the stability of the on-policy ones by combining off- and on-policy updates in the same loss function while satisfying performance bounds.
These algorithm unifies both techniques, has theoretical guarantees on the bias introduced by off-policy updates, and improves on the state-of-the-art model-free deep RL methods.

\subsubsection{Sample-Efficiency in Reinforcement Learning}

Current state-of-the-art RL algorithms heavily rely on multiple graphics and central processing units (GPUs and CPUs) to train intelligent agents on end-to-end approaches \cite{Mnih2013, Hasselt2015, Schaul2015, Mnih2016}.
End-to-end algorithms are initialized with no previous knowledge of the task nor the environment and and the action selection process (trial-and-error) develops almost randomly.
This approach lead to learning algorithms that are able to generalize to multiple problems at the cost of high number of interactions with the environment.
Agents learning simple continuous tasks (e.g., controlling a simulated two-link robotic arm) require on average more than 2.5 million samples to achieve satisfactory performance \cite{lillicrap2015continuous}. On more complex tasks (e.g., control of a simulated humanoid robot or robots with multiple degrees of freedom) they require on average 25 million samples \cite{Duan2016}. Discrete computer game tasks, such as learning how to play Atari games, may require tens of millions of interactions with the environment to achieve state-of-the-art results \cite{Mnih2013, Mnih2015a}.
Less complex algorithms, like REINFORCE \cite{Williams1992} and Cross-entropy Methods (CEM) \cite{Szita2006a}, achieve good performance optimizing policies parametrized as deep neural networks, but are likely to converge prematurely to local optima \cite{Peters2007}.
There is a need for sample-efficient learning architectures in which a learning agent would
quickly learn meaningful behavior, requiring fewer interactions with the environment, easing the
transition to robotic systems performing real-world tasks.

Model-based approaches are a common option to improve sample-efficiency of RL algorithms but it becomes challenging to learn a complex dynamical system with a low number of samples available.
\citeauthor{feinberg2018model} \cite{feinberg2018model} uses model-based approaches to estimate value functions for short term and model-free for long-term estimation of Q-values in reward dense environments. The distinction between how many time steps comprises short- or long-term relies on model value expansion (MVE) for value estimation error. Their approach can be generalized for any actor-critic algorithm.
\citeauthor{nagabandi2018neural} \cite{nagabandi2018neural} proposes to learn the dynamics using state and action as input and the variation in the state as output of a deep neural networks based on data generated by a random policy. With this learned model, at every time step the agent plans the trajectory using model predictive control (MPC) \cite{garcia1989model,morari1999model}, performs one step of the trajectory and re-plans. Reward is based on how close the agent follows the trajectory commanded. This on-policy data is aggregated with the initial data used to learn the model and the dynamics is retrained once in a while.
On their experiments with a real-robot, the model-based approach performs worse than a model-free benchmark. To solve this issues, the authors fine-tune the MPC controller together with learned dynamics using model-free RL. First, the agent's policy is trained to replicate the MPC controller (still mixes on- and off-policy data to re-train policy and dynamic model). At the final stage this policy is transferred to a actor-only algorithm, for example TRPO, and the training continues. This approach outperforms the model-based or model-free alone.

Another approach for leveraging simulated rollouts is proposed by \citeauthor{ha2018world} \cite{ha2018world}. On this work, the agent learns on a unsupervised manner, from random rollouts, a compressed representation of the environment using variational autoencoders (VAE) \cite{kingma2013auto,rezende2014stochastic}, mixture density networks (MDN) \cite{bishop1994mixture,bishop2006pattern}, and recurrent neural networks (RNN) \cite{pearlmutter1989learning,cleeremans1989finite,hochreiter1997long}, and use this learned model to update its policy which is later performs in the real environment. This research is inspired in human factors research that suggests that human brains only remembers certain features and what humans perceive depends on brain's internal model for future prediction. The models learned are mostly imperfect but the authors uses a weighting parameter to quantify the amount of uncertainty. The policy is trained using Covariance-Matrix Adaptation Evolution Strategy (CMA-ES) \cite{hansen1996adapting,hansen2003reducing} using multiple CPU cores to run multiple copies of the environment to parallelize and accelerate learning.
Another model-based approach for imperfect models by \citeauthor{racaniere2017imagination} \cite{racaniere2017imagination} called Imagination-Augmented Agents (I2A) does not rely exclusively on simulated returns. Their approach still learns a forward model but converts trajectories to embeddings, abstracting model imperfections.

Dynamic models can also be use to control exploration of RL environments.
\citeauthor{Pathak2017} \cite{Pathak2017} introduces an intrinsic reward module that rewards the agent based on how well it is able to predict the environment states. The better the prediction, lower the reward. This forces the agent to explore unexplored areas in the state space. In their approach, the agent creates a forward model based on latent-space and also learns an inverse model to predict which actions were used to change states.
Similar model-based approach was earlier proposed by \citeauthor{schmidhuber1991curious} \cite{schmidhuber1991curious} where the agent builds the model by actively provoking situation where it expects to learn something about the dynamics of the environment.

Other approaches to increase sample-efficiency of RL algorithms include combining them with classical controllers \cite{johannink2018residual}, adding auxiliary task and rewards \cite{jaderberg2016reinforcement}, and learning from episodes where the goal is not reached \cite{andrychowicz2017hindsight}.
In the work by \citeauthor{johannink2018residual} \cite{johannink2018residual} the authors add actions produced by the RL policy to actions of a hand-tuned classic controller. The majority of the task is solved by the classic controller and the rest is fine-tuned with RL. In their paper they use the TD3 algorithm to solve a block placing task with handcrafted dense reward function using simulated and real robots, also transferring policies learned in simulation to real world. The state-space for the RL agent includes the goal of the task and action-space is the position of the end-effector.
The addition of auxiliary tasks in the work by \citeauthor{jaderberg2016reinforcement} \cite{jaderberg2016reinforcement} introduces the UNsupervised REinforcement and Auxiliary Learning (UNREAL) agent, which translates to adding two additional reward signals (in addition to the one returned by the environment): maximize change in pixel intensity (generally correlated to important events in the screen) and correct reward prediction based on stacked input frames. It does increase the sample-efficiency because the agent is able to extract more information from the same amount of data plus converts any tasks with sparse reward signals to dense, aiding the optimization process.
The main insight to learn from episodes where the goal is not reached comes from \citeauthor{andrychowicz2017hindsight} \cite{andrychowicz2017hindsight} where the Hindsight Experience Replay (HER) algorithm is proposed. HER adds a goal vector to standard observation input for RL algorithms. At the end of each episode, it interprets the final state as a pseudo-goal so even when the agent fails to achieve the desired goal, it at least learns how to achieve this alternative state. It solves to problem of sparse rewards tasks with narrow successful regions where positive rewards are difficult to achieve.

\subsubsection{Human-in-the-loop Reinforcement Learning}

Reinforcement learning has been proven to work on scenarios with well-designed reward functions and easily available interactions with the environment \cite{goecks2018efficiently}. 
However, in real-world robotic applications, explicit reward functions are non-existent, and interactions with the hardware are expensive and susceptible to catastrophic failures \cite{goecks2018efficiently}. 
This motivates leveraging human interaction to supply this reward function and task knowledge, to reduce the amount of high-risk interactions with the environment, and to safely shape the behavior of robotic agents \cite{goecks2018efficiently}.

A common approach in human-in-the-loop reinforcement learning is modify the reinforcement learning loss function to leverage a human dataset of trajectories to solve the desired task.
\citeauthor{hester2018deep} \cite{hester2018deep} presents the iconic Deep Q-learning from Demonstrations (DQfD) algorithm where the agent is pre-trained and trained with four combined losses: $1$-step double Q-learning loss, $n$-step double Q-learning loss, supervised large margin classification loss, and L2 regularization on network weights and biases.
Authors argue that the combination of all four losses during pre-training is essential to learn a unified representation that is not destroyed when the loss function changes from the pre-training to the training phase.
Even after pre-training, the agent must continue using the expert data \cite{hester2018deep}.
They also modified the replay buffer to combine expert demonstrations and self-generated data (never overwriting the expert data) sampling proportional amounts of each type.
The authors shows that DQfD lead to more sample-efficient approach to solve the ATARI benchmarks.

Similar to DQfD but without the pre-training phase and the supervised loss in the loss function \citeauthor{vevcerik2017leveraging} \cite{vevcerik2017leveraging} present the Deep Deterministic Policy Gradient from Demonstration (DDPGfD) algorithm. It loads all expert demonstrations in a prioritized experience replay buffer before training, which is kept throughout the training process, and performs more than one learning step when sampling data from the buffers. The main contribution of this algorithm is that it can be used in continuous action-spaces, while DQfD is restricted to discrete ones.
Another extension to DQfD proposed by \citeauthor{pohlen2018observe} \cite{pohlen2018observe}, the Ape-X DQfD algorithm differs from DQfD in three aspects: no pre-training phase using only expert transitions, fixed ratio of actor and expert transitions, and the supervised loss is only applied to the best expert episode instead of all episodes. This was the first deep RL algorithm to solve the first level of Montezuma's Revenge, the ATARI game in which DQN had poor performance \cite{Mnih2015a}.
A pre-training strategy is also used by \citeauthor{cruz2019jointly} \cite{cruz2019jointly} feature and policy learning in RL. The authors tackle the feature learning during the pre-training phase with a combined supervised classification loss, an unsupervised reconstruction loss, and a value function loss. They also use a transformed Bellman operator to scale the whole action-value function without clipping or modifying the reward signal.
Another contribution is their self-imitation learning approach encourages the agent to imitate past decisions only when the returns are larger than the current estimated value.
Their approach is tested on ATARI games and shows improved sample-efficiency when compared to deep RL algorithms.

Building on top of Deep Deterministic Policy Gradients and Hindsight Experience Replay, \citeauthor{Nair2018ICRA} \cite{Nair2018ICRA} shows that when an actor is pre-trained with demonstration off-policy data and transferred to RL for training, the actor weights are destroyed by the untrained critic. Also, training the critic with only off-policy data it fails to correctly evaluate on-policy transitions, which motivates the combined loss approaches and mxiture between on- and off-policy data.
Similarly, the approach by \citeauthor{sun2017deeply} \cite{sun2017deeply} combines expert and agent data in a mix of on- and off-policy updates to reduce distribution mismatch between training and testing dataset and argue that interactive approaches leads to better performance through a reduction to no-regret online learning. This algorithm matches expert performance and even surpass it in case the expert is not optimal.
Multiple critics can also be used to constrain the actor's policy, as showed by \citeauthor{durugkar2019multi} \cite{durugkar2019multi} with the Multi-Preference Actor Critic (M-PAC) framework.

Mixing directly expert trajectories with RL, \citeauthor{schoettler2019deep} \cite{schoettler2019deep} extends the concept of Residual RL, where actions selected by the agent are added to actions performed by an expert controller, to the setting of vision-based manipulation tasks. The authors show that this approach can be used to train agents directly in the real-world.
\citeauthor{peng2018deepmimic} \cite{peng2018deepmimic} uses expert trajectories as reference to compute a reward function that penalizes mismatch between reference and performed trajectories. Constructed this reward function, the agent can be trained with any RL algorithm.
The authors show that multiple references can be integrate into the learning process to develop multi-skilled agents capable of performing multiple and combined skills.

Humans can also aid RL algorithms through natural language.
\citeauthor{kaplan2017beating} \cite{kaplan2017beating} trains an agent that learns the meaning of English words, translates them to game states, and learns how to execute these commands. Game images are used to train a policy and state value function using an state vector augmented with language instructions.
The authors generate training data by template matching the frames with the instructions, which is later used to verify if instructions were completed, giving the agent an additional reward. They also generated training data by playing the game and leaving the agent at specific positions or performing specific actions to cover gaps in the demonstration dataset. 
Similarly, \citeauthor{waytowich2019grounding} \cite{waytowich2019grounding} proposed a model that learns a mutual embedding between natural language commands and goal-states that can be used as a form of reward shaping and improve RL performance on environment with sparse rewards. Their approach is tested on the SC2LE (StarCraft II Learning Environment) \cite{vinyals2017starcraft}.
For dialogue applications, \citeauthor{jaques2019way} \cite{jaques2019way} train a generative model on known sequences of interaction data that is later used as reference during the RL training phase, penalizing divergence from this prior generative model with different forms of KL-control. Authors argue that this is necessary to remove the overestimation bias when agents are pre-trained directly with expert data.

Humans be queried to provide feedback, called active learning, and train a reward model for RL agents.
\citeauthor{daniel2014active} \cite{daniel2014active} propose to use humans to evaluate the agent's actions (assign numerical values to observed trajectories), instead of having them to design the reward function. Since human evaluation is noisy and non-repeatable, the authors also propose to learn a probabilistic model of the reward function using Gaussian processes and a noise hyperparameter.
The reward model is learned by sampling from a memory buffer the best performing trajectories plus the trajectories evaluated by the human expert. When to query the expert is based on the ratio of the predictive variance and observation noise, called lambda. If lambda is above a certain threshold, the agent queries the human and uses this evaluation to update the reward model.
The proposed approach is evaluated in a simulated environment and in a real robot for grasping tasks.
\citeauthor{su2016line} \cite{su2016line} use RL to train a dialogue policy at the same time as a reward model is trained based on human feedback. They also use Gaussian processes to quantify uncertainty to reduce number of queries to the user. This allow for online learning in real-world dialogue systems without manually annotated data.
\citeauthor{lopes2009active} \cite{lopes2009active} introduces active reward learning for inverse reinforcement learning where the agent queries the user for demostrations at specific areas of the state-space. It does that by measuring a state-wise entropy in the distribution and queries the user for the most informative states.
Including elements of active learning, \citeauthor{Hilleli2016} \cite{Hilleli2016} initially use human to generate expert trajectories and train a policy to imitate them. The human later labels the states in order to train a reward model and uses DDQN deep RL algorithm to optimize the reward, which learns to perform better than the human demonstrator.

More recent meta-learning approaches, as proposed by \citeauthor{zhou2019watch} \cite{zhou2019watch}, learns initially from few demonstration and later from binary feedback. This approach works with two separate policies: trial (learn from demonstrations) and retrial (learn from experience) policy. Trial policy is trained in a meta-imitation learning setup and is frozen to generate data to the retrial policy, which is trained on this stationary data.
\citeauthor{huang2019continuous} \cite{huang2019continuous} address the one-shot imitation learning problem, where the goal is to execute a previously unseen task based on only one demonstration. This is made possible by formulating the one-shot imitation learning as a symbolic planning and symbol grounding problem together with relaxation of the discrete symbolic planner to plan on the probabilistic outputs of the symbol grounding model.

\subsubsection{Reinforcement Learning in Robotics and in the Real World}
In robotics, it is often challenging to successfully integrate learning algorithms and controllers. \citeauthor{Valasek2005,Valasek2008} \cite{Valasek2005,Valasek2008}, developed a novel architecture called Adaptive-Reinforcement Learning Control (A-RLC) for morphing air vehicles. It defines optimal aircraft shape based on mission objectives, learns how to morph to the desired format, and controls variations on system's dynamics due to the new shape.

Reinforcement learning methods often fail in dynamical systems due to failure in properly representing the problem, inaccurate function approximation, and not accounting for time dependency in the selection of actions, as pointed by \citeauthor{Kirkpatrick2013a} \cite{Kirkpatrick2013a}. Controlling real continuous systems requires computer-based control, so sampling of the continuous system is necessary \cite{Kirkpatrick2013a}. If necessary to discretize these continuous systems, the size of the grid must be tuned to capture environment details and still achieve good convergence, as presented by \citeauthor{Lampton2011} \cite{Lampton2011}.

Often applied to simulated tasks and environments, RL algorithms need to overcome their extensive trial-and-error approach before they can be applied to real robotic systems. \citeauthor{Kober2015} \cite{Kober2015}, in their extensive survey of reinforcement learning applied to robotics, highlighted several open questions that need to be answered in order to have RL techniques reliably deployed to robotic systems, for example:
1) How to choose the best system representation in terms of approximate states, value functions, and policies, when dealing with continuous and multi-dimensional robotics problems?
2) How to generate reward functions straight from data, reducing the need for manually-engineered solutions?
3) How to incorporate prior knowledge and how much is needed to make robotics reinforcement learning problems tractable?
4) How to deal with parameter sensitivity and reduce the need for preprocessing sensory data, simplifying behavior programming?
5) How to approach model error and under-modeling during simulated training --- when the simulated environment cannot perfectly model every aspect of the real environment, since the policies learned only in simulation frequently cannot be transferred directly to the robot?
6) How to develop accurate simulation training environments which will ease transfer of policies learned from simulation to real-world hardware tasks?

\citeauthor{Amodei2016} \cite{Amodei2016}  addressed concrete risks of poorly-design AI systems deployed to real-world applications. Specifically in RL, \emph{reward hacking} and \emph{safe exploration} are the main issues. Reward hacking refers to developing unintended behavior by exploiting a faulty reward function. For example, a running robot that is rewarded according to its speed could learn to run in circles at maximum speed instead of completing the running course. Safe exploration refers to developing unsafe behavior (e.g., behavior that would damage the agent or its environment) while exploring the observation space. These problems are currently unsolved.

There is also a research gap on \emph{trust on autonomous systems}. Prior studies \cite{Shahrdar2019} indicate that humans have low levels of trust in semi and fully autonomous systems. This is directly correlated with the unpredictability of the robot \cite{Stormont2008}, their rate of success and reliability \cite{Carlson2004, Abd2017, Howard2014}, and how knowledgeable the user is about an autonomous system \cite{Uggirala2004} and the task to be completed \cite{Helldin2013}. This highlights the need to develop a platform to illustrate what future actions are being planned by the intelligent agent or any other form of education to demystify the intelligent behavior. This would improve human trust in these systems and allow humans to supervise future actions and consequences.

\section{Motivation and Objective} \label{sec:objectives}

\subsection{Motivation}

As discussed on Sections \ref{subsec:lit_human_interaction} and \ref{subsec:lit_rl}, current state-of-the-art research on learning algorithms focuses on end-to-end approaches.
The learning agent is initialized with no previous knowledge of the task nor the environment and the action selection process (trial-and-error) develops almost randomly.
End-to-end approaches abstract intermediate learning milestones at the cost of high number of interactions with the environment.
An important question is how RL agents could incorporate prior knowledge or learn directly from a trained policy.
For example, incorporate human knowledge or learn from an existing sub-optimal controller. Humans are preferred because they dictate the task to be accomplished and have traits that are in the frontier of learning algorithms research, as for example long-term autonomy \cite{Kunze2018} and continual lifelong learning \cite{Parisi2019}.

Currently, there is no definite approach to combining humans and RL agents.
Machines that can be taught by and seamlessly work in collaboration with humans will not only learn faster, but will potentially expand the frontier of having machines augmenting human performance.
The rapid increase of automation and development of AI techniques demands a better understanding of how humans and machines can better work in collaboration.

\subsection{Objective}

The objective of this research work is to develop the theoretical foundation to integrate multiple human input modalities (in the form of demonstration, intervention, and evaluation) to allow intelligent data-driven agents to learn in real-time, safely, and with fewer samples (increase sample-efficiency) by bootstrapping human interaction data and reinforcement learning theory.

\section{Research Contributions} \label{sec:contributions}

\begin{figure}[!t]
    \centering
    \includegraphics[width=.75\linewidth]{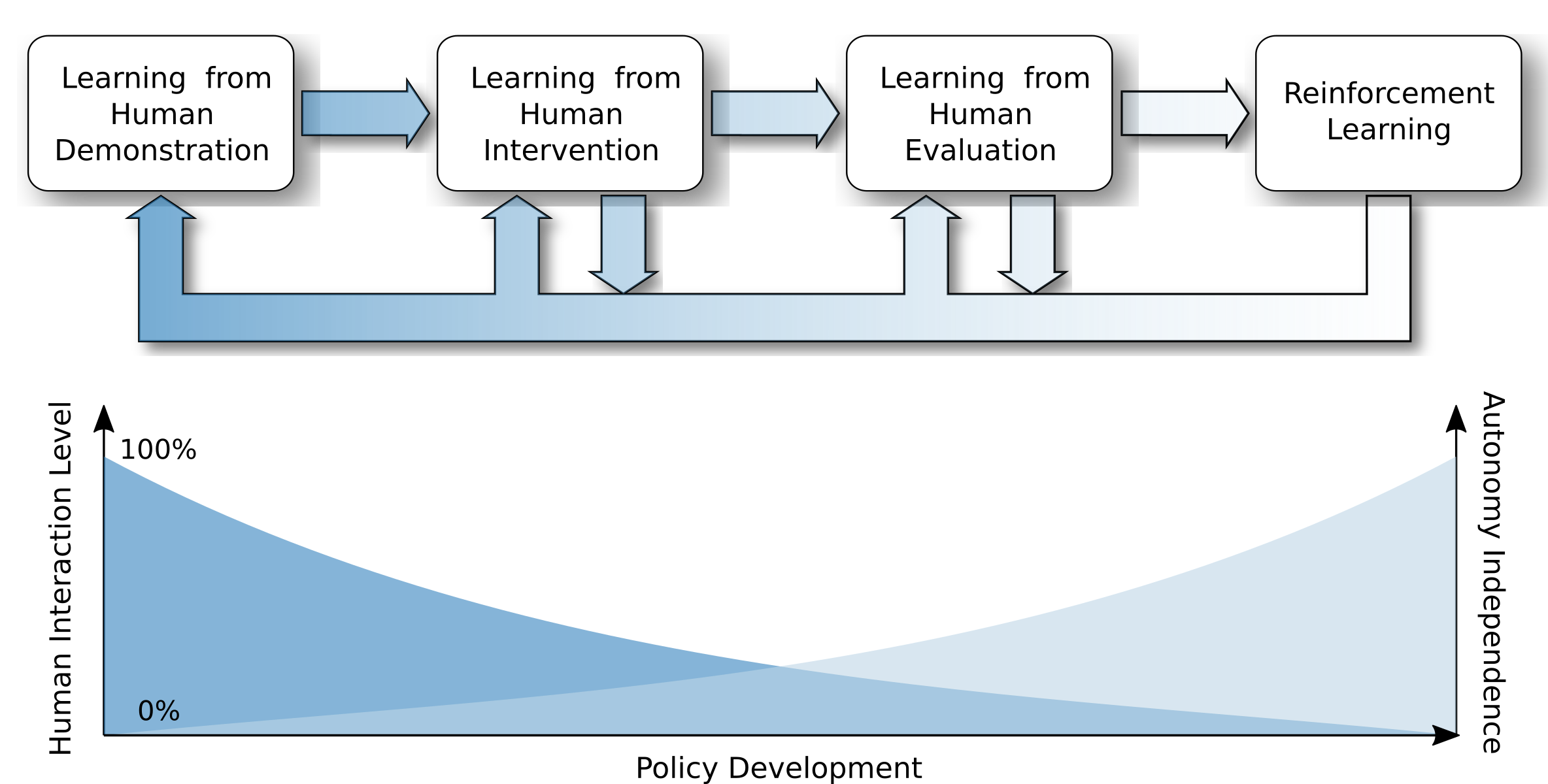}
    \caption{Cycle-of-Learning for Autonomous Systems from Human Interaction: as the policy develops, the autonomy independence increases while the human interaction level decreases. Adapted from \cite{Waytowich2018}.}
    \label{fig:diagram}
\end{figure}

This dissertation investigates how to efficiently and safely train autonomous systems in real-time by extending supervised and reinforcement learning theories and human input modalities. Intuitively named the \textbf{\nameProj} (CoL), this research investigates and develops training autonomous system policies through human interaction that is based on the evidence-based teaching methods of ``teaching for mastery" \cite{Block1974}.
On teaching for mastery, or mastery learning \cite{Block1974, Kulik1990, Anderson1994}, human interaction is provided systematically, from complete human control (and no autonomy) when learning from demonstrations to no human control (and full autonomy) at the reinforcement learning stage. After learning from demonstration the policy is evaluated in real-time and given feedback only on specific areas that needs to improve through interventions. After mastering an initial concept, the role of the human is only to evaluate the performance of the policy while it develops by itself using reinforcement learning.
This process is repeated as new concepts are introduced and the policy masters previous concepts, as illustrated in Figure \ref{fig:diagram}.

While extensive research has been conducted into each of these stages separately in the context of machine learning and robotics, a complete theoretical integration of these concepts has yet to be done. This theoretical integration will be important to fielding adaptable autonomous systems that can be trained in real-time to perform new behaviors depending on the task at hand, in a manner that does not require expert programming. 

The contributions of the research are:

\begin{enumerate}
    \item A novel theoretical contribution of incorporating multiple human input modalities to reinforcement learning algorithms to perform, simultaneously, on-policy and off-policy learning, leveraging their strengths and mitigating their weaknesses.
    \item Reduction of the data-complexity and the number of samples (interactions with the environment) required by state-of-the-art reinforcement learning algorithms to achieve satisfactory performance.
    \item Enabling reinforcement learning agents to be trained in real-time, with real-world data, safely.
\end{enumerate}

Results presented in this dissertation show that policies trained with human demonstrations and human interventions together outperform policies trained with just human demonstrations while simultaneously using fewer data samples. To the best of knowledge, this is the first result showing that training a policy with a specific sequence of human interactions (demonstrations, then interventions) outperforms training a policy with just human demonstrations (controlling for the total amount of time of human interactions). One can obtain this performance with significantly reduced data requirements, providing initial evidence that the role of the human should adapt during the training of safe autonomous systems.

\section{Organization} \label{sec:organization}

This dissertation is organized as follows.
Chapter \ref{ch:bc} details Behavior Cloning and Imitation Learning in discrete or continuous spaces and how these algorithms can be used to replicate human behavior.
Chapter \ref{ch:rl} formalizes the Reinforcement Learning problem as a Markov Decision Process and explains how this research field transitioned from shallow to deep representations for policy and value functions when using function approximation.
This chapter also includes an unusual application example: Deep Reinforcement Learning on Intelligent Motion Video Guidance for Unmanned Air System Ground Target Tracking.
Chapters \ref{ch:tt} and \ref{ch:wing} presents to case studies of Deep Reinforcement Learning applied to Aerospace problems: the tracking of a ground target using unmanned aerial systems and learning to control morphing wings.
Chapter \ref{ch:irl} initializes the human-in-the-loop work with a case study leveraging Inverse Reinforcement Learning methods to land a simulated lander and simulated unmanned aerial vehicle by learning from human demonstrations of the task.
Chapter \ref{ch:cybersteer} addresses how to leverage initial human demonstrations to learn the intrinsic reward function used by the human to pursue the task goal while performing the demonstrations.
This reward signal is then used to feed reinforcement learning algorithms in a unmanned aerial system collision-avoidance scenario.
Chapter \ref{ch:col} introduces the main theoretical contribution of this dissertation: Cycle-of-Learning for Autonomous Systems from Human Interaction.
It covers how to combine multiple forms of human interaction and integrate them to reinforcement learning.
Chapter \ref{ch:col_aaai19} explains how to efficiently combine human demonstrations and interventions on learning algorithms to increase task performance while using fewer human samples.
The main study case is learning from human interaction autonomous landing controllers for unmanned aerial systems.
Chapter \ref{ch:col_loss} finalizes the Cycle-of-Learning by transitioning from policies learned from human interaction to reinforcement learning.
This approach is validated on the previous unmanned aerial system landing scenario and standard reinforcement learning tasks for continuous control.
Conclusions from the various applications and recommendations for future work are summarized in Chapter
\ref{ch:conclusions}.

\chapter{MACHINE LEARNING AND CLONING BEHAVIORS} \label{ch:bc}

This chapter introduces the notation and basics of machine learning, specifically supervised learning, and how an algorithms can learn to clone behaviors by framing the learning problem as a sequential decision making problem.
This includes important concepts in machine learning, the mathematical formulation of sequential decision making problem framed as a Markov Decision Process, the relation between imitation learning and behavior cloning, success cases in the literature, and the limitations of this approach in isolation.

\section{Problem Definition}

According to \citeauthor{barber2012bayesian} \cite{barber2012bayesian}, Machine Learning is the research field concerned in automating large-scale data analysis, leveraging concepts of statistics with focus on mathematical modeling and prediction.
The dataset that one uses to learn from to create this mathematical model, or \emph{train the model}, is called the \emph{training set} \cite{bishop2006pattern}. Ideally, the model trained on this training set will be able to \emph{generalize} to new unseen inputs, often called the \emph{test set} \cite{bishop2006pattern}. The process of training the model depends on how this training set is structured, which further subdivides the field of machine learning in three groups: \emph{supervised learning}, \emph{unsupervised learning}, and \emph{reinforcement learning}.
In supervised learning, the training set contains data inputs and desired outputs so the learning model changes its internal parameters during training in order to produce outputs closer to the desired ones given the same input data. For example, a machine learning model for autonomous vehicles would be interested in identifying traffic signs and lights from image data so the vehicle can behave according to the traffic rules. The training set would consist in multiple images as input and the location of all traffic signs and lights in the image as output.
In unsupervised learning, the training set consists only of input data and no desired output. The main goal is to discover groups in similar category (clustering), estimate the input distribution (density estimation), project data in different dimensions to aid visualization \cite{bishop2006pattern}. An example of unsupervised learning would be a machine learning model that is trained to cluster customers of a bank based on their credit card transaction history.
In Reinforcement Learning there is no training set and machine learning model is trained by trial-and-error while performing a desired task in order to maximize a performance metric.
A example of reinforcement learning would be a machine learning model that receives images from a screen while playing a computer game and aims to maximize the game score by performing different commands and observing its effect in the next game screens and score.

Supervised learning is covered with more details on Section \ref{sec:sl} and Reinforcement Learning on Chapter \ref{ch:rl} of this research work. Unsupervised learning is out of scope of this dissertation however introductory material in the area can be found in \citeauthor{barber2012bayesian} \cite{barber2012bayesian} and \citeauthor{bishop2006pattern} \cite{bishop2006pattern}.

\section{Supervised Learning} \label{sec:sl}

\citeauthor{barber2012bayesian} \cite{barber2012bayesian} defines supervised learning problem as, given a set of data pairs

\begin{equation*}
\mathcal{D} = \{ (x^n,y^n), n=1, \dots, N \},     
\end{equation*}

learning the relationship between the input $x$ and output $y$ such that, given an unseen input $x'$ not present in $\mathcal{D}$ the predicted $y'$ output is accurate, assuming the pair $(x',y')$ is generated by the same unknown process that generated the set $\mathcal{D}$. In other words, it is desired to predict $y'$ conditioned on know $x'$ and the set $\mathcal{D}$, which can be represented by the conditional distribution $p(y'|x, \mathcal{D})$. If the output $y$ is one of a discrete number representing a possible discrete outcome, the supervised learning problem is called ``classification'' (for example, given computed tomography (CT) x-ray images of the lung, predicts if the person is likely or not to have cancer \cite{ardila2019end}). If the output $y$ is a continuous variable, the supervised learning problem is called ``regression'' (for example, predicting future market price of a given stock based on financial news articles \cite{vargas2017deep}). This distinction is important to define the loss function used to compute the misfit between true and predicted model, which guides the optimization process used to learn the model, to be discussed in more detail throughout the next sections.

\subsection{Model Representation and Learning}

The word \emph{model} has not been properly defined in this work yet. The model is the mathematical representation of the underlying function or distribution it is desired to approximate, from which the training set is sample from, in order to generalize to unseen data samples and predict future outcomes. Models differ by the number and the mathematical relationship between its parameters, which should be chosen based on the underlying function the model is desired to approximate and assumptions made. Examples of these assumptions are if the underlying process that generated the training set is linear or nonlinear, noise processes that affect the measurements, and others.

As explained by \citeauthor{bishop2006pattern} \cite{bishop2006pattern}, an example of model representation would be a polynomial function of the form

\begin{equation}
    y(x,\vec{w}) = w_0 + w_1x + w_2x^2 + \dots + w_Mx^M = \sum^M_{j=0} w_jx^j,
\end{equation}

where $M$ is the order of the polynomial and its coefficients $w_0, \dots, w_M$ are collectively denoted by the vector $\vec{w}$. Even though there is a linear relation between the polynomial coefficients $\vec{w}$, the model $y(x,\vec{w})$ is nonlinear and, consequently, is able to approximate nonlinear functions. The main limitation of models comprised of linear combination of fixed basis is that they do not scale well as the number of inputs $x$ increases, so called ``curse of dimensionality'' \cite{Sutton2018,bishop2006pattern}. For example, for the polynomial case of order $M$ with $N$ inputs, the number of coefficients grow following the power law $N^M$.

\subsubsection{Neural Networks}

\begin{figure}[!htb]%
  \centering
  \includegraphics[width=.55\linewidth]{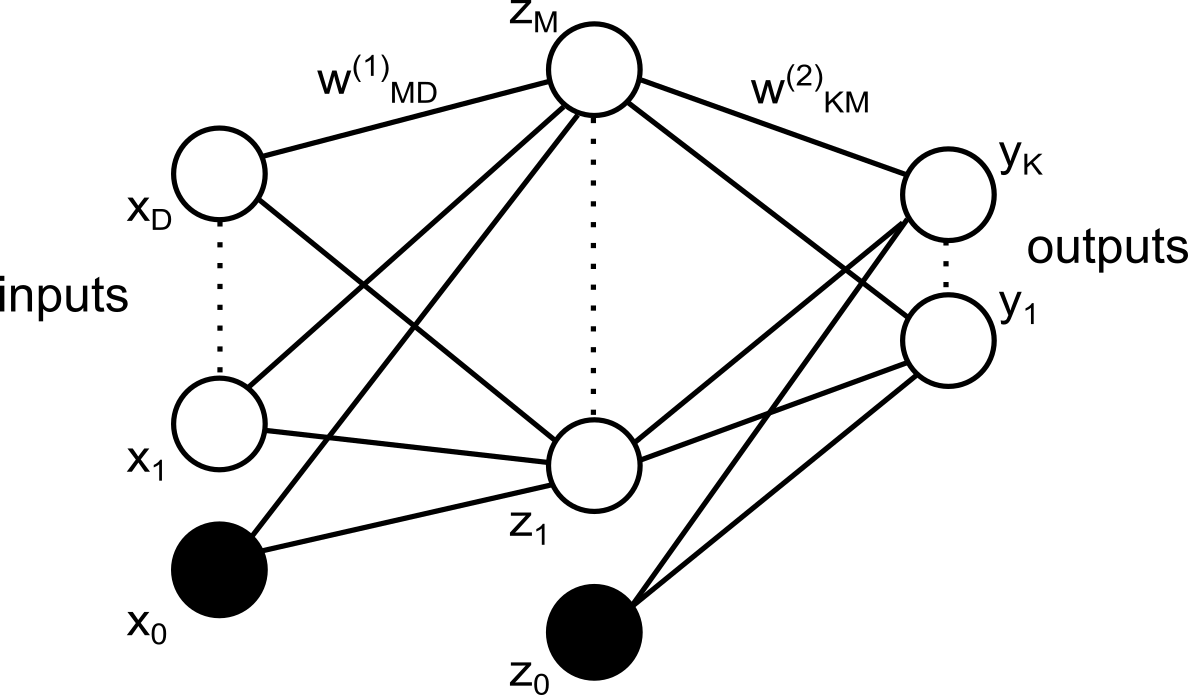}%
  \caption{A neural network with $D$ inputs, one hidden layer with $M$ units, and $K$ outputs. Adapted from \cite{bishop2006pattern}.}%
  \label{f:nn_bishop}%
\end{figure}

One of the model representations that counters these limitations are \emph{neural networks} \cite{bishop2006pattern,nielsen2015neural,Goodfellow2016}, also called feedforward neural networks, or multilayer perceptrons (MLPs) \cite{Goodfellow2016}.
Neural networks are said to be \emph{universal approximators} due to their approximation properties, which are able to approximate any function given enough network size \cite{funahashi1989approximate,cybenko1989approximation,hornik1989multilayer,stinchcombe1989universal,cotter1990stone,ito1991representation,kreinovich1991arbitrary,hornik1991approximation,ripley1996pattern}.
Originally inspired by biological neural networks \cite{mcculloch1943logical,widrow1960adaptive,rosenblatt1961principles,rumelhart1985learning}, neural networks are a series of functional transformations where a linear combination of parameters $\vec{w}$ are transformed for nonlinear activation functions $h(\cdot)$ in cascade along \emph{layers}, which consists of many \emph{units} that act in parallel representing a vector-to-scalar function \cite{Goodfellow2016}. A neural network with $D$ inputs, one hidden layer with $M$ units, and $K$ outputs, as seen in Figure \ref{f:nn_bishop}, is be represented by the equation

\begin{equation}
    y_k(\vec{x},\vec{w}) = h_2 \left( \sum^M_{j=1} w_{kj}^{(2)}h_1 \left( \sum^D_{i=1} w_{ji}^{(1)}x_i + w_{j0}^{(1)} \right) + w_{k0}^{(1)} \right).
\end{equation}

\begin{figure}[!htb]%
  \centering
  \includegraphics[width=1.0\linewidth]{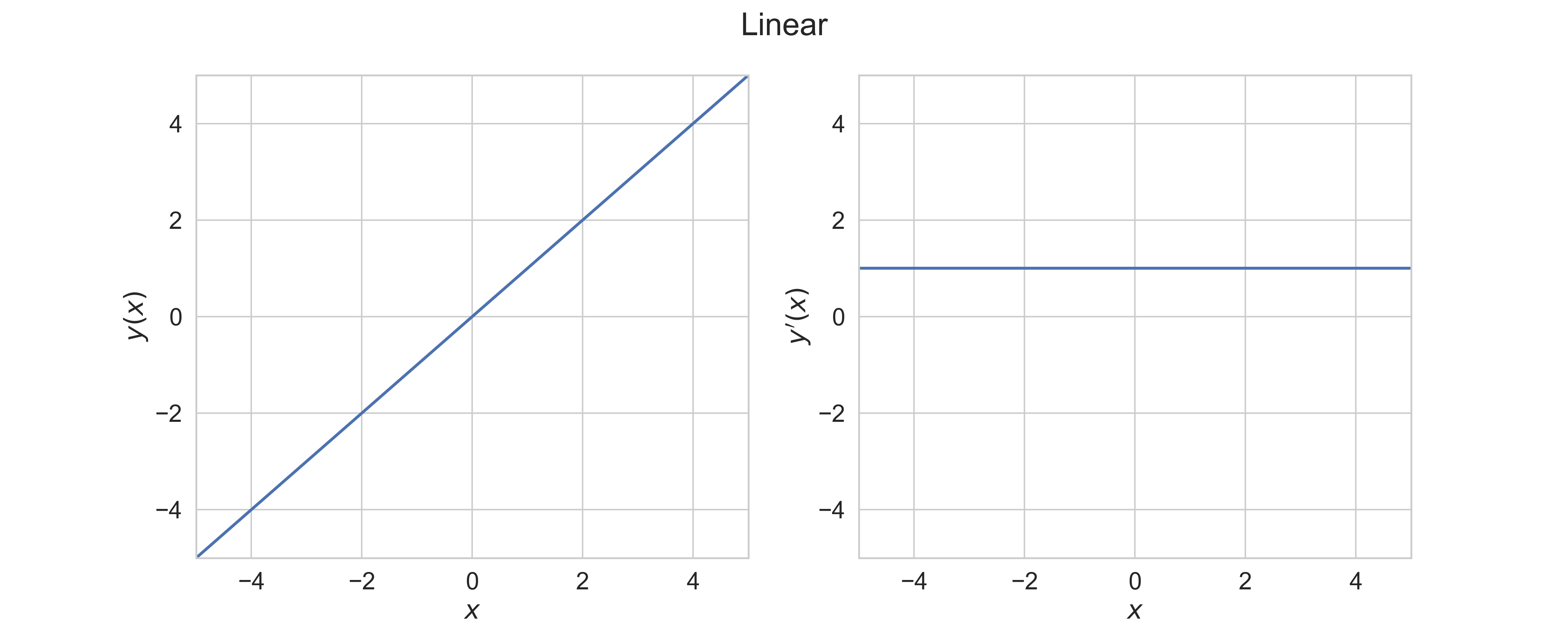}%
  \caption{Linear activation function $y(x)$ and its derivative $y'(x)$.}%
  \label{f:linear_act_fn}%
\end{figure}

\begin{figure}[!htb]%
  \centering
  \includegraphics[width=1.0\linewidth]{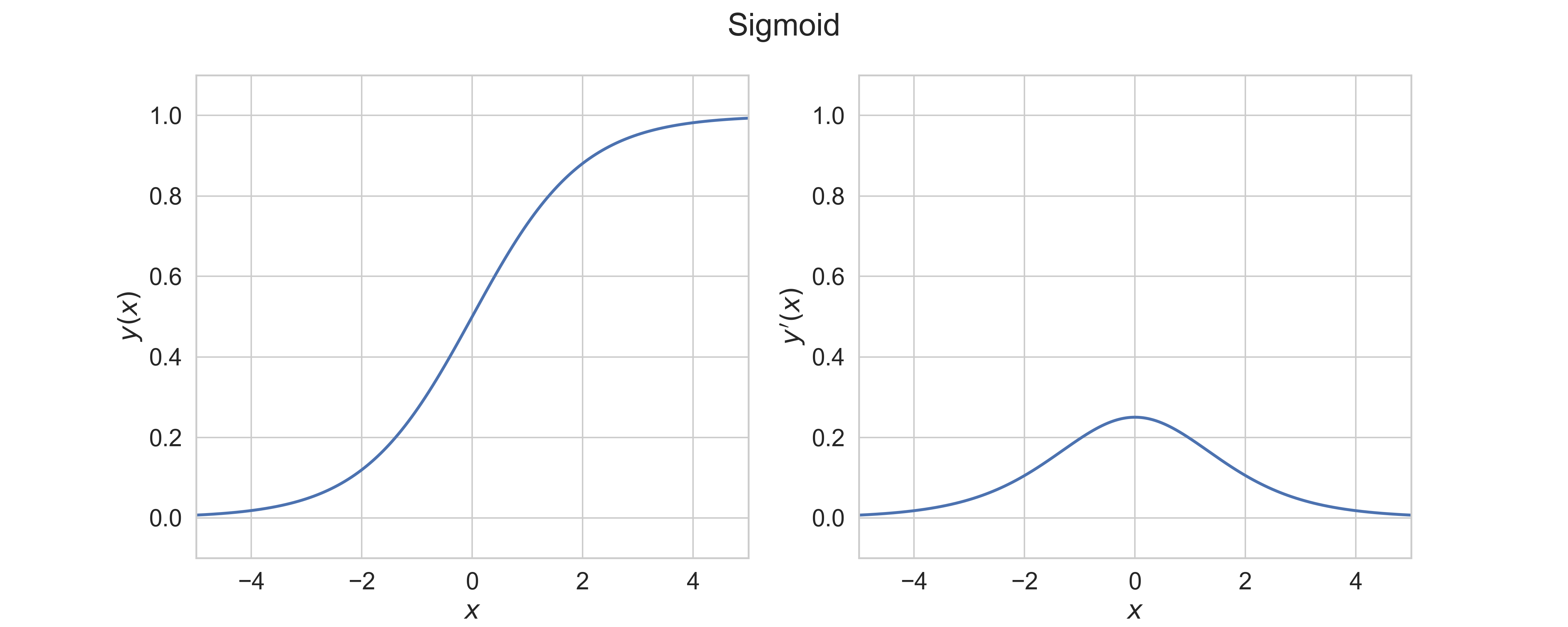}%
  \caption{Sigmoid activation function $y(x)$ and its derivative $y'(x)$.}%
  \label{f:sigmoid_act_fn}%
\end{figure}

\begin{figure}[!htb]%
  \centering
  \includegraphics[width=1.0\linewidth]{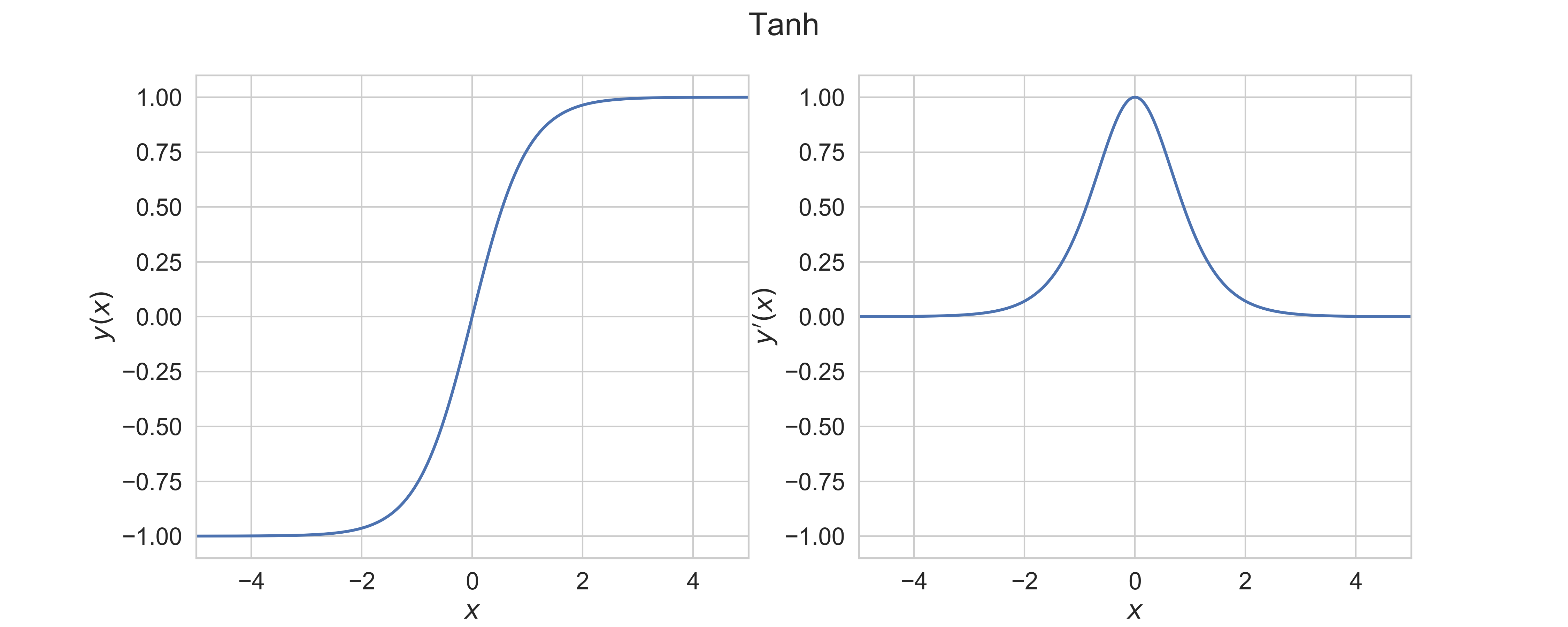}%
  \caption{Hyperbolic Tangent (Tanh) activation function $y(x)$ and its derivative $y'(x)$.}%
  \label{f:tanh_act_fn}%
\end{figure}

\begin{figure}[!htb]%
  \centering
  \includegraphics[width=1.0\linewidth]{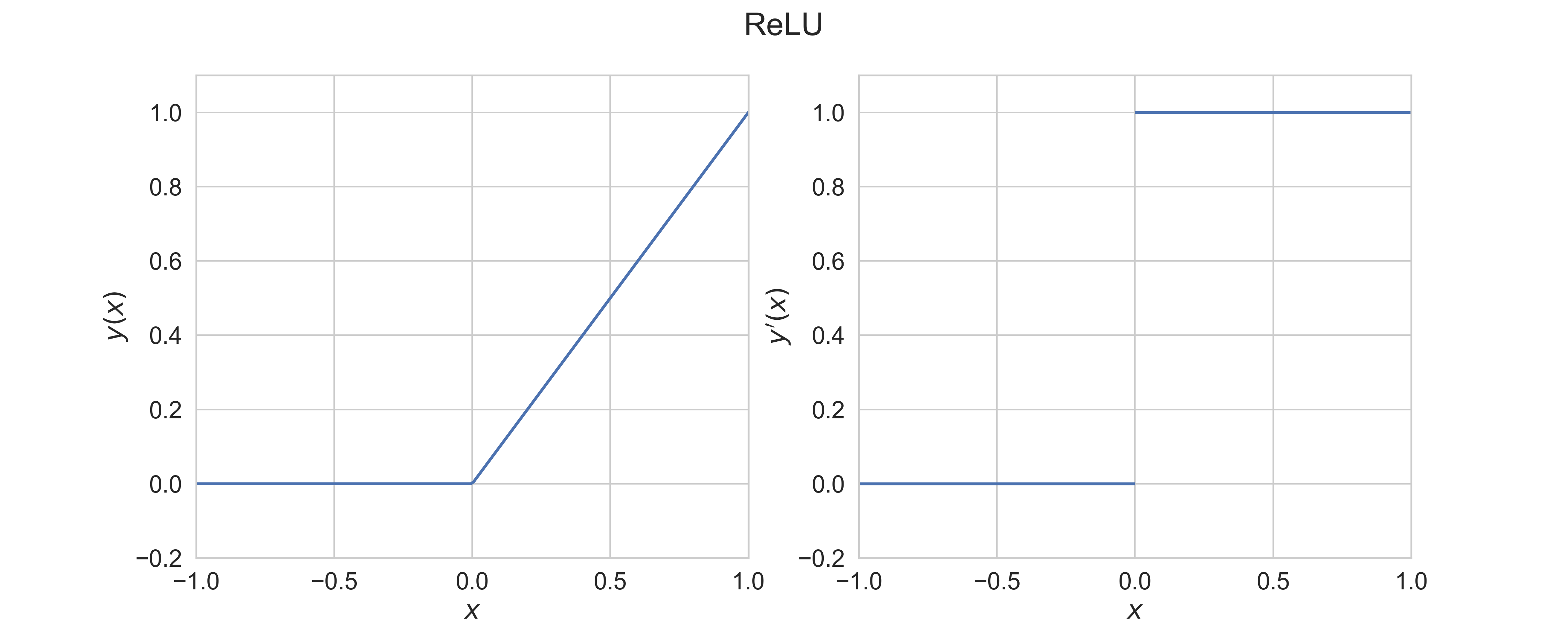}%
  \caption{Rectified Linear Unit (ReLU) activation function $y(x)$ and its derivative $y'(x)$.}%
  \label{f:relu_act_fn}%
\end{figure}

\begin{figure}[!htb]%
  \centering
  \includegraphics[width=1.0\linewidth]{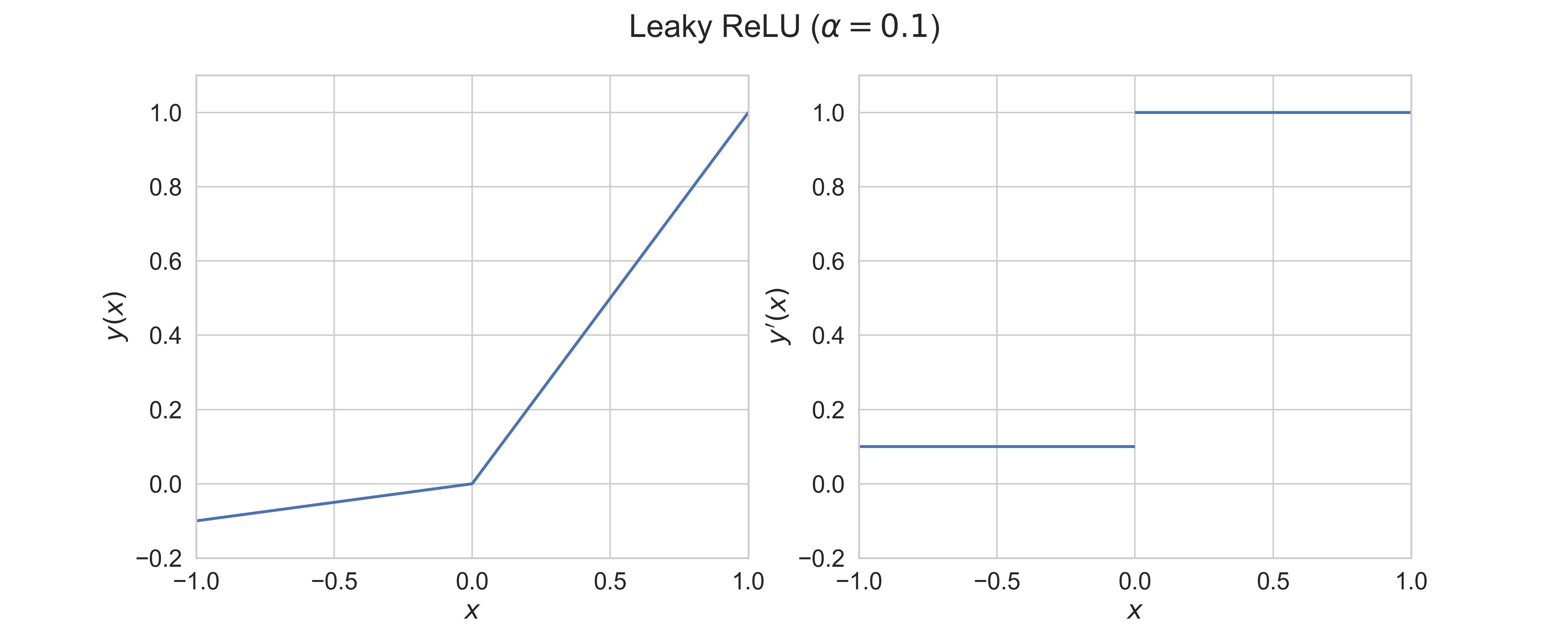}%
  \caption{Leaky Rectified Linear Unit (Leaky ReLU) activation function $y(x)$ and its derivative $y'(x)$.}%
  \label{f:leaky_relu_act_fn}%
\end{figure}

\begin{figure}[!htb]%
  \centering
  \includegraphics[width=1.0\linewidth]{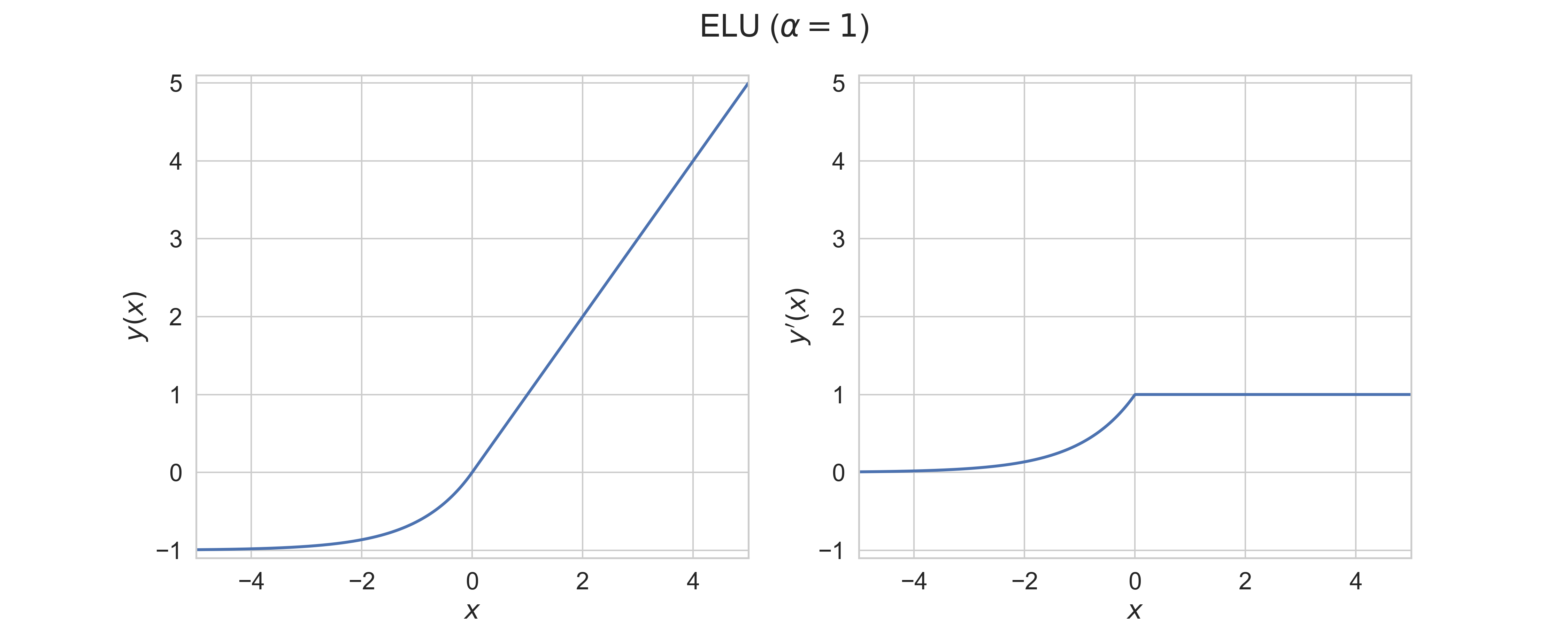}%
  \caption{Exponential Linear Unit (ELU) activation function $y(x)$ and its derivative $y'(x)$.}%
  \label{f:elu_act_fn}%
\end{figure}

\begin{figure}[!htb]%
  \centering
  \includegraphics[width=1.0\linewidth]{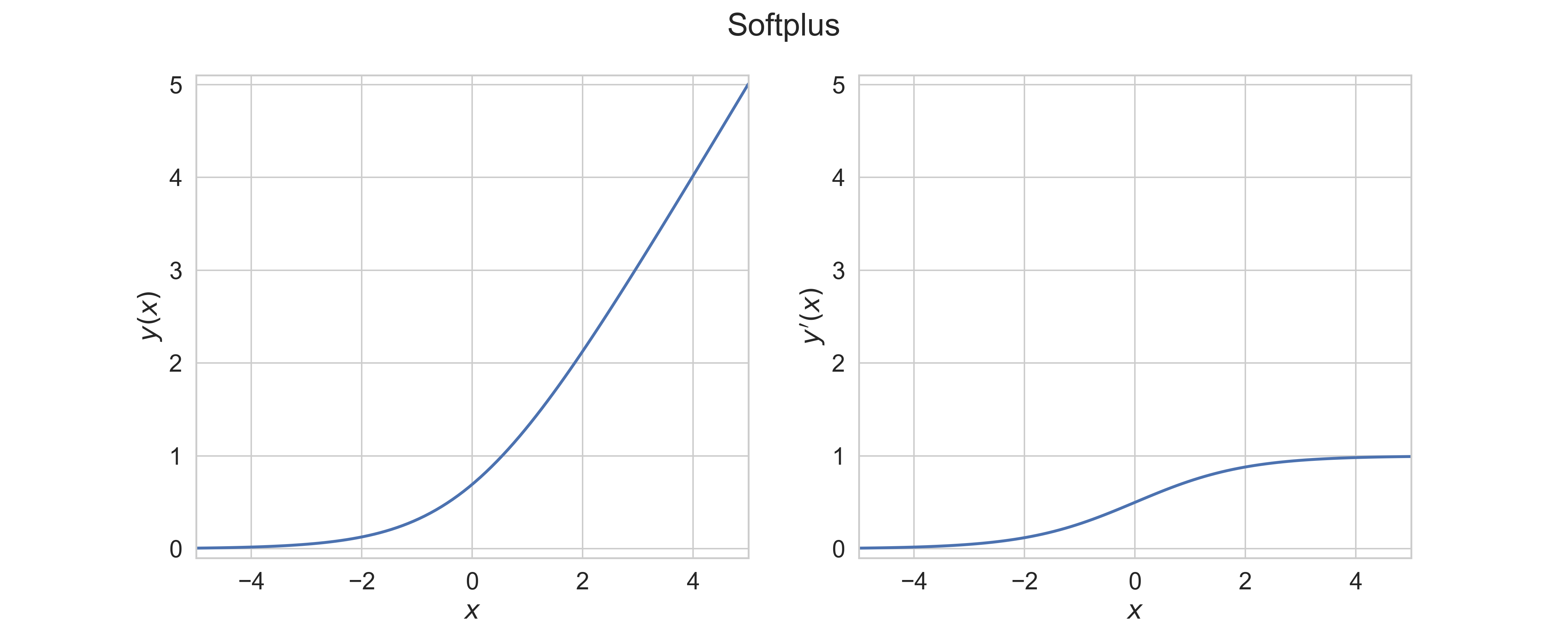}%
  \caption{Softplus activation function $y(x)$ and its derivative $y'(x)$.}%
  \label{f:softplus_act_fn}%
\end{figure}

The activation function $h(\cdot)$ plays an important role in neural networks since it gives nonlinear properties while propagating the inputs and also affects how the the gradients are back-propagated during training.
Example of the most common activation functions are: Linear (Figure \ref{f:linear_act_fn}), Sigmoid (Figure \ref{f:sigmoid_act_fn}), Hyperbolic Tangent (Tanh, Figure \ref{f:tanh_act_fn}), Rectified Linear Unit (ReLU, Figure \ref{f:relu_act_fn}), Leaky Rectified Linear Unit (Leaky ReLU, Figure \ref{f:leaky_relu_act_fn}), Exponential Linear Unit (Figure \ref{f:elu_act_fn}), and Softplus (Figure \ref{f:softplus_act_fn}).
Table \ref{tab:activation} details the equations and their derivatives and Table \ref{tab:activation_specs} details advantages and disadvantages for the select of activation functions commonly used in neural networks \cite{karpathy2016cs231n}.

\begin{table}
\centering
\caption{Select activation function equations and their derivatives.}
\begin{tabular}{@{}lll@{}}
\toprule
Name & Equation & Derivative \\
\midrule
Linear (Fig. \ref{f:linear_act_fn}) & $h(x) = x$ & $h'(x) = 1$ \\\\
Sigmoid (Fig. \ref{f:sigmoid_act_fn}) & $h(x) = \dfrac{1}{1+e^{-x}}$ & $h'(x) = h(x)(1 - h(x))$ \\\\
Tanh (Fig. \ref{f:tanh_act_fn}) & $h(x) = \dfrac{2}{1+e^{-2x}}-1$ & $h'(x) = 1 - h(x)^2$ \\\\
ReLU (Fig. \ref{f:relu_act_fn}) & 
$h(x) = \begin{cases}
    0, & \text{for $x<0$}.\\
    x, & \text{for $x\geq0$}.
  \end{cases}$ & 
$h'(x) = \begin{cases}
    0, & \text{for $x<0$}.\\
    1, & \text{for $x\geq0$}.
  \end{cases}$ \\\\
Leaky ReLU (Fig. \ref{f:leaky_relu_act_fn}) & 
$h(x) = \begin{cases}
    \alpha x, & \text{for $x<0$}.\\
    x, & \text{for $x\geq0$}.
  \end{cases}$ & 
$h'(x) = \begin{cases}
    \alpha, & \text{for $x<0$}.\\
    1, & \text{for $x\geq0$}.
  \end{cases}$\\\\
ELU (Fig. \ref{f:elu_act_fn}) & 
$ h(x) = \begin{cases}
    \alpha (e^x-1), & \text{for $x<0$}.\\
    x, & \text{for $x\geq0$}.
  \end{cases}$ & 
$h'(x) = \begin{cases}
    \alpha e^x, & \text{for $x<0$}.\\
    1, & \text{for $x\geq0$}.
  \end{cases}$\\\\
Softplus (Fig. \ref{f:softplus_act_fn}) &
$h(x) = \ln(1+e^x)$ & 
$h'(x) = \dfrac{1}{1+e^{-x}}$\\\\
\bottomrule
\end{tabular}
\label{tab:activation}
\end{table}

\begin{table}
  \centering
  \caption{Select activation function advantages and disadvantages.}
  \begin{tabular}{lll}
    \toprule
    Name & Advantages & Disadvantages \\
    \midrule
      Linear & Wide range of outputs, useful & Does not add nonlinearities to inputs, \\
      & when scaling is not desired. & only a linear transformation. \\
      & & Constant derivative, so gradient is not\\
      & & dependent on input when back-propagated. \\
    \\[.5\normalbaselineskip]
      Sigmoid & Bounded output [0,1]. & Saturates and kills gradients due to \\
      & Smooth derivative. & flat derivative towards the tail. \\
      & & Outputs not zero centered, changing \\
      & & dynamics of gradient descent. \\
    \\[.5\normalbaselineskip]
      Tanh & Bounded output [-1,1]. & Saturates and kills gradients due to \\
      & Smooth derivative. & flat derivative towards the tail. \\
    \\[.5\normalbaselineskip]
      ReLU & Simple implementation and  & Large gradients can shift weights \\
      & inexpensive operation. & in a way that disables neuron \\
      & Shown to accelerate learning & units permanently (never activate),\\
      & on vision-based tasks \cite{Krizhevsky2012}. & the ``dead neuron'' problem. \\
    \\[.5\normalbaselineskip]
      Leaky ReLU & Attempts to correct the ReLU & Large gradients can shift weights \\
      & ``dead neuron'' problem. & in a way that disables neuron \\
      & Simple implementation and & units permanently (never activate).\\
      & inexpensive operation. &  \\
    \\[.5\normalbaselineskip]
      ELU & Attempts to correct the ReLU & Output not bounded for $x > 0$. \\
      & ``dead neuron'' problem. &   \\
      & Can output negative values. &  \\
      & Strong alternative to ReLU. &  \\
    \\[.5\normalbaselineskip]
      Softplus & Similar to ReLU, but & Operation not as \\
      & with smooth gradient. & inexpensive as ReLU. \\
      & near zero. & \\
    \bottomrule
  \end{tabular}
  \label{tab:activation_specs}
\end{table}

\subsubsection{Learning in Discrete and Continuous Spaces}

Learning in neural networks occurs by systematically changing the network weights of each layer in order to drive the predicted output $\hat{y}_i$ closer to the target value or label $y_i$ for the same input sample. This is measured by a \emph{cost function}, generally represented by $\mathcal{L}(\cdot)$, which is different for discrete spaces (classification) in continuous (regression) cases.
Common cost function for classification are Cross-Entropy and Hinge loss while for regression we have Mean Squared Error (MSE), Mean Absolute Error (MAE), and Huber loss.
The Cross-Entropy (CE) loss, written as
\begin{equation*}
    \mathcal{L}_{CE}(y_i, \hat{y}_i) = - \frac{1}{N} \sum^N_i \sum^C_{j = 1} y_{ij} \log(\hat{y}_{ij}),
\end{equation*}
is used for classification with $C$ classes.
Binary Cross-Entropy (BCE) loss, also known as Log loss, is the special case of the CE loss when the number of classes $C$ is equal to $2$ (binary classification problem) and is written as
\begin{equation*}
    \mathcal{L}_{BCE}(y_i, \hat{y}_i) = - \frac{1}{N} \sum^N_i \sum^{C = 2}_{j = 1} y_{ij} \log(\hat{y}_{ij}) = -y_{i1} \log(\hat{y}_{i1}) - (1-y_{i1})\log(1-\hat{y}_{i1}).
\end{equation*}
Hinge loss, often used for maximum-margin classification in Support Vector Machines (SVM) \cite{barber2012bayesian}, is written as
\begin{equation*}
    \mathcal{L}_{Hinge}(y_i, \hat{y}_i) = \frac{1}{N} \sum^N_i \max (0,1-y_i \hat{y}_i).
\end{equation*}
For regression problems, Mean Squared Error (MSE) loss computes the average of the square of the errors between predicted $\hat{y}_i$ and true $y_i$ value, which results in an arithmetic mean-unbiased estimator:
\begin{equation*}
    \mathcal{L}_{MSE}(y_i, \hat{y}_i) = \frac{1}{N} \sum^N_i (y_i - \hat{y}_i)^2.
\end{equation*}
The Mean Absolute Error (MAE) loss, similar to MSE, computes the average of the absolute errors between predicted $\hat{y}_i$ and true $y_i$ value, which results in an median-unbiased estimator:
\begin{equation*}
    \mathcal{L}_{MAE}(y_i, \hat{y}_i) = \frac{1}{N} \sum^N_i |y_i - \hat{y}_i|.
\end{equation*}
Huber loss combines the nonlinearity of MSE and linearity of MAE in a single loss function, controlled by the hyperparameter $\delta$, less sensitive to high magnitude error that would normally be amplified by MSE loss.
The Huber loss is written as
\begin{equation*}
    \mathcal{L}_{Huber}(y, \hat{y}) = \begin{cases}
    \frac{1}{2} (y-\hat{y})^2, & \text{for $|y - \hat{y}| \leq \delta$}.\\
    \delta |y - \hat{y}| - \frac{1}{2}\delta^2, & \text{otherwise}.
  \end{cases}
\end{equation*}

Defined a cost function, the neural network weights are updated (also called ``trained'') using the \emph{backpropagation} algorithm.
In backpropagation each weight parameter is computed based on their proportional contribution to the cost for the given input $\vec{x}$ using the \emph{chain rule} for partial derivatives and gradient descent updates.
For example, in the case of the neural network presented in Figure \ref{f:nn_bishop} in a regression task, the cost or loss would be given by

\begin{equation*}
    \mathcal{L} = \sum_{j = 1}^{K} (a_j^{(L)} - y_j)^2,
\end{equation*}

where $L$ is the index of the hidden layer, $K$ the number of output neurons, and $a_j$ the activation output of the $j$ neuron in the layer. Explicitly, for the case we have three neurons and one bias in the hidden layer, $a_j$ is represented as

\begin{equation*}
    a_j^{(L)} = h( w_{10}^{(L)}a_1^{(L-1)} + w_{20}^{(L)}a_2^{(L-1)} + w_{30}^{(L)}a_3^{(L-1)} + b_j^{(L)} ),
\end{equation*}

where $h(\cdot)$ is a, ideally nonlinear, activation function.
Using the chain rule, the loss contribution of each previous activation output can be written as

\begin{equation*}
    \frac{\partial \mathcal{L}}{\partial a_k^{(L-1)}} = \sum_{j = 1}^{K}  \frac{\partial a_j^{(L)}}{\partial a_k^{(L-1)}} \frac{\partial \mathcal{L}}{\partial a_j^{(L)}}.
\end{equation*}

This process is repeated for every previous layer until we reach the input layer and compute $\frac{\partial \mathcal{L}}{\partial x_i}$ and for each weight component $\frac{\partial \mathcal{L}}{\partial w_{ij}}$ along the connection path. Each weight component at time $t$, denoted $w_{ij}^t$, is then updated with gradient descent as

\begin{equation*}
    w_{ij}^{t+1} = w_{ij}^t - \alpha \frac{\partial \mathcal{L}}{\partial w_{ij}^t},
\end{equation*}

where $\alpha$ is a step size hyperparemeter.

\section{Sequential Decision Making Problems}

\begin{figure}[!htb]%
  \centering
  \includegraphics[width=.5\linewidth]{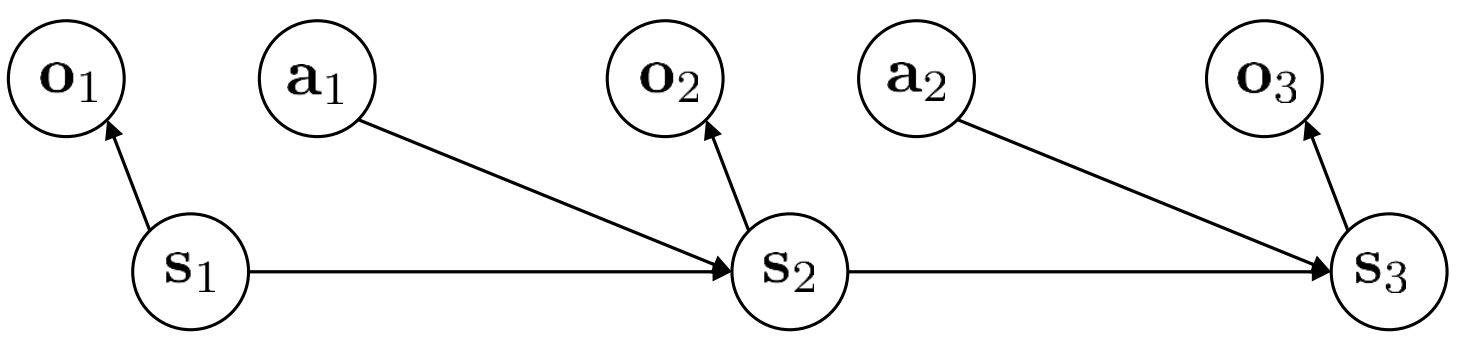}%
  \caption{Partially-Observable Markov Decision Process diagram. Adapted from \cite{DeepRLBerkeley2018}.}%
  \label{f:pomdp}%
\end{figure}

Mathematically and similarly to any sequential decision making process, in its most general form, the RL problem is formalized as a \emph{Partially Observable Markov Decision Process} (POMDP) \cite{Bertsekas2000}. A POMDP $\mathcal{M} = \{ \mathcal{S},\mathcal{O},\mathcal{E},\mathcal{A},\mathcal{T}, r \}$ is characterize by its state-space $\mathcal{S}$ (where a vector of states $\vec{s} \in \mathcal{S}$), an observation-space $\mathcal{O}$ (where a vector of observations $\vec{o} \in \mathcal{O}$), an emission probability $\mathcal{E}$ that controls the probability of observing $\vec{o}$ conditioned to the underlying states $\vec{s}$, an action-space $\mathcal{A}$ (where a vector of actions $\vec{a} \in \mathcal{A}$), a transition operator $\mathcal{T}$ (which defines the probability distribution $p(\vec{s}_{t+1}|\vec{s}_{t})$), and the reward function $r: \mathcal{S} \times \mathcal{A} \rightarrow \mathbb{R}$ (or $r(\vec{s},\vec{a}))$, as showed in Figure \ref{f:pomdp} of the previous chapter. In \emph{Fully Observable Markov Decision Process}, or simply a \emph{Markov Decision Process} (MDP), it is assumed full knowledge of the underlying states, removing the dependency to the observations and simplifying the process to $\mathcal{M} = \{ \mathcal{S},\mathcal{A},\mathcal{T}, r \}$.

\section{Cloning Behaviors with Imitation Learning} \label{sec:sl}

As discussed in Section \ref{sssec:lfd}, \emph{Imitation Learning} provides a directed path to learn demonstrated behaviors by utilizing examples of a secondary policy performing the task, quickly converging to more stable behaviors. This secondary policy can be another agent trained with reinforcement learning, a human, a classical controllers, or any other entity that is able to attempt to solve the desired task.

\emph{Behavior Cloning} (BC) is the subset of Imitation Learning where the demonstration data is directly used to clone the demonstrated behavior by training a model to generate actions as similar as possible to the demonstrated data, given similar inputs.
In BC, a policy $\pi$ is trained in order to generalize over a subset $\mathcal{D}$ of states and actions visited during a task demonstration over $T$ time steps:
\begin{align*} \label{eq:demonstration_dataset}
\mathcal{D} = \left\{\vec{a}_0, \vec{s}_0, \vec{a}_1, \vec{s}_1, ... , \vec{a}_T, \vec{s}_T\right\}.
\end{align*}

This demonstration can be performed by a human supervisor, optimal controller, or virtually any other pre-trained policy.
In the case of human demonstrations, the human is implicitly trying to maximize what may be represented as an internal reward function for a given task (Equation \ref{eq:imitation_rew_bc}), where $\pi^*(\vec{a}^*_t | \vec{s}_t)$ represents the optimal policy that is not necessarily known, in which the optimal action $\vec{a}^*$ is taken at state $\vec{s}$ for every time step $t$.

\begin{equation}\label{eq:imitation_rew_bc}
\max_{\vec{a}_0,...,\vec{a}_T} \sum_{t=0}^{T} r_t(\vec{s}_t, \vec{a}_t) = \sum_{t=0}^{T} \log p (\pi^*(\vec{a}_t^* | \vec{s}_t))
\end{equation}

Defining the policy of the supervisor as $\pi_{sup}$ and its estimate as $\hat{\pi}_{sup}$, behavior cloning can be achieved through standard supervised learning, where the parameters $\theta$ of a policy $\pi_\theta$ are trained in order to minimize a loss function, such as mean squared error, as shown in Equation \ref{eq:sup_mse_bc}.
Behavior cloning can be seen as supervised learning applied to sequential decision making problems.

\begin{equation}\label{eq:sup_mse_bc}
\hat{\pi}_{sup} = \argmin_{\pi_\theta} \sum_{t=0}^{T} ||\pi_\theta(\vec{s}_t) - \vec{a}_t ||^2
\end{equation}



\chapter{FROM SHALLOW TO DEEP REINFORCEMENT LEARNING} \label{ch:rl}

This chapter introduces the notation and defines the Reinforcement Learning (RL) problem, including its mathematical formulation as a Markov Decision Process, types of Reinforcement Learning algorithms, and the transition from ``Shallow" to ``Deep" Reinforcement Learning.
Reinforcement Learning is very closely related to the theory of classical optimal control, dynamic programming, stochastic programming, and optimization \cite{Powell2012}. However, while optimal control assumes perfect knowledge of the system's model, RL operates based on performance metrics returned as consequence of interactions with an unknown environment \cite{Kober2015}.

\section{Partially and Fully Observable Markov Decision Processes}

Mathematically and similarly to the supervised learnig problem presented on Chapter \ref{ch:bc}, in its most general form, the RL problem is formalized as a \emph{Partially Observable Markov Decision Process} (POMDP) \cite{Bertsekas2000}. A POMDP $\mathcal{M} = \{ \mathcal{S},\mathcal{O},\mathcal{E},\mathcal{A},\mathcal{T}, r \}$ is characterize by its state-space $\mathcal{S}$ (where a vector of states $\vec{s} \in \mathcal{S}$), an observation-space $\mathcal{O}$ (where a vector of observations $\vec{o} \in \mathcal{O}$), an emission probability $\mathcal{E}$ that controls the probability of observing $\vec{o}$ conditioned to the underlying states $\vec{s}$, an action-space $\mathcal{A}$ (where a vector of actions $\vec{a} \in \mathcal{A}$), a transition operator $\mathcal{T}$ (which defines the probability distribution $p(\vec{s}_{t+1}|\vec{s}_{t})$), and the reward function $r: \mathcal{S} \times \mathcal{A} \rightarrow \mathbb{R}$ (or $r(\vec{s},\vec{a}))$, as showed in Figure \ref{f:pomdp} of Chapter \ref{ch:bc}. In \emph{Fully Observable Markov Decision Process}, or simply a \emph{Markov Decision Process} (MDP), it is assumed full knowledge of the underlying states, removing the dependency to the observations and simplifying the process to $\mathcal{M} = \{ \mathcal{S},\mathcal{A},\mathcal{T}, r \}$.

\section{The Reinforcement Learning Problem}
\subsection{Problem Definition} \label{ss:pg}

\begin{figure}[htbp]
    \centering
    \includegraphics[width=2.5in]{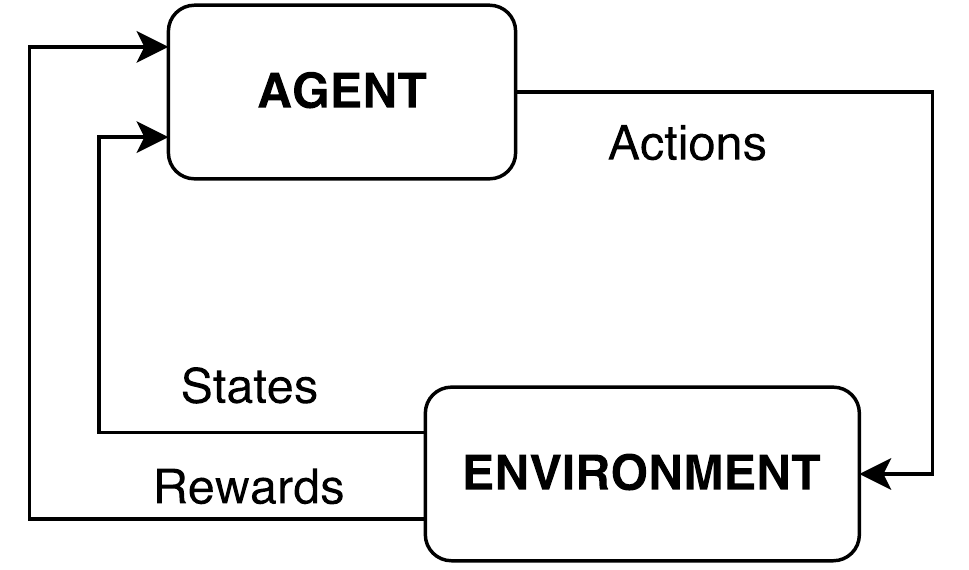}
    \caption{Reinforcement learning problem modeled as agent and environment interactions.}
    \label{fig:rl_classic}
\end{figure}

In \emph{Reinforcement Learning} (RL), it is desired to train an agent to learn the parameters $\theta$ of a \emph{policy} (or \emph{controller}) $\pi_{\theta}$, in order to map the partially-observable environment's \emph{observation} vector $\vec{o}$ (or \emph{state} vector $\vec{s}$ in fully-observable environments) to agent \emph{actions} $\vec{a}$. The performance of the agent is measured by a scalar \emph{reward signal} $r$ returned by the environment (external rewards, $r_e$) and/or returned by the agent itself (intrinsic rewards, $r_i$). At each time step $t$ the reward signal can be computed as the sum of all the extrinsic and intrinsic rewards received $r_t = r_{e_t} + r_{i_t}$. Figure \ref{fig:rl_classic} illustrates this description as a diagram.

This interaction between agent and environment occurs during an \emph{episode}, which is defined as the time interval between the initial time step $t_0$ and the maximum number of time steps allowed $T$ or until a previously established metric is achieved. 
At each time step $t$ of a finite time horizon $T$, the \emph{policy} (or controller) $\pi$, parametrized by $\theta_t$, maps the current environment's \emph{states} $\vec{s}_t$ to \emph{actions} $\vec{a}_t$: $\pi_{\theta}(\vec{a}_t|\vec{s}_t)$. This action affects the current environment's states $\vec{s}_t$ which evolves to $\vec{s}_{t+1}$ based on the environment's transition distribution (dynamics) $p(\vec{s}_{t+1}|\vec{s}_t,\vec{a}_t)$.
During an episode, the sequence of states observed and actions taken over a number of time steps $T$ can be represented by a \emph{trajectory} $\tau = \{ \vec{s}_0, \vec{a}_0, \vec{s}_1, \vec{a}_1, \ldots , \vec{s}_T, \vec{a}_T \}$.
The total reward per episode $R$ is defined as the sum of the rewards received for each time step, as shown in Equation \ref{eq:total_rew}.

\begin{equation}\label{eq:total_rew}
R = \sum_{t=0}^{T} r_t = \sum_{t=0}^{T} \left( r_{e_t} + r_{i_t} \right).
\end{equation}

Similarly, the expected total reward per episode received by a policy $\pi_\theta (\vec{a}_t | \vec{s}_t)$ can be defined by Equation \ref{eq:expected_rew}.

\begin{equation}\label{eq:expected_rew}
R_{\pi_\theta} = \sum_{t=0}^{T} \E_{\vec{a}_t \sim \pi_\theta} [r_t(\vec{s}_t, \vec{a}_t)].
\end{equation}

To simplify further derivation of the algorithm it is assumed a fully-observable environment, where the observation-space $\mathcal{O} = \mathcal{S}$ (state-space) and, consequently, vector of observations $\vec{o} = \vec{s}$ (states).

At each time step $t$ of a finite time horizon $T$, the \emph{policy} (or controller) $\pi$, parametrized by $\theta_t$, maps the current environment's \emph{states} $\vec{s}_t$ to \emph{actions} $\vec{a}_t$: $\pi_{\theta}(\vec{a}_t|\vec{s}_t)$. This action affects the current environment's states $\vec{s}_t$ which evolves to $\vec{s}_{t+1}$ based on the environment's transition distribution (dynamics) $p(\vec{s}_{t+1}|\vec{s}_t,\vec{a}_t)$. The environment also returns a scalar \emph{reward function} $r$ that evaluates the action taken $\vec{a}_t$ at the state $\vec{s}_t$: $r(\vec{s}_t,\vec{a}_t)$.

The probability of experiencing a given trajectory $\tau$ (sequence of state $\vec{s}$ and action $\vec{a}$ pairs) in a Markov Decision process can be written as:
\begin{align}
\pi_{\theta}(\tau) &= p_{\theta} (\vec{s}_0, \vec{a}_0, \vec{s}_1, \vec{a}_1, \ldots , \vec{s}_T, \vec{a}_T) \\
&= p(\vec{s}_1) \prod_{t=1}^T \pi_{\theta}(\vec{a}_t|\vec{s}_t) p(\vec{s}_{t+1} | \vec{s}_t, \vec{a}_t).
\end{align}

The goal in RL is to find the parameters $\theta^*$ that will maximize the objective $J(\theta)$

\begin{align}
J(\theta) &= \E_{\tau \sim p_{\theta} (\tau)} \left[\sum_{t=1}^{T} r(\vec{s}_t, \vec{a}_t)\right]\\
&= \E_{\tau \sim \pi_{\theta} (\tau)} [r(\tau)] = \int(\pi_{\theta} (\tau) r(\tau) d\tau),
\end{align}

which represents the expected total reward to be received by this policy $\pi_{\theta}(\tau)$:

\begin{align}
\theta^* &= \argmax_{\theta} J(\theta) \\
&= \argmax_{\theta} \E_{\tau \sim p_{\theta} (\tau)} \left[\sum_{t=1}^{T} r(\vec{s}_t, \vec{a}_t)\right].
\end{align}

To maximize $J(\theta)$, it is possible to apply the gradient operator $\nabla_{\theta}$ and compute its gradient with respect to its parameters $\theta$:

\begin{equation} \label{eq:grad_j}
\nabla_{\theta} J(\theta) = \int(\nabla_{\theta} \pi_{\theta}(\tau) r(\tau) d\tau).
\end{equation}

Using the following identity

\begin{equation} 
\pi_{\theta}(\tau) \nabla_{\theta} \log \pi_{\theta}(\tau) = 
\pi_{\theta}(\tau) \frac{\nabla_{\theta} \pi_{\theta}(\tau)}{\pi_{\theta}(\tau)} =
\nabla_{\theta} \pi_{\theta}(\tau)
\end{equation}

on Equation \ref{eq:grad_j} it is possible to simplify it to

\begin{equation}
\nabla_{\theta} J(\theta) = \int(\pi_{\theta}(\tau) \nabla_{\theta} \log \pi_{\theta}(\tau) r(\tau) d\tau)
\end{equation}

\begin{equation} \label{eq:pg}
\nabla_{\theta} J(\theta) = \E_{\tau \sim \pi_{\theta} (\tau)} [\nabla_{\theta} \log \pi_{\theta}(\tau) r(\tau)],
\end{equation}

known as the \emph{policy gradient}, which, by substituting $\tau$, can also be written in terms of states and actions over time steps $t$ in a given time horizon $T$:

\begin{equation} 
\nabla_{\theta} J(\theta) = \E_{\tau \sim \pi_{\theta} (\tau)} \left[\sum_{t=1}^T \nabla_{\theta} \log \pi_{\theta}(\vec{a}_t | \vec{s}_t ) \sum_{t=1}^T r(\vec{a}_t, \vec{s}_t )\right].
\end{equation}

Since the structure of the objective $J(\theta)$ and its gradient $\nabla_{\theta} J(\theta)$ are unknown, in practice the only way to evaluate them and approximate the expectation term is by sampling and averaging over $N$ samples:

\begin{equation}
\nabla_{\theta} J(\theta) \approx \frac{1}{N} \sum_{i=0}^N \left[ \left( \sum_{t=0}^T \nabla_{\theta} \log \pi_{\theta}(\vec{a}_t^{(i)} | \vec{s}_t^{(i)}) \right) \left( \sum_{t=0}^T r(\vec{a}_t^{(i)}, \vec{s}_t^{(i)}) \right) \right].
\end{equation}

\begin{figure}[!htb]%
  \centering
  \includegraphics[width=.45\linewidth]{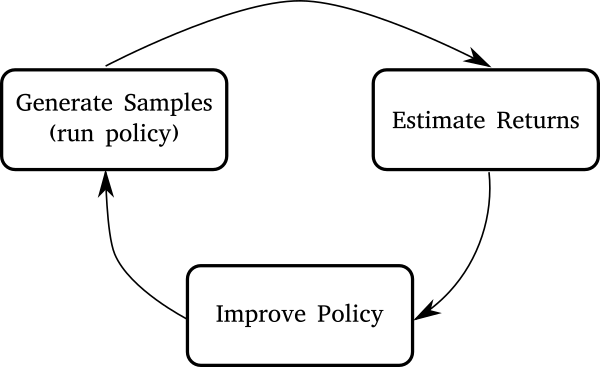}%
  \caption{Policy evaluation and optimization cycle.}%
  \label{fig:policy_imp}%
\end{figure}

The policy is improved by \emph{gradient ascent} according to a step size $\alpha$, as shown in Equation \ref{eq:grad_ascent} and Figure \ref{fig:policy_imp}, optimizing the agent's policy during the training time in what is called the \emph{vanilla policy gradient} (VPG) in RL.

\begin{equation} \label{eq:grad_ascent}
\theta \rightarrow \theta + \alpha \nabla_{\theta} J(\theta)
\end{equation}

\subsection{Types of Reinforcement Learning Algorithms}

As shown in the previous section, policy gradient methods in RL directly differentiate the objective function, which is to maximize the expected sum of rewards, and improve the policy by following gradient ascent.
Other types of RL algorithms directly estimate a \emph{value function} for a policy $\pi$ of a particular state $V^{\pi}(s)$, representing the expected sum of rewards to be received if starting at that state $s$, written as
\begin{equation*}
    V^{\pi}(s) = \E_{\pi} [R_t|s_t=s],
\end{equation*}
or the value function of a particular state-action pair $Q^{\pi}(s,a)$, representing the expected sum of rewards to be received if starting at that state $s$ and taking the action $a$ \cite{Sutton2018}, written as
\begin{equation*}
    Q^{\pi}(s,a) = \E_{\pi} [R_t|s_t=s, a_t=a].
\end{equation*}
One can also define an \emph{advantage function} $A(s,a)$ as
\begin{equation*}
    A^{\pi}(s,a) = Q^{\pi}(s,a) - V^{\pi}(s)
\end{equation*}
to quantify the advantage of taking a particular action $a$ at the state $s$ when compared to the average value of that state \cite{baird1993advantage,sutton2000policy}.
Bellman \cite{bellman1952theory,bellman1957markovian} showed that in Markovian Decision Processes there is a relationship between the value of a state and the value of it successor states \cite{Sutton2018}.
After averaging and weighting all probabilities, Bellman showed that ``the value of the start state must equal the (discounted) value of the expexted next state, plus the reward expected along the way''\cite{Sutton2018}:
\begin{equation*}
    V^{\pi}(s_t) = \E_{\pi} [r_t + \gamma V^{\pi}(s_{t+1})].
\end{equation*}
We can remove the dependency to the policy $\pi$ and show that, under the optimal policy, the value of the state $V^*(s_t)$ is equal the expected discount sum of rewards for the best action when the agent is in that state, know as the \emph{Bellman Optimality Equation} \cite{Sutton2018}:
\begin{align*}
    V^*(s_t) &= \max_a Q^{\pi^*} (s_t,a_t)\\
    &= \max_a \E [r_t + \gamma V^*(s_{t+1})].
\end{align*}
Similarly, the Bellman Optimality Equation can be written for $Q^*(s_t,a_t)$ as
\begin{align*}
    Q^*(s_t,a_t) = \E [r_t + \gamma \max_{a'} Q^*(s_{t+1},a')],
\end{align*}
where $a'$ is the action to be taken at state $s_{t+1}$.

Based on the Bellman equations, other RL algorithms can be derived. A classical example is \emph{Q-learning}, by \citeauthor{Watkins1989} \cite{Watkins1989}.
In Q-learning, by interacting with the environment, the agent continuously estimates the state-action value $Q(s_t,a_t)$ as
\begin{equation*}
    Q(s_t,a_t) \leftarrow Q(s_t,a_t) + \alpha[r_t + \gamma \max_{a'} Q(s_{t+1},a_t) - Q(s_t,a_t)],
\end{equation*}
where $\alpha$ is a learning rate hyperparameter.

Another type of RL algorithms called \emph{actor-critic} methods combine both policy gradient and value function concepts.
In Section \ref{ss:pg} it was shown that, in policy gradient methods, the gradient of the objective can be written as
\begin{equation} 
\nabla_{\theta} J(\theta) = \E_{\tau \sim \pi_{\theta} (\tau)} \left[ \sum_{t=1}^T \nabla_{\theta} \log \pi_{\theta}(\vec{a}_t | \vec{s}_t ) \sum_{t=1}^T r(\vec{a}_t, \vec{s}_t ) \right].
\end{equation}
Note that the term $\sum_{t=1}^T r(\vec{a}_t, \vec{s}_t$) is unknown until the end of the episode or a given number of time steps $T$, however it can be estimated by the value function $\hat{Q}^{\pi}_t$ as
\begin{equation} 
\nabla_{\theta} J(\theta) = \E_{\tau \sim \pi_{\theta} (\tau)} \left[ \sum_{t=1}^T \nabla_{\theta} \log \pi_{\theta}(\vec{a}_t | \vec{s}_t ) \hat{Q}^{\pi}_t \right]
\end{equation}
and work as a \emph{critic} in order to evaluate the actions taken by policy $\pi$, the \emph{actor}.
Hence, actor-critic algorithms incorporate value learning by updating the estimates of its value function and using it to compute the gradient of the policy.

Policy gradient, value learning, and actor-critic are also classified as \emph{model-free} algorithms because they assume the dynamic model of the environment is unknown to the agent. Other algorithms, called \emph{model-based}, leverage interaction data between agent and environment to learn the dynamics of the environment (or are given the model), which could be use for planning \cite{nagabandi2018neural}, perform policy updates using simulated data \cite{ha2018world}, and others. 

\section{From Shallow to Deep Representations}

As a side note, in Deep RL where deep neural networks are used as learning representation for the policies, a typical policy is represented by a neural network with two fully-connected layers with 64 neurons each (plus a bias term). If the dimension of the input is $N$ and the dimension of the output is $M$, this leads to $(N+1)*64 + (64+1)*64 + (64+1)*M$ parameters. For example, a policy for a continuous dynamical system with twelve inertial states and three control outputs would be represented by 5187 parameters.

\begin{figure}[htb]
    \centering
	\includegraphics[width=4in]{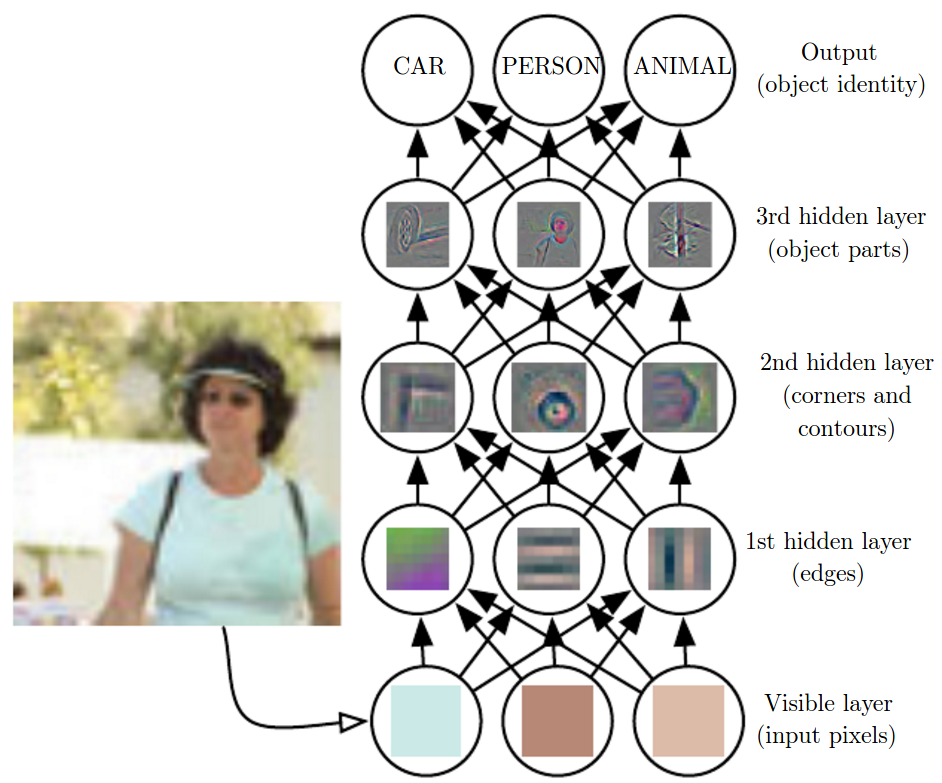}
	\caption{Visualization of each layer of a deep neural network use to classify different images. Reprinted from \cite{Zeiler2014}.}
	\label{fig:deep_net_image}
\end{figure}

The two key ideas that allow reinforcement learning to achieve the desired goal are: sampling through interactions to compactly represent the dynamics of an unknown stochastic environment, and the use of function approximations to represent policies \cite{Szepesvari2010} that guides the algorithm during the action selection. Recently, due to advances in computational power, deep neural networks, or Deep Learning, is being used for sensory processing and function approximation.
\citeauthor{Goodfellow2016} \cite{Goodfellow2016} define deep learning as the idea of training computers to gather knowledge by experience and to learn complex concepts by building them out of simpler ones. Feeding raw pixels of an image to a deep neural network is possible to see how it combines simple concepts of edges to create concepts of corners and contours, and combine the concepts of corners and contours to represent the concept of an object, as seen in Figure \ref{fig:deep_net_image}.
In RL, Deep Learning is used to represent polices and value functions (or the dynamics model, in model-based approaches) --- so-called Deep Reinforcement Learning (Deep RL) \cite{lillicrap2015continuous, Mnih2013}.
The higher complexity and plasticity of deep neural networks allows better quality on feature and data representations \cite{Goodfellow2016} and is directly related to the performance of reinforcement learning algorithms \cite{Mnih2013}.

\subsection{Inductive Bias}

A core concept of learning is to be able to generalize from past experiences (for example, a training dataset) to address unseen situations \cite{mitchell1980need}.
Inductive bias refers to ``any basis for choosing one generalization over another, other than strict consistency with the observed training instances." \cite{mitchell1980need}.

This research work assumes that the human brain and its underlying decision process is nonlinear and stochastic. Nonlinear due to the neural pathway arrangement and connections \cite{Grill-Spector2004}, and stochastic in a sense that different action outputs can be observed from the same observation inputs depending on the current human emotional states \cite{de2006frames}.
To address this nonlinearity and stochasticity, this research adopts deep neural networks with Gaussian outputs to represent the learning policy.
This representation also allows processing the high-dimensional continuous observation and action-space encountered in real-world robotic applications.
By using neural networks it is also assumed that the policy to be learned, and the function that approximates it, has a smooth differentiable gradient which can be learned through back-propagation.

\section{Expanding Q-learning to Continuous Space with Deep Reinforcement Learning}\label{ddpg_algo}

Tabular Q-learning has been a popular algorithm in reinforcement learning due to its simplicity and convergence guarantees \cite{Watkins1989}. According to \citeauthor{lillicrap2015continuous} \cite{lillicrap2015continuous}, it is not possible to directly apply Q-learning to continuous action spaces stating that the optimization is too slow to be practical with large and unconstrained function approximators (required for continuous spaces). The traditional tabular Q-learning is also unfeasible due to the ``curse of dimensionality'', when the discretized action space grows exponentially with the number of degrees-of-freedom of the problem \cite{Ng2003}. Due to these restrictions, \citeauthor{lillicrap2015continuous} \cite{lillicrap2015continuous} presented a novel model-free, off-policy actor-critic reinforcement learning algorithm using deep neural networks to learn policies in continuous action spaces based on \citeauthor{Silver2014} \cite{Silver2014} called the Discrete Policy Gradient (DPG) algorithm. It is model-free because it does not require model dynamics to learn the control policy. It is off-policy because it can follow a different control policy than the one that is being learned (can use different policies to explore the state space). Actor-critic because it has different structures to improve the policy (actor, through the gradient of the policy's performance with respect to its parameters) and evaluate it (critic, through function approximation to estimate expected future rewards).

According to \citeauthor{Silver2014} \cite{Silver2014}, the policy gradient is the gradient of the policy's performance, which on DPG is represented by the actor $\mu(s|\theta^\mu)$, where $\theta^\mu$ is the actor's network parameters. A critic $Q(s,a|\theta^Q)$ is learned using the traditional Q-learning approach \cite{Watkins1989}, where $\theta^Q$ is the critic's network parameters. The actor parameters are updated by applying the chain rule to the start distribution $J$ with respect to the actor parameters $\theta^\mu$ \cite{lillicrap2015continuous}, as showed in Eq. (\ref{eq:actor}).

\begin{equation}
\begin{aligned}\label{eq:actor}
\nabla_{\theta^\mu} J & \approx \E [\nabla_{\theta^\mu} Q(s,a|\theta^Q)|_{s=s_t,a=\mu(s_t|\theta^\mu)}] \\
& = \E [\nabla_{a} Q(s,a|\theta^Q)|_{s=s_t,a=\mu(s_t)} \nabla_{\theta^\mu} \mu(s|\theta^\mu)|_{s=s_t}]
\end{aligned}
\end{equation}

Prior to the work presented by Mnih et al. \cite{Mnih2013}, it was thought that using large nonlinear function approximators to learn critic functions was difficult and unstable \cite{lillicrap2015continuous}. However, using the techniques of replay buffer and target Q networks introduced by Deep Q-Networks (DQN) was possible to train actor-critic methods using neural function approximators. The resulting algorithm, Deep DPG (DDPG), is able to learn competitive policies using only low-dimension observations (e.g. coordinates, velocities, angles, or any other form of sensor reading). The main advantage of DDPG is being able to do off-policy exploration, one of the major challenges when learning on continuous spaces. The exploration policy was constructed by adding process noise $\mathcal{N}$ to the actor policy \cite{lillicrap2015continuous}:

\begin{equation}
\mu'(s_t) = \mu(s_t|\theta_t^\mu) + \mathcal{N}
\end{equation}

The Deep Deterministic Policy Gradient (DDPG) algorithm is outlined in Algorithm \ref{alg:ddpg} below and works as follow: two neural networks are initialized - one to work as the actor $\mu(s|\theta^\mu)$, and the second to work as the critic $Q(s,a|\theta^Q)$. Two copies of these networks are created (they will work as the target actor $\mu'(s|\theta^\mu)$ and the target critic $Q'(s,a|\theta^Q)$ networks). An empty replay buffer $\mathcal{R}$ is created to store what the agent is experiencing by interacting with the environment (states $s$, actions $a$, and rewards $r$). Each interaction between the agent and environment count as a time step (represented by $t$). A collection of a given number of time steps $T$ is called an episode. The DDPG algorithm loops for a desired number of episodes $M$. In the beginning of each episode, a random process $\mathcal{N}$ is initialized (for state exploration, explained below) and the agent is placed in a different initial state $s_1$ (from where it has a different initial observation). For each time step an action is sampled from the actor network and added with exploration noise (noisy actions allow the agent to explore the state space beyond its policy, which leads to higher future rewards). This action is executed in the environment, which returns the next state and the reward for the action. The most recent experience (reward, action applied, and the current and next state) is stored in the replay buffer. A batch of $N$ experiences is sampled from the replay buffer to train the learning algorithm. The training consists of the following steps: the critic $Q$ evaluates the sampled states and actions, adds the returned reward, and computes the target value $y$; the critic network computes a gradient step to update its parameters based on the target value; and the actor network $\mu$ is updated based on the critic's network gradient - the policy gradient. The parameters of the actor and critic networks ($\theta^\mu$ and $\theta^Q$, respectively) are copied to its respective targets ($\mu'(s|\theta^\mu)$ and $Q'(s,a|\theta^Q)$, respectively) based on the target update rate hyperparameter. Following this process, the actor network tends to increase the probability of suggesting actions that will be better evaluate by the critic, which will lead to higher rewards and higher controller performance.

\begin{algorithm}
\caption{Deep Deterministic Policy Gradient (DDPG) \cite{lillicrap2015continuous}}
\label{alg:ddpg}
\begin{algorithmic}[1]
\State {Randomly initialize critic $Q(s,a|\theta^Q)$ and actor $\mu(s|\theta^\mu)$ neural networks with weights $\theta^Q$ and $\theta^Q\mu$}
\State {Initialize target network $Q'$ and $\mu'$ with weights $\theta^{Q'} \leftarrow \theta^Q$, $\theta^{\mu'} \leftarrow \theta^\mu$}
\State {Initialize replay buffer $\mathcal{R}$}
\For {episode = 1, M}
	\State {Initialize a random process $\mathcal{N}$ for action exploration}
	\State {Receive initial observation state $s_1$}
	\For {t = 1, T}
		\State {Select action $a_t = \mu(s_t|\theta^u) + \mathcal{N}$ according to the current policy and exploration noise}
		\State {Execute action $a_t$ and observe reward $r_t$ and new state $s_{t=1}$}
		\State {Store transition $(s_t,a_t,r_t,s_{t+1})$ in R}
		\State {Sample a random minibatch of $N$ transitions $(s_t,a_t,r_t,s_{t+1})$ from R}
		\State {Set $y_i = r_i + \gamma Q'(s_{i+1}, \mu'(s_{i+1}|\theta^{\mu'})|\theta^{Q'})$}
		\State {Update critic by minimizing the loss, Eq. (\ref{eq:ddpg_loss})}
          \begin{equation} \label{eq:ddpg_loss}
          L = \frac{1}{N}\sum_i^N(y_i-Q(s_i,a_i|\theta^Q))^2
          \end{equation}
		\State {Update the actor policy using the sampled policy gradient, Eq. (\ref{eq:ddpg_grad})}
          \begin{equation} \label{eq:ddpg_grad}
          \nabla_{\theta^\mu}J \approx \frac{1}{N}\sum_i^N \nabla_a Q(s,a|\theta^Q)|_{s=s_i,a = \mu()s_i}    \nabla_{\theta^\mu} 		  \mu(s|\theta^\mu)|_{s_i}
          \end{equation}
		\State {Update target networks, Eq. (\ref{eq:ddpg_target})}
          \begin{equation} \label{eq:ddpg_target}
          \begin{split}
          \theta^{Q'} \leftarrow \tau\theta^Q + (1-\tau)\theta^{Q'} \\
          \theta^{\mu'} \leftarrow \tau\theta^\mu + (1-\tau)\theta^{\mu'}
          \end{split}
          \end{equation}
	\EndFor
\EndFor
\end{algorithmic}
\end{algorithm}


\chapter[CASE STUDY: DEEP REINFORCEMENT LEARNING ON INTELLIGENT MOTION VIDEO GUIDANCE FOR UNMANNED AIR SYSTEM GROUND TARGET TRACKING]{CASE STUDY: DEEP REINFORCEMENT LEARNING ON INTELLIGENT MOTION VIDEO GUIDANCE FOR UNMANNED AIR SYSTEM GROUND TARGET TRACKING\footnote{Adapted with permission from ''Deep Reinforcement Learning on Intelligent Motion Video Guidance for Unmanned Air System Ground Target Tracking”, by Vinicius G. Goecks and John Valasek, presented at the AIAA Scitech 2019 Forum \cite{goecks2019deep}, Copyright 2019 by the American Institute of Aeronautics and Astronautics.}}\label{ch:tt}

Tracking motion of ground targets based on aerial images can benefit commercial, civilian, and military applications. On small fixed-wing unmanned air systems that carry strapdown instead of gimbaled cameras, it is a challenging problem since the aircraft must maneuver to keep the ground targets in the image frame of the camera. Previous approaches for strapdown cameras achieved satisfactory tracking performance using standard reinforcement learning algorithms.  However, these algorithms assumed constant airspeed and constant altitude because the number of states and actions was restricted.  This paper presents an approach to solve the ground target tracking problem by proposing the Policy Gradient Deep Reinforcement Learning controller. The learning is based on the continuous full-state aircraft states and uses multiple states and actions.  Compared to previous approaches, the major advantage of this controller is the ability to handle the full-state ground target tracking case.  Policies are trained for three different target cases: static, constant linear motion, and random motion. Results presented in the paper on a simulated environment show that the trained Policy Gradient Deep Reinforcement Learning controller is able to consistently keep a randomly maneuvering target in the camera image frame. Learning algorithm sensitivity to hyperparameters selection is investigated in the paper, since this can drastically impact the tracking performance. 


\section{Problem Definition}

Following breakthroughs in artificial intelligence and cost reduction of unmanned air system (UAS) platforms, novel applications combining UAS and computer vision for tracking ground targets are on the rise.
Examples of civilian and military applications that directly benefit from autonomous UAS with ground target tracking capabilities are commercial package delivery and disaster relief response equipped with sense-and-avoid (SAA) technology \cite{ProjectWing2018, PrimeAir2018}, aerial filming and photography \cite{TheDroneCo2018}, and military intelligence, surveillance, and reconnaissance (ISR) programs \cite{DoD2007, ISR_Congress2005}.

Ground target tracking can be performed by using cameras fixed to the UAS body (camera frame fixed with respect to the aircraft frame) or gimbaled cameras (camera frame independent of the aircraft frame).
The problem is simplified by using gimbaled cameras, in which the image frame can be controlled independently of the aircraft trajectory. Unfortunately, small fixed-wing UAS have restricted payload capabilities, which restricts the use of onboard gimbaled cameras.
The immediate solution is to equip small fixed-wing UAS with cameras fixed to the fuselage. This requires aircraft maneuvers to keep the target of interest in the image frame.

Current ground target tracking approaches using fixed cameras \cite{Valasek2016a, Noren2018} are able to achieve satisfactory tracking performance using reinforcement learning algorithms to train a control policy on a simulated environment, which is later transferred to the physical hardware.
These approaches rely on training the control policy on simplified planar motion simulated environments, with constant altitude and airspeed, restricting the controller to a single-input to the aircraft due to limitations of the learning algorithm applied.

This research builds upon the previously presented work \cite{Valasek2016a,Dunn2012} by proposing the development of a learning controller that handles the full-state continuous target tracking case by incorporating concepts of deep reinforcement learning (Deep RL) and policy gradient (PG) algorithms. The main contributions of this research are i) development of a learning controller that handles the full-state continuous target tracking case by incorporating concepts of deep reinforcement learning (Deep RL) and policy gradient (PG) algorithms; ii) investigation of different learning controller designs to achieve better performance on the tracking problem; and iii) comparison of the tracking performance between previous work that relied on discretization of the state-space and current work the uses the full-state continuous state-space.  The PG Deep RL controller uses the 6-Degree-of-Freedom (DOF) linear aircraft simulation to extract target position on the image frame.  The aircraft states consist of linear velocities and position; roll, pitch, and yaw attitude angles and rates.  The controls consist of body-axis pitch, roll and yaw rates; throttle and nozzle position.

\section{Related Work}

Valasek et al. \cite{Valasek2016a} developed a machine learning algorithm approach for learning control policies based on target position in the image frame (pixel position) and current aircraft roll angle. They developed both Q-learning and Q-learning-with-eligibility-traces policies to solve the tracking problem. Dunn and Valasek \cite{Dunn2012} developed a similar approach to localize atmospheric thermal locations and then guide an UAS to soar from one to another. This research was inspired on how they framed the target tracking problem and it also builds upon their camera and target dynamics.

Noren et al. \cite{Noren2018} detailed the initial flight testing of a previously trained control policy \cite{Valasek2016a} for the autonomous tracking of ground targets. Their work showed that it is possible to train a controller using machine learning algorithms by using a simulated environment and later deploy the learned policy to hardware. Even limited by a planar motion simulation and single-output controller, they were able to achieve satisfactory tracking performance on the tracking problem when flying the algorithm trained on a simulated environment. The authors expected to follow the same development route presented by \cite{Noren2018} and deploy the presented Deep RL PG algorithm to hardware.

There were also alternative approaches besides machine learning algorithms to solve the target tracking problem with a fixed strap-down camera on a small UAS. Beard and Egbert \cite{Beard2007} derived a explicit roll-angle and altitude-above-ground-level (AGL) constraints for road tracking that that guarantees the target will remain in the camera frame. Beard and Saunders \cite{Beard2011} later extended their previous work proposing a non-linear guidance law using range-to-target and bearing-to-target information obtained from target motion in the image plane to plan trajectories for the aircraft. Both approaches are validated on simulation and flight testing.

Advances in modern Deep RL algorithms greatly contributed to the presented research. Schulman et al. presented a series of improvement to classic policy gradient algorithms in order to make them tractable and suitable to optimize policies represented by deep neural networks. Trust Region Policy Optimization (TRPO) \cite{Schulman2015} and Proximal Policy Optimization (PPO) \cite{Schulman2017} both rely on restricting the gradients during the policy update, either by a trust region (defined according to the Kullback-Leibler divergence between current and previous policy) or by a clipped objective (defined by the log probability ratio of current and previous policy). Both approaches are able to stabilize learning of policies represented by deep neural networks, although PPO leads to faster results in practice. Schulman et al. also reduced the variance of the policy gradient estimates by proposing a exponentially-weighted estimator, known as Generalized Advantage Estimation (GAE) \cite{schulman2015high}. These three insights were incorporated to the present research and greatly improved the learning performance.

The main weakness of the presented approach is it increase computational complexity. Even  if the control policy is not update in real-time during flight, the policy is still represented by deep neural networks, which require more computational when compared to classical controllers. The current approach also requires extensive validation and verification, since there is no formal proof of convergence of the learning algorithm. Additionally, hard constraints should be added to aircraft states and controls in order to guarantee operation within safety limits.

\section{Learning Tracking Policies with Reinforcement Learning} \label{sec:tracking_intro}

\begin{figure}[hbt!]%
    \centering
    \includegraphics[width=.75\linewidth]{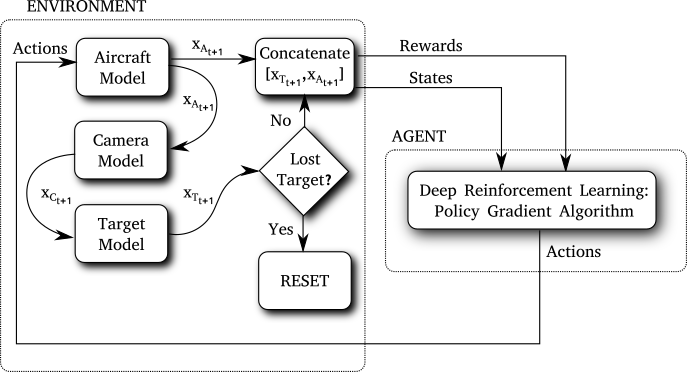}%
    \caption{Intelligent motion video guidance for unmanned air system ground target tracking modeled as a reinforcement learning problem. Reprinted from \cite{goecks2019deep}.}%
    \label{fig:main_diag}%
\end{figure}

The proposed PG Deep RL controller frames ground target tracking as a reinforcement learning problem. In Reinforcement Learning (RL), it is desired to train an agent to learn the parameters $\theta$ of a \emph{policy} (or \emph{controller}) $\pi_{\theta}$, in order to map the partially-observable environment's \emph{observation} vector $\vec{o}$ (or \emph{state} vector $\vec{s}$ in fully-observable environments) to agent \emph{actions} $\vec{a}$. The performance of the agent is measured by a scalar \emph{reward signal} $r$ returned by the environment. Figure \ref{fig:rl_classic} illustrates this description as a diagram and Figure \ref{fig:main_diag} shows how it can be applied to the target tracking case.


In the target tracking case proposed by this work, the environment comprises the aircraft, camera, and ground target model, explained in more details in the following paragraphs. Environment states are represented by the aircraft states (linear position and velocity, roll, pitch, yaw angles and rates) and target pixel position in the image plane --- see Equation \ref{eq:states}. The agent is represented by the PG Deep RL controller, explained in more details in Section \ref{ssec:agent_details}. The agent controls the aircraft through elevator, throttle, nozzle position, aileron, and rudder actions, $\delta_e$, $\delta_T$, $\delta_n$, $\delta_a$, $\delta_r$ respectively --- see Equation \ref{eq:actions}. Rudder control was removed to reduce complexity of the model for the PG Deep RL controller. The agent's actions are evaluated by the a scalar value, the reward signal. The reward signal is computed based on the normalized Euclidean distance of the target to the center of the image plane --- see Equation \ref{eq:rewards}.

\begin{equation} \label{eq:states}
\mathcal{S} = \{ X_T,Y_T,u,w,q,\theta,x,z,v,p,r,\phi,\psi,v \}
\end{equation}

\begin{equation} \label{eq:actions}
\mathcal{A} = \{ \delta_e, \delta_T, \delta_n, \delta_a, \delta_r \}
\end{equation}

\begin{equation} \label{eq:rewards}
r = \Bigg( 1 -\frac{\sqrt[]{X_T^2 + Y_T^2}} {\sqrt[]{{X_T^2}_{MAX} + {Y_T^2}_{MAX} }} \Bigg)^2
\end{equation}

The PG Deep controller optimizes the action selection by trial-and-error process on a simulated environment. The simulation is broke down in episodes, starting with the aircraft on a random position in the inertial space and ground target initially in the image frame. The episode resets if the target leaves the image frame or if aircraft performs any maneuver outside the safety limits, manually defined by the controller design.

\section{Policy Gradient Deep Reinforcement Learning Controller}\label{ssec:agent_details}

This section details how each part is modeled and how the tracking policies are learned.

\begin{figure}[!htb]%
  \centering
  \includegraphics[width=.5\linewidth]{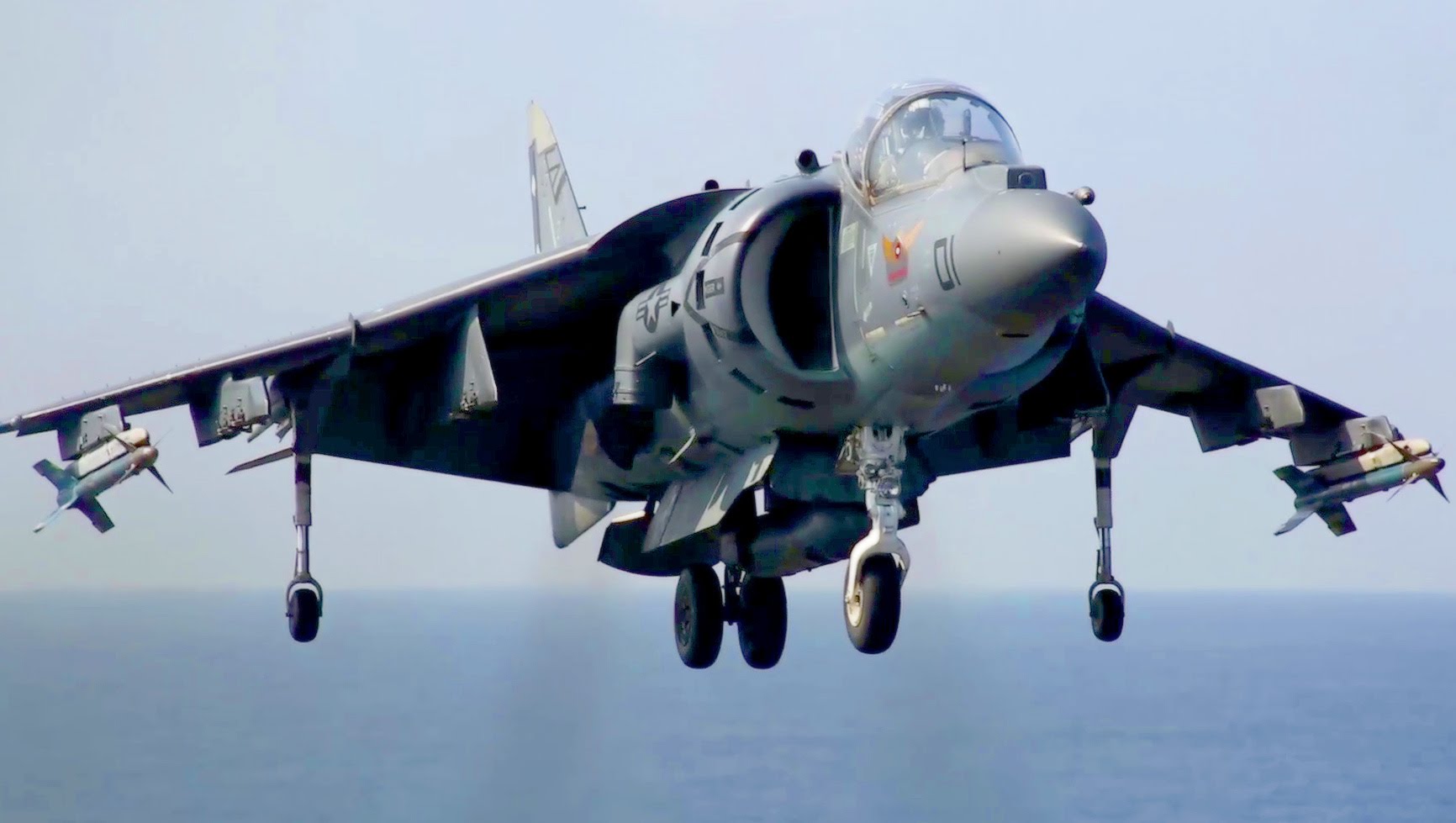}%
  \caption{Illustration of AV-8B Harrier, whose linear model is used to validate the proposed PG Deep RL controller. Reprinted from \cite{goecks2019deep}.}%
  \label{f:f18a}%
\end{figure}

One of the main contributions of this research is the development of a learning controller that handles the full-state continuous target tracking case, which includes the full-state continuous simulation of the aircraft, camera, and target. The aircraft is represented by a linear model of the AV-8B Harrier --- Figure \ref{f:f18a} illustrates the aircraft, Equations \ref{eq:long_model} and \ref{eq:latd_model} represent the longitudinal and lat/d model, respectively, and Table \ref{tab:trim_params} its trim parameters. Camera specifications are shown in Table \ref{tab:camera_specs}. The target is modeled as a point mass with planar motion on the XY plane. Policies are trained for three different target cases: static, constant linear motions, and random motion.

\begin{equation} \label{eq:long_model}
  \begin{bmatrix}
    \dot{u} \\
    \dot{w} \\
    \dot{q} \\
    \dot{\theta} \\
    \dot{x} \\
    \dot{z}
  \end{bmatrix}
=
[A_{LON}]
  \begin{bmatrix}
    u \\
    w \\
    q \\
    \theta \\
    x \\
    z
  \end{bmatrix}
+
[B_{LON}]
  \begin{bmatrix}
    \delta_e \\
    \delta_T \\
    \delta_n
  \end{bmatrix},
\end{equation}

where

\begin{equation*} 
[A_{LON}] =
  \begin{bmatrix}
 	-0.0693 & -0.0006 & -0.132 & -32.171 & 0 & 0 \\
    0.0199 & 0.0005 & 10.1644 & -0.3936 & 0 & 0 \\
    0 & 0 & -0.0409 & 0.0373 & 0 & 0 \\
    0 & 0 & 1 & 0 & 0 & 0 \\
    1 & 0 & 0 & 0 & 0 & 0 \\
    0 & 1 & 0 & 0 & 0 & 0 
  \end{bmatrix}
\end{equation*}
and
\begin{equation*} 
[B_{LON}] =
\begin{bmatrix}
    3.816 & 0.0237 & -32.667 \\
    4.255 & -1.008 & -0.724 \\
    -4.648 & -0.0094 & -0.0679 \\
    0 & 0 & 0 \\
    0 & 0 & 0 \\
    0 & 0 & 0
  \end{bmatrix}.
\end{equation*}

\begin{equation} \label{eq:latd_model}
  \begin{bmatrix}
    \dot{v} \\
    \dot{p} \\
    \dot{r} \\
    \dot{\phi} \\
    \dot{\psi} \\
    \dot{y}
  \end{bmatrix}
=
  [A_{LATD}]
  \begin{bmatrix}
    v \\
    p \\
    r \\
    \phi \\
    \psi \\
    y
  \end{bmatrix}
+
  [B_{LATD}]
  \begin{bmatrix}
    \delta_a \\
    \delta_r
  \end{bmatrix}
\end{equation}

where

\begin{equation*}
[A_{LATD}] =
  \begin{bmatrix}
    0 & 0.131 & -10.252 & 32.171 & 0 \\
    0 & -0.105 & 0.0264 & 0 & 0 \\
    0 & -0.0027 & -0.0489 & 0 & 0 \\
    0 & 1 & 0.0122 & 0 & 0 \\
    0 & 0 & 1 & 0 & 0 \\
    1 & 0 & 0 & 0 & 0
  \end{bmatrix}
\end{equation*}
and
\begin{equation*}
[B_{LATD}] =
  \begin{bmatrix}
    -1.0263 & 4.972 \\
    13.735 & 0.454 \\
    0.959 & -1.347 \\
    0 & 0 \\
    0 & 0 \\
    0 & 0
  \end{bmatrix}.
\end{equation*}

\begin{table}[h]
\centering
\caption{Aircraft trim parameters. Reprinted from \cite{goecks2019deep}.}
\label{tab:trim_params}
\begin{tabular}{cc}
\hline
\textbf{Trim Parameter} & \textbf{Value} \\ \hline
$M_1$ & 0.009 \\ \hline 
$U_1$ (ft/s) & 10 \\ \hline 
$H_1$ (ft) & 100 \\ \hline 
$\delta_e$ (deg) & 1.82 \\ \hline 
$\delta_n$ (deg) & 88 \\ \hline 
$\delta_T$ (deg) & 89.5 \\ \hline 
\end{tabular}
\end{table}

\begin{table}[h]
\centering
\caption{Modeled camera specifications. Reprinted from \cite{goecks2019deep}.}
\label{tab:camera_specs}
\begin{tabular}{cc}
\hline
\textbf{Parameter} & \textbf{Value} \\ \hline
Resolution (pixels) & 1024x768 \\ \hline 
Aspect Ratio & 4:3 \\ \hline 
Horizontal Field of View (deg) & 90 \\ \hline 
Vertical Field of View (deg) & 30 \\ \hline 
Pan Angle w.r.t. Aircraft Frame (deg) & -90 \\ \hline 
Tilt Angle w.r.t. Aircraft Frame (deg) & -20 \\ \hline 
\end{tabular}
\end{table}


\emph{Reinforcement Learning} is concerned about learning in an unknown stochastic environment and can be formalized as a \emph{Partially Observable Markov Decision Process} (POMDP) \cite{Bertsekas2000}. A POMDP $\mathcal{M}$ can be characterize by its state-space $\mathcal{S}$ (where a vector of states $\vec{s} \in \mathcal{S}$), an observation-space $\mathcal{O}$ (where a vector of observations $\vec{o} \in \mathcal{O}$), an action-space $\mathcal{A}$ (where a vector of actions $\vec{a} \in \mathcal{A}$), a transition operator $\mathcal{T}$ (which defines the probability distribution $p(\vec{o}_{t+1}|\vec{o}_{t})$), and the reward function $r: \mathcal{S} \times \mathcal{A} \rightarrow \mathbb{R}$ (or $r(\vec{s},\vec{a})$. This definition is illustrated by Equation \ref{eq:pomdp} and Figure \ref{f:pomdp}:

\begin{equation} \label{eq:pomdp}
\mathcal{M} = \{ \mathcal{S},\mathcal{A},\mathcal{O},\mathcal{T}, r \}.
\end{equation}


To simplify further derivation of the algorithm it is assumed a fully-observable environment, where the observation-space $\mathcal{O} = \mathcal{S}$ (state-space) and, consequently, vector of observations $\vec{o} = \vec{s}$ (states).

In Reinforcement Learning, at each time step $t$ of a finite time horizon $T$, a \emph{policy} (or controller) $\pi$, parametrized by $\theta_t$, maps the current environment's \emph{states} $\vec{s}_t$ to \emph{actions} $\vec{a}_t$: $\pi_{\theta}(\vec{a}_t|\vec{s}_t)$. This action affects the current environment's states $\vec{s}_t$ which evolves to $\vec{s}_{t+1}$ based on the environment's transition distribution (dynamics) $p(\vec{s}_{t+1}|\vec{s}_t,\vec{a}_t)$. The environment also returns a scalar \emph{reward function} $r$ that evaluates the action taken $\vec{a}_t$ at the state $\vec{s}_t$: $r(\vec{s}_t,\vec{a}_t)$.

During an episode, the sequence of states observed and actions taken over a number of time steps $T$ can be represented by a \emph{trajectory} $\tau$:

\begin{equation}
\tau = \{ \vec{s}_0, \vec{a}_0, \vec{s}_1, \vec{a}_1, \ldots , \vec{s}_T, \vec{a}_T \}
\end{equation}

The probability of experiencing a given trajectory $\tau$ in a Markov Decision process can be written as:
\begin{align}
\pi_{\theta}(\tau) &= p_{\theta} (\vec{s}_0, \vec{a}_0, \vec{s}_1, \vec{a}_1, \ldots , \vec{s}_T, \vec{a}_T) \\
&= p(\vec{s}_1) \prod_{t=1}^T \pi_{\theta}(\vec{a}_t|\vec{s}_t) p(\vec{s}_{t+1} | \vec{s}_t, \vec{a}_t)
\end{align}

The goal in RL is to find the parameters $\theta^*$ that will maximize the objective $J(\theta)$, which represents the expected total reward to be received by this policy $\pi_{\theta}(\tau)$:

\begin{align}
\theta^* &= \argmax_{\theta} J(\theta) \\
&= \argmax_{\theta} \E_{\tau \sim p_{\theta} (\tau)} [\sum_{t=1}^{T} r(\vec{s}_t, \vec{a}_t)]
\end{align}

To maximize $J(\theta)$, it is possible to compute its gradient with respect to its parameters $\theta$. Since the structure of the objective $J(\theta)$ and its gradient $\nabla_{\theta} J(\theta)$ are unknown, in practice the only way to evaluate them and approximate the expectation term is by sampling and averaging over $N$ samples:

\begin{equation}
\nabla_{\theta} J(\theta) \approx \frac{1}{N} \sum_{i=1}^N [ ( \sum_{t=1}^T \nabla_{\theta} \log \pi_{\theta}(\vec{a}_t^{(i)} | \vec{s}_t^{(i)}) ) ( \sum_{t=1}^T r(\vec{a}_t^{(i)}, \vec{s}_t^{(i)}) )]
\end{equation}

The policy is improved by \emph{gradient ascent} according to a step size $\alpha$, as shown in Equation \ref{eq:grad_ascent} and Figure \ref{fig:policy_imp}, optimizing the agent's policy during the training time.

\begin{equation} \label{eq:grad_ascent}
\theta \rightarrow \theta + \alpha \nabla_{\theta} J(\theta)
\end{equation}


\section{Numerical Results}  \label{ssec:experiments}

Machine learning algorithms, specially reinforcement learning using deep neural network as model representation, require thousands or even millions of training episodes to converge to a satisfactory policy. These learning algorithms are also sensitive to hyperparameter values, which tune the learning performance. Due to this reason, it is a good practice to performance a hyperparameter search before running long-term experiments. During a hyperparameter search, the same algorithm is tested for a short number of episodes with multiple hyperparameter values. Since, intuitively, more data leads to better performance in machine learning, it was decided to investigate the impact of the batch size hyperparameter, which controls the number of episodes that are used to compute the gradient ascent step during the policy improvement phase.

\begin{figure}[hbt!]%
    \centering
    \subfloat[Mean reward achieved during a training batch.]
    {{\includegraphics[width=.45\linewidth]{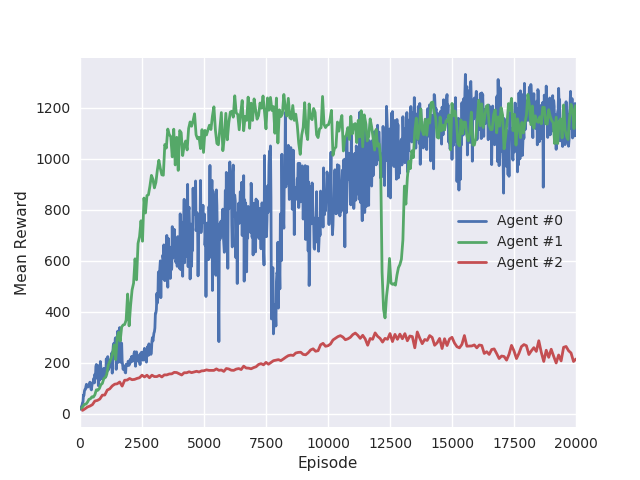} }}%
    \qquad
    \subfloat[Maximum number of steps during a training batch.]
    {{\includegraphics[width=.45\linewidth]{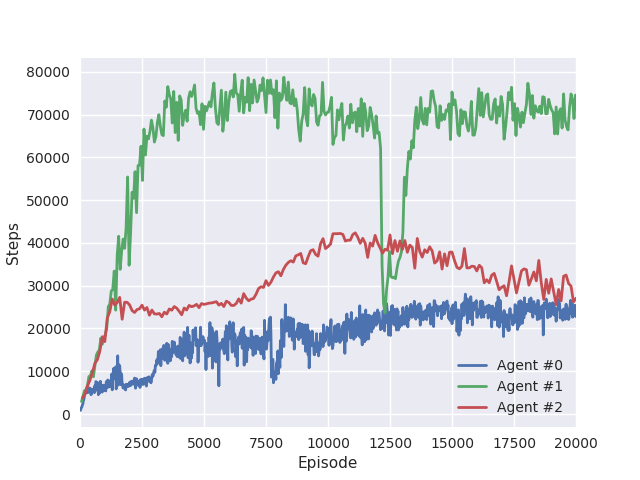} }}%
    \caption{Performance evaluation with respect to reward values and maximum number of steps achieved for different learning agents with different hyperparameters (see Table \ref{tab:hyperparameters_tt}) during each training episode. Reprinted from \cite{goecks2019deep}.}%
    \label{fig:rew_steps}%
\end{figure}

\begin{figure}[hbt!]%
    \centering
    \subfloat[Mean discounted reward achieved during a training batch.]
    {{\includegraphics[width=.45\linewidth]{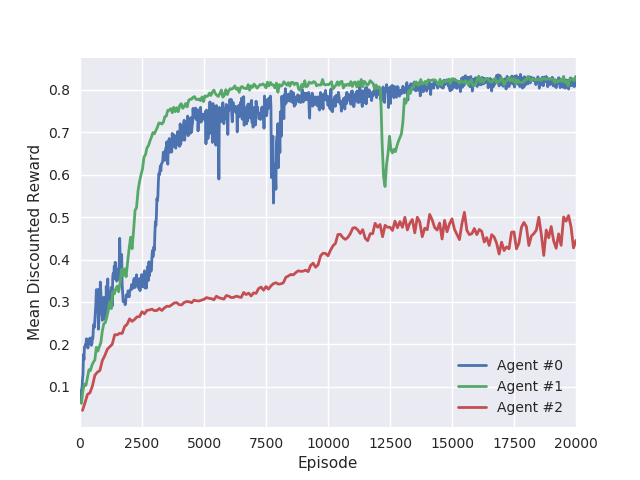} }}%
    \qquad
    \subfloat[Standard deviation of discounted reward achieved during a training batch.]
    {{\includegraphics[width=.45\linewidth]{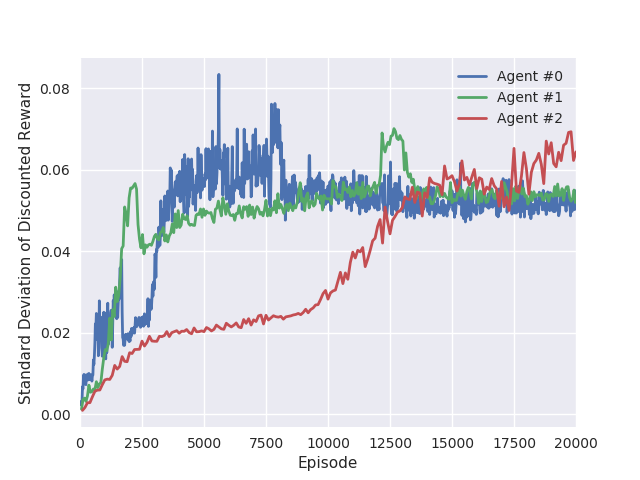} }}%
    \caption{Performance evaluation with respect to mean discounted reward and its standard deviation for different learning agents with different hyperparameters (see Table \ref{tab:hyperparameters_tt}) during each training episode. Reprinted from \cite{goecks2019deep}.}%
    \label{fig:mean_std_disc}%
\end{figure}

Figure \ref{fig:rew_steps} shows the reward values and maximum number of steps achieved per episode for different learning agents with different batch sizes, as summarized in Table \ref{tab:hyperparameters_tt}. A moderate batch size of 60, compare to 20 and 100, lead to longer tracking time as faster reward convergence. Counter-intuitively, larger batch size does not necessarily leads to faster convergence of the algorithm. Figure \ref{fig:mean_std_disc} shows the mean discounted reward achieved during a training batch, averaged by the batch size.   Larger batch sizes seems to lead to high variance performance.

\begin{table}[h]
\centering
\caption{Hyperparameter values used for the policy gradient deep reinforcement learning algorithm. Reprinted from \cite{goecks2019deep}.}
\label{tab:hyperparameters_tt}
\begin{tabular}{cccc}
\hline
\textbf{Hyperparameter} & \textbf{Agent \#0} & \textbf{Agent \#1} & \textbf{Agent \#2} \\ \hline
Number of Episodes per Training Batch & 20 & 60 & 100 \\ \hline 
Number of Episodes & 20000 & 20000 & 20000 \\ \hline
Discount Factor $\gamma$ & 0.995 & 0.995 & 0.995 \\ \hline
Generalized Advantage Estimation $\lambda$ & 0.98 & 0.98 & 0.98 \\ \hline
KL Divergence Target Value & 0.003 & 0.003 & 0.003 \\ \hline
\end{tabular}
\end{table}

\begin{figure}[hbt!]%
    \centering
    \includegraphics[width=.65\linewidth]{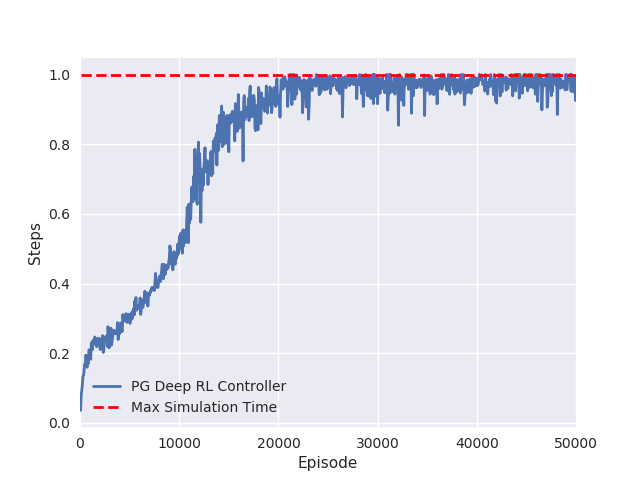}
    \caption{Consistent tracking performance of PG Deep RL agent on a simulated environment. Reprinted from \cite{goecks2019deep}.}%
    \label{fig:agent_steps}%
\end{figure}

After the hyperparameter search, it was desired to evaluate the tracking performance of the best performing agent. Using the hyperparameters of Agent \#1, the agent was retrained during 50000 episodes. The main goal was to keep the target in the image frame throughout the simulation period (one minute). Figure \ref{fig:agent_steps} shows the the PG Deep RL controller is able to consistently keep the target on the image frame throughout the maximum simulated time after 23,000 training episodes.

\section{Summary}


This paper presented an approach to solve the ground target tracking problem by proposing the Policy Gradient Deep Reinforcement Learning controller. The major advantage of the proposed controller with respect to previous approaches is being able to handle the full-state ground target tracking case, learning based on the full-state continuous aircraft states and controlling multiple outputs. Results on a simulated environment show that, after trained, the controller is able to consistently keep the target in the image frame.

Current results shows that the performance of the controller is drastically affected by the choice of the agent's hyperparameters, specifically the number of episodes per training batch --- batch size. Contrary to intuition, increasing batch size doesn't lead to improve controller performance. Batch size in the order of 60 episodes leads to faster learning and improved tracking time when compared to 20 and 100 episodes per batch.

After selected the best performing hyperparameters, current results shows that after 23,000 episodes of training the controller is able to learn how to maneuver the aircraft and keep the target in the image frame during the whole simulation period (60 seconds).

\chapter[CASE STUDY: CONTROL OF MORPHING WING SHAPES WITH DEEP REINFORCEMENT LEARNING]{CASE STUDY: CONTROL OF MORPHING WING SHAPES WITH DEEP REINFORCEMENT LEARNING\footnote{Adapted with permission from ``Control of Morphing Wing Shapes with Deep Reinforcement Learning'', by Vinicius G. Goecks, Pedro B. Leal, Trent White, John Valasek, and Darren J. Hartl, presented at the 2018 AIAA Information Systems-AIAA Infotech@ Aerospace \cite{goecks2018control}, Copyright 2018 by the American Institute of Aeronautics and Astronautics.}}\label{ch:wing}

Traditional model-based feedback control techniques are of limited utility for the control of many shape changing systems due to the high reconfigurability, high dimensionality, and nonlinear properties of the plant and actuators of these systems. Computational intelligence and learning techniques offer the promise of effectively leveraging the use of both smart materials and controls for application in aerospace systems such as morphing air vehicles. This paper addresses the challenge of controlling morphing air vehicles by developing a deep neural networks and reinforcement learning technique as a control strategy for shape-memory alloy (SMA) actuators in the context of a morphing wing.  The control objective is to minimize the error between an objective and the actual airfoil trailing edge deflection. The proposed controller is evaluated on a simple inverted pendulum for validation, on a 3D printed wing section that is actuated by a composite SMA actuator in a wind tunnel, and on a simulation based on wind tunnel data. Results show that the learning algorithm is capable of learning how to morph the wing. It is also able to control shape changes from arbitrary initial shapes to arbitrary goal shapes using the same trained learning algorithm. The results provide a proof of concept for the use of learning algorithms to control more complex morphing aircraft with continuous states and actions for the outer mold line configuration. 

\section{Problem Definition}
Aircraft design is typically a performance trade-off between different flight conditions while satisfying certain safety requirements~\cite{valasek_morphing_2012}. Morphing aircraft have the potential to allow an aircraft to change its shape and optimize performance for different mission objectives \cite{Lampton2008}. Recent efforts in the development of shape-memory alloy (SMA) actuators~\cite{leal_design_2017-1} have shown that SMA actuators enable small-scale camber morphing that can increase performance and versatility for different mission objectives \cite{Kumar2009, Lampton2008}. However, these actuators have strong thermomechanical coupling, hysteretic dynamical behavior, and can change response characteristics over time~\cite{lagoudas_shape_2008}. Traditional model-based feedback control techniques are thus of limited utility especially when multi actuation is considered, however, computational intelligence and learning techniques offer the promise of effectively synthesizing robust and reliable controllers for systems with smart materials.

Academic interest in morphing structures has existed for decades~\cite{valasek_morphing_2012}, and federal agencies such as NASA have shown strong interest over the years. An example of this is the Mission Adaptive Digital Composite Aerostructure Technologies (MADCAT)~\cite{Swei2016}; a NASA program to develop a novel aerostructure concept that takes advantage of emerging digital composite materials and manufacturing methods. The objective is to build high stiffness-to-density ratio, ultra-light structures that will facilitate the design of adaptive and aerodynamically efficient air vehicles. Another explored concept was the Spanwise Adaptive Wing Concept~\cite{NASA2016} that introduced a folding surface at the outboard part of the wing. According to flight conditions, the surface is folded to reduce drag, increase lift, and improve lateral-directional stability. This would allow an increase of performance during taxiing, takeoff, cruise, and supersonic flight.

Considering morphing wings, they not only present challenges from a material science and structure standpoint, but also from a flight control design standpoint due to changing dynamics and inherent uncertainties. A solution proposed by Valasek et al. \cite{Valasek2005,Valasek2008} is the Adaptive-Reinforcement Learning Control technique (A-RLC) that learns near optimal shape changes from arbitrary initial configuration to other arbitrary configuration that maximizes the performance of the aircraft for a given flight condition or maneuver. At the same time, it uses Structured Adaptive Model Inversion (SAMI) as a trajectory tracking controller to handle the time-variant properties, parametric uncertainties, and disturbances. The control law learns how to shape the airfoil by changing two discrete degrees-of-freedom: thickness and camber. The goal of the learning algorithm is to meet nominal aerodynamic goals in terms of lift \cite{Lampton2010}. The work was later extended to handle four degrees-of-freedom while morphing: wing tip chord, root chord, span, and leading edge sweep angle \cite{Valasek2012}.

This paper addresses the camber control of morphing wings using modern reinforcement learning techniques and deep neural networks. The main advantage of this method is the use of continuous states and actions for the outer mold line (OML) configuration and control inputs. The same learning algorithm is validated in different cases of a pendulum upswing task (different mass, gravity, length) to validate that a nominal learning algorithm can learn a satisfactory policy even when the model changes, which will happen to a morphing airfoil. The validated algorithm is then applied to morph a scaled wing prototype with embedded SMA actuators tested in a wind tunnel environment, and on a simulated morphing wing model created based on the wind tunnel data. Performance of the algorithm is evaluated by commanding different desired airfoil trailing edge deflections from different initial states.

\section{Learning Morphing Between Wing Shapes}

\begin{figure}[htb]
    \centering
	\includegraphics[width=4.8in]{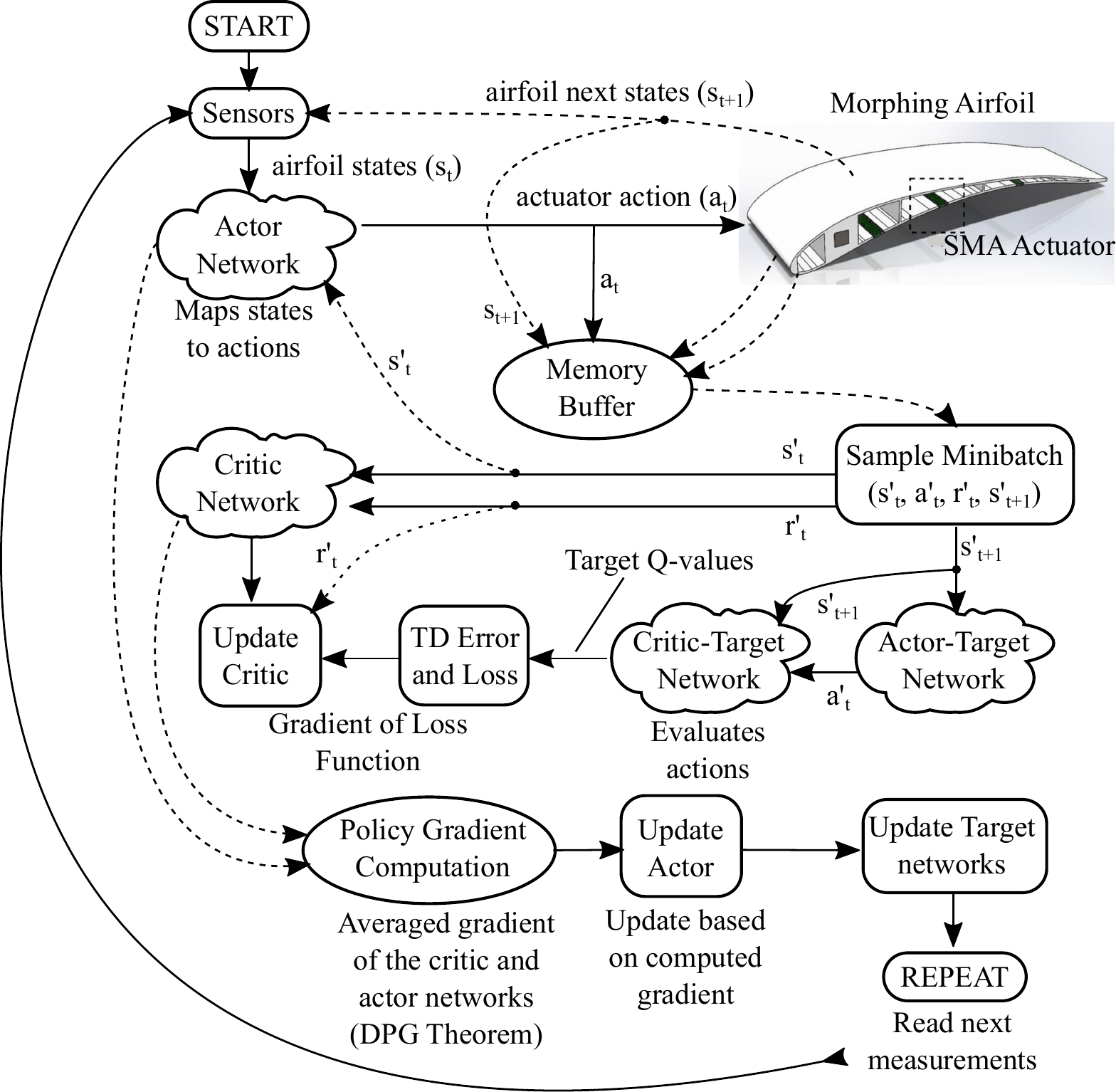}
	\caption{Diagram of the learning algorithm and interface with the morphing wing model. Reprinted from \cite{goecks2018control}.}
	\label{fig:alg_diagram}
\end{figure}

In this paper, the configuration of the wing is defined by the deflection of its trailing edge. The goal of the learning algorithm is to adapt the deep neural network parameters that represent the controller to successfully map the input deflections to output voltages necessary to move from any arbitrary wing configuration to another. The Deep Reinforcement Learning controller receives sensor readings that indicate the current deflection of the wing and a reward signal based off of an evaluation of the control applied. The output voltage is used to heat up the SMA wires via Joule effect leading to austenite transformation. Consequently, the transformation strain is recovered and the wing morphs. Figure \ref{fig:alg_diagram} shows how the learning algorithm interfaces with the morphing wing model, stores past experiences in a memory buffer, maps current states to actions (learns the policy), and computes the gradients based on the received rewards to update the network parameters, as explained in Algorithm \ref{alg:ddpg} and Subsection \ref{ddpg_algo}.

\begin{figure}[h]
    \centering
	\includegraphics[width=5in]{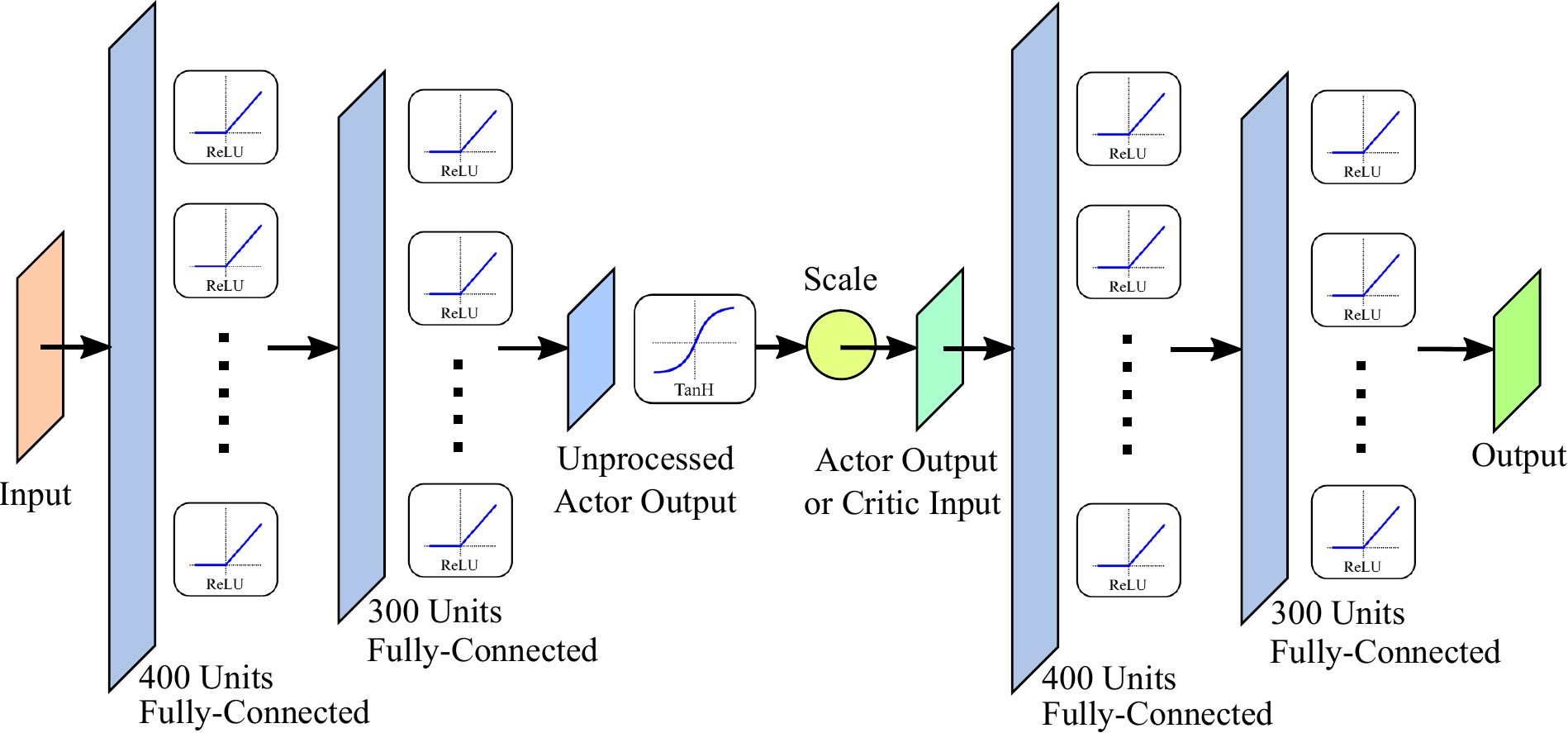}
	\caption{Deep neural network architecture that maps the current wing configuration to control inputs. Reprinted from \cite{goecks2018control}.}
	\label{fig:net_architecture}
\end{figure}

As mentioned in Section II.B, the network architecture consists of neurons organized as hidden layers. The neural network herein implemented is depicted in Figure \ref{fig:net_architecture} which in addition shows how these neurons are connected between layers, the specific activation functions used after each layer output, and additional operations to transform the network output. The learning algorithm is evaluated based on the control effort required to morph, time to go from the initial to the final configuration, and number of iterations required to train the learning algorithm to achieve the best performance.

Another important feature of the DDPG algorithm are the hyperparameters. These parameters are used for extra tuning of the learning algorithm and direct affect its learning performance. The hyperparameter values used in this work are depicted in Table \ref{tab:hyperparameters}. All the parameters and their function are herein described. The Target Update Rate defines how often (in episodes) the target network parameters are copied from the original network. Target Update Factor is a relaxation factor for the copied network parameters ($\theta$). Actor and Critic Learning Rates define the step size of the gradient update when performing the optimization of the network parameters. Memory Buffer Size defines the number of past experiences (rewards, states, and actions observed from the interaction between agent and environment) that are stored and mixed with current experiences. Minibatch Size controls the number of experiences used during each gradient update on the networks. Discount Factor, bounded between zero and one, defines how future expected rewards are accounted during the learning phase.

\begin{table}[h]
\centering
\caption{Hyperparameter values used for DDPG algorithm. Reprinted from \cite{goecks2018control}.}
\label{tab:hyperparameters}
\begin{tabular}{cc}
\hline
\textbf{Hyperparameter} & \textbf{Value} \\ \hline
Target Update Rate (Episodes)                & 100              \\ \hline
Target Update Factor ($\tau$)                & 0.001              \\ \hline
Actor Learning Rate               & 0.0001              \\ \hline
Critic Learning Rate                & 0.001              \\ \hline
Memory Buffer Size                & 100,000              \\ \hline
Minibatch Size                & 64              \\ \hline
Discount Factor $\gamma$                & 0.99              \\ \hline
\end{tabular}
\end{table}


\section{Experimental Setup}

\begin{figure}[!ht]
    \centering
    \includegraphics[width=0.65\textwidth]{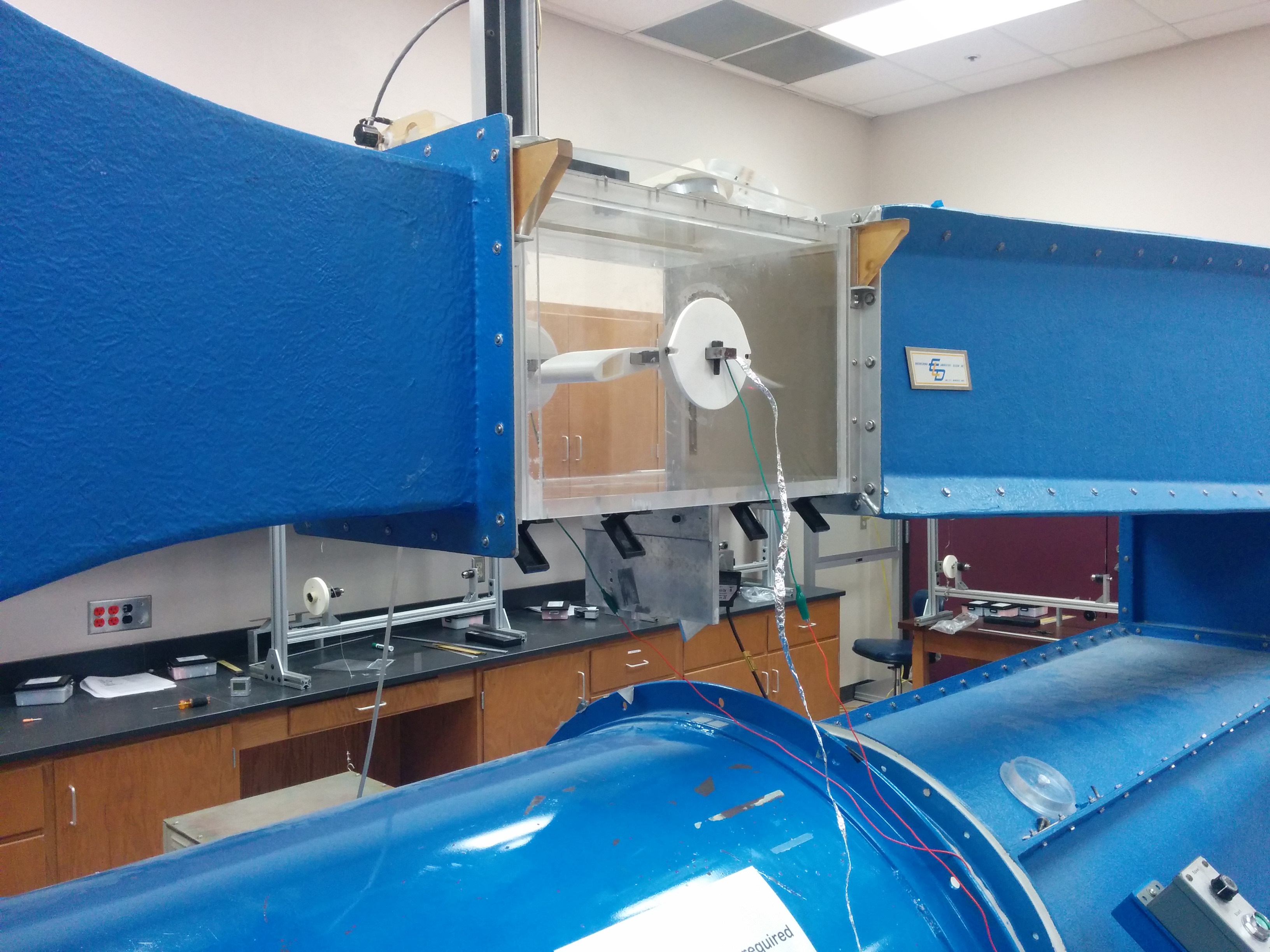}
    \caption{Experimental setup for the learning algorithm: prototype in the wind tunnel. Reprinted from \cite{goecks2018control}.}
    \label{f:wind_tunnel_a}%
\end{figure}

\begin{figure}[!ht]
    \centering
    \includegraphics[width=0.5\textwidth]{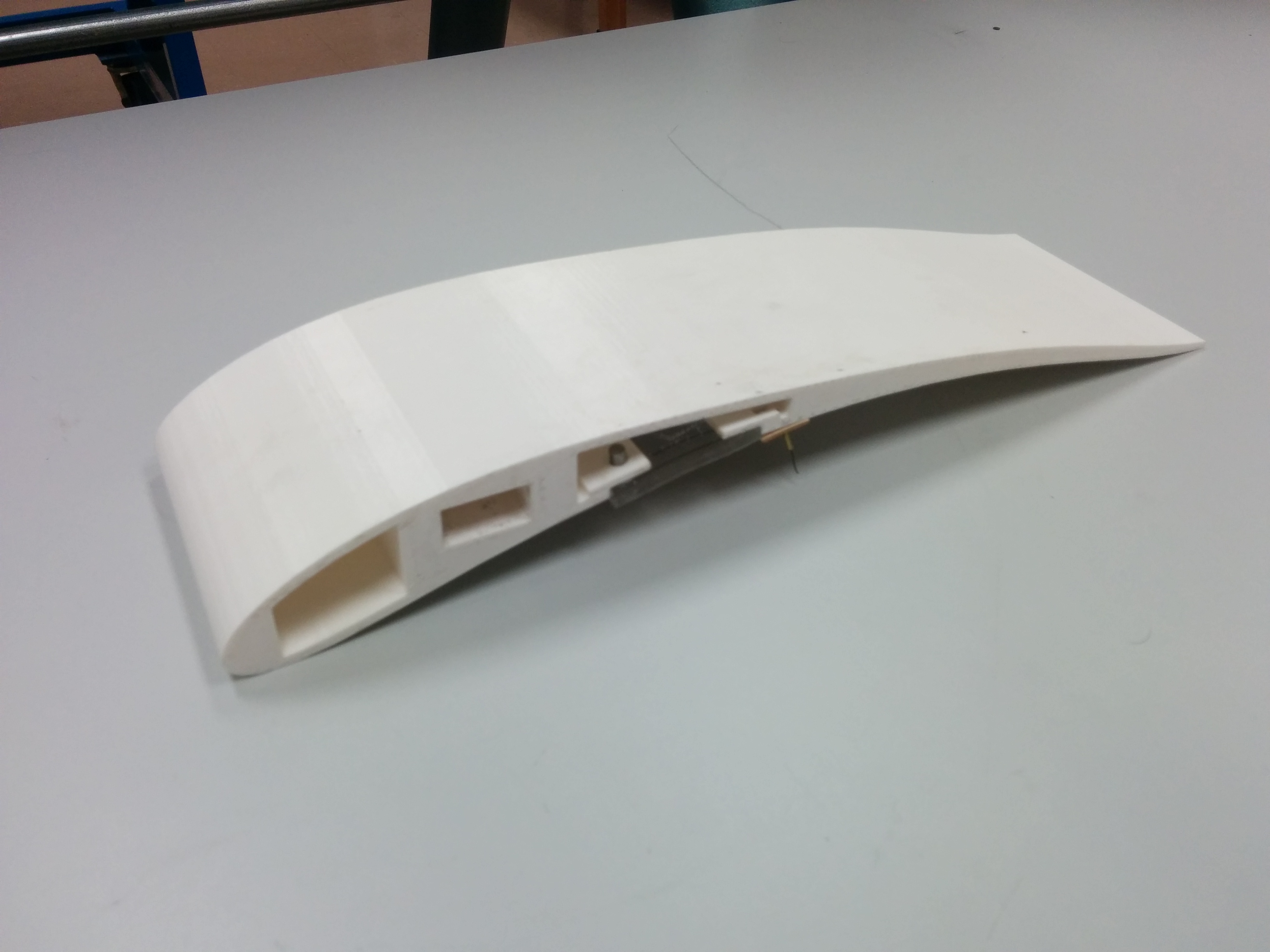}
    \caption{Experimental setup for the learning algorithm: top view of the prototype. Reprinted from \cite{goecks2018control}.}
    \label{f:wind_tunnel_b}%
\end{figure}

\begin{figure}[!ht]
    \centering
    \includegraphics[width=0.5\textwidth]{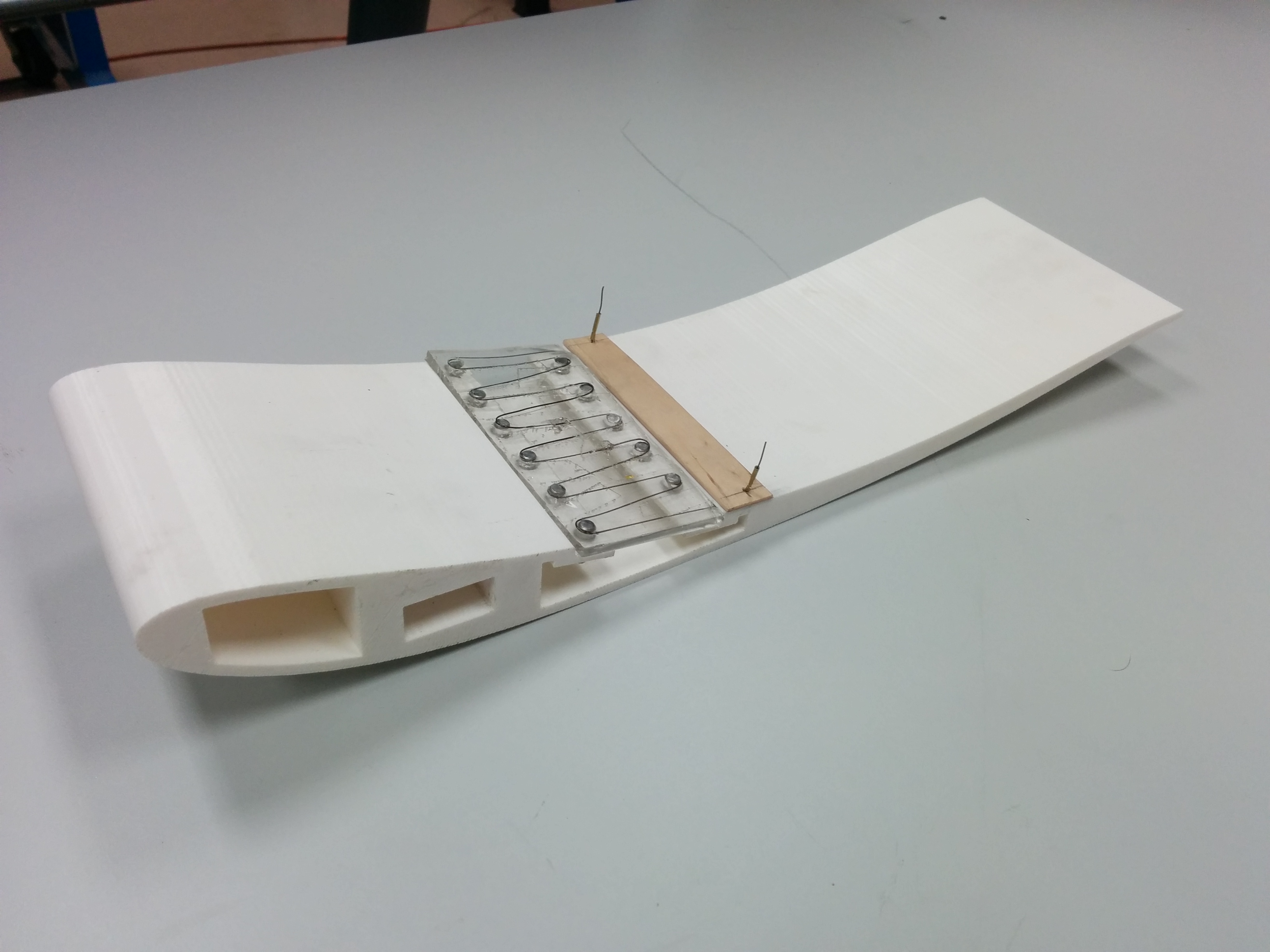}
    \caption{Experimental setup for the learning algorithm: bottom view of the prototype. Reprinted from \cite{goecks2018control}.}
    \label{f:wind_tunnel_c}%
\end{figure}

The main experimental setup comprised of wind tunnel testing a 3D-printed, avian-inspired, wing prototype actuated by a composite SMA actuator powered by an external power supply, as seen in Figures \ref{f:wind_tunnel_a}, \ref{f:wind_tunnel_b}, and \ref{f:wind_tunnel_c}. The actuator consists of a PDMS matrix with embedded SMA wires. Trailing edge deflection of the wing, voltage applied to the SMA wire, and its temperature were measured during the experiment. 


The experiment consists in defining a random setpoint and initial position for the trailing edge of the wing. The learning algorithm reads the current deflection and setpoint applied voltage to the SMA wire in order to morph the wing to the desired state in less than 200 measurement iterations that comprise the learning episode. At the beginning of each episode, the goal and initial states are randomized and the learning algorithm has to perform the task again. During each episode, the learning algorithm is trying to maximize the objective function $R$ (``total reward'' of the episode) dependent on the magnitude of control applied and error between current and desired deflection given as:

\begin{equation}\label{eq:reward_funciton}
R = \frac{1}{T} \sum_{t=0}^{T} e^{(x^*-x)^2 - u/8},
\end{equation}

\noindent where $T$ is the total number of time steps $t$, $x$ is the current deflection, $x^{*}$ the setpoint, and $u$ is the voltage applied being normalized by its maximum value of 8 volts.

The network used to learn the dynamics comprises of three fully connected layers (each input is connected to every parameter of that layer) with 220, 160, and 130 neurons, respectively. The fully connected layers are followed by dropout connections (randomly disconnects 20\% of the units and its input connections during the training phase for better generalization \cite{Srivastava2014a}). The connection between each layer is made through rectified linear units (ReLU) activation functions, which discards the input if it is negative and adds nonlinearities to the network when more hidden layers are stacked. The deep neural network if optimized using the ``Adam'' (Adaptive Momentum Estimation) \cite{Kingma2015} optimizer based on a standard mean squared error loss function of the predicted and true temperature and displacement values. ``Adam'' is an algorithm for first-order gradient-based optimization of stochastic objective functions, based on adaptive estimates of lower-order moments. It is computationally efficient, little memory requirements, is invariant to diagonal rescaling of the gradients, and is well suited for problems that are large in terms of data and/or parameters. It stores exponentially decaying average of past squared gradients and past gradients \cite{Kingma2015}.

\begin{figure}[H]
    \centering\subfloat[]{{\includegraphics[width=0.75\linewidth]{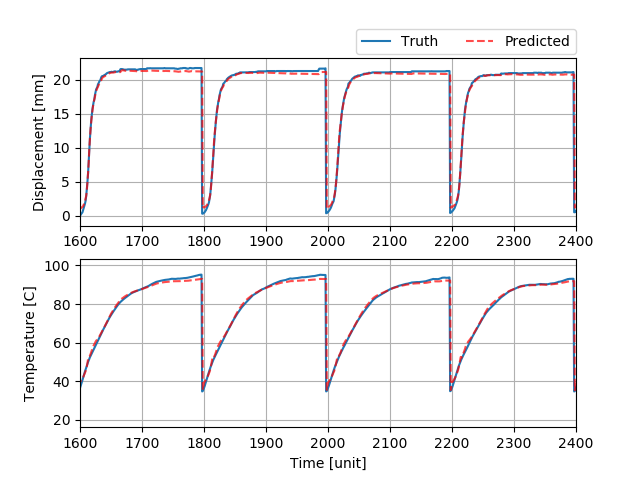}\label{fig:learned_a} }}%
    \hfill
    \centering\subfloat[]{{\includegraphics[width=0.75\linewidth]{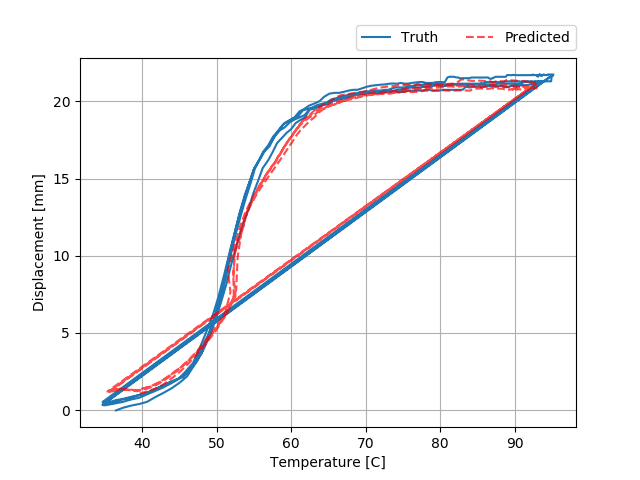}\label{fig:learned_b} }}%
    \caption{Preliminary results modeling the simple spring system showing (a) Displacement and temperature changes over time; and (b) Displacement as a function of temperature. Reprinted from \cite{goecks2018control}.}
    \label{fig:learned}
\end{figure}

Applying data-driven learning algorithms directly to hardware is challenging. Deep Reinforcement Learning algorithms are well known to heavily depend on multiple graphic and central processing units (GPUs and CPUs) \cite{Mnih2013, Hasselt2015, Schaul2015, Mnih2016} and training time on the order of days \cite{Mnih2013, Mnih2015a} to achieve meaningful results. Simple continuous tasks (e.g., controlling a two-link robotic arm), requires on average more than 2.5 million steps \cite{lillicrap2015continuous} and more complex tasks (e.g., control of a humanoid robot or robots with multiple degrees of freedom), requires on average 25 million steps using different reinforcement learning algorithms \cite{Duan2016}. Since it was infeasible to perform a wind tunnel test in the order of weeks, it was desired to create a high-fidelity simulation model that would have the same dynamics of the avian-inspired airfoil section actuated by a SMA wire under a wind tunnel test. To achieve this goal, a deep neural network was implemented using initial wind tunnel data to learn to approximate the hardware dynamics. Figure \ref{fig:learned} shows the learned model compared to the truth model in terms of temperature and displacement.

\section{Numerical Results}


\subsection{Validation of the Learning Algorithm}
\begin{figure}[H]
    \subfloat[]{{\includegraphics[width=0.45\linewidth]{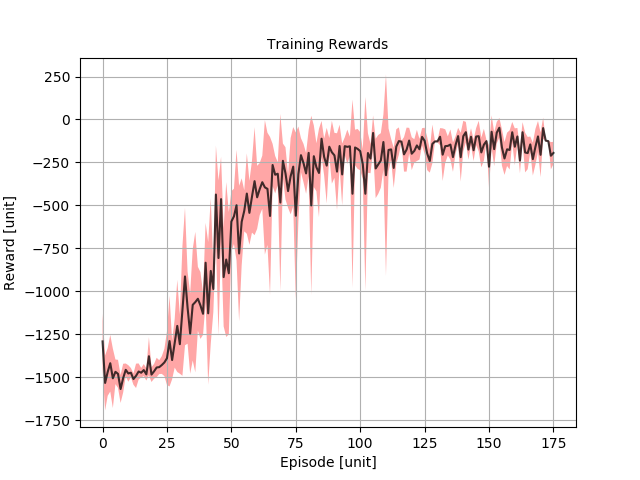}\label{fig:val_ddpg_a} }}%
    \subfloat[]{{\includegraphics[width=0.45\linewidth]{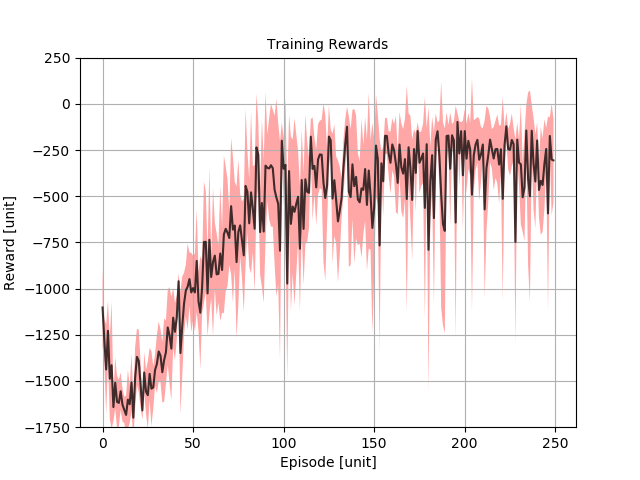}\label{fig:val_ddpg_b} }}%
    \hfill
    \subfloat[]{{\includegraphics[width=0.45\linewidth]{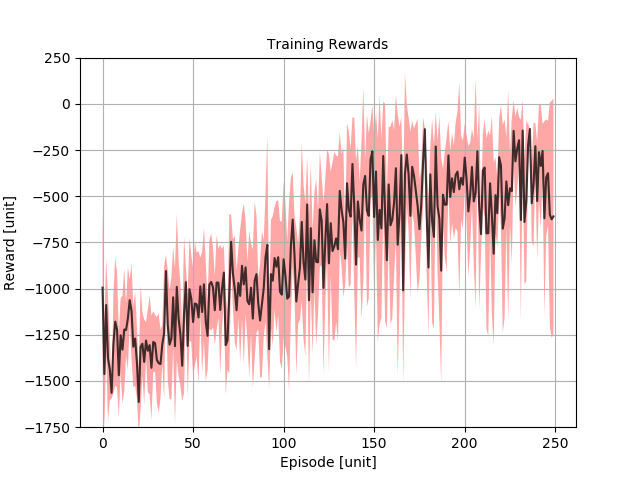}\label{fig:val_ddpg_c} }}%
    \subfloat[]{{\includegraphics[width=0.45\linewidth]{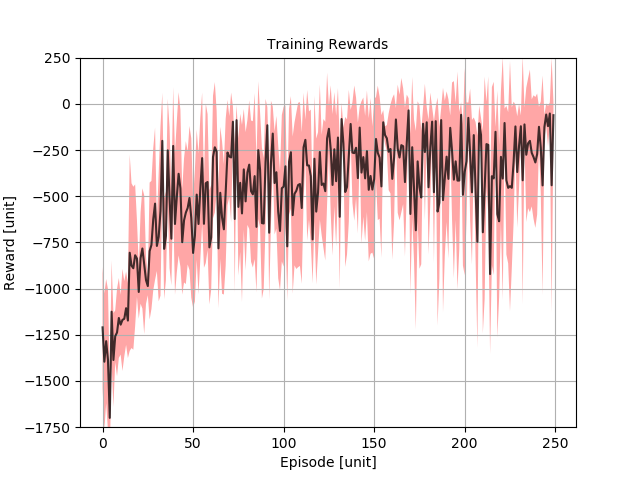}\label{fig:val_ddpg_d} }}%
  \caption{Results of the validation of the Deep Deterministic Policy Gradient algorithm solving the inverted
pendulum upswing. Average of five runs and standard deviation of rewards per episode. (a) Unitary mass pendulum case; (b) Double mass pendulum case; (c) Double length pendulum case; and (d) Half gravity pendulum case. Reprinted from \cite{goecks2018control}.}
  \label{val_ddpg_total} 
\end{figure}

Before implementing the DDPG learning algorithm on the morphing wing experiment, it was the implementation was tested on a simulated inverted pendulum upswing. In this system, the algorithm must apply continuous positive or negative torque ($u$) to the pendulum shaft to keep it in the upright position. The only states available to the learning algorithm is the current angular position and velocity ($\theta$ and $\dot{\theta}$) of the pendulum. A reward signal was given to algorithm to represent its performance, as shown in Eq. (\ref{eq:cost_ddpg}), for every step. Reward results of more than -200 per episode are considered satisfactory because they numerically represent that the pendulum is held in vertical equilibrium after the swing. Figure \ref{fig:val_ddpg_a} shows the resultant reward values per episode. It can be seen that the algorithm achieved the desired behavior of obtaining reward results of more than -200 on the final learning episodes, converging to a constant result. This can be interpreted as the learning algorithm, based on measured position and velocity and torque applied to the pendulum, being able to maximize the reward signal returned by the environment and complete the maneuver of an upswing in within 10 seconds. 

To validate that the algorithm with same hyperparameters is able to achieve satisfactory behavior even with changes in the model, the same agent was trained on an inverted pendulum double the original mass. The algorithm has no knowledge that the environment was changed. Figure \ref{fig:val_ddpg_b} shows the resultant reward values per episode for the double mass case. It can be observed that the same algorithm and neural network were able to learn a policy robust to changes in the environments and still approach the defined mark of -200 reward points that represent a complete inverted pendulum upswing maneuver. As an additional evaluation, Fig. \ref{fig:val_ddpg_c} and \ref{fig:val_ddpg_d} show the resultant reward values per episode for a double length pendulum case and for the half gravity case, respectively.

\begin{equation}\label{eq:cost_ddpg}
r = -(||\theta^2|| + .1(\dot{\theta}^2) + .001(u^2))
\end{equation}

\subsection{The Learning Algorithm in the Wind Tunnel}\label{s:experimental}

\begin{figure}[H]
    \subfloat[]{{\includegraphics[width=0.5\linewidth]{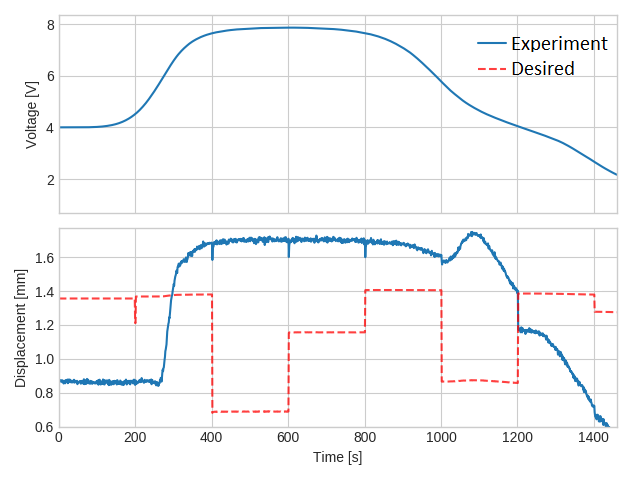}\label{fig:wind_tunnel_res1} }}%
    \subfloat[]{{\includegraphics[width=0.5\linewidth]{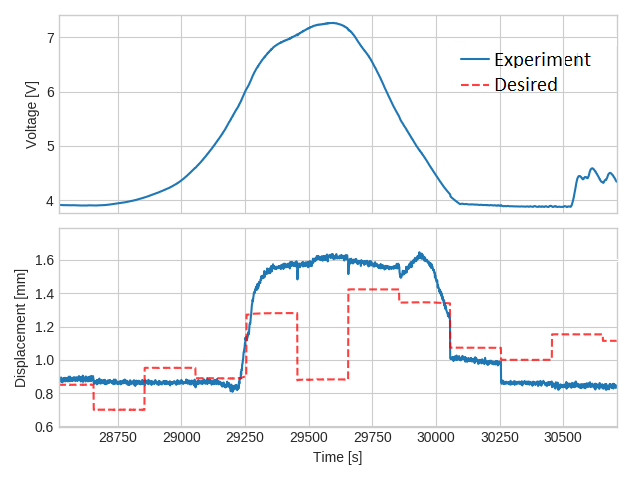}\label{fig:wind_tunnel_res2} }}%
  \caption{Training on the Wind Tunnel: Deep Reinforcement Learning controller performance for training time consisting of: (a) 100 and (b) 200 episodes. Reprinted from \cite{goecks2018control}.}
  \label{fig:wind_tunnel_res} 
\end{figure}

The main hypothesis is that the same algorithm used to learn to control different cases of the upswing pendulum task can be used to learn an shape change policy for morphing airfoils. The deep reinforcement learning algorithm was deployed to the hardware platform to test the hypothesis. The algorithm ran in the wind tunnel for more than 300 episodes. Every 200 time steps, or about 200 seconds, the commanded setpoint was randomly changed and the learning algorithm had to adapt the applied voltage to match the commanded value. It was intended to perform the training session for more than 300 episodes, but the wind tunnel operation was limited. Results achieved after 100 and 200 episodes are shown in Fig. \ref{fig:wind_tunnel_res}. As expected, 300 episodes were not enough to train a policy that would satisfactory command the current displacement to the desired reference value, but it is possible to observe an improvement after 100 and 200 training episodes.

\subsection{The Learning Algorithm in the Simulation Model}

\begin{figure}[H] 
    \subfloat[]{{\includegraphics[width=0.5\linewidth]{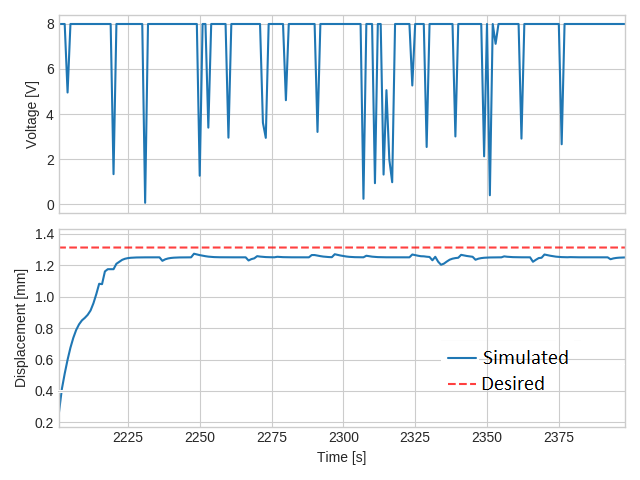}\label{fig:sim_wind_tunnel_res1} }}%
    \subfloat[]{{\includegraphics[width=0.5\linewidth]{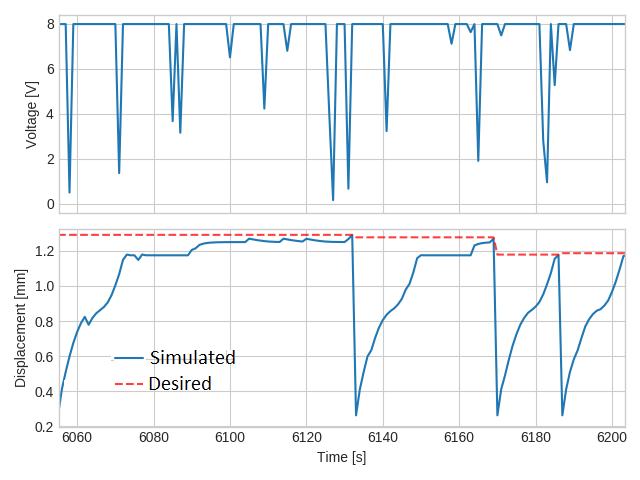}\label{fig:sim_wind_tunnel_res2} }}%
  \caption{Training on the Modeled Airfoil: Deep Reinforcement Learning controller performance after 900 (a) and 1500 (b) episodes of training time, respectively. Reprinted from \cite{goecks2018control}.}
  \label{fig:sim_wind_tunnel_res} 
\end{figure}

As explained in Section \ref{s:experimental}, due to limited time at the wind tunnel, the learning algorithm was also tested using a simulated wing model created based on data collected from the wind tunnel. The algorithm ran for 1500 episodes. Every 200 time steps the commanded setpoint was randomly changed and the learning algorithm had to adapt the applied voltage to match the commanded value. Figure \ref{fig:sim_wind_tunnel_res} shows the controller performance after 900 and 1500 episodes of training. After 900 episodes there is still steady-state error between the current and commanded displacement. After 1500 the deep reinforcement learning algorithm is able to complete the task multiple times in a row for different commanded morphing shapes.

\section{Summary}


This paper addressed the challenge of controlling morphing air vehicles by developing a deep neural network and reinforcement learning technique for the control of shape memory alloy actuators that adapt the wing outer mold line. The objective was to investigate if the learning algorithm could also be used to learn satisfactory control policies for morphing wing.  These are characterized by a plant which has continuous states and actions and time-varying dynamics.  Based upon the results presented in the paper, the following conclusions are drawn. 

\begin{enumerate}

\item Data collected from wind tunnel testing was shown to be effective in training a deep neural network that accurately mimics the dynamic behavior of the wing section actuated by a shape memory alloy.  The same learning algorithm and hyper-parameters were also shown to control shape changes from arbitrary initial shapes to arbitrary goal shapes when trained with the high-fidelity simulation model.

\item The deep reinforcement learning algorithm using the Deep Deterministic Policy Gradient learned to compute torques and perform an upswing maneuver on four different versions of an inverted pendulum. This algorithm was also shown to control a 3D-printed, avian-inspired airfoil section actuated by a shape-memory alloy operating in a wind tunnel.  However, a fully satisfactory control policy was not learned due to time limitations on the testing platform.

\item The results provided insights that deep neural networks can be successfully used to model the dynamics of complex morphing air vehicle systems, followed by deep reinforcement learning algorithms to control these dynamic systems with continuous states and actions.

\end{enumerate}

Further work on deep reinforcement learning algorithms are needed to reduce the extensive offline training required, and reduce the number of interactions with the environment.  This would ease the transition to direct training on relevant hardware platforms.

\chapter{CASE STUDY: INVERSE REINFORCEMENT LEARNING APPLIED TO GUIDANCE AND CONTROL OF SMALL UNMANNED AERIAL VEHICLES} \label{ch:irl}

This chapter was submitted under the name of ``Inverse Reinforcement Learning Applied to Guidance and Control of Small Unmanned Aerial Vehicles'' as final project of the 2018 Reinforcement Learning class at Texas A\&M University, in Department of Electrical and Computer Engineering, taught by Dr. Dileep Kalathil, wrote by Vinicius G. Goecks, Akshay Sarvesh, Kishan P. Badrinath and is partially reproduced here.

In this chapter we investigate traditional Inverse Reinforcement Learning approaches and their extensions Guided Cost Learning and Generative Adversarial Imitation Learning to learn how to land a small Unmanned Aerial Vehicle based on human demonstration. This task is complex to traditional Inverse Reinforcement Learning algorithms due to the continuous nature of the observation and action-space. Our approach is first validated on less complex continuous tasks available on OpenAI Gym \cite{Brockman2016} --- the Pendulum-v0 and LunarLanderContinuous-v2 environments --- and later transferred to a high-fidelity UAV simulator, Microsoft AirSim \cite{Airsim2017}.
Our results show that Generative Adversarial Imitation Learning is able to learn how to land an unmanned aerial vehicle and surpass human mean performance after 100 learning iterations. This approach learns from 100 human demonstrations of the landing task, equivalent to about 30 minutes of demonstration. The learning converges to the upper-bound of human performance after 400 learning iterations.

\section{Problem Definition}
\label{sec:introduction}
It is true that humans can no longer claim to be the experts in tasks like image classification and recognition \cite{Microsoft2018}. At the same time there are also a number of tasks where humans can easily outperform the machines, for now, in more complex tasks as, for example, driving and playing sports, mostly due to challenges in perception, motion control, and mechanical actuation.

A traditional Reinforcement Learning (RL) problem focuses on an agent learning a process of decisions (policies) to produce a output which maximizes a reward function return by the environment. In Inverse Reinforcement Learning (IRL), the goal is to extract a reward functions by observing the behavior of an policy or any set of demonstrations.
In this project we investigate traditional IRL approaches and their extensions Guided Cost Learning (GCL) and Generative Adversarial Imitation Learning (GAIL) to learn how to land a small Unmanned Aerial Vehicle (UAV) based on human demonstration. This task is complex for traditional IRL algorithms due to the continuous nature of the observation and action-space. Our approach is first validated on less complex continuous tasks available on OpenAI Gym \cite{Brockman2016} --- the Pendulum-v0 and LunarLanderContinuous-v2 environments --- and later transferred to a high-fidelity UAV simulator, Microsoft AirSim \cite{Airsim2017}.

Section \ref{sec:notation} of this chapter introduces the notation used throughout this document and the theoretical background of the algorithms used for this work. Section \ref{sec:related_work} details previous approaches for the IRL problem, highlighting their strengths and limitations. Section \ref{sec:uav} details the application of IRL to learn the UAV landing task based on human demonstration, followed by results and discussion on Section \ref{sec:results}. A summary is described in Section \ref{sec:conclusions}.

\section{Notation and Background}
\label{sec:notation}
\subsection{Notation}
A finite Markov Decision process (MDP) is a tuple $(S,A,\{P_{sa}\},\gamma,R)$ where : $S$ is a finite set of N $\mathbf{states}$, $A=\{a_1,a_2,..,a_k\}\ \ \mathbf{actions}$, $P_{sa}$ are the \emph{transition probabilities} for taking an action $a$ in state $s$, $\gamma \ \ \epsilon\ \ (0,1] $ is the \emph{discount factor}, R is the \emph{reinforcement function} which is bounded by $R_{max}$. 
\par
A \emph{policy} $\pi$ is defined as a map from $\pi : S\rightarrow A $ and the value fn evaluated for any policy $\pi$ is given by $V^{\pi}(s) = E[R(s_1)+\gamma R(S_2)+] + {\gamma}^2 R(s_3)+...+{\gamma}^k R(s_k)]$. The Q fn can also be defined as : $Q^{\pi}(s,a)=R(s)+ \gamma E[V^{\pi}(s)]$. The \emph{optimal value fn} is $V^{*}(s) = sup_{\pi}V_{\pi}(s)$ and the \emph{optimal Q fn is} $Q^*(s,a) = sup_{\pi}Q^{\pi}(s,a)$.
\par 
In a standard RL problem we find a policy $\pi$ such that $V^{pi}(s)$ is maximized. In the IRL problem, the goal is to find the best reward function for a set of observations.

\subsection{Guided Cost Learning}
Guided Cost Learning (GCL) \cite{Finn2016} introduces an iterative sample-based method for estimating the probability normalization term $Z$ in the Maximum entropy IRL formulation \cite{Ziebart2008}, and can scale to high-dimensional state and action spaces and non-linear reward structures. The algorithm estimates $Z$ by training a new sampling distribution $q(\tau)$, with estimates of the expert trajectories as $\tilde{p}(\tau)$, general reward function parameterized by $\theta$, and using importance sampling:
\begin{eqnarray*}
    \mathcal{L}_{\textit{reward}}(\theta) & = & \mathrm{E}_{\tau\sim p}[-r_\theta(\tau)] + \log \left( \mathrm{E}_{\tau\sim q}[\frac{\exp(r_\theta(\tau))}{ \frac{1}{2}\tilde{p}(\tau) + \frac{1}{2}q(\tau) }] \right).
\end{eqnarray*}
GCL alternates between optimizing $r_\theta$ using this estimate, and optimizing $q(\tau)$ to minimize the variance of the importance sampling estimate.

The optimal importance sampling distribution for estimating the partition function $Z$ is proportional to the exponential family of the reward function. During GCL, the sampling policy $q(\tau)$ is updated to match this distribution by minimizing the Kullback-Leibler (KL or $D_{KL}$) divergence between $q(\tau$ and $\frac{1}{Z}\exp(r_\theta(\tau))$, or equivalently maximizing the learned reward and entropy:
\begin{eqnarray*}
    \mathcal{L}_{\textit{sampler}}(q) = \mathrm{E}_{\tau\sim q}[-r_\theta(\tau)] + \mathrm{E}_{\tau\sim q}[\log q(\tau) ].
\end{eqnarray*}

\subsection{Generative Adversarial Imitation Learning}
Generative Adversarial Imitation Learning (GAIL) \cite{Ho2016} follows the generative modeling literature, in particular, Generative Adversarial Networks (GANs). Here, two models, a generator $G$ and a discriminator $D$, are trained simultaneously. The discriminator is tasked with classifying its inputs as either the output of the generator, or actual samples coming from an expert's data distribution $p(\tau)$; where $\tau$ is sample trajectories for GAIL. The goal of the generator is to produce outputs that are classified by the discriminator as coming from the underlying data distribution.

To elucidate, the generator is subsumed by the environment generating trajectories ($\tau \sim G$) from a random policy, while the discriminator takes as input a sample $\tau$ and outputs the probability $D(\tau)$ that the sample was from the data distribution. Since $D$ and $G$ are playing the same game but with opposing goals, their loss functions can be written as the following respectively:
\begin{eqnarray*}
    \mathcal{L}(D) = \mathrm{E}_{\tau\sim p}[-\log D(\tau)] & + & \mathrm{E}_{\tau\sim G}[-\log (1-D(\tau))] \\
    \mathcal{L}(G) = \mathrm{E}_{\tau\sim G}[-\log D(\tau)] & + & \mathrm{E}_{\tau\sim G}[\log (1-D(\tau))].
\end{eqnarray*}
\section{Related Work}
\label{sec:related_work}

\citeauthor{Ng2000} \cite{Ng2000} defined the problem of IRL almost two decades ago. They state that the reward function is a more robust definition of a task rather than the policies themselves. The authors formulate the problem of IRL as a optimization problem and demonstrate algorithms to solve the Linear Programming (LP) problem on simple discrete, finite and continuous and infinite state problems.

\citeauthor{Ziebart2008} \cite{Ziebart2008} takes a different approach to matching feature counts which allow their formulation to avoid the ambiguity in getting many reward functions to derive policies; hence giving a unique, but randomized, solution as opposed to works like \citeauthor{Abbeel2004} \cite{Abbeel2004}. They employ a tool, maximum entropy, from statistical mechanics developed by \citeauthor{jaynes57} \cite{jaynes57}, which prefers higher reward functions at an exponential rate in probability. So, maximizing the sum, meaning all possible observed trajectories, of logarithm of these probabilities across the reward function parameters, yields us with a stochastic solution for the reward function. But the suggested gradient descent algorithm has a drawback of recalculating the Markov Decision Process (MDP) in every step. So, we move on to the next work which incorporates these ideas and perform better.

\citeauthor{Wulfmeier2015a} \cite{Wulfmeier2015a} states that the objective of maximum entropy inverse reinforcement learning defined by \citeauthor{Ziebart2008} \cite{Ziebart2008}, maximizing the joint posterior distribution of observing the expert demonstration under a given reward structure and model parameters, is fully differentiable with respect to deep neural network weights.
\citeauthor{Wulfmeier2015a} \cite{Wulfmeier2015a} uses fully-connected convolutional networks (FCCN) to learn the reward model and evaluate the algorithm using \emph{expected value difference}: difference between value function for the optimal policy obtained using the learned reward model and using the ground truth reward. This approach achieves better results when compared to traditional maximum entropy IRL \cite{Ziebart2008}.

\citeauthor{Finn2016} \cite{Finn2016} extended MaxEntropy \cite{Ziebart2008} formulation for solving high dimensional problems by making use of statistic's tool of importance sampling to estimate the partition function, which was the cause of hindrance in MaxEntropy problem.
\citeauthor{Ho2016} \cite{Ho2016} presented a GAN-like algorithm for imitation learning, where the goal is to recover the policy that matched the export trajectories.
\citeauthor{FinnConn2016} \cite{FinnConn2016} show that both GCL and GAIL are mathematically equivalent, that is, both converge to the same policy which imitates the expert policy. However, \citeauthor{Ho2016} \cite{Ho2016} use the typical unconstrained form of the discriminator and do not use the generator's density, and thus the reward function remains implicit within the discriminator and cannot be recovered. 

\section{Application to Guidance and Control of Small Unmanned Aerial Vehicles}
\label{sec:uav}
More traditional IRL approaches as \cite{Ng2000,Abbeel2004,Ziebart2008} requires the MDP to be solved at each learning iteration in order to optimize the policy for the learned reward function. In our UAV scenario, since the MDP is unknown, it is not possible to solve it at each iteration. Two recent algorithms, as explained in Section \ref{sec:notation}, proposes to address these issues: GCL and GAIL.

The GCL and GAIL algorithms are first validated on less complex continuous tasks available on OpenAI Gym \cite{Brockman2016} --- the Pendulum-v0 and LunarLanderContinuous-v2 environments --- and later transferred to a high-fidelity UAV simulator, Microsoft AirSim \cite{Airsim2017}.
For the UAV scenario, the data was collected using the Microsoft AirSim environment \cite{Airsim2017} modified to simulate an small UAV perching task (landing) on top of a static vehicle, as seen in Figure \ref{fig:Screenshot}. For this task, a UAV equipped with a bottom-facing RGB camera starts on a random location flying over the target vehicle and have to land in a certain amount of time. The states consists of inertial data of the UAV (estimated x, y, and z position, linear and angular velocities) and three additional features of the landing pad extracted using a custom computer vision module running on the background (radius, x and y pixel position of the center of the landing pad) --- total of 15 features.


\section{Results and Discussion}
\label{sec:results}
This section details the validation of the GCL and GAIL algorithms using the Pendulum-v0 and LunarLanderContinuous-v2 OpenAI Gym environments and the transition to Microsoft Airsim simulator to train an agent to perform the UAV landing task.

\subsection{Numerical Results}
This work started with simpler environments to evaluate and understand how the GCL and GAIL algorithms worked, what hyperparameters had the most impact in the performance, and how many expert demonstrations were necessary to learn a satisfactory policy.

\begin{figure}[!htb]
    \centering
    \includegraphics[width=0.45\linewidth]{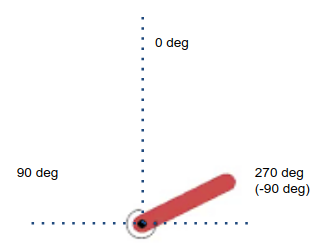}
    \caption{Illustration of the Pendulum environment. The agent controls the torque on the base of the pendulum in order to maintain it in the vertical position.}
    \label{fig:pendulum}
\end{figure}

This evaluation started by using the GCL algorithm on the OpenAI Gym Pendulum-v0 environment. The Pendulum-v0 task consists of applying a torque on the pendulum joint (one-dimensional continuous action) in order to keep it in a inverted vertical position. The observartion-space comprises three continuous features: $sin(\theta)$, $cos(\theta)$, and $\dot{\theta}$, where $\theta$ is the pendulum angle measured from the inverted vertical position, as seen in Figure \ref{fig:pendulum}.

\begin{figure}[!htb]
    \centering
    \includegraphics[width=0.65\linewidth]{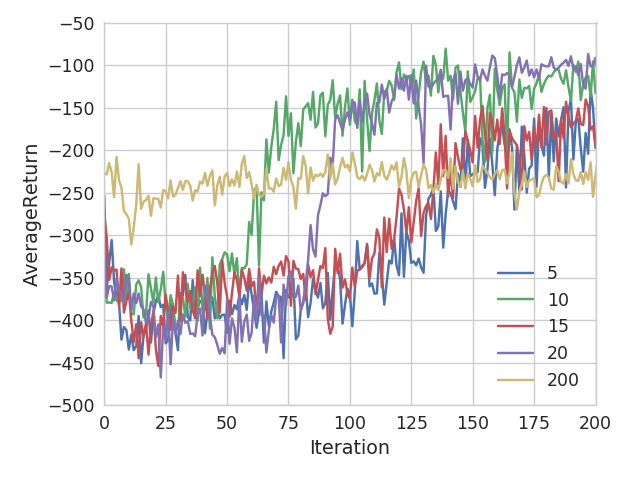}
    \caption{Performance in terms of original task reward for different number of expert trajectories for Pendulum GCL.}
    \label{fig:pendulum_gcl}
\end{figure}

In order to evaluate the impact of the quantity and quality of the expert demonstrations to initialize the GCL algorithm, we ran GCL on Pendulum-v0 initializing it with 5, 10, 15, and 20 close-to-optimal demonstrations and a final run with 200 demonstrations, in which half of them were close-to-optimal and the other half was sub-optimal. Results in terms of average return can be seen in Figure \ref{fig:pendulum_gcl} when trained for 200 learning iterations. The algorithm performed the best with 10 close-to-optimal demonstration, showing that more demonstrations does not necessarily translates to better task performance.

\begin{figure}[!htb]
    \centering
    \includegraphics[width=0.65\linewidth]{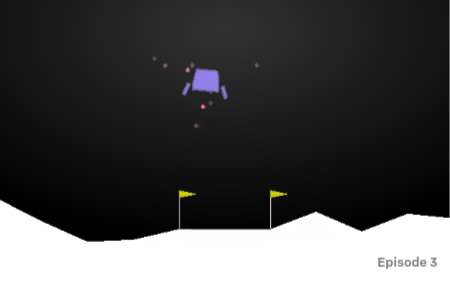}
    \caption{Illustration of the LunarLanderContinuous environment. The agent controls the main and side thrusts of the vehicle in order to land safely between he flags.}
    \label{fig:lunar_lander}
\end{figure}

The GAIL algorithm was evaluated using the OpenAI Gym Lunar Lander Continuous environment. The LunarLanderContinuous-v2 task consists of controlling the main and side thrusts (two-dimensional continuous action) in order to safely land a spacecraft between to flags. The observartion-space comprises eight continuous and discrete features related to the spacecraft's position and attitude and contact to the ground, as illustrated in Figure \ref{fig:lunar_lander}.

\begin{figure}[!htb]
    \centering
    \includegraphics[width=0.65\linewidth]{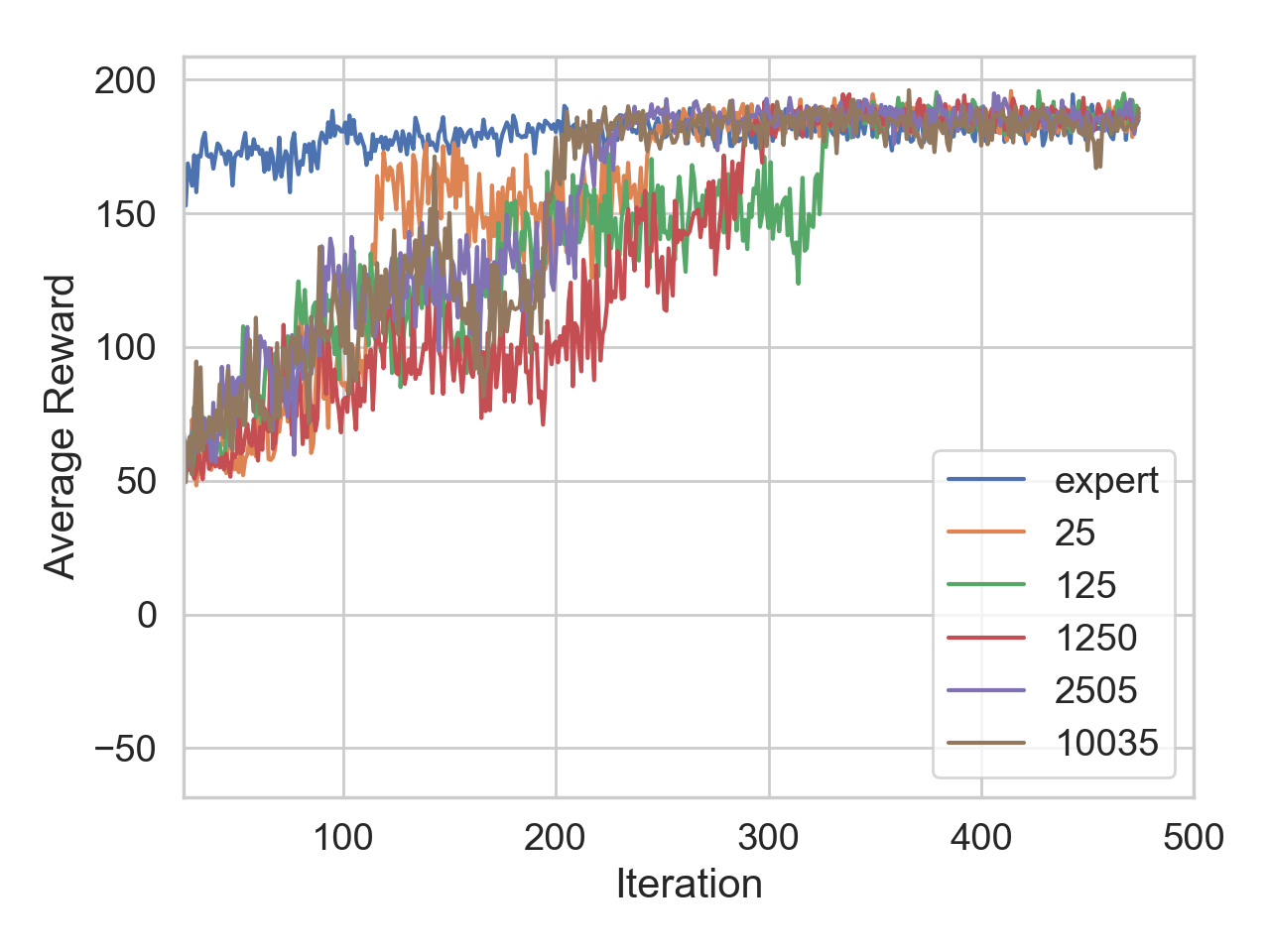}
    \caption{Performance in terms of original task reward for different number of expert trajectories for LunarLanderContinuous GAIL.}
    \label{fig:gail_lunar}
\end{figure}

\begin{figure}[!h]
    \centering
    \includegraphics[width=0.65\linewidth]{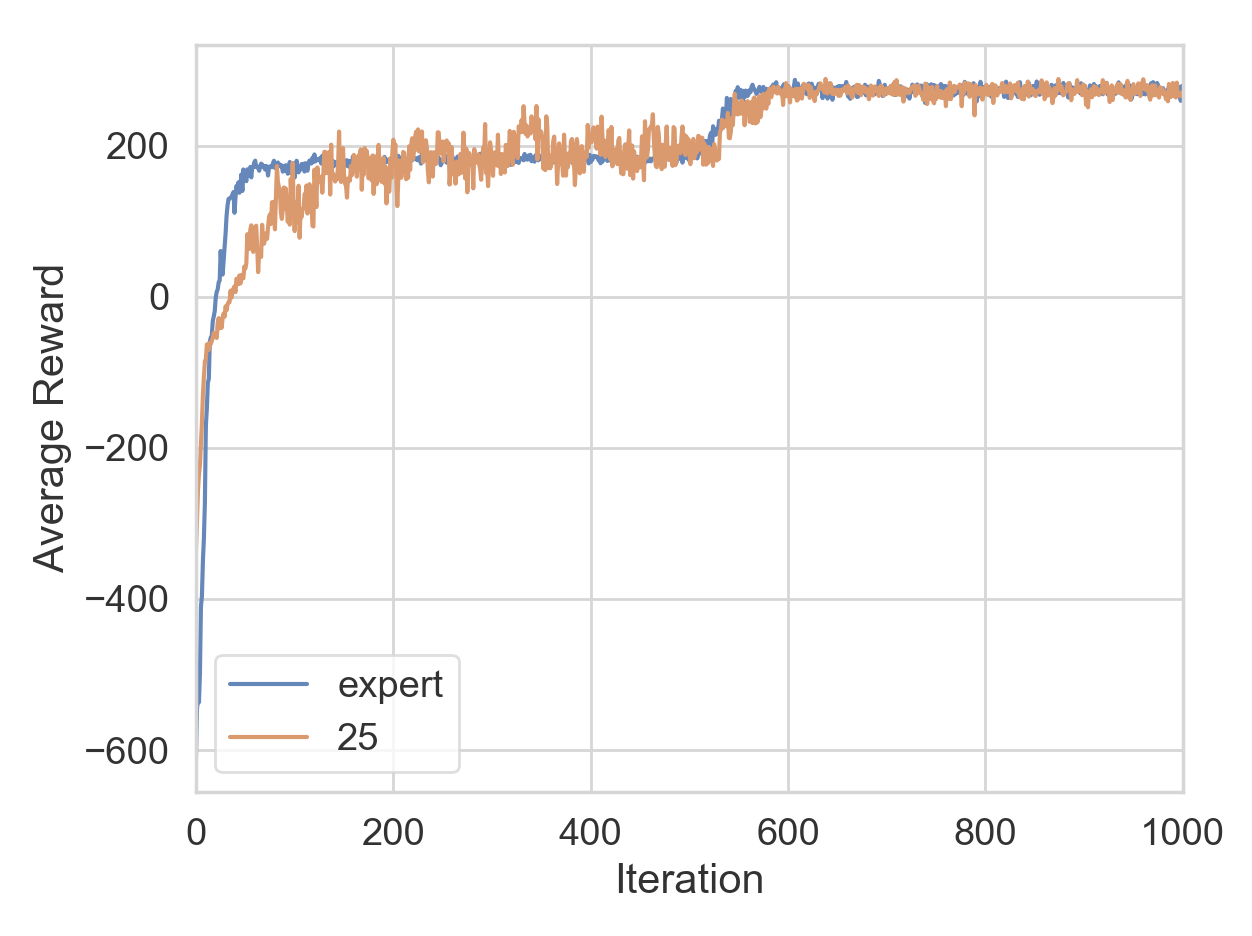}
    \caption{Performance in terms of original task reward for the best performing number of expert trajectories for LunarLanderContinuous GAIL.}
    \label{fig:gail_lunar_final}
\end{figure}

In order to evaluate the impact of the quantity and quality of the expert demonstrations to initialize the GAIL algorithm, we ran GAIL on LunarLanderContinuous-v2 initializing it with 25, 125, 1,250, 2,505, and 10,035 close-to-optimal demonstrations. Results in terms of average return can be seen in Figure \ref{fig:gail_lunar} when trained for 500 learning iterations. The algorithm performed the best with 25 close-to-optimal demonstration, showing one more time that more demonstrations does not necessarily translates to better task performance. Figure \ref{fig:gail_lunar_final} shows the performance of this better performing hyperparameter when trained for 1,000 learning iterations. The complete list of hyperparameters can be see in Table \ref{tab:lunar_hyper1}.

\subsection{Unmanned Aerial Vehicle Results}
\begin{figure}[!h]
    \centering
    \includegraphics[width=0.65\linewidth]{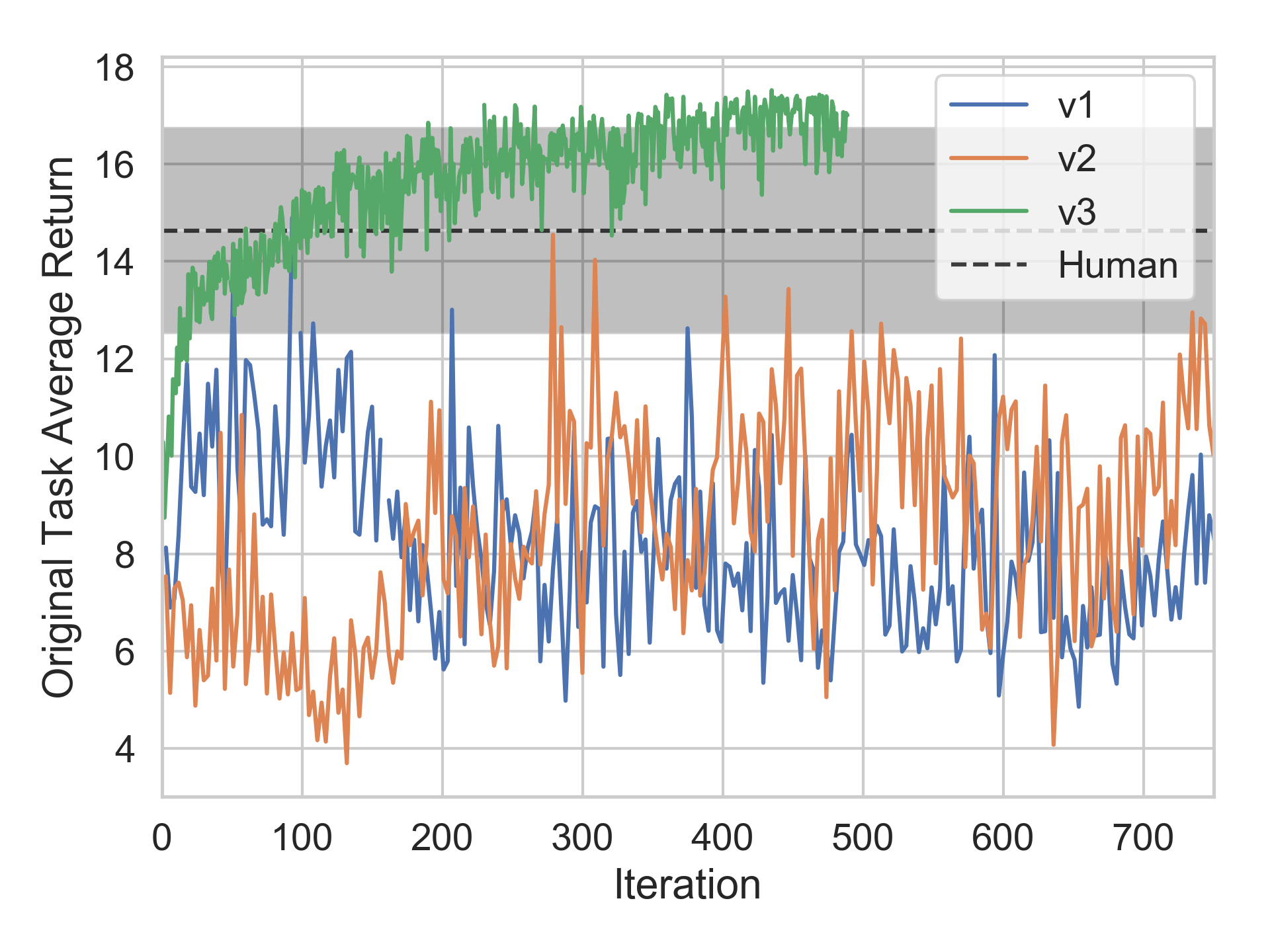}
    \caption{Performance in terms of original task reward for different hyperparameters for AirSim GAIL.}
    \label{fig:airsim_gail_final}
\end{figure}

\begin{figure}[!h]
    \centering
    \includegraphics[width=0.65\linewidth]{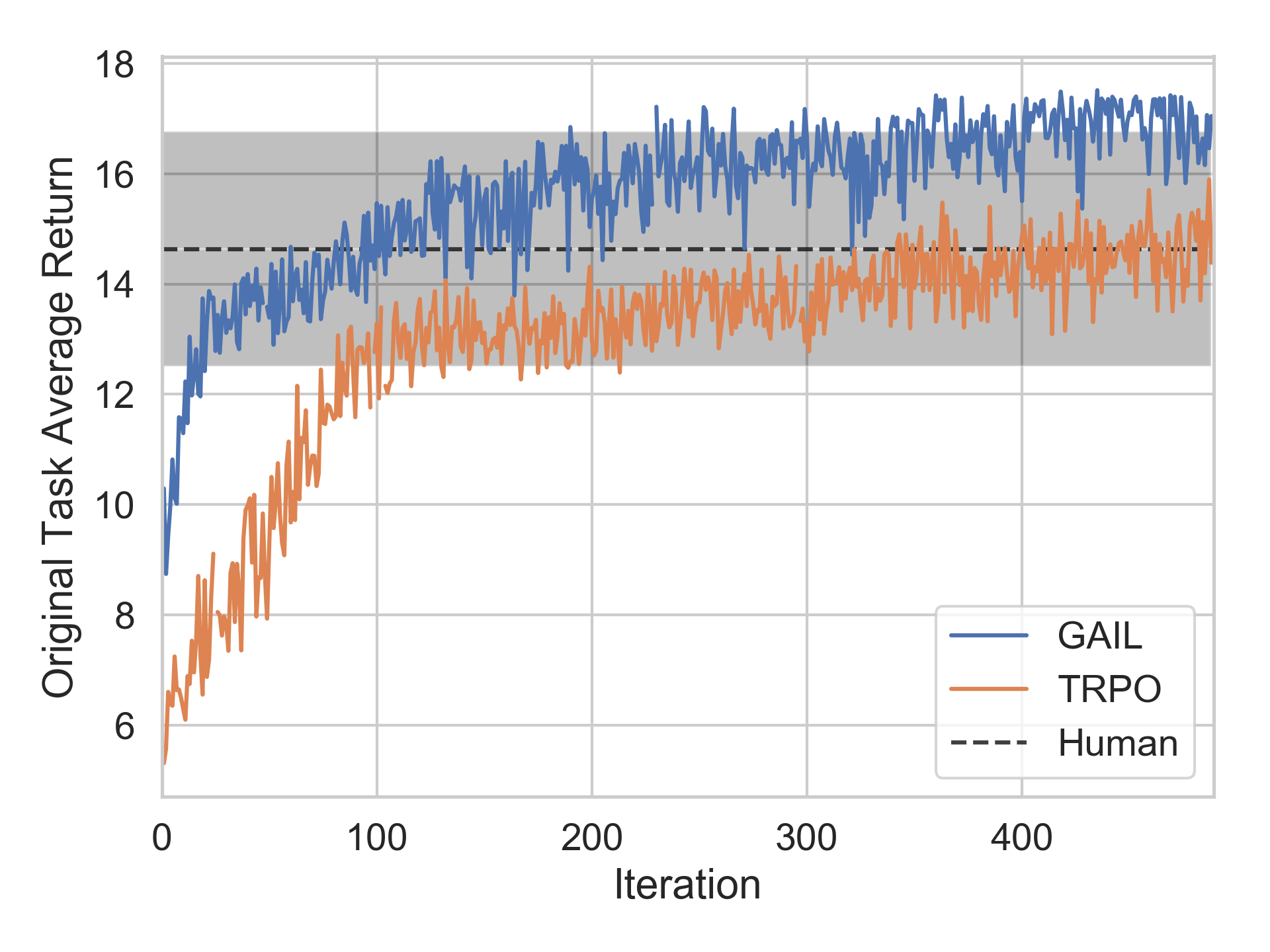}
    \caption{Performance in terms of original task reward for the best performing hyperparameters for AirSim GAIL.}
    \label{fig:airsim_gail_detail_final}
\end{figure}

After validation of the GAIL algorithm on a similar but less complex task, the Lunar Lander Continuous, we applied GAIL to solve the UAV landing task in Microsoft AirSim. A human pilot performed the landing task for approximately 30 minutes (or 100 demonstrations). This data was saved and used to initialize the GAIL algorithm. Results in terms of original task average return during the hyperparameter tuning phase can be seen in Figure \ref{fig:airsim_gail_final} when trained for 500 learning iterations. Figure \ref{fig:airsim_gail_detail_final} details the performance of the better performing hyperparameter when trained for 500 learning iterations. For the UAV case, GAIL was able to surpass human mean performance after about 100 training iterations and converged to the human performance upper-bound after about 400 learning iterations. We also compared GAIL to a pure RL approach using TRPO (Trust Region Policy Optimization), as also seen in Figure \ref{fig:airsim_gail_detail_final}. TRPO only reached human performance by training iteration 500. The complete list of hyperparameters can be see in Table \ref{tab:uav_hyper} and training statistics in Table \ref{tab:uav_training}.

\section{Summary}
\label{sec:conclusions}
This chapter presented the application of IRL, specifically GCL and GAIL, to learn from previous task demonstration and solve continuous observation and action-space problems. Our approach using GCL and GAILwas initially validated in less complex environments, as the OpenAI Gym Pendulum-v0 and LunarLanderContinuous-v2, where we were able to solve the tasks with less than 200 learning iterations using as little as 10 demonstrations. The same GAIL approach was applied to a UAV landing task, where we were able to surpass human mean performance after 100 learning iterations. The same performance using pure RL, specifically TRPO, takes 400 learning iterations, or 400\% more iterations. 
GAIL, as demonstrated for LunarLander (see Figure \ref{fig:gail_lunar}) and UAV (see Figures \ref{fig:airsim_gail_final} and \ref{fig:airsim_gail_detail_final}), is sample efficient in terms of number of expert trajectories needed it to reach a policy which is closer to the expert policy.

\subsection{Training Hyperparameters}\label{ss:irl_params}

\begin{table}[!h]
\centering
\caption{Hyperparemeter search for LunarLanderContinuous GAIL}
\label{tab:lunar_hyper1}
\begin{tabular}{l|l|l|l}
No. Expert Traj. & 25 & 125 & 1250 \\ \cline{1-1}
Hyperparameter & & & \\ \hline
Policy NN & 20, 20 & 30, 30 & 50, 50, 10 \\
Batch Size & 4000 & 7000 & 7000 \\
Max Path Length & 200 & 200 & 200 \\
Discrim. Iters & 100 & 100 & 100 \\
Discount Factor & 0.99 & 0.99  & 0.99 \\ 
Policy NN     & 100, 100, 20           & 200, 200, 20            & 20, 20 \\
Batch Size & 7000    & 7000     & 7000   \\
Max Path Length           & 200     & 200      & 200    \\
Discrim. Iters & 100     & 100      & 0      \\
Discount Factor           & 0.99    & 0.99     & 0.99  
\end{tabular}
\end{table}

\begin{table}[!h]
\centering
\caption{Hyperparemeter search for AirSim GAIL}
\label{tab:uav_hyper}
\begin{tabular}{l|l|l|l}
Hyperparameter   & v1 & v2 & v3 \\ \hline
Policy NN     & 50, 50, 10  & 32, 32      & 32, 32      \\
Batch Size & 60          & 60          & 1200        \\
Max Path Length           & 60          & 60          & 60          \\
Discrim. Iters & 100         & 100         & 100         \\
Discount Factor           & 0.99        & 0.99        & 0.99       
\end{tabular}
\end{table}

\begin{table}[!h]
\centering
\caption{Training Statistics for AirSim GAIL}
\label{tab:uav_training}
\begin{tabular}{l|l}
Training Time  & 48.12 hours     \\
Trajectories Generated        & 10,429          \\
GAIL Iterations               & 489             \\
Timer per GAIL Iterations     & $\sim$6 minutes \\
Time per Trajectory (average) & 16.6 seconds    \\
Human Trajectories            & 100             \\
Human Time     & 28 minutes     
\end{tabular}
\end{table}






\chapter[CYBER-HUMAN APPROACH FOR LEARNING HUMAN INTENTION AND SHAPE ROBOTIC BEHAVIOR BASED ON TASK DEMONSTRATION]{CYBER-HUMAN APPROACH FOR LEARNING HUMAN INTENTION AND SHAPE ROBOTIC BEHAVIOR BASED ON TASK DEMONSTRATION\footnote{Adapted with permission from ``Cyber-human approach for learning human intention and shape robotic behavior based on task demonstration'', by Vinicius G. Goecks, Gregory M. Gremillion, Hannah C. Lehman, and William D. Nothwang, presented at the 2018 International Joint Conference on Neural Networks (IJCNN) \cite{goecks2018cyber}, Copyright 2018 by the Institute of Electrical and Electronics Engineers.}}\label{ch:cybersteer}

Recent developments in artificial intelligence enabled training of autonomous robots without human supervision. Even without human supervision during training, current models have yet to be human-engineered and have neither guarantees to match human expectation nor perform within safety bounds.
This paper proposes \nameProj to leverage human-robot interaction and align goals between humans and robotic intelligent agents. Based on human demonstration of the task, \nameProj learns an intrinsic reward function used by the human demonstrator to pursue the goal of the task. The learned intrinsic human function shapes the robotic behavior during training through deep reinforcement learning algorithms, removing the need for environment-dependent or hand-engineered reward signal.
Two different hypotheses were tested, both using non-expert human operators for initial demonstration of a given task or desired behavior: one training a deep neural network to classify human-like behavior and other training a behavior cloning deep neural network to suggest actions.
In this experiment, \nameProj was tested in a high-fidelity unmanned air system simulation environment, Microsoft AirSim. The simulated aerial robot performed collision avoidance through a clustered forest environment using forward-looking depth sensing.
The performance of \nameProj is compared to behavior cloning algorithms and reinforcement learning algorithms guided by handcrafted reward functions.
Results show that the human-learned intrinsic reward function can shape the behavior of robotic systems and have better task performance guiding reinforcement learning algorithms compared to standard human-handcrafted reward functions. 

\section{Problem Definition}
Intelligent robots have the potential to positively impact and augment human activities in various tasks and modalities. Part of this success will depend on how they are integrated and the underlying motivation that drives the behavior of these intelligent robots. To comply with this need, it is beneficial to design a framework to train and shape the behavior of intelligent robotic agents to comply with human intention.

One of the approaches to shape the behavior of robotic agents is to formulate the process as a reinforcement learning problem. The robot (called \emph{agent}) senses its surrounding environment (\emph{observation}) through onboard sensors. The robot's controller (\emph{policy}) selects that adequate control input (\emph{action}) in order to maximize a given metric of performance (\emph{reward signal}). The policy is trained based on its interaction with the environment in order to maximize the reward signal, which can be thought of as the goal of the intelligent agent.

This paper proposes the CyberSteer framework to better understand how to use human resources to train robotic agents in an environment without an explicit metric of performance (\emph{reward signal}).
The main research problem addressed by this paper is how to leverage initial human demonstration of the task to learn an intrinsic reward function used by the human to pursue the goal of the task. This learned intrinsic human function then shapes the robotic behavior during training through deep reinforcement learning (Deep RL) algorithms, aligning goals between humans and these intelligent agents.


Forming a reward signal is a challenge in real-world tasks. While games have clearly defined rewards from the game score, rewards in real-worlds tasks are currently handcrafted and hard-coded for each task based on some a priori knowledge of the possible state and clear goal. The challenge is in developing a method to translate human intention to a reward function in a framework that can be applied to arbitrary tasks.

CyberSteer provides a novel framework to integrate humans to robotic learning agents providing task demonstration and operating as an intervention mechanism to provide safety during learning and exploration. CyberSteer acts as an Intrinsic Reward Module (IRM) which provides the reward signal for the Deep RL agent.

Two different hypotheses were tested, both using non-expert human operators for initial demonstration of a given task or desired behavior. CyberSteer \#1 collected these demonstrated trajectories and trained a deep neural network to classify human-like behavior. CyberSteer \#2 trained a behavior cloning deep neural network that asynchronously ran in the background suggesting actions to the Deep RL module.

The human is not required to be an expert on the task, only to have limited knowledge of its high-level goal and be able to perform rudimentary elements of the task.

The framework is designed for real-world robotic applications where external rewards are not present and safe environment exploration is a concern. CyberSteer is tested in a high-fidelity unmanned air system (UAS) simulation environment, Microsoft AirSim. The simulated aerial robot performs collision avoidance through a clustered forest environment using forward-looking depth sensing and roll, pitch, and yaw references angle commands to the flight controller.

The proposed approach is compared to a direct behavior cloning deep neural network trained using the human demonstration dataset and a deep reinforcement learning algorithm, Deep Deterministic Policy Gradient, guided by a handcrafted reward signal. Evaluation is quantified by the alignment of the agents actions with human inputs and completion of the task. 

\section{Background} \label{sec:background}

\subsection{Preliminaries}
In reinforcement learning problems it is desired to train an agent to learn the parameters $\theta$ of a policy $\pi_{\theta}$, in order to map the environment's state vectors $\vec{s}$ (sampled from a distribution $\mathcal{S}$) to agent actions $\vec{a}$ (sampled from a distribution $\mathcal{A}$). The performance of the agent is measured by a reward signal returned by the environment (external rewards, $r_e$) and/or returned by the agent itself (intrinsic rewards, $r_i$). At each time step $t$ the reward signal can be computed as the sum of all the extrinsic and intrinsic rewards received (Equation \ref{eq:rew_step}).

\begin{equation}\label{eq:rew_step}
r_t = r_{e_t} + r_{i_t}
\end{equation}

An episode is defined as the time interval between the initial time step $t_0$ and the maximum number of time steps allowed $T$ or until a previously established objective is achieved. The total reward per episode $R$ is defined as the sum of the rewards received for each time step, as shown in Equation \ref{eq:total_rew}.

\begin{equation}\label{eq:total_rew}
R = \sum_{t=0}^{T} r_t = \sum_{t=0}^{T} (r_{e_t} + r_{i_t})
\end{equation}

The expected total reward per episode received by a policy $\pi_\theta (\vec{a}_t | \vec{s}_t)$ can be defined by Equation \ref{eq:expected_rew}:

\begin{equation}\label{eq:expected_rew}
R_{\pi_\theta} = \sum_{t=0}^{T} \E_{\vec{a}_t \sim \pi_\theta} [r_t(\vec{s}_t, \vec{a}_t)]
\end{equation}

\subsection{Behavior Cloning} \label{ssec:bc}

\emph{Behavior Cloning}, or \emph{Imitation Learning}, is defined by training a policy $\pi$ in order to replicate an expert's behavior given states and actions visited during an expert demonstration
\begin{align*}
\mathcal{D} = \left\{\vec{a}_0, \vec{s}_0, \vec{a}_1, \vec{s}_1, ... , \vec{a}_T, \vec{s}_T\right\}.
\end{align*}
This demonstration can be performed by a human supervisor, optimal controller, or virtually any other pre-trained policy $\pi$.

In the case of human demonstrations, the human expert is implicitly trying to maximize the reward function of a given task, as shown in Equation \ref{eq:imitation_rew}, where $\pi^*(\vec{a}^*_t | \vec{s}_t)$ represents the optimal policy (not necessarily known) in which the optimal action $\vec{a}^*$ is taken at state $\vec{s}$ for every time step $t$.

\begin{equation}\label{eq:imitation_rew}
\max_{\vec{a}_0,...,\vec{a}_T} \sum_{t=0}^{T} r_t(\vec{s}_t, \vec{a}_t) = \sum_{t=0}^{T} \log p (\pi^*(\vec{a}_t^* | \vec{s}_t))
\end{equation}

Defining the policy of the expert supervisor as $\pi_{sup}$ and its estimated policy as $\hat{\pi}_{sup}$, behavior cloning can be achieved through standard supervised learning (where the parameters $\theta$ of a policy $\pi_\theta$ are fit in order to minimize a loss function - such as mean squared error - as shown in Equation \ref{eq:sup_mse}) or using more advanced methods, such as Dataset Aggregation \cite{Ross2011} (DAgger, where data collected by the estimated policy is aggregated with data provided by the expert), Generative Adversarial Imitation Learning \cite{Ho2016} (where generative adversarial networks are used to fit distributions of states and actions defined by expert behavior), or Guided Cost Learning \cite{Finn2016} (where regularized neural networks and a cost learning algorithm based on policy optimization are used to learn the expert's cost function under unknown dynamics and high-dimensional continuous systems).

\begin{equation}\label{eq:sup_mse}
\hat{\pi}_{sup} = \argmin_{\pi_\theta} \sum_{t=0}^{T} ||\pi_\theta(\vec{s}_t) - \vec{a}_t ||^2
\end{equation}

\subsection{Deep Deterministic Policy Gradient}

The \emph{Deep Deterministic Policy Gradient} (DDPG) algorithm \cite{lillicrap2015continuous} is a model-free, off-policy, actor-critic algorithm that combines the \emph{Deterministic Policy Gradient} (DPG) algorithm \cite{Silver2014} to compute the policy's gradient and the insights gained by the Deep Q-Network (DQN) algorithm \cite{Mnih2013,Mnih2015a} to train the critic using Q-updates as, for example, the \emph{Experience Replay} and \emph{Target Networks}. Experience replay and target networks has been proven \cite{Mnih2013,Mnih2015a,Lin1992} to improve and stabilize policies learned using reinforcement learning. Both actor and critic are represented by deep neural networks.

The DDPG algorithm was selected for being able to process high-dimensional continuous state space, characteristic of robotic environments, and command continuous actions, similar to the human supervisor. To improve the state exploration performance in continuous spaces a noise component, sampled from a noisy process $\mathcal{N}$, is added to the agent's policy which affects the selected action.

\section{Learning Human Intention and Shaping Robotic Behavior}\label{sec:solution}

\subsection{The CyberSteer Mechanics}\label{ss:overview}

\begin{figure}[!b]
\centering
\includegraphics[width=0.5\textwidth]{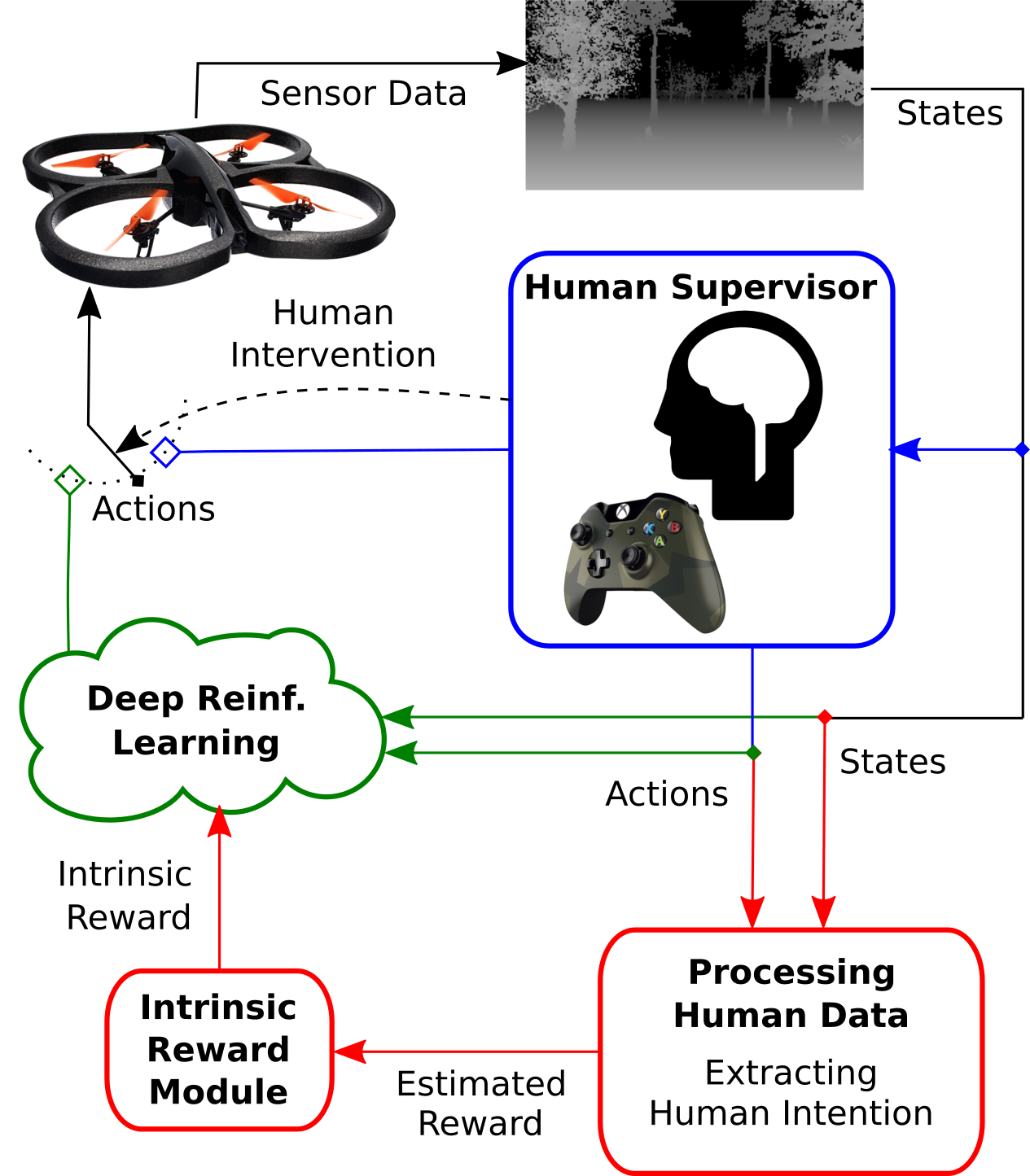}
\caption{Overall Diagram of the CyberSteer framework. Reprinted from \cite{goecks2018cyber}.}
\label{f:overall_diagram}
\end{figure}

A summarized diagram of the mechanics of CyberSteer can be seen in Figure \ref{f:overall_diagram}, which will support the explanation below.

Given an autonomous robotic agent initialized with an untrained policy (which maps the current observations to the next action), the human supervisor performs the desired task in order to collect demonstration data (observations and actions used by the human) safely performing an initial environment exploration. From this point, two hypotheses are being tested to efficiently use this human data:

\begin{enumerate}

\item \textbf{CyberSteer \#1} (Detailed in Algorithm \ref{alg:cs1}) Trains a convolutional recurrent neural network based on collected human data and random behavior to classify actions taken as being provided by human or not (human-like actions), Figure \ref{f:imit_deeprl_human}. The estimated reward $\hat{r}$ to be returned to the Deep RL algorithm is calculated based on the likelihood that the action taken is similar to the action executed by a human $p(\vec{a} = \vec{a}_{human} | (\vec{s},\vec{a}))$, as shown in Equation \ref{eq:prob_human}.

\begin{equation} \label{eq:prob_human}
\hat{r} = r_{max} (2 \cdot p(\vec{a} = \vec{a}_{human} | (\vec{s},\vec{a}) ) - 1)
\end{equation}

The estimated reward is bounded between $r_{max}$ (when $p(\vec{a} = \vec{a}_{human} | (\vec{s},\vec{a})) = 1$) and $-r_{max}$ (when $p(\vec{a} = \vec{a}_{human} | (\vec{s},\vec{a})) = 0$). CyberSteer \# 1 uses a convolutional network with two convolutional layers (32 and 16 filters, 3x3 kernels, 1 pixel stride, no padding, using rectified linear unit (ReLU) activation function and max pooling to reduced the size of the input in half after each layer), 8 LSTM recurrent units, two fully-connected layers with 64 neurons each, followed by a ReLU activation function, and the fully connected output layer, followed by a sigmoid function. \textit{Adam} algorithm \cite{Kingma2015} is used for optimization of a binary cross-entropy loss function. 

\begin{figure}[!t]
\centering
\includegraphics[width=0.5\textwidth]{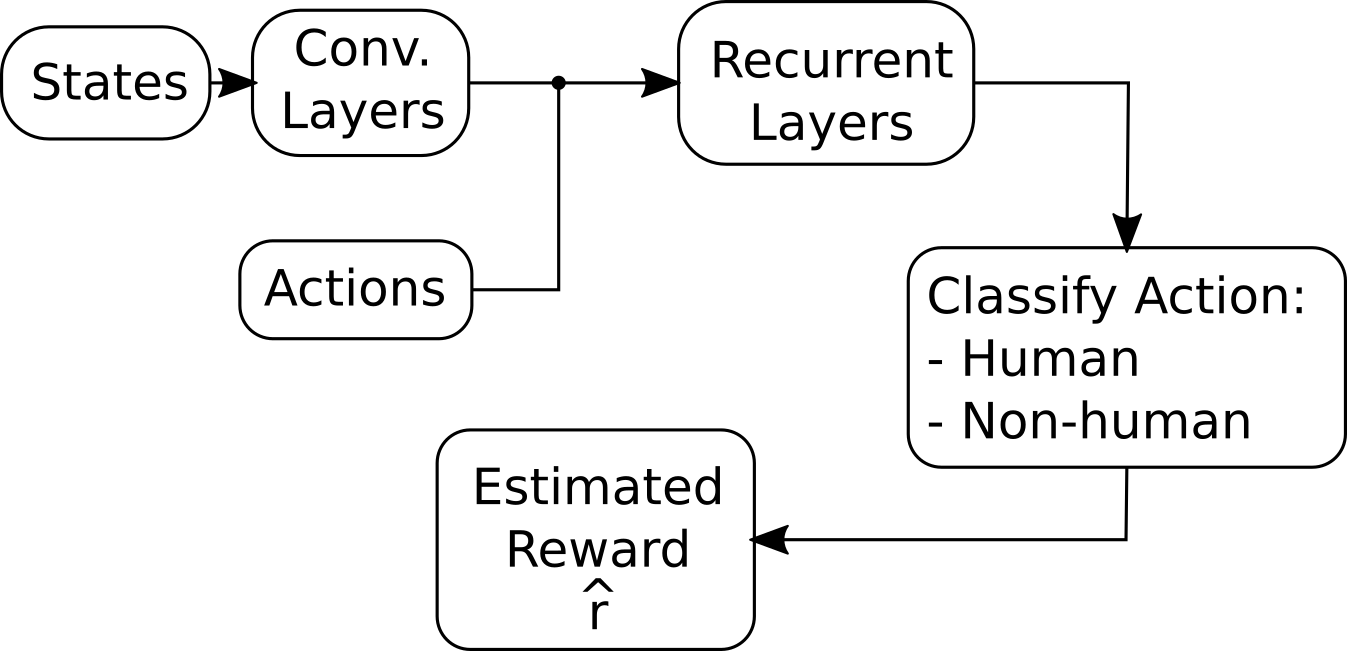}
\caption{CyberSteer \#1: Computing estimated rewards based on the likelihood of the action taken being similar to previous actions taken by the human supervisor. Reprinted from \cite{goecks2018cyber}.}
\label{f:imit_deeprl_human}
\end{figure}

\begin{algorithm}
\caption{CyberSteer \#1}
\label{alg:cs1}
\begin{algorithmic}[1]
\State {Human controls the robot and demonstrates the task to be executed. States ($\vec{s}$) and actions ($\vec{a}$) are recorded using onboard sensors and compiled as a human-controlled dataset $\mathcal{D}_{H} = \left\{\vec{a}_0, \vec{s}_0, \vec{a}_1, \vec{s}_1, ... , \vec{a}_T, \vec{s}_T\right\}$.}
\State {Robot performed magnitude-controlled random actions. States ($\vec{s}$) and actions ($\vec{a}$) are recorded using onboard sensors and compiled as a non-human-controlled dataset $\mathcal{D}_{NH}$.}
\State {Datasets $\mathcal{D}_{H}$ and $\mathcal{D}_{NH}$ are reorganized in order to have three sequences of actions stacked as one data point.}
\State {Datasets $\mathcal{D}_{H}$ and $\mathcal{D}_{NH}$ are aggregated and shuffled.}
\State {Convolutional recurrent neural network is trained using the aggregated dataset in order to classify between human and non-human behavior - the CyberSteer \#1 network.}
\State {Controls are completely transferred to robot, and its Deep Reinforcement Learning (Deep RL) module is initialized. Robot is trained:}
\While {Desired behavior is not achieved}
	\State {Robot's Deep RL module selects next action.}
    \If {Human judges robot's behavior is inadequate}
    	\State {Human intervention. Provides new demonstration.}
    	\State {Batch update CyberSteer \#1 network.}
    \Else
    	\State {CyberSteer \#1 network evaluates from 0 to 100\% how similar the action is to human behavior.}
    	\State {Estimated reward is computed based on Equation \ref{eq:prob_human} by the Intrinsic Reward Module (IRM).}
    	\State {Update Robot's Deep RL module.}
    \EndIf
\EndWhile
\end{algorithmic}
\end{algorithm}

\item \textbf{CyberSteer \#2} (Detailed in Algorithm \ref{alg:cs2}) Trains a behavior cloning neural network using convolutional and recurrent layers to suggest actions to the Deep RL algorithm that would have been taken by the human supervisor, Figure \ref{f:imit_deeprl}. The mean square error ($\vec{a}_{MSE}$) between the action suggested by the behavior cloning ($\vec{a}_{BC}$) and taken by the Deep RL algorithm ($\vec{a}_{DRL}$) is used to compute an estimated reward (Equation \ref{eq:prob_imit}, where $\vec{a}_{MSE} = (\vec{a}_{BC} - \vec{a}_{DRL})^2$) to guide the Deep RL algorithm.

\begin{equation} \label{eq:prob_imit}
\hat{r} = -r_{max}\left(1 - \sqrt{\vec{a}_{MSE}}\right)
\end{equation}

The estimated reward is bounded between $r_{max}$ (when $\vec{a}_{MSE} = 0$) and $-r_{max}$ (when $\vec{a}_{MSE} = 4$). CyberSteer \# 2 uses a convolutional network with two convolutional layers (32 and 16 filters, 3x3 kernels, 1 pixel stride, no padding, using rectified linear unit (ReLU) activation function and max pooling to reduce the size of the input in half after each layer), 8 LSTM recurrent units, one fully-connected layers with 64 neurons followed by a ReLU activation function, and the fully connected output layer followed by a sigmoid function. \textit{Adam} algorithm \cite{Kingma2015} is used for optimization of a binary cross-entropy loss function.

\begin{figure}[!t]
\centering
\includegraphics[width=0.5\textwidth]{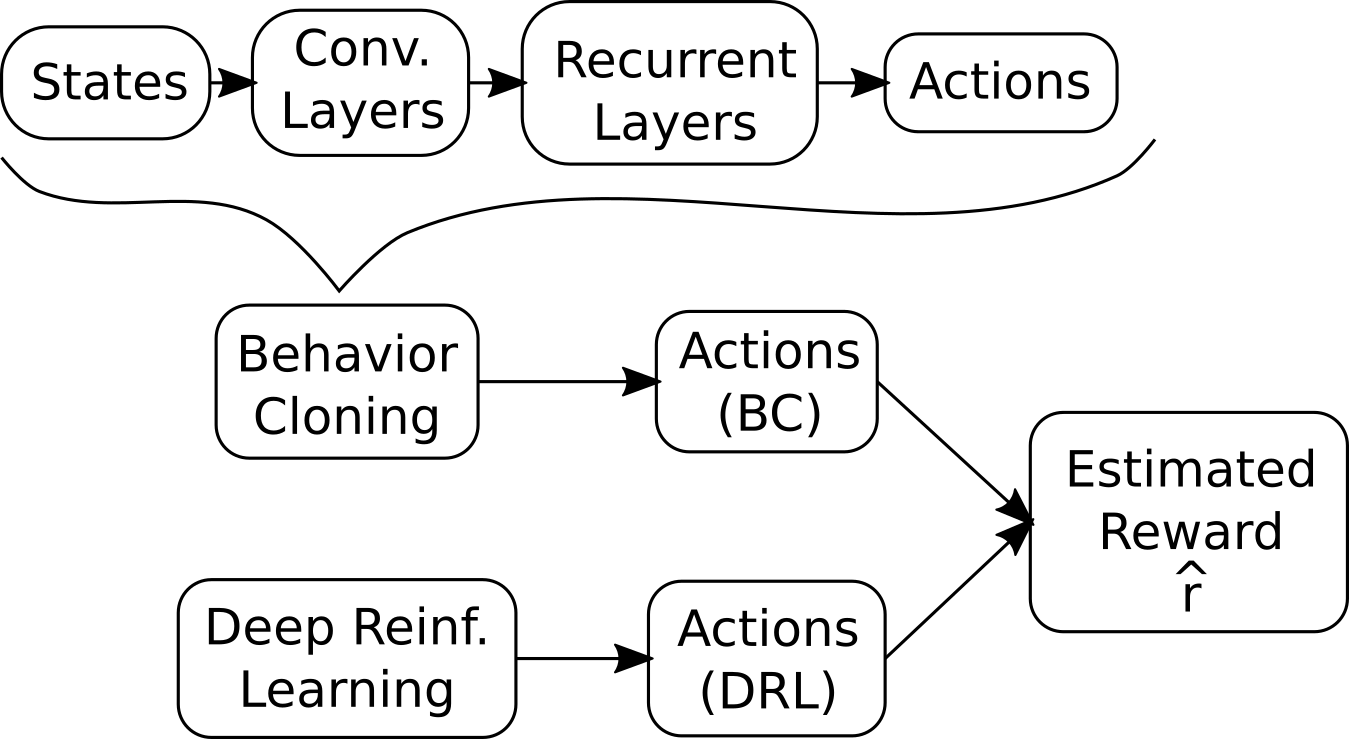}
\caption{CyberSteer \#2: Computing estimated rewards based on how similar the actions taken by the agent are when compared to a behavior cloning network trained with demonstration form the human supervisor. Reprinted from \cite{goecks2018cyber}.}
\label{f:imit_deeprl}
\end{figure}

\begin{algorithm}
\caption{CyberSteer \#2}
\label{alg:cs2}
\begin{algorithmic}[1]
\State {Human controls the robot and demonstrates the task to be executed. States ($\vec{s}$) and actions ($\vec{a}$) are recorded using onboard sensors and compiled as a human-controlled dataset $\mathcal{D}_{H} = \left\{\vec{a}_0, \vec{s}_0, \vec{a}_1, \vec{s}_1, ... , \vec{a}_T, \vec{s}_T\right\}$.}
\State {Dataset $\mathcal{D}_{H}$ is reorganized in order to have three sequences of actions stacked as one data point and dataset is shuffled.}
\State {Behavior Cloning network is trained is order to suggest actions based on past human demonstration.}
\State {Controls are completely transferred to robot, and its Deep Reinforcement Learning (Deep RL) module is initialized. Robot is trained:}
\While {Desired behavior is not achieved}
    \State {Robot's Deep RL module selects next action.}
    \If {Human judges robot's behavior is inadequate}
        \State {Human intervention. Provides new demonstration.}
        \State {Batch update CyberSteer \#2 network (Behavior Cloning).}
    \Else
        \State {CyberSteer \#2 network (Behavior Cloning) suggests action based on current robot's states.}
        \State {Estimated reward is computed based on Equation \ref{eq:prob_imit} by the Intrinsic Reward Module (IRM).}
        \State {Update Robot's Deep RL module.}
    \EndIf
\EndWhile
\end{algorithmic}
\end{algorithm}

\end{enumerate}

\subsection{Environment and Task Modeling}
\subsubsection{Unmanned Air System Simulation Environment}

The proposed approach was first developed and tested using the Microsoft AirSim platform \cite{Airsim2017} - a high-fidelity simulated environment for unmanned air systems (UAS) flight and control. This platform simulates the UAS flight dynamics, onboard inertial sensors (GPS, barometer, gyroscope, accelerometer, and magnetometer), onboard camera data (monocular RGB, ground truth depth sensor, and random image segmentation), and collision and environment physics.


The authors refer the reader to the original paper \cite{Airsim2017} for more details about the architecture, simulation parameters, accuracy of the model, and flight characteristics when compared to a real UAS flying in real-world.

\subsubsection{Collision Avoidance Scenario}

\begin{figure}[!t]
\centering
\includegraphics[width=0.70\textwidth]{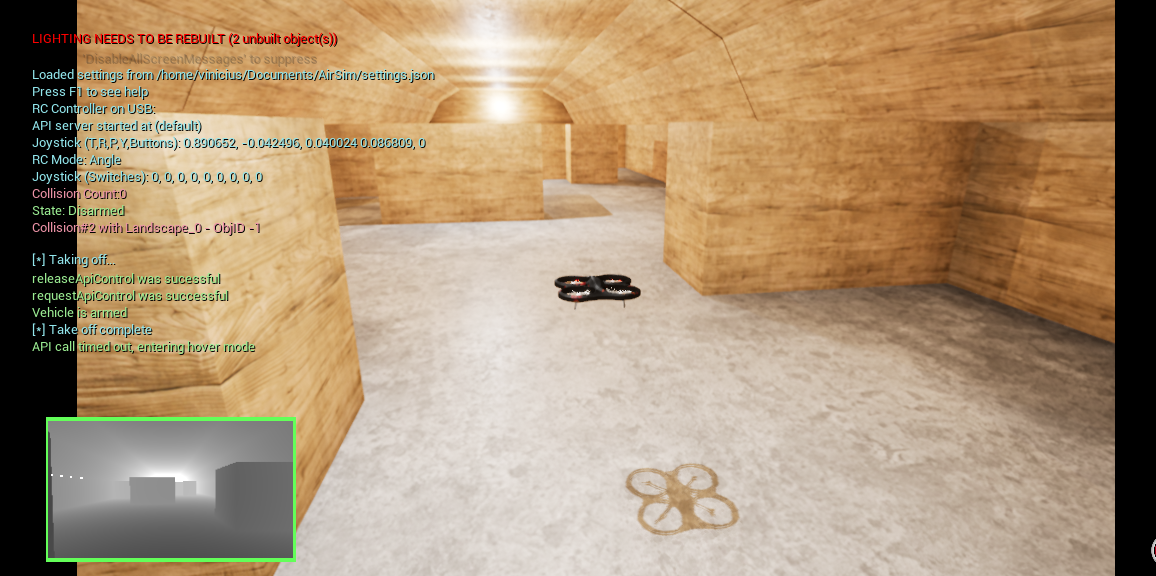}
\caption{Warehouse scenario created for the collision avoidance task using Unreal Engine for visually realistic textures and objects. Reprinted from \cite{goecks2018cyber}.}
\label{f:warehouse_scenario}
\end{figure}

The proposed algorithm was first tested on a collision avoidance scenario where the agent had to maneuver the unmanned air system using roll commands while automatic flying forward at constant altitude. A warehouse environment was created using Unreal Engine for visually realistic textures and objects. A sample of the forest scenario used can be seen in Figure \ref{f:warehouse_scenario}.

\section{Numerical Results}\label{sec:results}

Building onto the behavior cloning concepts explained in Subsection \ref{ssec:bc}, it was desired to validated that the autonomous agent would be able to learn, to some degree, to perform the demonstrated collision avoidance maneuvers using the behavior cloning network. The behavior cloning network is a critical component of the proposed framework as it is subsequently used to evaluate the actions proposed by the Deep RL algorithm for one of the hypotheses.

A non-expert human piloted the unmanned air system performing collision avoidance maneuvers for 12,000 time steps (approximately 20 minutes, performing actions at 20 Hz). The UAS was set to have constant altitude and constant forward velocity and images were collected from the onboard depth sensor and resized to 36 by 64 pixels. The human sent roll angle control inputs to steer laterally and avoid incoming obstacles. These images and control inputs were compiled as human-controlled dataset $\mathcal{D}_{H}$.

A convolutional network with three convolutional layers (32, 64, and 128 filters, 3x3 kernels, 1 pixel stride, no padding, using rectified linear unit (ReLU) activation function and max pooling to reduced the size of the input in half after each layer) and two fully-connected layers connected by a 50\% dropout layer during training was trained to imitate human performance. Three frames were stacked together to extract velocity of the scene. For the first experiment, the network was trained offline and deployed to the agent for testing. Video of the learning performance can be seen at: https://www.youtube.com/watch?v=rEbNytWz3c8.

The behavior cloning experiment was followed by the evaluation of both hypotheses. The main goal was to create an Intrinsic Reward Module (IRM) that would be able to guide a Deep RL algorithm.


CyberSteer \# 1 combined the human-controlled dataset $\mathcal{D}_{H}$ (collected for the behavior cloning network) to a dataset of images and control inputs generated at random by the robot performing the task: the non-human-controlled dataset $\mathcal{D}_{NH}$. While the agent was performing random actions, the human had total control authority to intervene and control the robot to prevent any damage to the platform. Both datasets, $\mathcal{D}_{H}$ and $\mathcal{D}_{NH}$, were aggregated and shuffled and the CyberSteer \# 1 network was trained to classify image-action pairs between human and non-human actions associating a number between 0 and 1 based on how similar the image-action pairs were compared to the collected human behavior. CyberSteer \# 1 network was deployed to the IRM in order to guide a Deep RL algorithm (Deep Deterministic Policy Gradient: DDPG) by using this similarity metric as a reward function.

CyberSteer \# 2 uses the collected human-controlled dataset $\mathcal{D}_{H}$ to train a behavior cloning network. When the Deep RL agent is controlling the robot, CyberSteer \# 2 IRM runs asynchronously in the background comparing the actions taken by the agent and what would have been taken by the CyberSteer \# 2 network (behavior cloning). Based on how similar the action taken is when compared to the suggested one, a reward signal is generated and guides the Deep RL algorithm.

A human reward baseline was established based on the maximum possible total reward for the task. In the case of CyberSteer \#1, all actions proposed by the Deep RL would be classified with 100\% confidence by the IRM as being human-like actions. In the case of CyberSteer \#2, all actions proposed by the Deep RL would 100\% match the actions suggested by the IRM running a behavior cloning network based on the initial demonstration of the task. As a baseline, the proposed solution was compared to the same Deep RL algorithm being fed by a standard human-engineered reward function returned by the environment based on the task completion, i.e., the fraction of the environment's total distance traversed.

\begin{figure}[!t]
\centering
\includegraphics[width=0.65\textwidth]{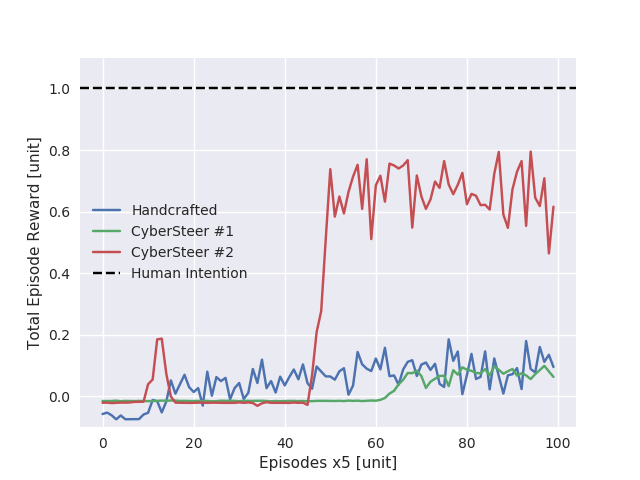}
\caption{Comparison of the proposed solutions to achieve human-like performance on the task with no feedback from the environment when compared to a baseline dependent on environment reward signals. Reprinted from \cite{goecks2018cyber}.}
\label{f:iclr_2018_learning_plot}
\end{figure}

Figure \ref{f:iclr_2018_learning_plot} shows the result of this first comparison for 500 episodes of the task being performed for about 30 seconds. CyberSteer \#2 performed seven orders of magnitude better than CyberSteer \#1 and the baseline.

\begin{figure}[!t]
\centering
\includegraphics[width=0.65\textwidth]{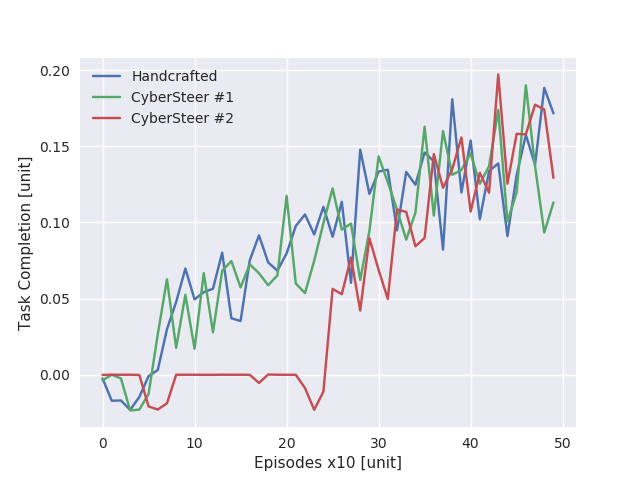}
\caption{Task completion performance of the proposed solutions with no feedback from the environment when compared to the established baseline dependent on environment reward signals. Reprinted from \cite{goecks2018cyber}.}
\label{f:iclr_2018_performance_plot}
\end{figure}

Figure \ref{f:iclr_2018_performance_plot} shows the comparison in terms of task completion using the proposed solution and the established baseline. Even though motivated by different reward signals both proposed approaches and the baseline performed similar in terms of task completion during training.

Since we are using deep neural networks as learning representation, there is still no theoretical convergence guarantee that the algorithm will lead to the optimal task behavior.

\section{Summary}\label{sec:conclusions}

This paper proposed CyberSteer to leverage human-robot interaction to train autonomous robots aligning their goals to comply with human intention. CyberSteer frames the training task as a reinforcement learning problem: an agent interacting with the environment to maximize a reward signal that represents the goal of the task. The learned human intention is instantiated as an intrinsic reward signal to shape the robotic behavior during training when combined with Deep Reinforcement Learning algorithms.

Two different hypotheses were tested to translate human intention as reward signal, both using non-expert human operators for initial demonstration of a given task or desired behavior. CyberSteer \#1 collected these demonstrated trajectories and trained a deep neural network to classify human-like behavior. CyberSteer \#2 trained a behavior cloning deep neural network that asynchronously ran in the background suggesting actions to the Deep Reinforcement Learning module.

In this experiment, CyberSteer was tested in a high-fidelity unmanned air system simulation environment, Microsoft AirSim. The simulated aerial robot performed collision avoidance through a clustered forest environment using forward-looking depth sensing.

CyberSteer \#2 initially showed seven-fold improvement in performance when compared to a human-engineered approach when training autonomous agents. The approach showed that modern Deep Reinforcement Learning algorithms can be adapted to be guided by reward signals reconstructed from human demonstration, with no need to have a human expert to handcraft a reward function. This enables the same algorithm to be deployed in different real-world applications, as long as the task can be demonstrated beforehand.

Both CyberSteer approaches were still able to perform similarly well to standard approaches dependent on environment-returned and human-engineered reward signals when driven by modern Deep Reinforcement Learning algorithms. 

\begin{figure}[!t]
\centering
\includegraphics[width=0.65\textwidth]{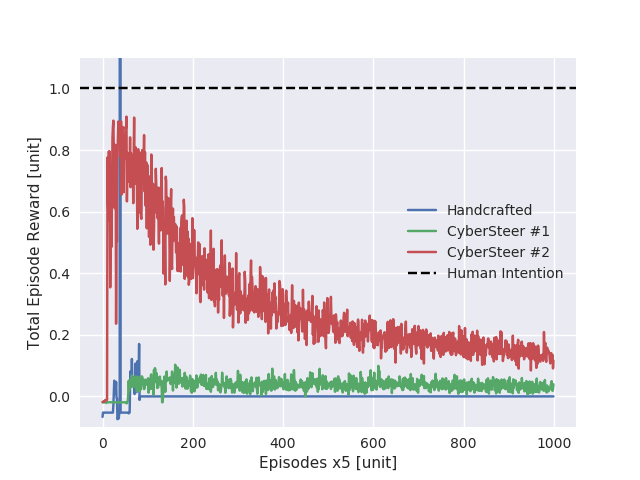}
\caption{Extended plots of the comparison of the proposed solutions to achieve human-like performance on the task with no feedback from the environment when compared to a baseline dependent on environment reward signals. Reprinted from \cite{goecks2018cyber}.}
\label{f:iclr_2018_learning_plot_ext}
\end{figure}

\begin{figure}[!t]
\centering
\includegraphics[width=0.65\textwidth]{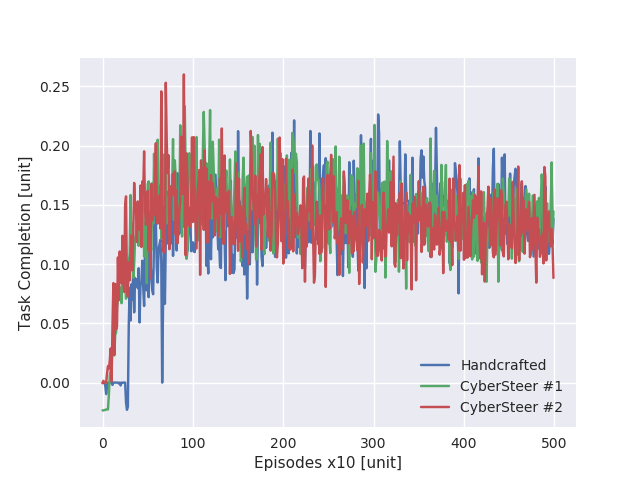}
\caption{Extended plots of the task completion performance of the proposed solutions with no feedback from the environment when compared to the established baseline dependent on environment reward signals. Reprinted from \cite{goecks2018cyber}.}
\label{f:iclr_2018_performance_plot_ext}
\end{figure}

Currently, both approaches show decrease of performance after the first 100 episodes, as seen in Figures \ref{f:iclr_2018_learning_plot_ext} and \ref{f:iclr_2018_performance_plot_ext}. This will be further investigated when performing different tasks.


\chapter[CYCLE-OF-LEARNING FOR AUTONOMOUS SYSTEMS FROM HUMAN INTERACTION]{CYCLE-OF-LEARNING FOR AUTONOMOUS SYSTEMS FROM HUMAN INTERACTION\footnote{Adapted with permission from ``Cycle-of-Learning for Autonomous Systems from Human Interaction'', by Nicholas R. Waytowich, Vinicius G. Goecks, Vernon J. Lawhern, presented at the 2018 Artificial Intelligence and Human-Interaction (AI-HRI) AAAI Fall Series Symposium \cite{Waytowich2018}, available under a Creative Commons license at the AI-HRI/2018/05 proceedings at  https://arxiv.org/html/1809.06606.}}\label{ch:col}

We discuss different types of human-robot interaction paradigms in the context of training end-to-end reinforcement learning algorithms. We provide a taxonomy to categorize the types of human interaction and present our Cycle-of-Learning framework for autonomous systems that combines different human-interaction modalities with reinforcement learning. Two key concepts provided by our Cycle-of-Learning framework are how it handles the integration of the different human-interaction modalities (demonstration, intervention, and evaluation) and how to define the switching criteria between them. 

\section{Problem Definition}

Reinforcement learning (RL) has been successfully applied to solve challenging problems from playing video games to robotics. In simple scenarios, a reward function or model of the environment is typically available to the RL algorithm, and standard RL techniques can be applied. In real-world physical environments where reward functions are not available or are too intractable to design by hand, these standard RL methods often tend to fail. Using humans to train robotic systems is a natural way to overcome these burdens. 

There have been many examples in the field of human-robot interaction where human interaction is used to train autonomous systems in the context of end-to-end reinforcement learning. For example, imitation learning is an approach where demonstrations of a task provided from a human are used to initially train an autonomous system to learn a policy that ``imitates'' the actions of the human. A recent approach termed Human Centered Reinforcement learning (HCRL) trains autonomous systems using positive and negative feedback provided from a human trainer, showing promising strides for learning policies in the absence of a reward function. Approaches that learn control policies jointly with humans and autonomous systems acting together in a shared autonomy framework have also been developed.

Each of these human-interaction approaches, which have their own unique advantages and disadvantages, have mostly been utilized in isolation without taking into consideration the varying degrees of human involvement; the human is a valuable, but finite, source of information that can communicate information in many ways. A key question to consider is not only \textit{what} information the human should convey, but \textit{how} this information should be conveyed. A method for combining these different paradigms is needed to enable efficient learning from human interaction. 

In this paper, we present a new conceptual framework for using humans to train autonomous systems called the Cycle-of-Learning for Autonomous Systems from Human Interaction. This framework fuses together different human-interaction modalities into a single learning paradigm inspired by how humans teach other humans new tasks. We believe this intuitive concept should be employed whenever humans are interacting with autonomous systems. Our contributions in this paper are as follows: we first describe a taxonomy of learning from human interaction (with corresponding literature) and list the advantages and disadvantages of each learning modality. We then describe the Cycle-of-Learning (Figure~\ref{fig:diagram}), our conceptual framework for intuitively combining these interaction modalities for efficient learning for autonomous systems. We describe the different stages of our framework in the context of the degree of human involvement (and, conversely, the amount of autonomy of the system). We conclude with potential avenues for future research.   


\section{Types of Learning from Human Interaction}
There are many different ways a human may be used to train an autonomous system to perform a task, each of which can be broadly categorized by the modality of human-interaction. Here we present a taxonomy that categorizes most types of human interactions into one of three methodologies: Learning from human demonstrations (LFD), learning from human interventions (LFI) and learning from human evaluations (LFE). This taxonomy is partitioned based on the amount of control that the human or Autonomous system has during the learning process. In LFD the human is providing demonstrations and is in full control. In LFI, where the human occasionally intervenes, both the human and autonomous system share control. In LFE, where the human is providing evaluative feedback as the autonomous system performs a task, the autonomous systems is in control. We describe rec   ent research efforts for each of these categories and present their relative advantages and disadvantages. 

\subsection{Learning from Human Demonstrations}
Learning from human Demonstrations (LFD) is from a broad class of techniques called Imitation Learning (IL) where the aim is to train autonomous systems to mimic human behavior in a given task. In this interaction paradigm, the human takes the role of a demonstrator to provide examples of the task in the terms of sequences of states and actions. Using these demonstrations, the autonomous system learns a policy (a mapping from states to actions) that mimics the human demonstrations. 

There are many empirical successes of using imitation learning to train autonomous systems. For self-driving cars, Bojarski et al. successfully used IL to train a policy that mapped from front-facing camera images to steering wheel commands using around one hundred hours of human driving data \cite{Bojarski2016}. A similar approach was taken to train a small unmanned air system (sUAS) to navigate through forest trails, but in this case data was collected by a human following the trail \cite{Giusti2015}.

These results highlight both advantages and disadvantages of imitation learning approaches. IL can be used end-to-end to learn a new task, overcoming initial exploration of randomly-initialized learning algorithms while having better convergence properties because it trains on static datasets. However, the major drawback is that policy performance and generalization rely on the diversity and quality  and size of the training dataset --- often requiring large amounts of demonstrated behavior. 

Inverse Reinforcement Learning (IRL), also known as apprenticeship learning, is another form of learning from human demonstrations where a reward function is learned based on task demonstration. The idea is that during the demonstration, the human is using a policy that is optimal with respect to some internal reward function.
However, in IRL, the quality of the reward function learned is still dependent on the demonstration performance. In addition, IRL is fundamentally underdefined (degenerated), in a sense that different reward functions can lead to the same behavior \cite{Ng2000,Finn2016}. IRL focuses on learning the reward function, which is believed to be a more robust and transferable definition of the task when compared to the policy learned \cite{Ng2000}.

\subsection{Learning from Human Interventions}
Learning From human Interventions (LFI) is a much less explored method of human interaction for training autonomous systems. Simply put, the human acts as an overseer and intervenes (i.e. takes control) when the autonomous system is about to enter a catastrophic state. This is especially important for training embodied autonomous systems, such as quadrotors or ground robots, that are learning and interacting in a real environment where sub-optimal actions could lead to damage to the systems itself or the surrounding environment. Recently, this concept was formalized in a framework called learning from Human Interventions for safe Reinforcement Learning (HIRL) \cite{Saunders2017}. 
In HIRL, when the human observes the agent perform sub-optimal actions that could lead to a failure state, the human ``intervenes'' by blocking the current action and providing an alternative action. Additionally, a negative reward signal is given to the AI system during the intervention to learn from. 

By using humans to perform an initial state exploration and prevent catastrophic actions, LFI is a promising approach to solve the exploitation-exploration dilemma and increase safety in RL. Additionally, there is evidence that shared autonomy scenarios can also improve human performance \cite{reddy2018shared}. However, LFI by itself does not scale to more complex environments where millions of observations are required for successful learning, increasing the required amount of human supervision \cite{Saunders2017}).

\subsection{Learning from Human Evaluations}
Another human-interaction modality that we consider is Learning From human Evaluations (LFE). In LFE, the human acts as a supervisor and provides real-time evaluations (or critiques) to interactively shape the behavior of the autonomous system.  There are several different approaches for how to provide the human feedback for LFE. One of the simplest approaches is fitting a reward function based on binarized feedback; for example, ``good'' vs ``bad'' actions, indicating positive and negative reward, respectively. Existing frameworks that use this approach include TAMER \cite{knox2009interactively,Warnell2018} and COACH \cite{MacGlashan2017}. Another approach asks the human to rank-order a given set of \textit{trajectories}, or short movie clips representing a sequence of state-action pairs, and learning a reward function in a relational manner based on human preferences \cite{Christiano2017}. Both approaches have been shown to be effective across a variety of domains and illustrates the utility of using human-centered reward shaping for shaping the policy of autonomous systems.

An advantage of LFE techniques is that they do not require the human to be able to perform demonstrations and only require an understanding of the task goal. However, if the time-scale of the autonomous system is faster than human reaction time, then it can be challenging for the autonomous system to attribute which actions correspond to the provided feedback.  In addition, the human reward signals are generally non-stationary and policy-dependent, i.e.: what was a good action in the past may not be a good action in the present depending on the humans perception of the autonomous system's policy.

\section{The Cycle-of-Learning Concept}

The Cycle-of-Learning is a framework for training autonomous systems through human interaction that is based on the intuition on how a human would teach another human to perform a new task. For example, teachers conveying new concepts to their students proceed first by demonstrating the concept, intervening as needed while students are learning the concept, then providing critique after students have started to gain mastery of the concept. This process is repeated as new concepts are introduced. While extensive research has been conducted into each of these stages separately in the context of machine learning and robotics, to the best of our knowledge, a model incorporating each of these aspects into one learning framework has yet to be proposed. We believe such a framework will be important to fielding adaptable autonomous systems that can be trained on-the-fly to perform new behaviors depending on the task at hand, in a manner that does not require expert programming. 


\begin{algorithm}[!tb]
\caption{Cycle-of-Learning Framework}\label{alg:col}
\begin{algorithmic}[1]

\Procedure{Learning from Demonstration}{}
    \While {$SwitchingFunction_{LFD}:$}
        \State Collect human demonstration data $D_H$
        \State Train imitation learning policy $\pi_{\theta D_H}$
        \State Learn human reward function $R_H$ from $D_H$
    \EndWhile
\EndProcedure

\Procedure{Learning from Intervention}{}
    \While {$SwitchingFunction_{LFI}:$}
        \State Autonomous system performs the task
        \If {Human intervenes}:
            \State Collect human intervention data $D_I$
            \State Aggregate $D_H \gets D_H \cup D_I$
            \State Update imitation learning policy $\pi_{\theta D_H}$
            \State Update human reward function $R_H$
        \EndIf
        \State Compute intervention reward $R_I$
        \State Train critic $Q_{H}$ using $R_I$ and TD error
        \State Update policy $\pi_{\theta D_H}$ using policy gradient
    \EndWhile
\EndProcedure

\Procedure{Learning from Evaluation}{}
    \While {$SwitchingFunction_{LFE}:$}
        \State Collect human evaluation reward $r_H$
        \State Update critic $Q_{H}$ using $r_H$ and TD error
        \State Update policy $\pi_{\theta D_H}$ using policy gradient
        \State Update human reward function $R_H$ with $r_H$
    \EndWhile
\EndProcedure

\Procedure{Reinforcement Learning}{}
    \While {$SwitchingFunction_{RL}:$}
        \State Autonomous system performs the task
        \State Compute rewards using $R_H$
        \State Update critic $Q_{H}$ using $R_H$ and TD error
        \State Update policy $\pi_{\theta D_H}$ using policy gradient
    \EndWhile
\EndProcedure

\end{algorithmic}
\end{algorithm}

Under the proposed Cycle-of-Learning framework (Figure \ref{fig:diagram}), we start with LFD where a human would be asked to provide several demonstrations of the task. This demonstration data (observations received and actions taken) constitute the initial human dataset $D_H$. The dataset $D_H$ feeds an imitation learning algorithm to be trained via supervised learning, resulting in the policy $\pi_{\theta D_H}$. In parallel to the policy training, the dataset $D_H$ is used by an Inverse Reinforcement Learning (IRL) algorithm to infer the reward function $R_H$ used by the human while demonstrating the task (Algorithm \ref{alg:col}, line 1).

On LFI (Algorithm \ref{alg:col}, line 6) the autonomous system performs the task according to the policy $\pi_{\theta D_H}$. During the task the human is able to intervene by taking over the control of the autonomous system, perhaps to avoid catastrophic failure, and provides more demonstrations during the intervention. This new intervention dataset $D_I$ is aggregated to the previous human dataset $D_H$. Using this augmented dataset, the policy $\pi_{\theta D_H}$ and the reward model $R_H$ are updated. An intervention reward $R_I$ is computed based on the degree of the intervention. The reward signal $R_I$ and the temporal-difference (TD) error associated with it are used to train a value function $Q_H$ (the critic) and evaluate the actions taken by the actor. At this point, the policy $\pi_{\theta D_H}$ is updated using actor-critic policy gradient methods. 

After the human demonstration and intervention stages, the human assumes the role of a supervisor who evaluates the autonomous system actions through a reward signal $r_H$ --- Learning from Evaluation (LFE, Algorithm \ref{alg:col}, line 17). Similarly to the LFI stage, the reward signal $r_I$ and the TD error associated with it are used to update the critic $Q_H$ and the policy $\pi_{\theta D_H}$. The reward model $R_H$ is also updated according to the signal $r_H$ plus the observations and actions associated with it.

The final stage is pure Reinforcement Learning (RL). The autonomous system performs the task and its performance is evaluated using the learned reward model $R_H$ (Algorithm \ref{alg:col}, line 23). Similar to the LFI and LFE stages, the reward signal $R_H$ and the TD error associated with it are used to update the critic $Q_H$ and the policy $\pi_{\theta D_H}$. This sequential process is repeated as new tasks are introduced. 

\subsection{Integrating and Switching Between Human Interaction Modalities}
Two key concepts of the Cycle-of-Learning framework are how to handle the integration of the learned models from the different interaction modalities (demonstration, intervention, and evaluation) and how to define the criteria to switch between them. First, to integrate the different interaction modalities, we propose using an actor-critic architecture \cite{Sutton2018}: initially training only the actor, and later adding the critic. Training the actor first allows the framework to leverage the initial demonstration and intervention provided by the human. The critic is then trained as the human assumes the role of supervisor. After enough human demonstration data has been collected we can infer a reward function through IRL. At the end, the actor and critic are combined on a standard actor-critic reinforcement learning architecture driven by the learned reward model.

Second, we propose different concepts to define a criteria to switch between interaction modalities: performance metrics, data modality limitation, and advantage functions.
\textit{Performance metrics: }A pre-defined performance metric can be used to indicate when to switch modalities once the policy reaches a certain level. Alternatively, the human interacting with the system could manually switch between different interaction modalities as s/he observes that the autonomous system performance is not increasing.
\textit{Data modality limitations: }Depending on the task, there can be a limited amount of demonstration, intervention, or evaluation that can be provided by humans. In this case, the framework switches between modalities according to data availability.
\textit{Advantage functions: }After training the reward model $R_H$, advantages $A(s,a)$ (the difference between the state-action value function $Q(s,a)$ and the state value function $V(s)$, which compares the expected return of a given state-action pair to the expected return on that state) can be computed and used for expected return comparison between human and autonomous systems actions. With this information, the framework could switch interaction modalities whenever the advantage function of the autonomous system surpasses the advantage function of the human.  These, as well as other potential concepts for modality switching, need to be further investigated and can be adapted to meet task requirements.

\section{Summary}

This paper presents the Cycle-of-Learning framework, envisioning the integration between different human-interaction modalities and reinforcement learning algorithms in an efficient manner. The main contributions of this work are (1) the formalization of the underlying learning architecture --- first leveraging human demonstrations and interventions to train an actor policy and reward model, then gradually moving to training a critic and fine-tuning the reward model based on the same interventions and additional evaluations, to finally combining these different parts on an actor-critic architecture driven by the learned reward model and optimized by a reinforcement learning algorithm --- and (2) the switching between these human-interaction modalities based on performance metrics, data modality limitations, and/or advantage functions.

We believe the proposed Cycle-of-Learning framework is most suitable for robotic applications, where both human and autonomous system resources are valuable and finite. As future work, it is planned to demonstrate these techniques on a human-sUAS (small unmanned air system) scenario.

\chapter[EFFICIENTLY COMBINING HUMAN DEMONSTRATIONS AND INTERVENTIONS FOR SAFE TRAINING OF AUTONOMOUS SYSTEMS IN REAL-TIME]{EFFICIENTLY COMBINING HUMAN DEMONSTRATIONS AND INTERVENTIONS FOR SAFE TRAINING OF AUTONOMOUS SYSTEMS IN REAL-TIME\footnote{Adapted with permission from ``Efficiently combining human demonstrations and interventions for safe training of autonomous systems in real time'', by Vinicius G. Goecks, Gregory M. Gremillion, Vernon J. Lawhern, John Valasek, and Nicholas R. Waytowich, presented at the 2019 AAAI Conference on Artificial Intelligence \cite{goecks2018efficiently}, Copyright 2019 by the Association for the Advancement of Artificial Intelligence.}}\label{ch:col_aaai19}

This paper investigates how to utilize different forms of human interaction to safely train autonomous systems in real-time by learning from both human demonstrations and interventions. 
We implement two components of the Cycle-of-Learning for Autonomous Systems, which is our framework for combining multiple modalities of human interaction.
The current effort employs human demonstrations to teach a desired behavior via imitation learning, then leverages intervention data to correct for undesired behaviors produced by the imitation learner to teach novel tasks to an autonomous agent safely, after only minutes of training. 
We demonstrate this method in an autonomous perching task using a quadrotor with continuous roll, pitch, yaw, and throttle commands and imagery captured from a downward-facing camera in a high-fidelity simulated environment. 
Our method improves task completion performance for the same amount of human interaction when compared to learning from demonstrations alone, while also requiring on average 32\% less data to achieve that performance.
This provides evidence that combining multiple modes of human interaction can increase both the training speed and overall performance of policies for autonomous systems. 

\section{Problem Definition}
\noindent The primary goal of learning methodologies is to imbue intelligent agents with the capability to autonomously and successfully perform complex tasks, when \emph{a priori} design of the necessary behaviors is intractable. 
Most tasks of interest, especially those with real-world applicability, quickly exceed the capability of designers to handcraft optimal or even successful policies.
It can even be infeasible to construct appropriate objective or reward functions in many cases.
Instead, learning techniques can be used to empirically discover the underlying objective function for the task and the policy required to satisfy it, typically utilizing state, action, or reward data.
Several classes of these techniques have yielded promising results, including learning from demonstration, learning from evaluation, and reinforcement learning.

Reinforcement learning has been proven to work on scenarios with well-designed reward functions and easily available interactions with the environment \cite{Mnih2015a}. 
However, in real-world robotic applications, explicit reward functions are non-existent, and interactions with the hardware are expensive and susceptible to catastrophic failures. 
This motivates leveraging human interaction to supply this reward function and task knowledge, to reduce the amount of high-risk interactions with the environment, and to safely shape the behavior of robotic agents.

Learning from evaluation is one such way to leverage human domain knowledge and intent to shape agent behavior through sparse interactions in the form of evaluative feedback, possibly allowing for the approximation of a reward function \cite{knox2009interactively,MacGlashan2017,Warnell2018}.
This technique has the advantage of minimally tasking the human evaluator and can be used when training behaviors they themselves cannot perform.
However, it can be slow to converge as the agent can only identify desired or even stable behaviors through more random exploration or indirect guidance from human negative reinforcement of unwanted actions, rather than through more explicit examples of desired behaviors.

In such a case, learning from demonstration can be used to provide a more directed path to these intended behaviors by utilizing examples of the humans performing the task.
This technique has the advantage of quickly converging to more stable behaviors.
However, given that it is typically performed offline, it does not provide a mechanism for corrective or preventative inputs when the learned behavior results in undesirable or catastrophic outcomes, potentially due to unseen states.
Learning from demonstration also inherently requires the maximal burden on the human, requiring them to perform the task many times until the state space has been sufficiently explored, so as to generate a robust policy. 
Also, it necessarily fails when the human is incapable of performing the task successfully at all.

Learning from interventions, where a human acts as an overseer while an agent is performing a task and periodically takes over control or intervenes when necessary, can provide a method to improve the agent policy while preventing or mitigating catastrophic behaviors \cite{Saunders2017}.
This technique can also reduce the amount of direct interactions with the agent, when compared to learning from demonstration.
Similar to learning from evaluation, this technique suffers from the disadvantage that desired behaviors must be discovered through more variable exploration, resulting in slower convergence and less stable behavior.

Most of these human interaction methods have been studied separately, and there is very little work combining multiple modalities to leverage strengths and mitigate weaknesses. 
In this paper, we work towards our conceptual framework that combines multiple human-agent interaction modalities into a single framework, called the Cycle-of-Learning for Autonomous Systems from Human Interaction \cite{Waytowich2018}. 
Our goal is to unify different human-in-the-loop learning techniques in a single framework to overcome the drawbacks of training from different human interaction modalities in isolation, while also maintaining data-efficiency and safety. 

In this paper, we present our initial work towards this goal with a method for combining learning from demonstrations and learning from interventions for safe and efficient training of autonomous systems. We seek to develop a real-time learning technique that combines demonstrations as well as interventions provided from a human to outperform traditional imitation learning techniques while maintaining agent safety and requiring less data. We validate our method with an aerial robotic perching task in a high-fidelity simulator using a quadrotor that has continuous roll, pitch, yaw and throttle commands and a downward facing camera. In particular, the contributions of our work are twofold:
\begin{enumerate}
	\item  We propose a method for efficiently and safely learning from human demonstrations and interventions in real-time. 
	\item We empirically investigate both the task performance and data efficiency associated with combining human demonstrations and interventions.
\end{enumerate}
We show that policies trained with human demonstrations and human interventions together outperform policies trained with just human demonstrations while simultaneously using less data. To the best of our knowledge this is the first result showing that training a policy with a specific sequence of human interactions (demonstrations, then interventions) outperforms training a policy with just human demonstrations (controlling for the total amount of human interactions), and that one can obtain this performance with significantly reduced data requirements, providing initial evidence that the role of the human should adapt during the training of safe autonomous systems.

\begin{figure}[!t]
    \centering
    \includegraphics[width=.75\columnwidth]{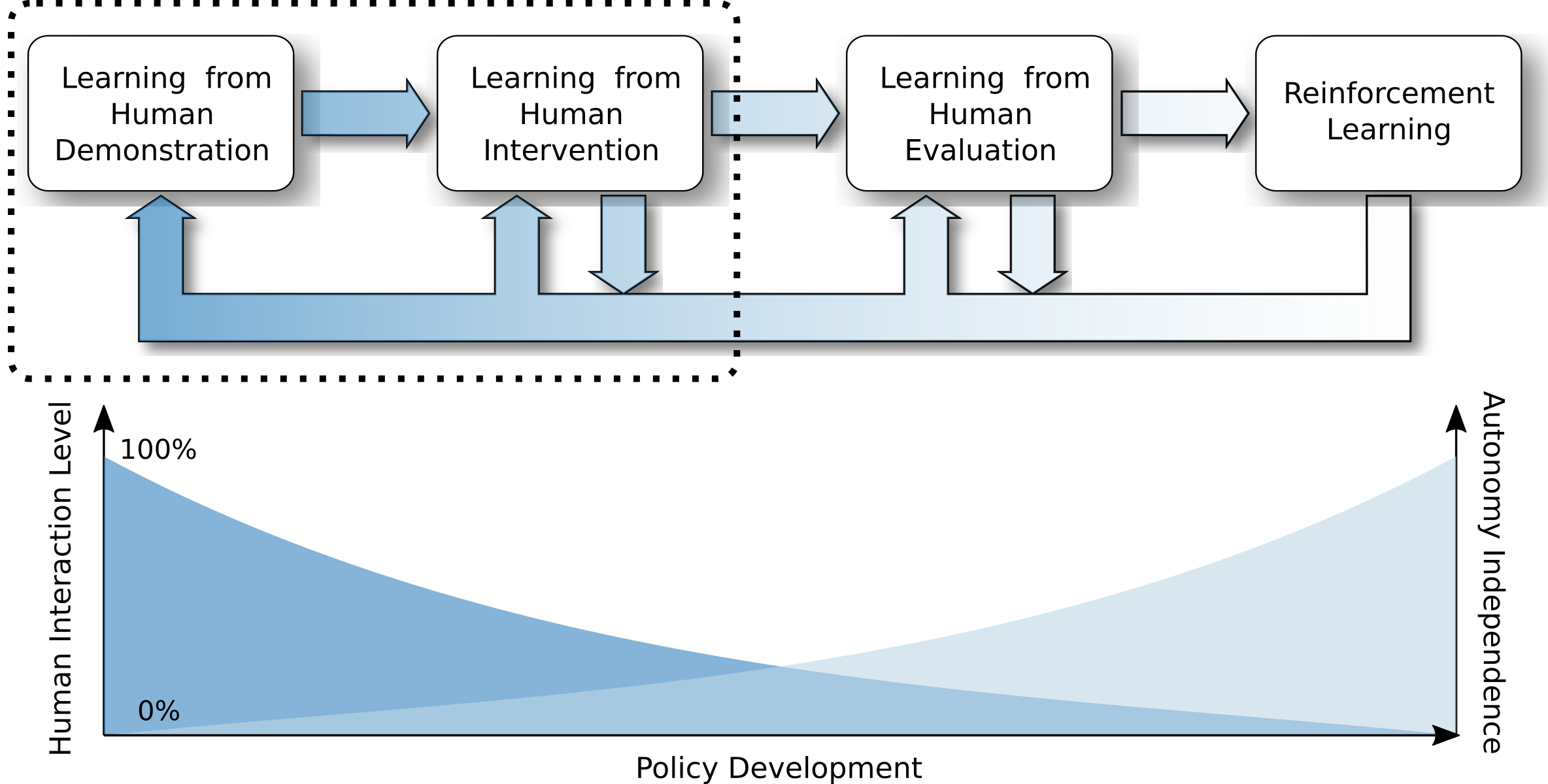}
    \caption{Cycle-of-Learning for Autonomous Systems from Human Interaction: a concept for combining multiple forms of human interaction with reinforcement learning. As the policy develops, the autonomy independence increases and the human interaction level decreases. This work focuses on the first two components of the cycle (dashed box): Learning from Demonstration and Learning from Intervention. Reprinted from \cite{goecks2018efficiently}.}
    \label{fig:col_diagram}
\end{figure}

\section{Background and Related Work}

\subsection{Learning from Demonstrations}
Here we provide a brief summary of Learning from Demonstrations (LfD); a more comprehensive review can be found in \cite{Argall2009}. 
Learning from Demonstrations, sometimes referred to as \emph{Imitation Learning}, is defined by training a policy $\pi$ in order to generalize over a subset $\mathcal{D}$ of states and actions visited during a task demonstration over $T$ time steps:
\begin{align*} \label{eq:demonstration_dataset}
\mathcal{D} = \left\{\vecb{a}_0, \vecb{s}_0, \vecb{a}_1, \vecb{s}_1, ... , \vecb{a}_T, \vecb{s}_T\right\}.
\end{align*}
This demonstration can be performed by a human supervisor, optimal controller, or virtually any other pre-trained policy.

In the case of human demonstrations, the human is implicitly trying to maximize what may be represented as an internal reward function for a given task (Equation \ref{eq:imitation_rew}), where $\pi^*(\vecb{a}^*_t | \vecb{s}_t)$ represents the optimal policy that is not necessarily known, in which the optimal action $\vecb{a}^*$ is taken at state $\vecb{s}$ for every time step $t$.

\begin{equation}\label{eq:imitation_rew}
\max_{\vecb{a}_0,...,\vecb{a}_T} \sum_{t=0}^{T} r_t(\vecb{s}_t, \vecb{a}_t) = \sum_{t=0}^{T} \log p (\pi^*(\vecb{a}_t^* | \vecb{s}_t))
\end{equation}

Defining the policy of the supervisor as $\pi_{sup}$ and its estimate as $\hat{\pi}_{sup}$, imitation learning can be achieved through standard supervised learning, where the parameters $\theta$ of a policy $\pi_\theta$ are trained in order to minimize a loss function, such as mean squared error, as shown in Equation \ref{eq:sup_mse}.

\begin{equation}\label{eq:sup_mse}
\hat{\pi}_{sup} = \argmin_{\pi_\theta} \sum_{t=0}^{T} ||\pi_\theta(\vecb{s}_t) - \vecb{a}_t ||^2
\end{equation}

There are many empirical successes of using imitation learning to train autonomous systems. For self-driving cars, \citeauthor{Bojarski2016} successfully used human demonstrations to train a policy that mapped from front-facing camera images to steering wheel commands using around one hundred hours of human driving data \cite{Bojarski2016}. 
Similar approaches have been taken to train small unmanned air system (sUAS) to navigate through cluttered environments while avoiding obstacles, where demonstration data was collected by human oracles in simulated \cite{goecks2018cyber} and real-world environments \cite{Giusti2015}.

\subsection{Learning from Interventions}
In \emph{Learning from Interventions} (LfI) the human takes the role of a supervisor and watches the agent performing the task and intervenes (i.e. overriding agent actions with human actions) when necessary, in order to avoid unsafe behaviors that may lead to catastrophic states. 
Recently, this learning from human intervention concept was used for safe reinforcement learning (RL) that could train model-free RL agents without a single catastrophe \cite{Saunders2017}. Similar work has proposed using human interaction to train a classifier to detect unsafe states, which would then trigger the intervention by a safe policy previously trained based on human demonstration of the task \cite{Hilleli2018}. This off-policy data generated by the safe policy is aggregated to the replay buffer of a value-based reinforcement learning algorithm (Double Deep Q-Network, or DDQN \cite{Hasselt2015}). The main advantage of this method is being able to combine the off-policy data generated by the interventions to update the current policy.

\subsection{Related Work}

Several existing works have studied, in isolation, the use of different human interaction modalities to train policies for autonomous systems, either in the form of demonstrations \cite{Akgun2012}, \cite{Argall2009}, interventions \cite{Akgun2012a}, \cite{Saunders2017} or evaluations \cite{knox2009interactively}.  However, there has been relatively little work on how to effectively combine multiple human interaction modalities into a single learning framework. Several cases include the combination of demonstrations and mixed initiative control for training robot polices \cite{Grollman2007} as well as the recent work by \citeauthor{Hilleli2018} where imitation learning was combined with interactive reward shaping in a simulated racing game  \cite{Hilleli2018} and the recent work \cite{peng2018deepmimic} where deviation from the expert demonstration is added to a reward function to be optimized with reinforcement learning. 

Another example of work that attempts to augment learning from demonstrations with additional human interaction is the Dataset Aggregation (DAgger) algorithm \cite{Ross2011}.
DAgger is an iterative algorithm that consists of two policies, a primary agent policy that is used for direct control of a system, and a reference policy that is used to generate additional labels to fine-tune the primary policy towards optimal behavior. Importantly, the reference policy's actions are not taken, but are instead aggregated and used as additional labels to re-train the primary policy for the next iteration. In \cite{Ross2013} DAgger was used to train a collision avoidance policy for an autonomous quadrotor using imitation learning on a set of human demonstrations to learn the primary policy and using the human observer as a reference policy. 
There are some drawbacks to this approach that are worth discussing. As noted in \cite{Ross2013}, because the human observer is never in direct control of the policy, safety is not guaranteed, since the agent has the potential to visit previously unseen states, which could cause catastrophic failures. 
Additionally, the subsequent labeling by the human can be suboptimal both in the amount of data recorded (perhaps recording more data in suboptimal states than is needed to learn an optimal policy) as well as in capturing the intended result of the human observer's action (as in distinguishing a minor course correction from a sharp turn, or the appropriate combination of actions to perform a behavior). 
Another limitation of DAgger is that the human feedback was provided \emph{offline} after each run while viewing a slower replay of the video stream to improve the resulting label quality. 
This prevents the application to tasks where real-time interaction between humans and agents are required. 

\section{Combining Learning from Human Demonstrations and Interventions}
This work demonstrates a technique for efficiently training an autonomous system safely and in real-time by combining learning from demonstrations and interventions. It is the first part of the Cycle-of-Learning concept (Figure~\ref{fig:col_diagram}) which aims to combine multiple forms of human-agent interaction for learning a policy that mimics the human trainer in a safe and efficient manner. Although this paper focuses on the first two parts of the Cycle-of-Learning, for brevity, we will refer to the algorithm presented here as the Cycle-of-Learning (CoL). 

The CoL starts by training an initial policy $\pi_0$ from a set of task demonstrations provided by the human trainer using a standard supervised learning technique (regression in this case since the action-space for our task is continuous). 
Next, the agent is given control and executes $\pi_0$ while the human takes the role of overseer and supervises the agent's actions. 
Using a joystick controller, the human intervenes whenever the agent exhibits unwanted behavior that diverges from the policy of the human trainer, and provides corrective actions to drive the agent back on course, and then releases control back to the agent. 
The agent then learns from this intervention by augmenting the original training dataset with the states and actions from the intervention, and then fine-tuning $\pi_0$. 
The agent then executes the new policy $\pi_n$ while the human continues to oversee and provides interventions as necessary. 
In practice, the human trainer can easily switch between providing demonstrations and interventions by switching control between the human and the agent as shown in Figure \ref{fig:airsim_col_diagram_compact}.
Combining demonstration and intervention data in this way should not only improve the policy over what learning from demonstration can do alone but also require less training data to do so. 
The intuition is that the agent will inevitably end up in states previously unexplored with the original demonstration data which will cause it's policy to fail and that intervening from those failure states allows the agent to quickly learn from those deficiencies or "blind spots" in its own policy in a more targeted fashion than from demonstration data alone \cite{Ramakrishnan2018}. 
In this way, we learn only from the critical states, which is more data efficient, instead of using all states for training as is done in DAgger \cite{Ross2011}.

\subsection{Data Efficiency}
A demonstration is defined as a human-produced trajectory of state-action pairs for the entire episode, while an intervention is defined as a trajectory of state-action pairs for only the subset of the episode where corrective action is deemed necessary by the human. 
Thus, the amount of data provided via intervention is nearly always less than the amount provided via demonstration. 
Training routines that incorporate more episodes utilizing learning from intervention rather than learning from demonstration will in general be more data sparse, assuming comparable task performance.
Therefore, by utilizing components of the CoL to learn from both demonstration and intervention, we can train with less data than if demonstrations had been used in isolation for an equivalent number of episodes, resulting in a more efficient training framework.
This concept generalizes to the full CoL, as the agent naturally requires less input from the human as its policy develops, its task proficiency increases, and it becomes more autonomous (indicated in Figure \ref{fig:col_diagram}).

\subsection{Safe Learning}
The notion of \emph{safe} learning here refers to the ability of a human oracle to intervene in cases where catastrophic failure may be imminent.
Thus, the agent is able to explore higher risk regions of the state space with a greater degree of safety.
This approach leverages human domain knowledge and ability to forecast such boundary states, which the agent cannot do early in the training process when the state space is less explored.
By allowing the policy to explore less seen regions and then provide training data of how to correct from those states, human interventions provide a richer dataset that improves the policy in those regimes.
This is contrasted to a method based solely on demonstration, which may only see states and observations along a nominal trajectory and have a policy poorly fit to data outside that envelope.
The result is a policy that is more robust, through greater data diversity, while not risking damage to the agent that is typical with methods that rely on random exploration of the state space.
This provides a method to safely train an autonomous system.

\subsection{Real-Time Interaction}
The utility of the demonstrated approach is partially linked to the ability of the agent to consume data as it is provided by the human oracle and update its policy online.
The current system accomplishes this by storing all subject state-action pairs in the training dataset, which is queried in real-time to update the policy, and then fine-tuning that policy whenever new samples are added to the dataset.
During intervention, this allows for interaction with an agent using a policy trained on the most recent corrective actions provided by the human.
The short time between novel human intervention data and behavioral roll-outs from the agent policy prevents significant delay in this feedback loop that might result from more infrequent, batch learning.
As in closed loop systems, large temporal delays between feedback inputs and their resultant output behaviors can lead to instability.
In this context, that would manifest as unstable training as the human oracle would need to correct for undesired actions for significantly longer before seeing any effect on the agent behavior.
This shortcoming was exhibited in DAgger, where policy correction was a delayed, offline process.

\begin{figure}[!t]
    \centering
    \includegraphics[width=.7\columnwidth]{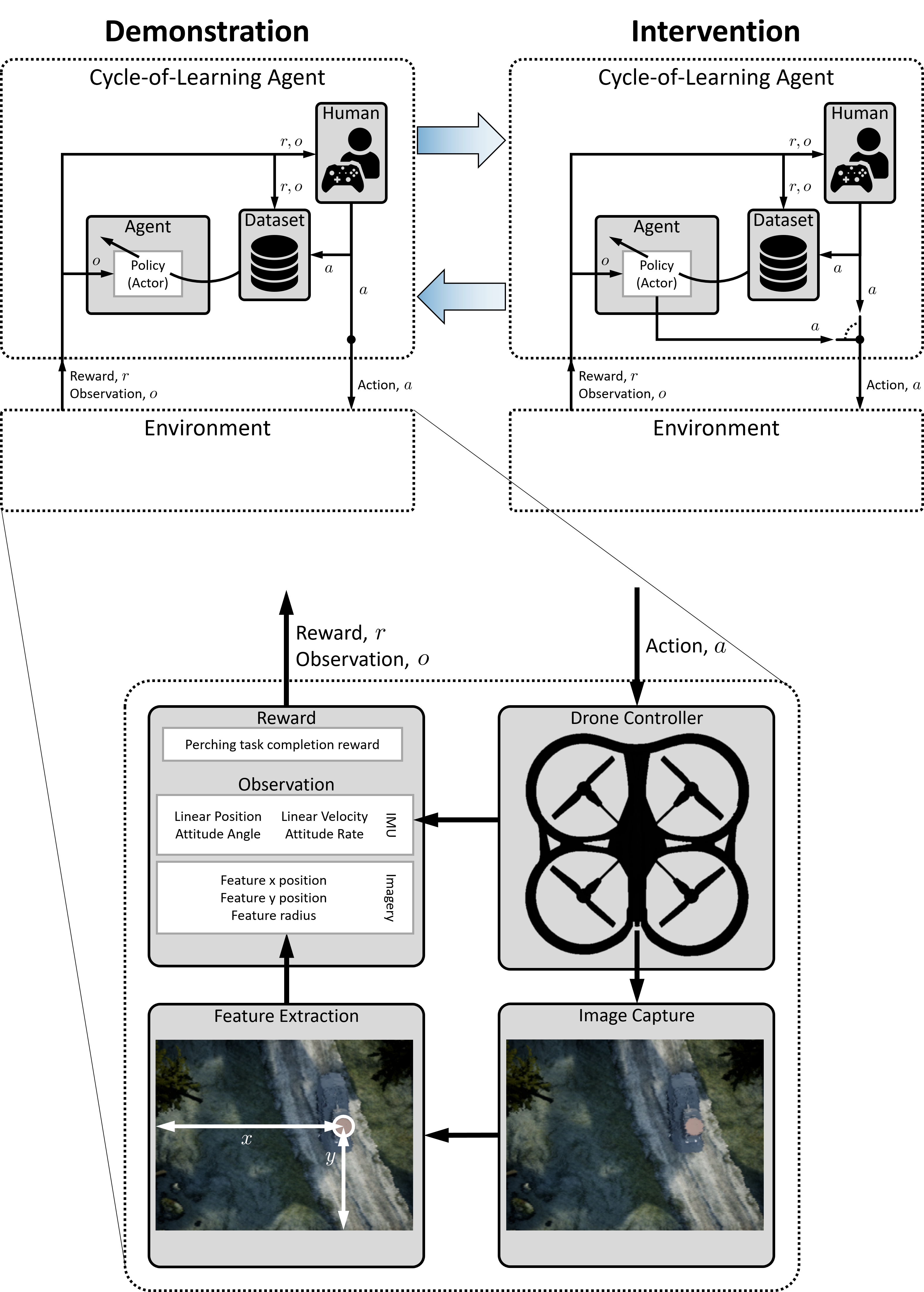}
    \caption{Flow diagram illustrating the learning from demonstration and intervention stages in the CoL for the quadrotor perching task. Reprinted from \cite{goecks2018efficiently}.}
    \label{fig:airsim_col_diagram_compact}
\end{figure}

\section{Implementation}
The next sections address the experimental methodology used to evaluate the proposed approach and the implementation of the learning algorithm (shown in Figures \ref{fig:airsim_col_diagram_compact} and Algorithm \ref{alg:algo}).

\subsection{Environment Modeling}
\begin{figure}[!t]
    \centering
    \includegraphics[width=.7\columnwidth]{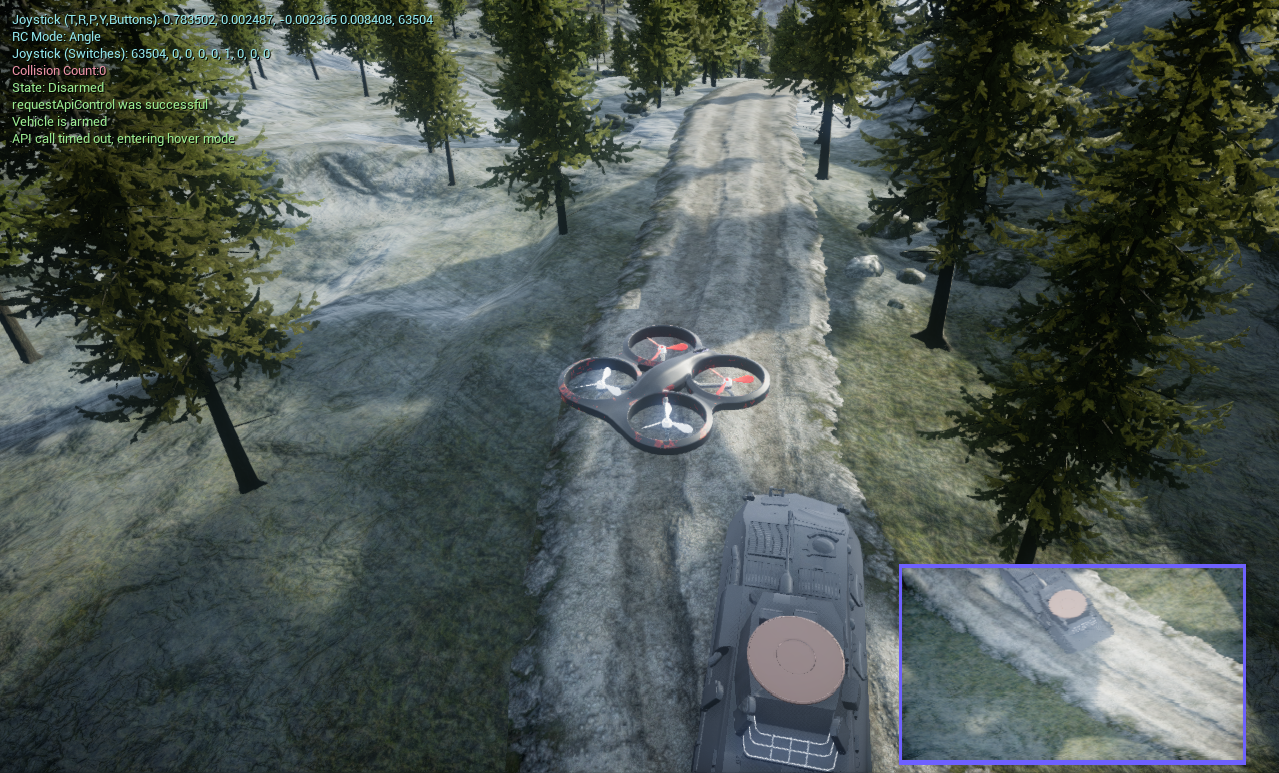}
    \caption{Screenshot of AirSim environment and landing task. Inset image in lower right corner: downward-facing camera view used for extracting the position and radius of the landing pad which is part of the observation-space that the agent learns from. Reprinted from \cite{goecks2018efficiently}.}
    \label{fig:Screenshot}
\end{figure}

We tested our CoL approach (Figure~\ref{fig:airsim_col_diagram_compact}) in an autonomous quadrotor perching task using a high-fidelity drone simulator based on the Unreal Engine called AirSim developed by Microsoft \cite{Airsim2017}. 
AirSim provides realistic emulation of quad-rotor vehicle dynamics while the Unreal Engine allows for the development of photo-realistic environments. 
In this paper, we are concerned with training a quadrotor to autonomously land on a small landing platform placed on top of a ground vehicle (see Figure \ref{fig:Screenshot}). 

The current observation-space consists of vehicle inertial and angular positions, linear and angular velocities, and pixel position and radius of the landing pad extracted using a hand-crafted computer vision module (15 dimensional continuous observation-space).
The vehicle is equipped with a downward-facing RGB camera that captures $320 \times 240$ pixel resolution images.
The camera framerate and agent action frequency is \SI{10.5}{\Hz}, while the human observer views the video stream at approximately \SI{35}{\Hz}.  
The action-space comprises the four continuous joystick commands (roll, pitch, yaw, and throttle), which are translated to reference velocity commands (lateral, longitudinal, heading rate, and heave) by the vehicle autopilot. 

For the perching task, the goal is to land the quadrotor as close to the center of the landing pad as possible. We define a landing a success if the quadrotor lands within 0.5m radius of the center of the platform and a failure otherwise. At the beginning of each episode, the quadrotor starts in a random x,y location at a fixed height above the landing pad and the episode ends when either the quadrotor reaches the ground (or landing pad) or after 500 time-steps have elapsed. 

\subsection{The Cycle-of-Learning Algorithm}

\begin{algorithm}[!tb]
\caption{Combining Human Demonstrations and Interventions (Cycle-of-Learning)}\label{alg:algo}
\begin{algorithmic}[1]

\Procedure{Main}{}
    \State Initialize agent's policy $\pi$
    \State Initialize human dataset $\mathcal{D}_H$
    \State Initialize \emph{Update Policy} procedure
    \State Define performance threshold $\alpha$
    \While {task performance $< \alpha$}
        \State Read observation $o$
        \State Sample action $a_{\pi} \sim \pi$
        \If {Human Interaction ($a_H$)}:
            \State Perform human action $a_H$
            \State Add $o$ and $a_H$ to $\mathcal{D}_H$
        \Else
            \State Perform $a_{\pi}$ 
        \EndIf
        \If {End of Episode}
            \State Evaluate task performance
        \EndIf
    \EndWhile
\EndProcedure

\Procedure{Update Policy}{}
    \State Spawn separate thread
    \State Initialize $loss$ threshold $loss_{TH}$
    \While {\emph{Main} procedure running}
        \State Load human dataset $\mathcal{D}_H$
        \If {New Samples}:
            \While {$loss > loss_{TH}$ or $n < n_max$ }
                \State Sample $N$ samples $o, a$ from $\mathcal{D}_H$
                \State Sample $\hat{a} \sim \pi$
                \State Compute $loss =
                \frac{1}{N} \sum_{i}^{N} (\hat{a}_i - a_i)^2$
                \State Perform gradient descent update on $\pi$
            \EndWhile
        \EndIf
    \EndWhile
\EndProcedure

\end{algorithmic}
\end{algorithm}

As shown in Algorithm~\ref{alg:algo} the main procedure starts by initializing the agent's policy $\pi$, the human dataset $\mathcal{D}_H$, the \emph{Update Policy} subroutine, and task performance threshold. 
The main loop consists of either executing actions provided by the agent or actions provided by the human. The agent reads an observation from the environment and an action is sampled based on the current policy. At any moment the human supervisor is able to override the agent's action by holding a joystick trigger. When this trigger is held, the actions performed by the human $a_h$ are sent to the vehicle to be executed and are added to the human dataset $\mathcal{D}_H$ to update the policy according to the \emph{Update Policy} subroutine.

The agent's policy $\pi$ is a fully-connected, three hidden-layer, neural network with 130, 72, and 40 neurons, respectively. The network is randomly initialized with weights sampled from a normal distribution. The policy is optimized by minimizing the mean squared error loss using the Root Mean Square Propagation (RMSProp) optimizer with learning rate of 1e-4.
Unless defined otherwise, the human dataset $\mathcal{D}_H$ is initialized as an empty comma-separated value (CSV) file. Its main goal is to store the observations and actions performed by the human.
The procedure to update the policy in real-time spawns a separate CPU thread to perform policy updates in real-time while the human either demonstrates the task or intervenes. 
This separate thread continuously checks for new demonstration or intervention data based on the size of the human dataset.
If new samples are found, this thread samples a minibatch of 64 samples of observations and actions from the human dataset and is used to perform policy updates based on the mean squared error loss until it reaches the loss threshold of 0.005 or maximum number of epochs (in this case, 2000 epochs).
This iterative update routine continues until the task performance threshold $\alpha$ is achieved, which can vary from task to task depending on the desired performance. For this work, we set $\alpha$ to 1 and only stop training after a pre-specified number of episodes defined in our experimentation methodology to empirically evaluate our method over a controlled number of human interactions, here defined as either human demonstrations or human interventions.

\subsection{Experimental Methodology}
Using the AirSim landing task, we tested our proposed CoL framework against several baseline conditions where we compared against using only a single human interaction modality (i.e. only demonstrations or only interventions) using equal amounts of human interaction time for each condition. By controlling for the human interaction time, we can assess if our method of utilizing multiple forms of human interaction provides an improvement over a single form of interaction given the same amount of human effort. 

Each human participant (n=4) followed the same experimental protocol: given an RGB video stream from the downward-facing camera, the participant controlled the continuous roll, pitch, yaw, and throttle of the vehicle using an Xbox One joystick to perform 4, 8, 12 and 20 complete episodes of the perching task for three experimental conditions: demonstrations only, interventions only, and demonstrations plus interventions with the CoL method, with each condition starting from a randomly initialized policy. For the CoL condition, participants performed an equal number of demonstrations and interventions to match the total number of episodes for that condition. 
For example, given 4 episodes of training, our CoL approach would train with learning from demonstrations in the first 2 episodes and then switch to learning from interventions for the last 2 episodes. 
We compared this to learning from demonstration for all 4 episodes as well as learning from interventions for all 4 episodes. 
This was repeated for 8, 12 and 20 episodes to study the effect of varying amounts of human interaction on task performance. 
Following the diagram in Figure \ref{fig:airsim_col_diagram_compact} and Update Policy procedure on Algorithm \ref{alg:algo}, the agent's policy is trained on a separate thread in real-time, and a model is saved for each complete episode together with the human-observed states and the actions they performed. 
These saved models are later evaluated to assess task performance according to our evaluation procedure described in the next section.

We also compared our approach to a random agent as well as an agent trained using a state-of-the-art reinforcement learning approach. The reinforcement learning agent used a publicly available implementation of Proximal Policy Optimization (PPO) \cite{Schulman2017} with a four degree-of-freedom action space (pitch, roll, throttle, yaw) and was trained for 1000 episodes, using only task completion as a binary sparse reward signal. To investigate the effect of action-space complexity on task performance, we also implemented the PPO where only two actions (pitch and roll) and three actions (pitch, roll and throttle) were available; for both cases, all other actions were held to a constant value. For the two actions condition (pitch and roll), the agent was given constant throttle and descended in altitude at a constant velocity. For both conditions yaw was set to 0. Training time of the reinforcement learning agent was limited to the simulated environment running in real-time. 

\section{Numerical Results}

We evaluated our method in terms of task completion percentage, defined as the number of times the drone successfully landed on the landing pad over 100 evaluation runs, for each training method as well as for different amounts of human training data. Additionally, We compared the number of human data samples, i.e. observation-action pairs, used during training for each condition.

\begin{figure}[t]
\centering
  \includegraphics[width=.8\columnwidth]{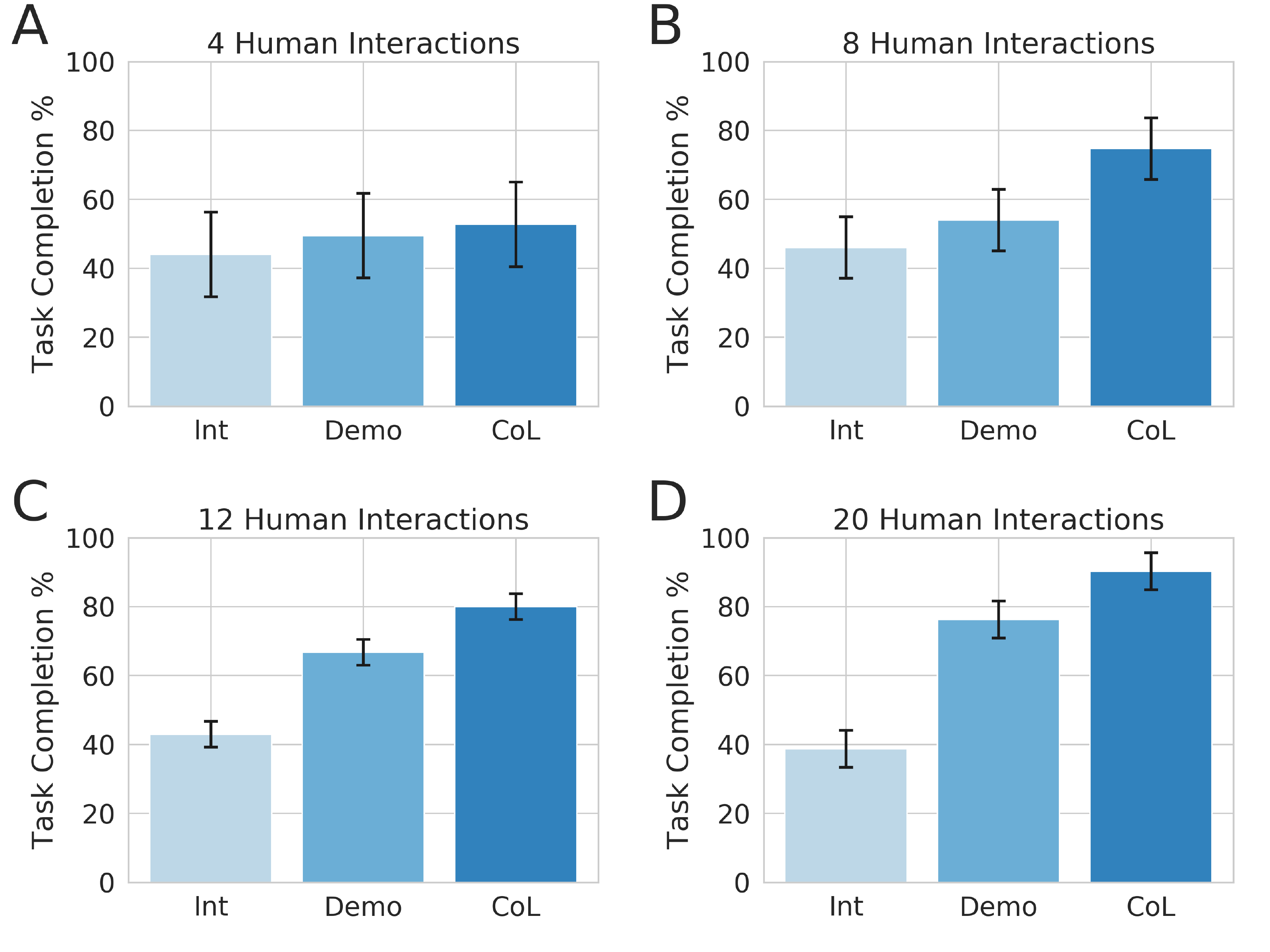}
\caption{Performance comparison in terms of task completion with Interventions (Int), Demonstrations (Demo) and the Cycle-of-Learning (CoL) framework for (A) 4 human interactions, (B) 8 human interactions, (C) 12 human interactions and (D) 20 human interactions, respectively. Here, an interaction equates to a single demonstration or intervention and roughly corresponds to the number of episodes. Error bars denote 1 standard error of the mean.  We see that CoL outperforms Int and Demo across nearly all human interaction levels. Reprinted from \cite{goecks2018efficiently}.}
\label{fig:performance_all}
\end{figure}

Figure \ref{fig:performance_all} compares the performance of the models trained using only interventions (Int), the models trained using only demonstrations (Demo), and the models trained using the Cycle-of-Learning approach (CoL). 
We show results for only these conditions as the random policy condition and the RL condition trained using PPO with the full four degree-of-freedom action space were not successful given the small amount of training episodes, as explained later in this section. 
Barplots show the task completion performance from each condition averaged over all participants with error bars representing 1 standard error. 
Subpanels show the performance for varying amounts of human interaction: 4, 8, 12 and 20 episodes.
For the 4 human interaction condition (Figure \ref{fig:performance_all}A), all methods show similar task completion conditions. 
However, for the 8, 12 and 20 human interaction conditions, we see that the CoL approach achieves higher task completion percentages compared to the demonstration-only and intervention-only conditions, with the intervention condition performing the worst. 

For the final condition of 20 episodes our proposed approach achieves 90.25\% ($\pm$ 5.63\% std. error) task completion as compared to 76.25\% ($\pm$ 2.72\% std. error) task completion using only demonstrations.
In comparison, for the 8 episodes condition, our proposed approach achieves 74.75\% ($\pm$ 9.38\% std. error) task completion in contrast to 54.00\% ($\pm$ 8.95\% std. error) task completion when using only demonstrations. 

\begin{figure}[t]
\centering
  \includegraphics[width=.8\columnwidth]{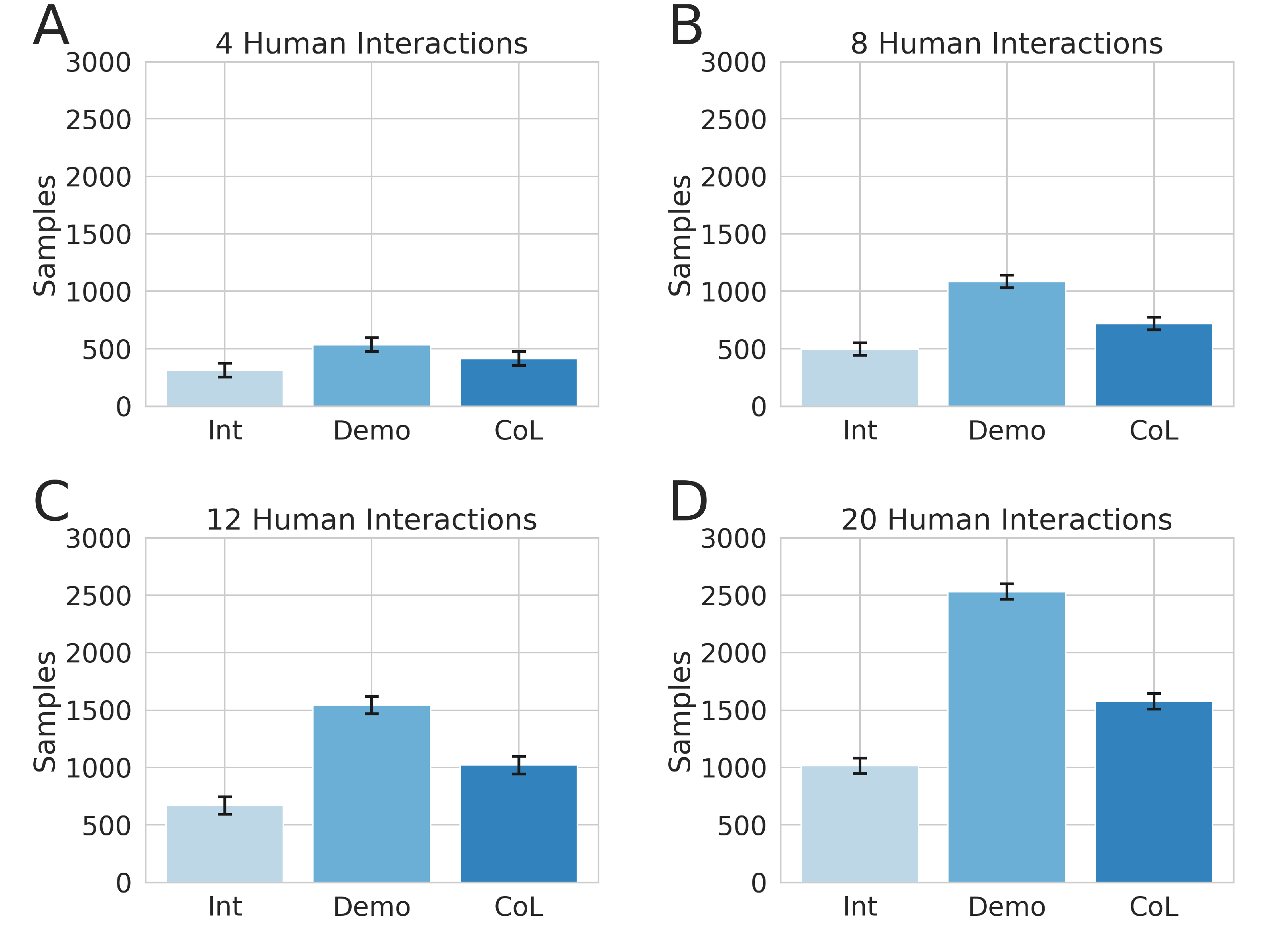}
\caption{Comparison of the number of human samples used for training with Interventions (Int), Demonstrations (Demo) and the Cycle-of-Learning (CoL) framework for (A) 4 human interactions, (B) 8 human interactions, (C) 12 human interactions and (D) 20 human interactions, respectively. Error bars denote 1 standard error of the mean. We see that CoL uses less data than the demonstration-only condition and only slightly more data than the intervention-only condition. Reprinted from \cite{goecks2018efficiently}.}
\label{fig:samples_all}
\end{figure}


Figure \ref{fig:samples_all} compares the number of human data samples used to train the models for the same conditions and datasets as in Figure \ref{fig:performance_all}. 
For the final condition of 20 episodes our proposed approach used on average 1574.50 ($\pm$ 54.22 std. error) human-provided samples, which is 37.79\% fewer data samples when compared to using only demonstrations.
Note that the policies generated from this sparser dataset were able to increase task completion by 14.00\%. 
These results yield a CoL agent that has 1.90 times the rate of task completion performance per sample when compared to learning from demonstrations alone.
This value is computed by comparing the ratios of task completion rate to data samples utilized between the CoL agent and the demonstration-only agent, respectively. 
Averaging the results over all presented conditions and datasets, the task completion increased by 12.81\% ($\pm$ 3.61\% std. error) using 32.05\% ($\pm$ 3.25\% std. error) less human samples, which results in a CoL agent that overall has a task completion rate per sample 1.84 times higher than its counterparts.

\begin{figure}[t]
\centering
  \includegraphics[width=.7\columnwidth]{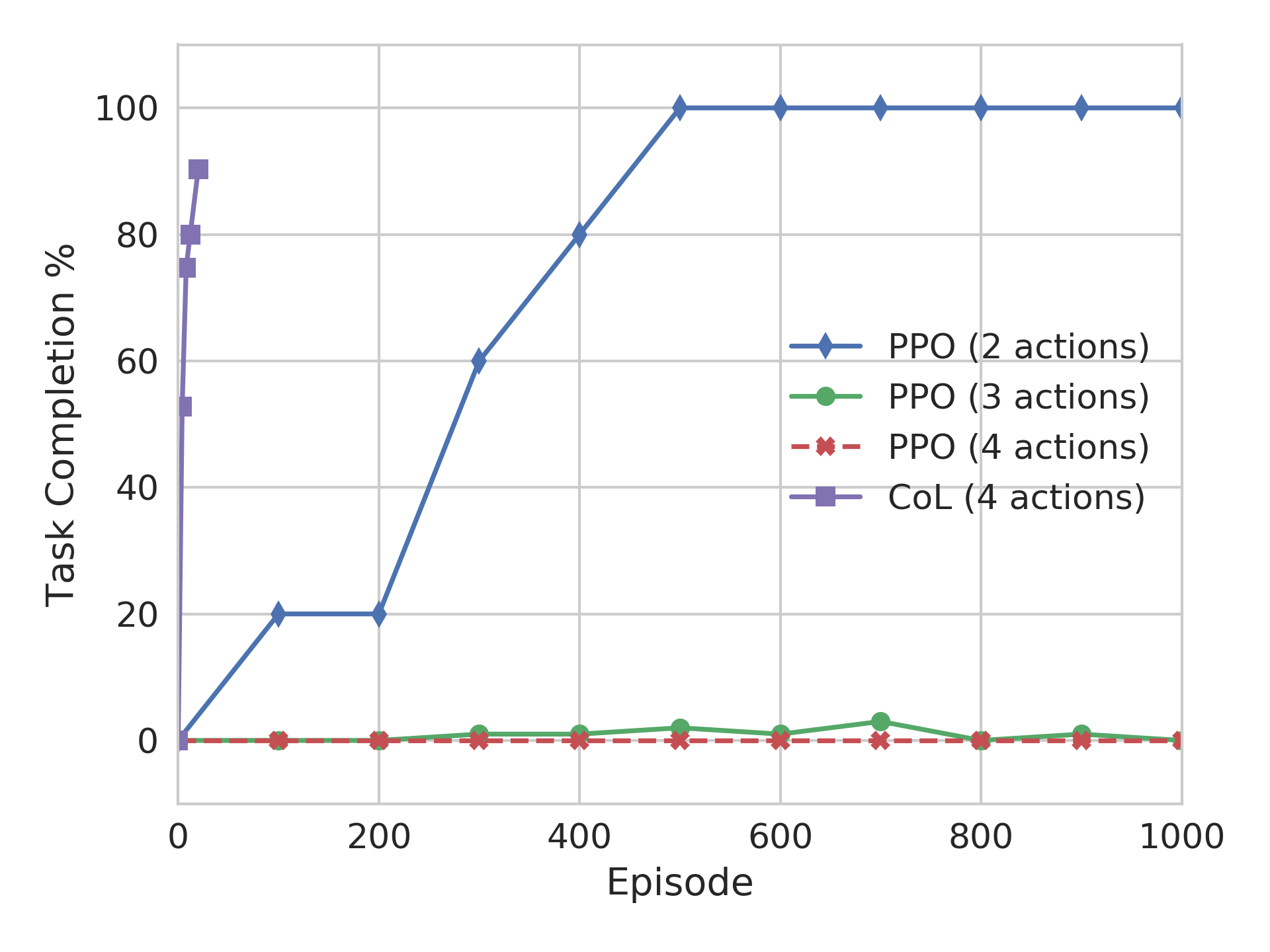}
\caption{Performance comparison between the Cycle-of-Learning (CoL) with four continuous actions and the Deep Reinforcement Learning algorithm Proximal Policy Optimization (PPO) trained for three different task complexities using 2, 3, and 4 continuous actions. Reprinted from \cite{goecks2018efficiently}.}
\label{fig:ppo_curve}
\end{figure}

Figure \ref{fig:ppo_curve} shows a comparison between the performance of the CoL method as well as the PPO baseline comparisons using a 2, 3 and 4 degree-of-freedom action space. For PPO with two actions, the agent was able to achieve 100\% task completion after 500 episodes, on average. However, when the action space complexity increases to three actions, the PPO agent performance was significantly reduced, now completing the task less than 5\% of the time after training for 1000 episodes. As expected, the PPO agent with the full four degree-of-freedom action space fails to complete the task after training for 1000 episodes. In contrast, the CoL method, with the same four degree-of-freedom action space, achieves about 90\% task completion in only 20 episodes, representing significant gains in sample efficiency compared to PPO.

\section{Summary}

Learning from demonstrations in combination with learning from interventions yields a more proficient policy based on less data, when compared to either approach in isolation.
It is likely that the superior performance of the CoL is due to combining the two methods in sequence so as to leverage their strengths while attenuating their deficiencies.

Having been initialized to a random policy, learning from interventions alone produced more random behaviors, making convergence to a baseline behavior much slower.
Overall performance is thus slower to develop, resulting in lower percent completion for the same number of interaction episodes.
Conversely, learning from demonstrations alone was quicker to converge to stable behavior, but it was consistently outperformed by the CoL across varying numbers of interactions, while having more training data to utilize.
This seems initially counter-intuitive as more training data should result in a more accurate and presumably more proficient policy.
However, as the demonstrations typically follow stable trajectories, the agent is less likely to encounter regions of the state space outside these trajectories.
When enacting the policy at test time, any deviations from these previously observed states is not captured well in the policy, resulting in poor generalization performance.
By allowing the agent to act under its current policy, in conjunction with adaptively updating the policy with corrective human-provided actions needed to recover from potentially catastrophic states, the dataset and subsequent policy is improved.
Thus, the CoL allows for both rapid convergence to baseline stable behavior and then safe exploration of state space to make the policy more refined \emph{and} robust.
    
The results shown in Figure \ref{fig:samples_all} confirm the expectation that the combination of learning from demonstrations and interventions requires less data than the condition of learning from demonstrations alone, for the same number of episodes.
This supports the notion that the CoL is a more data efficient approach to training via human inputs.
When additionally considering the superior performance exhibited in Figure \ref{fig:performance_all}, the data efficiency provided by this technique is even more significant.
This result further supports the notion that a combinatorial learning strategy inherently samples more data rich inputs from the human observer.
 
It should be emphasized here that rather than providing an incremental improvement to a specific demonstration or intervention learning strategy, this work proposes an algorithmically agnostic methodology for combining modes of human-based learning.
The primary assertion of this work is that learning is made more robust, data efficient, and safe through a fluid and complementary cycling of these two modes, and would similarly be improved with the addition of the later stages of the CoL (i.e. learning from human evaluation and reinforcement learning).

As seen in Figure~\ref{fig:ppo_curve}, the PPO baseline comparison method was tested across varying complexity with different numbers of action dimensions. A striking result that can be seen is the significant drop-off in performance when going from two-actions, in which the drone had a constant downward throttle and only controlled roll and pitch, to three and four actions, in which the drone also had to control its own throttle. An obvious characteristic of a successful policy for the perching task is that the drone needs to descend in a stable and smooth manner, which is already provided in the two-action condition, as the downward throttle was set \textit{apriori}. This makes the task of solving for an optimal policy much simpler. In the three and four action condition, however, this behavior must be learned from a sparse reward signal (success or failure to land), which is very difficult given limited episodes. 

When implementing the CoL in real world environments, catastrophic failures may be seriously damaging to the autonomous agent, and thus unacceptable.
Having a human observer capable of intervening provides a mechanism to prevent this inadmissible outcome.
Further, techniques that might be applied to enforce a similar level of safety automatically might limit the exploration of the state space, yielding a less robust or less capable policy.
Analogously to the shift of policy design from roboticists or domain experts to human users and laypersons, which is yielded by human-in-the-loop learning, the technique of learning from interventions shifts the implementation of system fail safes away from developers toward users.
This shift leverages human abilities to predict outcomes, adapt to dynamic circumstances, and synthesize contextual information in decision making.

\subsection{Current Limitations and Future Directions}

Our current implementation is limited to the first two stages of the CoL: learning from demonstrations and interventions.
Our planned future work will include adding in more components of the CoL; for example, learning from human evaluative feedback as done in \cite{knox2009interactively,MacGlashan2017,Warnell2018}. 
Additionally, we aim to incorporate reinforcement learning techniques to further fine-tune the learning performance after learning from human demonstrations, interventions and evaluations using an actor-critic style RL architecture \cite{Sutton2018}.

A second limitation of the current implementation is that it requires the human to supervise the actions taken by the agent at all times. 
Future work aims to incorporate confidence metrics in our learning models so that the autonomous system can potentially halt its own actions when it determines it has low confidence and query the human directly for feedback in a mixed-initiative style framework \cite{Grollman2007}, similar to active learning techniques. 
Furthermore, our results clearly indicate that a two-stage process - with a primary stage with a large proportion of human-provided actions followed by a secondary stage with a smaller proportion of those actions - outperforms processes with uniformly large or small amounts of human data throughout. 
This suggests there is perhaps an optimal point in the learning process at which to vary in the amount of human input from full demonstrations to interventions.
Figure \ref{fig:performance_all} illustrates this notion across the varying number of interactions shown in the subfigures, i.e. through the change in relative performance between the three conditions.
In future work we will examine if such an optimal mixture or sequencing of demonstrations and interventions exists, such that learning speed and stability are maximized, and if so, whether it is operator dependent.
Rather than having a predetermined staging of the demonstrations and interventions that is potentially suboptimal, a mixed initiative framework could determine this optimal transition point.
This could further reduce the burden on the human observer, allow for faster training, and even provide a mechanism to generate more robust policies through guided exploration of the state space.

This work demonstrates the first two stages of the CoL in a simulation environment with the goal of eventually transitioning to physical systems, such as an sUAS.
The CoL framework was implicitly designed for use in real world systems, where interactions are limited, and catastrophic actions are unacceptable. 
As can be seen in Figure~\ref{fig:ppo_curve}, our method learns to perform the perching task in several orders of magnitude less time than traditional RL approaches, potentially allowing for feasible on-the-fly training of real systems. 
Therefore, we expect that the application of the CoL to sUAS platforms, or other physical systems, should operate in effectively the same manner as demonstrated in this work.
Future efforts will focus on transitioning this framework onto such physical platforms to study its efficacy in real world settings.
One critical hurdle that must be overcome, is the implementation of the learning architecture on embedded hardware, constrained by the limited payload of an sUAS.

Additionally, given that we are utilizing a relatively high fidelity simulation environment, i.e. AirSim, it may be beneficial to bootstrap a real world system with a policy learned in simulation.
Although there are numerous challenges in transferring a policy learned in simulation into the real world, the CoL itself should allow for significantly smoother transfer due to its cyclic nature in which the user can revert to more direct and user intensive inputs at any point during the learning to allow for adaptation to previously unobserved states. 
This capability inherently provides a method of transfer learning in the case of disparities between simulated and real world properties of the vehicle, sensors, and environment.
For example, if the perching behavior learned in simulation was transferred to an actual sUAS, the vehicle dynamics may have unmodeled non-linearities, the imagery may have dynamic range limitations, or the environment may present exogenous gust disturbances.
In such cases, the baseline policy would be monitored and corrected via learning from intervention, if these discrepancies yielded undesirable or possibly catastrophic behaviors.

\chapter[INTEGRATING BEHAVIOR CLONING AND REINFORCEMENT LEARNING FOR IMPROVED PERFORMANCE IN SPARSE AND DENSE REWARD ENVIRONMENTS]{INTEGRATING BEHAVIOR CLONING AND REINFORCEMENT LEARNING FOR IMPROVED PERFORMANCE IN SPARSE AND DENSE REWARD ENVIRONMENTS\footnote{Adapted with permission from ``Integrating Behavior Cloning and Reinforcement Learning for Improved Performance in Sparse and Dense Reward Environments'', by Vinicius G. Goecks, Gregory M. Gremillion, Vernon J. Lawhern, John Valasek, and Nicholas R. Waytowich, presented at the 2020 International Conference on Autonomous Agents and Multi-Agent Systems \cite{goecks2019integrating} and is partially reproduced here, Copyright 2020 by the International Foundation for Autonomous Agents and MultiAgent Systems.}} \label{ch:col_loss}

This paper investigates how to efficiently transition and update policies, trained initially with demonstrations,  using off-policy actor-critic reinforcement learning. 
It is well-known that techniques based on Learning from Demonstrations, for example behavior cloning, can lead to proficient policies given limited data. 
However, it is currently unclear how to efficiently update that policy using reinforcement learning as these approaches are inherently optimizing different objective functions. 
Previous works have used loss functions, which combine behavior cloning losses with reinforcement learning losses to enable this update.
However, the components of these loss functions are often set anecdotally, and their individual contributions are not well understood. 
In this work, we propose the Cycle-of-Learning (CoL) framework that uses an actor-critic architecture with a loss function that combines behavior cloning and 1-step Q-learning losses with an off-policy pre-training step from human demonstrations.
This enables transition from behavior cloning to reinforcement learning without performance degradation and improves reinforcement learning in terms of overall performance and training time. 
Additionally, we carefully study the composition of these combined losses and their impact on overall policy learning. 
We show that our approach outperforms state-of-the-art techniques for combining behavior cloning and reinforcement learning for both dense and sparse reward scenarios. 
Our results also suggest that directly including the behavior cloning loss on demonstration data helps to ensure stable learning and ground future policy updates.

\section{Problem Definition}


Reinforcement Learning (RL) has yielded many recent successes in solving complex tasks that meet and exceed the capabilities of human counterparts, demonstrated in video game environments \cite{Mnih2015a}, robotic manipulators \cite{andrychowicz2018learning}, and various open-source simulated scenarios \cite{lillicrap2015continuous}.
However, these RL approaches are sample inefficient and slow to converge to this impressive behavior, limited significantly by the need to explore potential strategies through trial and error, which produces initial performance significantly worse than human counterparts.
The resultant behavior that is initially random and slow to reach proficiency is poorly suited to various situations, such as physically embodied ground and air vehicles or in scenarios where sufficient capability must be achieved in short time spans.
In such situations, the random exploration of the state space of an untrained agent can result in unsafe behaviors and catastrophic failure of a physical system, potentially resulting in unacceptable damage or downtime. 
Similarly, slow convergence of the agent's performance requires exceedingly many interactions with the environment, which is often prohibitively difficult or infeasible for physical systems that are subject to energy constraints, component failures, and operation in dynamic or adverse environments.
These sample efficiency pitfalls of RL are exacerbated even further when trying to learn in the presence of sparse rewards, often leading to cases where RL can fail to learn entirely. 

One approach for overcoming these limitations is to utilize demonstrations of desired behavior from a human data source (or potentially some other agent) to initialize the learning agent to a significantly higher level of performance than is yielded by a randomly initialized agent. 
This is often termed Learning from Demonstrations (LfD) \cite{Argall2009}, which is a subset of imitation learning that seeks to train a policy to imitate the desired behavior of another policy or agent. 
LfD leverages data (in the form of state-action tuples) collected from a demonstrator for supervised learning, and can be used to produce an agent with qualitatively similar behavior in a relatively short training time and with limited data. 
This type of LfD, called Behavior Cloning (BC), learns a mapping between the state-action pairs contained in the set of demonstrations to mimic the behavior of the demonstrator. 

Though BC techniques do allow for the relatively rapid learning of behaviors that are comparable to that of the demonstrator, they are limited by the quality and quantity of the demonstrations provided and are only improved by providing additional, high-quality demonstrations. 
In addition, BC is plagued by the distributional drift problem in which a mismatch between the learned policy distribution of states and the distribution of states in the training set can cause errors that propagate over time and lead to catastrophic failures. 
By combining BC with subsequent RL, it is possible to address the drawbacks of either approach, initializing a significantly more capable and safer agent than with random initialization, while also allowing for further self-improvement without needing to collect additional data from a human demonstrator. 
However, it is currently unclear how to effectively update a policy initially trained with BC using RL as these approaches are inherently optimizing different objective functions.
Previous works have used loss functions that combine BC losses with RL losses to enable this update, however, the components of these loss functions are often set anecdotally and their individual contributions are not well understood.

In this work, we propose the Cycle-of-Learning (CoL) framework, which uses an actor-critic architecture with a loss function that combines behavior cloning and 1-step Q-learning losses with an off-policy algorithm, and a pre-training step to learn from human demonstrations. 
Unlike previous approaches to combine BC with RL, such as \cite{Rajeswaran-RSS-18}, our approach uses an actor-critic architecture to learn both a policy and value function from the human demonstration data, which we show, speeds up learning. 
Additionally, we perform a detailed component analysis of our method to investigate the individual contributions of pre-training, combined losses, and sampling methods of the demonstration data and their effects on transferring from BC to RL.
To summarize, the main contribution of this work are:
\begin{itemize}
    \item We introduce an actor-critic based method, that combines pre-training as well as combined loss functions to learn both a policy and value function from demonstrations, to enable transition from behavior cloning to reinforcement learning.
    \item We show that our method can transfer from BC to RL without performance degradation while improving upon existing state-of-the-art BC to RL algorithms in terms of overall performance and training time.
    \item We perform a detailed analysis to investigate the contributions of the individual components in our method.
\end{itemize}

Our results show that our approach outperforms BC, Deep Deterministic Policy Gradients (DDPG), and Demonstration Augmented Policy Gradient (DAPG) in two different application domains for both dense- and sparse-reward settings.
Our results also suggest that directly including the behavior cloning loss on demonstration data helps to ensure stable learning and ground future policy updates, and that a pre-training step enables the policy to start at a performance level greater than behavior cloning.


\section{Preliminaries}

We adopt the standard Markov Decision Process (MDP) formulation for sequential decision making \cite{Sutton2018}, which is defined as a tuple $(S, A, R, P, \gamma)$, where $S$ is the set of states, $A$ is the set of actions, $R(s,a)$ is the reward function, $P(s'|s, a)$ is the transition probability function and $\gamma$ is a discount factor. 
At each state $s \in S$, the agent takes an action $a \in A$, receives a reward $R(s,a)$ and arrives at state $s'$ as determined by $P(s'|s, a)$. 
The goal is to learn a behavior policy $\pi$ which maximizes the expected discounted total reward. 
This is formalized by the Q-function, sometimes referred to as the state-action value function:
\begin{equation*}
    Q^{\pi} (s,a) = \mathbb{E}_{a_t\sim\pi}\left[\sum_{t=0}^{+\infty}\gamma^tR(s_t,a_t)\right]
\end{equation*}

\noindent taking the expectation over trajectories obtained by executing the policy $\pi$ starting at $s_0 = s$ and $a_0 = a$.

Here we focus on actor-critic methods which seek to maximize
\begin{equation*}
    J(\theta) = \mathbb{E}_{s\sim\mu}[Q^{\pi(.|\theta)}(s,\pi(s|\theta))]
\end{equation*}

\noindent with respect to parameters $\theta$ and an initial state distribution $\mu$.
The Deep Deterministic Policy Gradient (DDPG) \cite{lillicrap2015continuous} is an off-policy actor-critic reinforcement learning algorithm for continuous action spaces, which calculates the gradient of the Q-function with respect to the action to train the policy. 
DDPG makes use of a replay buffer to store past state-action transitions and target networks to stabilize Q-learning \cite{Mnih2015a}. 
Since DDPG is an off-policy algorithm, it allows for the use of arbitrary data, such as demonstrations from another source, to update the policy. 
A demonstration trajectory is a tuple $(s, a, r, s')$ of state $s$, action $a$, the reward $r = R(s,a)$ and the transition state $s'$ collected from a demonstrator's policy.
In most cases these demonstrations are from a human observer, although in principle these demonstrations can come from any existing agent or policy. 

\section{Related Work}

Several works have shown the efficacy of combining behavior cloning with reinforcement learning across a variety of tasks. 
Recent work by \cite{hester2018deep}, known as Deep Q-learning from Demonstrations (DQfD), combined behavior cloning with deep Q-learning \cite{Mnih2015a} to learn policies for Atari games by leveraging a loss function that combines a large-margin supervised learning loss function, 1-step Q-learning loss, and an $n$-step Q-learning loss function that helps ensure the network satisfies the Bellman equation. 
This work was extended to continuous action spaces by \cite{vevcerik2017leveraging} with DDPG from Demonstrations (DDPGfD), who proposed an extension of DDPG \cite{lillicrap2015continuous} that uses human demonstrations, and applied their approach to object manipulation tasks for both simulated and real robotic environments. 
The loss functions for these methods include the $n$-step Q-learning loss, which is known to require on-policy data to accurately estimate. 
Similar work by \cite{Nair2018ICRA} combined behavior cloning-based demonstration learning, goal-based reinforcement learning, and DDPG for robotic manipulation of objects in a simulated environment. 

A method that is very similar to ours is the Demonstration Augmented Policy Gradient (DAPG) \cite{Rajeswaran-RSS-18}, a policy-gradient method that uses behavior cloning as a pre-training step together with an augmented loss function with a heuristic weight function that interpolates between the policy gradient loss, computed using the Natural Policy Gradient \cite{Kakade2001}, and behavior cloning loss.
They apply their approach across four different robotic manipulations tasks using a 24 Degree-of-Freedom (DoF) robotic hand in a simulator and show that DAPG outperforms DDPGfD \cite{vevcerik2017leveraging} across all tasks. 
Their work also showed that behavior cloning combined with Natural Policy Gradient performed very similarly to DAPG for three of the four tasks considered, showcasing the importance of using a behavior cloning loss both in pre-training and policy training. 

\section{Integrating Behavior Cloning and Reinforcement Learning}
The Cycle-of-Learning (CoL) framework is a method for leveraging multiple modalities of human input to improve the training of RL agents. 
These modalities can include human demonstrations, i.e. human-provided exemplar behaviors, human interventions, i.e. interdictions in agent behavior with subsequent partial demonstrations, and human evaluations, i.e. sparse indications of the quality of agent behavior.
These individual mechanisms of human interaction have been previously shown to provide various benefits in learning performance and efficiency \cite{knox2009interactively,MacGlashan2017,Warnell2018,goecks2018efficiently,Saunders2017}. 
The successful integration of these disparate techniques, which would leverage their complementary characteristics, requires a learning architecture that allows for optimization of common objective functions and consistent representations.
An actor-critic framework with a combined loss function, as presented in this work, is such an architecture.

In this paper, we focus on extending the Cycle-of-Learning framework to tackle the known issue of transitioning BC policies to RL by utilizing an actor-critic architecture with a combined BC+RL loss function and pre-training phase for continuous state-action spaces, that can learn in both dense- and sparse-reward environments.
The main advantage of our method is the use of an off-policy, actor-critic architecture to pre-train both a policy and value function, as well as continued re-use of demonstration data during agent training, which reduces the amount of interactions needed between the agent and environment. 
This is an important aspect especially for robotic applications or real-world systems where interactions can be costly.


The combined loss function consists of the following components: an expert behavior cloning loss that drives the actor's actions toward previous human trajectories, $1$-step return Q-learning loss to propagate values of human trajectories to previous states, the actor loss, and a $L_2$ regularization loss on the actor and critic to stabilize performance and prevent over-fitting during training. 
The implementation of each loss component and their combination are defined as follows:

\begin{itemize}
    \item \textbf{Expert behavior cloning loss ($\mathcal{L}_{BC}$): } Given expert demonstration subset $\mathcal{D}_E$ of continuous states and actions $s^E$ and $a^E$ visited by the expert during a task demonstration over $T$ time steps
        \begin{align}
        \mathcal{D}_E = \left\{s^E_0, a^E_0, s^E_1, a^E_1, ... , s^E_T, a^E_T\right\},
        \end{align}
        a behavior cloning loss (mean squared error) from demonstration data $\mathcal{L}_{BC}$ can be written as
        \begin{equation} \label{eq:bc_loss}
            \mathcal{L}_{BC}(\theta_\pi) = \frac{1}{2} \left(\pi(s_t|\theta_\pi) - a^E_t) \right)^2
        \end{equation}
        in order to minimize the difference between the actions predicted by the actor network $\pi(s_t)$, parametrized by $\theta_\pi$, and the expert actions $a_{E_t}$ for a given state vector $s_t$.
    
    \item \textbf{$1$-step return Q-learning loss ($\mathcal{L}_1$): }The $1$-step return $R_1$ can be written in terms of the critic network $Q$, parametrized by $\theta_Q$, as
        \begin{equation}
            R_1 = r_t + \gamma Q(s_{t+1},\pi(s_{t+1}|\theta_\pi)|\theta_Q).
        \end{equation}
        In order to satisfy the Bellman equation, we minimize the difference between the predicted Q-value and the observed return from the $1$-step roll-out for a batch of sampled states $\textbf{s}$:
        \begin{equation} \label{eq:1step_loss}
            \mathcal{L}_{Q_{1}}(\theta_Q) = \frac{1}{2} \left( R_1 - Q(\textbf{s}, \pi(\textbf{s}|\theta_\pi)|\theta_Q) \right)^2.
        \end{equation}
    
    \item \textbf{Actor Q-loss ($\mathcal{L}_A$): } It is assumed that the critic function $Q$ is differentiable with respect to the action. Since we want to maximize the Q-values for the current state, the actor loss became the negative of the Q-values predicted by the critic for a batch of sampled states $\textbf{s}$:
        \begin{equation} \label{eq:actor_qloss}
            \mathcal{L}_A(\theta_\pi) = - Q(\textbf{s}, \pi(\textbf{s}|\theta_\pi)|\theta_Q).
        \end{equation}
    
\end{itemize}

Combining the above loss functions for the Cycle-of-Learning becomes

\begin{align} \label{eq:col_loss}
    \mathcal{L}_{CoL}& (\theta_Q, \theta_\pi) = \lambda_{BC} \mathcal{L}_{BC} (\theta_\pi) + \lambda_A \mathcal{L}_A (\theta_\pi) \nonumber \\
     & +  \lambda_{Q_1} \mathcal{L}_{Q_{1}} (\theta_Q) + \lambda_{L2Q} \mathcal{L}_{L2} (\theta_Q) + \lambda_{L2\pi} \mathcal{L}_{L2} (\theta_\pi).
\end{align}

Our approach starts by collecting contiguous trajectories from expert policies and stores the current and subsequent state-actions pairs, reward received, and task completion signal in a permanent expert memory buffer $\mathcal{D}_E$.
During the pre-training phase, the agent samples a batch of trajectories from the expert memory buffer $\mathcal{D}_E$ containing expert trajectories to perform updates on the actor and critic networks using the same combined loss function (Equations \ref{eq:col_loss}).
This procedure shapes the actor and critic initial distributions to be closer to the expert trajectories and eases the transition from policies learned through expert demonstration to reinforcement learning.

After the pre-training phase, the policy is allowed to roll-out and collect its first on-policy samples, which are stored in a separate first-in-first-out memory buffer with only the agent's samples.
After collecting a given number of on-policy samples, the agent samples a batch of trajectories comprising 25\% of samples from the expert memory buffer and 75\% from the agent's memory buffer.
This fixed ratio guarantees that each gradient update is grounded by expert trajectories.
If a human demonstrator is used, they can intervene at any time the agent is executing their policy, and add this new trajectories to the expert memory buffer.

The proposed method is shown in Algorithm \ref{alg:algo}.

\begin{algorithm}[!tb]
\caption{Cycle-of-Learning (CoL): Transitioning from Demonstration to Reinforcement Learning}\label{alg:algo}
\begin{algorithmic}[1]

    \State Input:
        \begin{description}
            \item Environment $env$, number of training steps $T$, number of training steps per batch $M$, number of pre-training steps $L$, number of gradient updates $K$, and CoL hyperparameters $\lambda_{Q_1}$, $\lambda_{BC}$, $\lambda_A$, $\lambda_{L2Q}$, $\lambda_{L2\pi}$, $\tau$.
        \end{description}
    \State Output:
        \begin{description}
            \item Trained actor $\pi(s|\theta_\pi)$ and critic $Q(s,\pi|\theta_Q)$ networks.
        \end{description}
    \State Randomly initialize:
        \begin{description}
            \item Actor network $\pi(s|\theta_\pi)$ and its target $\pi'(s|\theta_{\pi'})$.
            \item Critic network $Q(s,\pi|\theta_Q)$ and its target $Q'(s,\pi'|\theta_{Q'})$.
        \end{description}
    \State Initialize agent and expert replay buffers $\mathcal{R}$ and $\mathcal{R}_E$.
    \State Load $\mathcal{R}$ and $\mathcal{R}_E$ with expert dataset $\mathcal{D}_E$.
    \For {pre-training steps = 1, \dots, $L$}
        \State Call \emph{TrainUpdate}() procedure.
    \EndFor
    \For {training steps = 1, \dots, $T$}
        \State Reset $env$ and receive initial state $s_0$.
        \For {batch steps = 1, \dots, $M$}
            \State Select action $a_t = \pi(s_t|\theta_\pi)$ according to policy.
            \State Perform action $a_t$ and observe reward $r_t$ and next state $s_{t+1}$.
            \State Store transition $(s_t, a_t, r_t, s_{t+1})$ in $\mathcal{R}$.
        \EndFor
        \For {update steps = 1, \dots, $K$}
            \State Call \emph{TrainUpdate}() procedure.
        \EndFor
    \EndFor

\Procedure{TrainUpdate()}{}
    \If {Pre-training}
        \State Randomly sample $N$ transitions $(s_i, a_i, r_i, s_{i+1})$ from the expert replay buffer $\mathcal{R}_E$.
    \Else
        \State Randomly sample $N*0.25$ transitions $(s_i, a_i, r_i, s_{i+1})$ from the expert replay buffer $\mathcal{R}_E$ and $N*0.75$ transitions from the agent replay buffer $\mathcal{R}$.
    \EndIf
    \State Compute $\mathcal{L}_{Q_1}(\theta_{Q})$, $\mathcal{L}_{BC}(\theta_{\pi})$, $\mathcal{L}_{A}(\theta_{\pi})$, $\mathcal{L}_{L2}(\theta_{Q})$, $\mathcal{L}_{L2}(\theta_{\pi})$
    \State Update actor and critic for $K$ steps according to Equation \ref{eq:col_loss}.
    \State Update target networks:
    \begin{align*}
        \theta_{\pi'} \leftarrow \tau\theta_{\pi} + (1-\tau)\theta_{\pi'}, \\
        \theta_{Q'} \leftarrow \tau\theta_{Q} + (1-\tau)\theta_{Q'}.
    \end{align*}
\EndProcedure

\end{algorithmic}
\end{algorithm}

\section{Numerical Results}

\subsection{Experimental Setup}

As described in the previous sections, in our approach, the Cycle-of-Learning (CoL), we collect contiguous trajectories from expert policies and store them in a permanent memory buffer.
The policy is allowed to roll-out and is trained with a combined loss from a mix of demonstration and agent data, stored in a separate first-in-first-out buffer.
We validate our approach in three environments with continuous observation- and action-space: Lunar\-Lander\-Continuous-v2 \cite{Brockman2016} (dense and sparse reward cases) and a custom quadrotor landing task \cite{goecks2018efficiently} implemented using Microsoft AirSim \cite{Airsim2017}.
The dense reward case of Lunar\-Lander\-Continuous-v2 is the standard environment provided by OpenAI Gym library \cite{Brockman2016}: the state space consists of a eight-dimensional continuous vector with inertial states of the lander, the action space consists of a two-dimensional continuous vector controlling main and side thrusters, and the reward is given at every step based on the relative motion of the lander with respect to the landing pad (bonus reward is given when the landing is completed successfully).
The sparse reward case is a custom modification with the same reward scheme and state-action space, however the reward is stored during the policy roll-out and is only given to the agent when the episode ends and is zero otherwise.
The custom quadrotor landing task is a modified version of the environment proposed by \citeauthor{goecks2018efficiently} \cite{goecks2018efficiently}, implemented using Microsoft AirSim \cite{Airsim2017}, which consists of landing a quadrotor on a static landing pad in a simulated gusty environment, as seen in Figure \ref{fig:Screenshot}.
The state space consists of a fifteen-dimensional continuous vector with inertial states of the quadrotor and visual features that represent the landing pad image-frame position and radius as seen by a downward-facing camera.
The action space is a four-dimensional continuous vector that sends velocity commands for throttle, roll, pitch, and yaw.
Wind is modeled as noise applied directly to the actions commanded by the agent and follows a temporal-based, instead of distance-based, discrete wind gust model \cite{moorhouse1980us} with 65\% probability of encountering a wind gust at each time step.
This was done to induce additional stochasticity in the environment. 
The gust duration is uniformly sampled to last between one to three real time seconds and can be imparted in any direction, with maximum velocity of half of what can be commanded by the agent along each axis.
This task has a sparse-reward scheme (reward $R$ is given at the end of the episode, and is zero otherwise) based on the relative distance $r_{rel}$ between the quadrotor and the center of the landing pad at the final time step of the episode:
\begin{equation*}
    R = \frac{1}{1+r_{rel}^2}.
\end{equation*}

The hyperparameters used in CoL for each environment are described in table \ref{tab:exp_hyperparams}.

\begin{table}[!h]
  \caption{Cycle-of-Learning hyperparemeters for each environment: (a) LunarLanderContinuous-v2 and (b) Microsoft AirSim. Reprinted from \cite{goecks2019integrating}.}
  \label{tab:exp_hyperparams}
  \centering
  \begin{tabular}{lll}\toprule[1.5pt]
    & \multicolumn{2}{l}{Environments}                  \\
    \cmidrule(r){2-3}
    Hyperparameter     & \rotatebox{0}{(a)}  & \rotatebox{0}{(b)} \\
    \midrule
    $\lambda_{Q_1}$ factor & 1.0 & 1.0 \\
    $\lambda_{BC}$ factor & 1.0 & 1.0 \\
    $\lambda_A$ factor & 1.0 & 1.0 \\
    $\lambda_{L2Q}$ factor & 1.0$e^{-5}$ & 1.0$e^{-5}$ \\
    $\lambda_{L2\pi}$ factor & 1.0$e^{-5}$ & 1.0$e^{-5}$ \\
    Batch size & 512 & 512 \\
    Actor learning rate  & 1.0$e^{-3}$ & 1.0$e^{-3}$ \\
    Critic learning rate & 1.0$e^{-4}$ & 1.0$e^{-4}$ \\
    Memory size  & 5.0$e^{5}$ & 5.0$e^{5}$ \\
    Expert trajectories & 20 & 5 \\
    Pre-training steps & 2.0$e^{4}$ & 2.0$e^{4}$ \\
    Training steps & 5.0$e^{6}$ & 5.0$e^{5}$ \\
    Discount factor $\gamma$ & 0.99 & 0.99 \\
    Hidden layers & 3 & 3 \\
    Neurons per layer & 128 & 128 \\
    Activation function & ELU & ELU \\
    \bottomrule[1.25pt]
  \end{tabular}
\end{table}



The baselines that we compare our approach to are Deep Deterministic Policy Gradient (DDPG) \cite{lillicrap2015continuous,Silver2014}, Demonstration Augmented Policy Gradient (DAPG) \cite{Rajeswaran-RSS-18}, and traditional behavior cloning (BC).
For the DDPG baseline we used an open-source implementation by Stable Baselines \cite{stable-baselines}. 
The hyperparameters used concur with the original DDPG publication \cite{lillicrap2015continuous}: actor and critic networks with 2 hidden layers with 400 and 300 units respectively, optimized using Adam \cite{Kingma2015} with learning rate of $10^{-4}$ for the actor and $10^{-3}$ for the critic, discount factor of $\gamma = 0.99$, trained with minibatch size of 64, and replay buffer size of $10^{6}$. 
Exploration noise was added to the action following an Ornstein-Uhlenbeck process \cite{uhlenbeck1930theory} with mean of 0.15 and standard deviation of 0.2.
For the DAPG baseline we used an official release of the DAPG codebase from the authors \footnote{Code available at \url{https://github.com/aravindr93/hand_dapg}.}. The policy is represented by a deep neural network with three hidden layers of 128 units each, pre-trained with behavior cloning for 100 epochs, with a batch size of 32 samples, and learning rate of $10^{-3}$, $\lambda_0 = 0.01$, and $\lambda_1 = 0.99$.
The BC policies are trained by minimizing the mean squared error between the expert demonstrations and the output of the model. 
The policies consist of a fully-connected neural network with 3 hidden layers with 128 units each and exponential linear unit (ELU) activation function \cite{clevert2015fast}. 
The BC policy was evaluated for 100 episodes which was used to calculate the mean and standard error of the performance of the policy.

All baselines that rely on demonstrations, namely BC, DAPG, and CoL, use the same human trajectories collected in the Lunar\-Lander\-Continuous-v2 and custom Microsoft AirSim environment.

\subsection{Experimental Results}

\begin{figure}[!htb]%
    \centering
    \includegraphics[width=.7\linewidth]{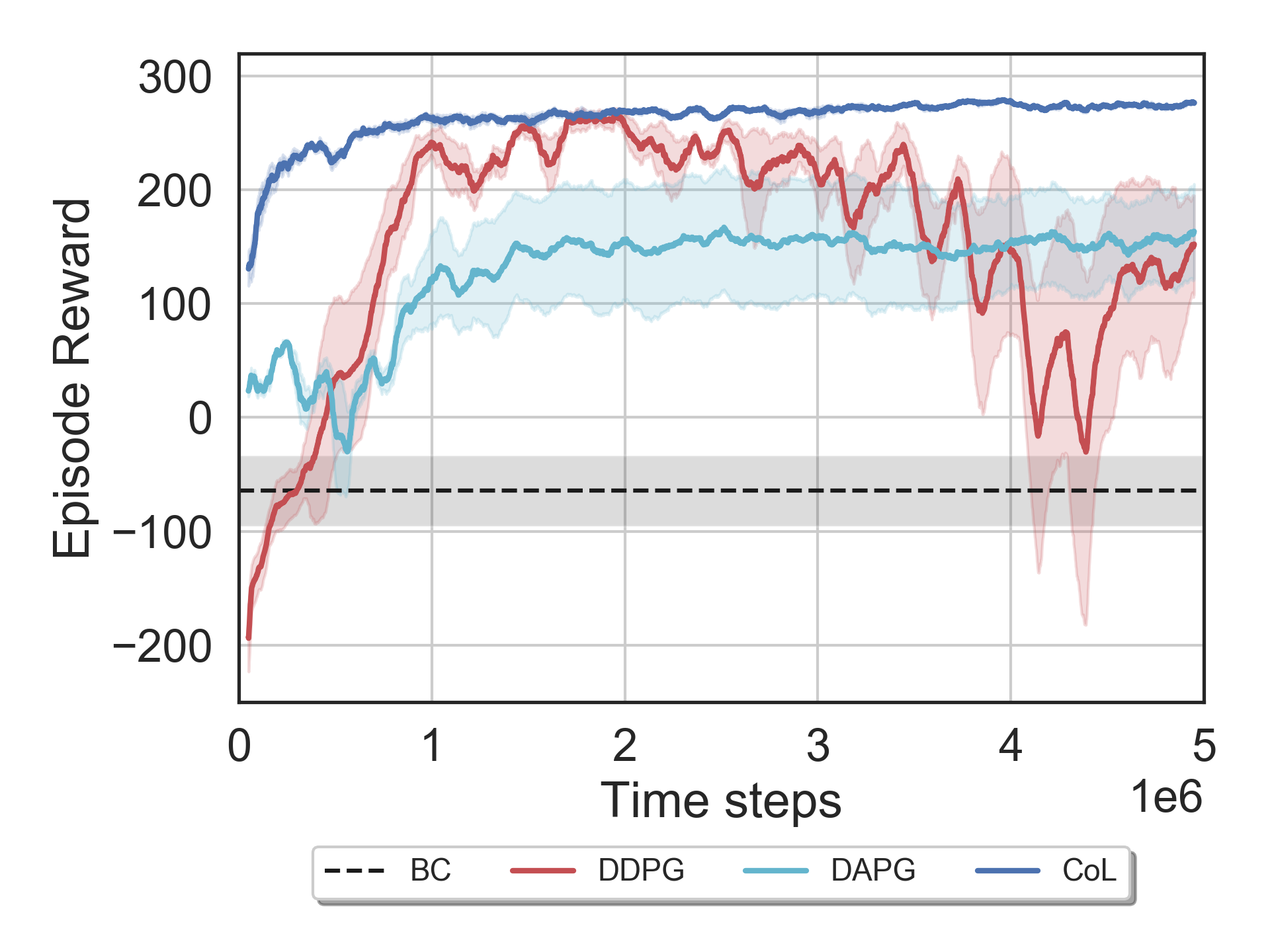}
    \caption{Comparison of CoL, BC, DDPG, and DAPG for 3 random seeds (bold line representing the mean and shaded area the standard error) in the dense--reward Lunar\-Lander\-Continuous-v2 environment. Reprinted from \cite{goecks2019integrating}.}%
    \label{fig:col_results_compact_a}%
\end{figure}

The comparative performances of the CoL against the baseline methods (BC, DDPG and DAPG) for the Lunar\-Lander\-Continuous-v2 environment are presented via their training curves in Figure \ref{fig:col_results_compact_a}, using the standard dense reward.
The mean reward of the BC pre-trained from the human demonstrations is also shown for reference, and its standard error is shown by the shaded band.
The CoL reward initializes to values at or above the BC and steadily improves throughout the reinforcement learning phase.
Conversely, the DDPG RL baseline initially returns rewards lower than the BC and slowly improves until its performance reaches similar levels to the CoL after approximately one million steps.
However, this baseline never performs as consistently as the CoL and eventually begins to diverge, losing much of its performance gains after about four million steps.
The DAPG baseline initial performance, similar to the CoL, surpasses behavior cloning due to the pre-training phase and slowly converges to a high score, although slower than the CoL.

\begin{figure}[!htb]%
    \centering
    \includegraphics[width=.7\linewidth]{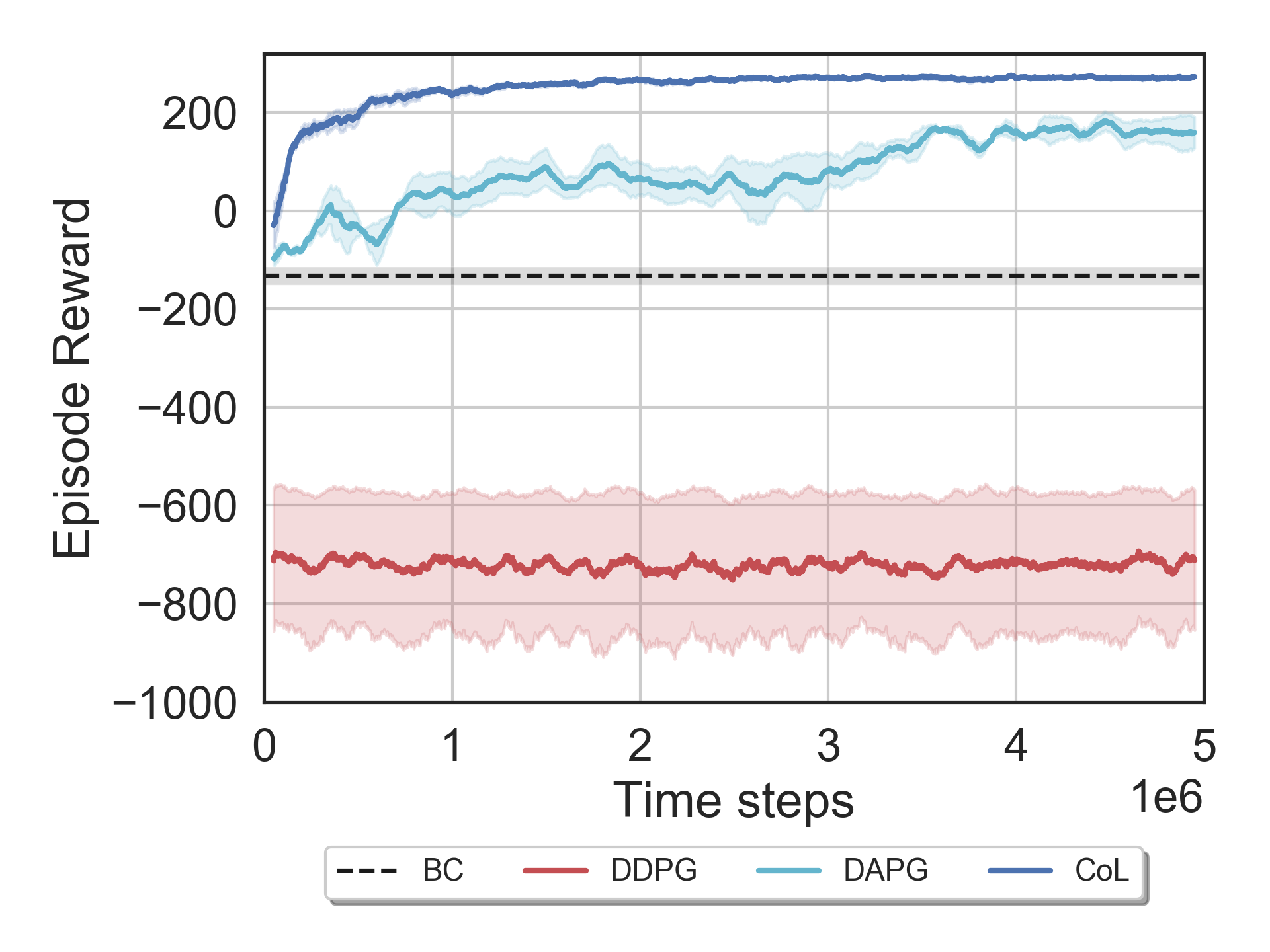}
    \caption{Comparison of CoL, BC, DDPG, and DAPG for 3 random seeds (bold line representing the mean and shaded area the standard error) in the sparse-reward Lunar\-Lander\-Continuous-v2 environment. Reprinted from \cite{goecks2019integrating}.}%
    \label{fig:col_results_compact_b}%
\end{figure}

When using sparse rewards, meaning the rewards generated by the Lunar\-Lander\-Continuous-v2 environment are provided only at the last time step of each episode, the performance improvement of the CoL relative to the DDPG and DAPG baselines is even greater (Figure \ref{fig:col_results_compact_b}). 
The performance of the CoL is qualitatively similar during training to that of the dense case, with an initial reward roughly equal to or greater than that of the BC and a consistently increasing reward.
Conversely, the performance of the DDPG baseline is greatly diminished for the sparse reward case, yielding effectively no improvement throughout the whole training period.
The training of the DAPG does not deteriorate when compared to the dense reward case but still the performance does not match CoL for the specified training time.

\begin{figure}[!htb]%
    \centering
    \includegraphics[width=.7\linewidth]{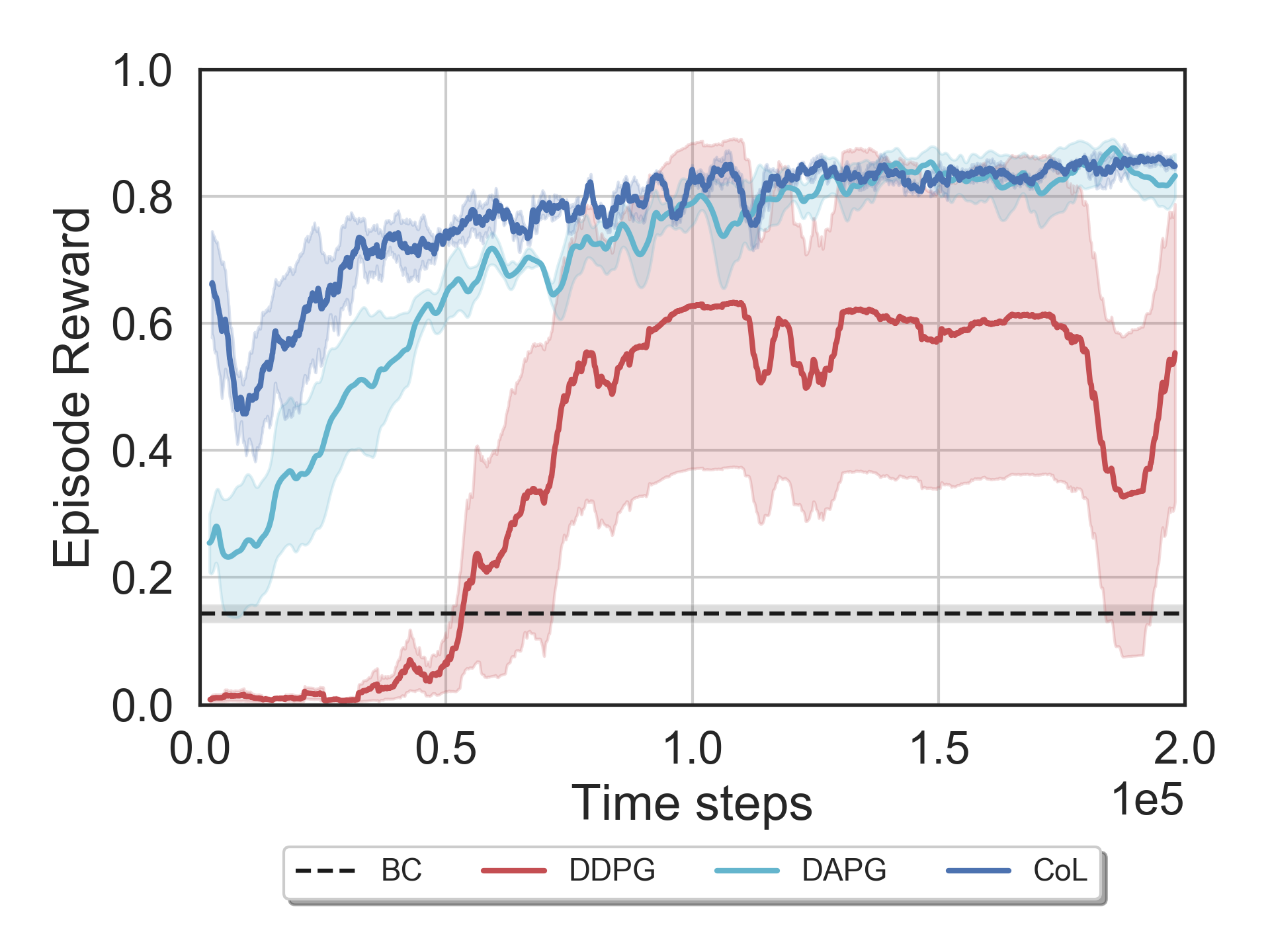}
    \caption{Comparison of CoL, BC, DDPG, and DAPG for 3 random seeds (bold line representing the mean and shaded area the standard error) in the sparse-reward Microsoft AirSim quadrotor landing environment. Reprinted from \cite{goecks2019integrating}.}%
    \label{fig:col_results_compact_c}%
\end{figure}

The results for the more realistic and challenging AirSim quadrotor landing environment (Figure \ref{fig:col_results_compact_c}) illustrate a similar trend.
The CoL initially returns rewards above the BC, DDPG, and DAPG baselines and steadily increases its performance, with DAPG converging at end to a similar level of performance. 
The DDPG baseline practically never succeeds and subsequently fails to learn a viable policy, while displaying greater variance in performance when compared to CoL and DAPG. 
Noting that successfully landing on the target would generate a sparse episode reward of approximately 0.64, it is clear that these baseline algorithms, with exception of DAPG, rarely generate a satisfactory trajectory for the duration of training.

\subsection{Component Analysis}

\begin{table*}[!htb]
\begin{threeparttable}
\centering
\caption{Method Comparison on Lunar\-Lander\-Continuous-v2 environment, dense-reward case. Reprinted from \cite{goecks2019integrating}.}
\label{tab:ablation_table} 
\begin{tabular}{lllll}\toprule
\bf Method & \bf Pre-Training Loss & \bf Training Loss  & \bf Buffer Type & \bf Average Reward \\\midrule
CoL        & $\mathcal{L}_{Q_{1}} + \mathcal{L}_A + \mathcal{L}_{BC}$     & $\mathcal{L}_{Q_{1}} + \mathcal{L}_A + \mathcal{L}_{BC}$ & Fixed Ratio & 261.80 $\pm$ 22.53  \\
CoL-PT     & None                                                 & $\mathcal{L}_{Q_{1}} + \mathcal{L}_A + \mathcal{L}_{BC}$  & Fixed Ratio & 253.24 $\pm$ 46.50 \\
CoL+PER     & $\mathcal{L}_{Q_{1}} + \mathcal{L}_A + \mathcal{L}_{BC}$     & $\mathcal{L}_{Q_{1}} + \mathcal{L}_A + \mathcal{L}_{BC}$ & PER & 245.24 $\pm$ 37.66 \\
DAPG     & $\mathcal{L}_{BC}$     & Augmented Policy Gradient & None & 127.99 $\pm$ 37.28 \\
DDPG    & None                                    & $\mathcal{L}_{Q_{1}} + \mathcal{L}_A$ & Uniform & 152.98 $\pm$ 69.45  \\
BC         & $\mathcal{L}_{BC}$                                     & None  & None & -48.83 $\pm$ 27.68*  \\
BC+DDPG    & $\mathcal{L}_{BC}$                                     & $\mathcal{L}_{Q_{1}} + \mathcal{L}_A$ & Uniform & -57.38 $\pm$ 50.11  \\
CoL-BC     & $\mathcal{L}_Q{_{1}} + \mathcal{L}_A$                        & $\mathcal{L}_{Q_{1}} + \mathcal{L}_A$  & Fixed Ratio & -105.65 $\pm$ 196.85  \\
\bottomrule
\end{tabular}
\begin{tablenotes}
    \item Summary of learning methods. Enumerated for each method are all non-zero loss components (excluding regularization), buffer type, and average and standard error of the reward throughout training (after pre-training) across the three seeds, evaluated with dense reward in Lunar\-Lander\-Continuous-v2 environment. $^*$For BC, these values are computed from 100 evaluation trajectories of the final pre-trained agent.
\end{tablenotes}
\end{threeparttable}%
\end{table*}

Several component analyses were performed to evaluate the impact of each of the critical elements of the CoL on learning.
These respectively include the effects of pre-training, the combined loss function, and the sample composition of the experience replay buffer.
The results of each analysis are shown in Figures \ref{fig:col_pt_cp}-\ref{fig:ablation_results_per} and are summarized in Table \ref{tab:ablation_table}.

\subsubsection{Effects of Pre-Training}
\begin{figure}
    \centering
    \includegraphics[width=0.7\columnwidth]{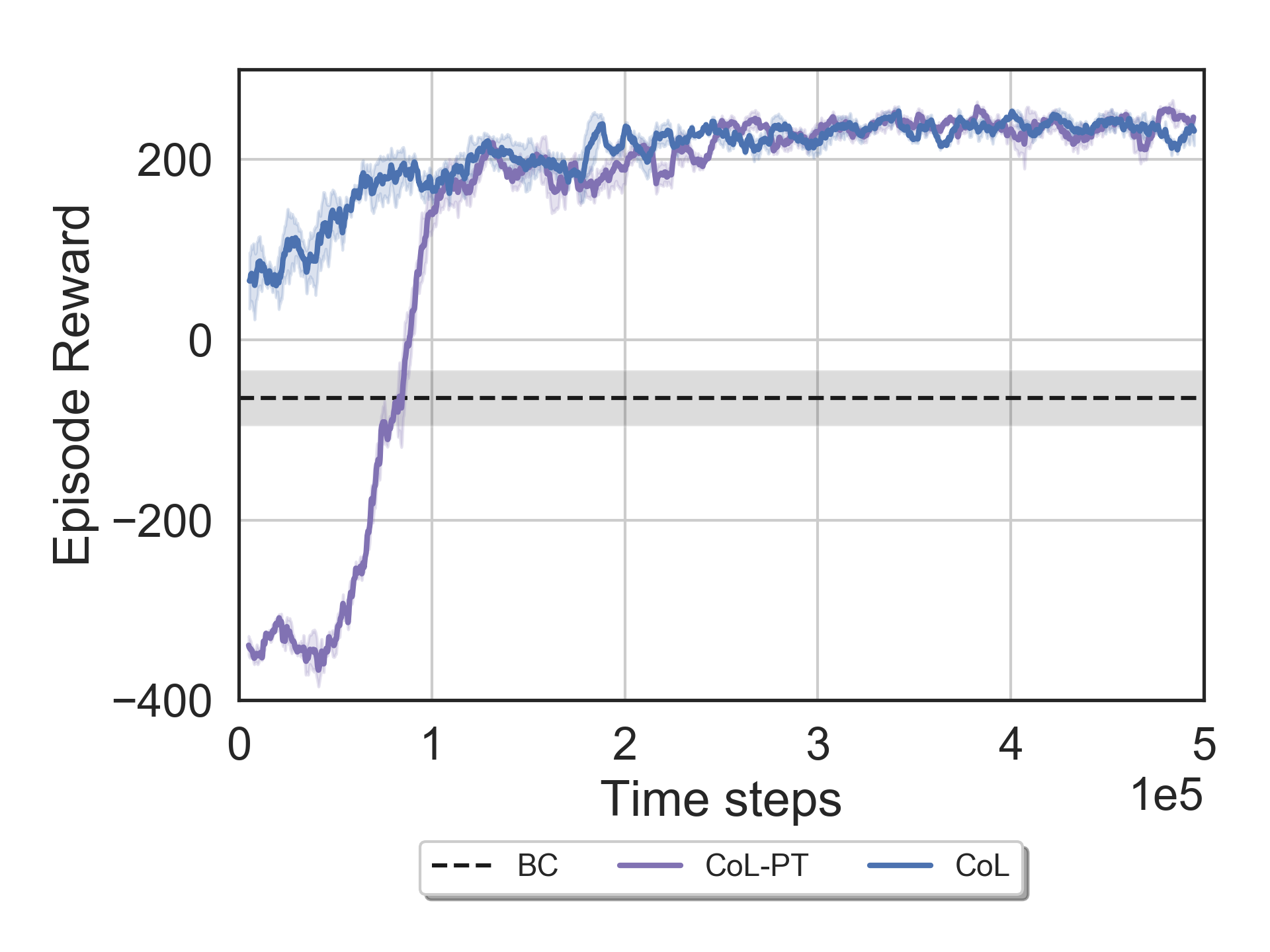}
    \caption{Effects of the pre-training phase in the Cycle-of-Learning. Results for 3 random seeds (bold line representing the mean and shaded area the standard error) showing component analysis in Lunar\-Lander\-Continuous-v2 environment comparing pre-trained Cycle-of-Learning (CoL curve) against the Cycle-of-Learning without the pre-training phase (CoL-PT curve) and the behavior cloning (BC) baseline. Reprinted from \cite{goecks2019integrating}.}
    \label{fig:col_pt_cp}
\end{figure}

To determine the effects of pre-training on performance we compare the standard CoL against an implementation without this pre-training phase, where the number of pre-training steps $L=0$, denoted as \emph{CoL-PT}. 
The complete combined loss, as seen in Equations \ref{eq:col_loss} is used during the reinforcement learning phase.
This condition assesses the impact on learning performance of not pre-training the agent, while still using the combined loss in the RL phase.
As seen in Figure \ref{fig:col_pt_cp}, this condition differs from the baseline CoL in its initial performance being worse, i.e. significantly below the BC, but does reach similar rewards after several hundred thousand steps, exhibiting the same consistent response during training thereafter.
Effectively, this highlights that the benefit of pre-training is improved initial response and significant speed gain in reaching steady-state performance level, without qualitatively impacting the long-term training behavior.

\subsubsection{Effects of Combined Loss}
\begin{figure}
    \centering
    \includegraphics[width=0.7\columnwidth]{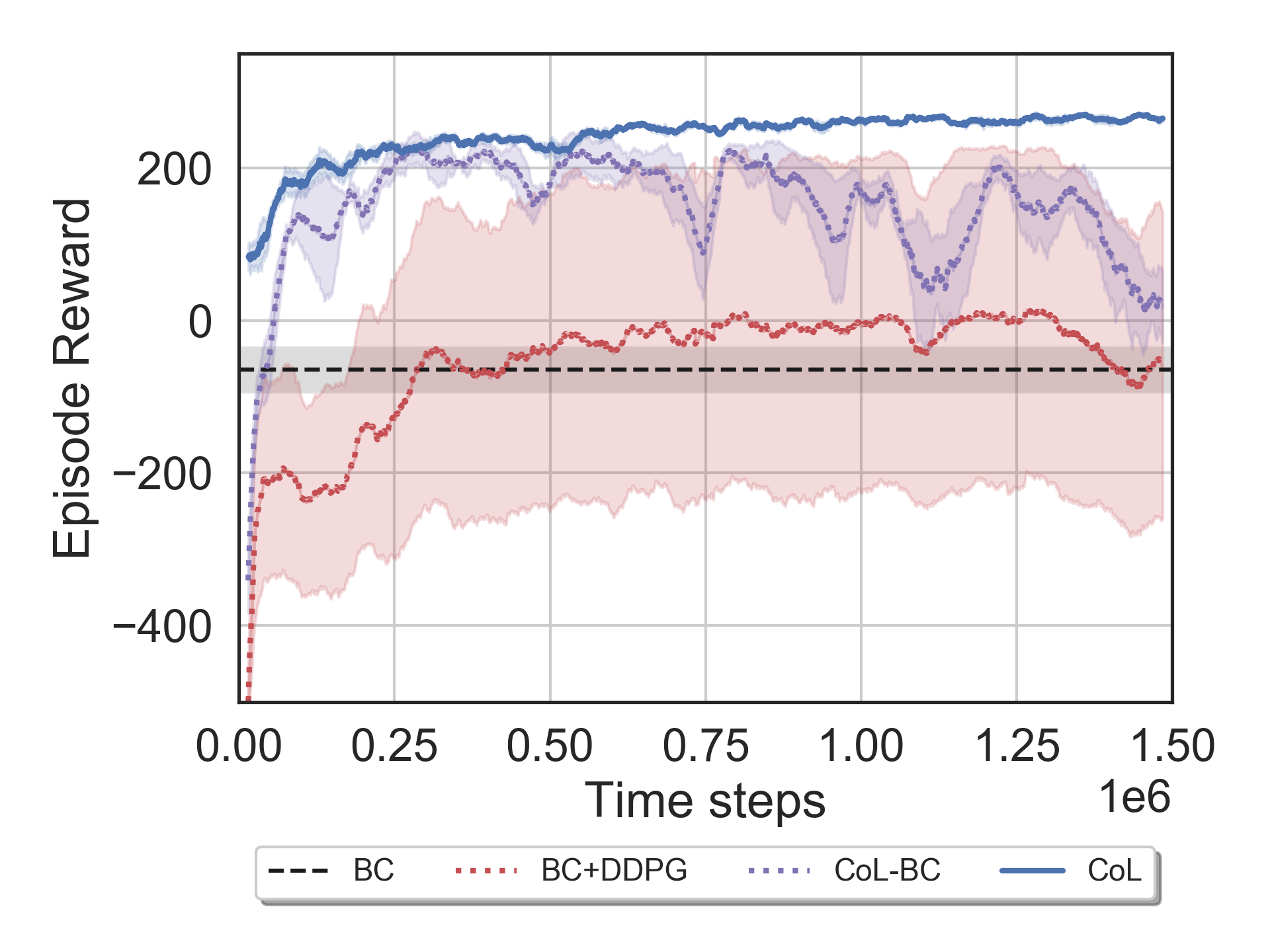}
    \caption{Effects of the combined loss in the Cycle-of-Learning. Results for 3 random seeds (bold line representing the mean and shaded area the standard error) showing component analysis in Lunar\-Lander\-Continuous-v2 environment comparing complete Cycle-of-Learning (CoL), CoL without the expert behavior cloning loss (CoL-BC), and pre-training with BC followed by DDPG without combined loss (BC+DDPG). Reprinted from \cite{goecks2019integrating}.}
    \label{fig:col_bc_cp}
\end{figure}

To determine the effects of the combined loss function on performance we compare the standard CoL against two alternate learning implementations: 1) the CoL without the behavioral cloning expert loss on the actor ($\lambda_{BC} := 0$) during both pre-training and RL phases, denoted as \emph{CoL-BC}, and 2) standard BC followed by DDPG using standard loss functions, denoted as \emph{BC+DDPG}. 
For the implementation of the CoL without the behavior cloning loss (\emph{CoL-BC}), the critic loss remains the same as in Equation \ref{eq:col_loss} for both training phases.
This condition assesses the impact on learning performance of the behavior cloning loss component $\mathcal{L}_{BC}$, given otherwise consistent loss functions in both pre-training and RL phases.
As seen in Figure \ref{fig:col_bc_cp}, this condition (purple, dashed) improves upon the CoL-PT condition (Figure \ref{fig:col_pt_cp}) in its initial reward return and similarly achieves comparable performance to the baseline CoL in the first few hundred thousand steps, but then steadily deteriorates as training continues, with several catastrophic losses in performance.
This result makes clear that the behavioral cloning loss is an essential component of the combined loss function toward maintaining performance throughout training, anchoring the learning to some previously demonstrated behaviors that are sufficiently proficient.

The second of these comparative implementations that illustrate the effects of the combined loss, behavior cloning with subsequent DDPG (\emph{BC+DDPG}), utilized standard loss functions (Equations \ref{eq:bc_loss}, \ref{eq:1step_loss}, and \ref{eq:actor_qloss}) rather than the CoL combined loss in both phases (Equation \ref{eq:col_loss}). 
Pre-training of the actor with BC uses only the regression loss, as seen in Equation \ref{eq:bc_loss}.
DDPG utilizes standard loss functions for the actor and critic, as seen in Equation \ref{eq:ddpg_loss}.
This condition assesses the impact on learning performance of standardized loss functions rather than our combined loss functions across both training phases.
This condition (Figure \ref{fig:col_bc_cp}; red, dashed) produces initial rewards below the BC response and subsequently improves in performance only to an average level similar to that of the BC and is much less stable in its response throughout training, as indicated by the wide standard error band. 
This result indicates that simply sequencing standard BC and RL algorithms results in significantly worse performance and stability even after millions of training steps, emphasizing the value of a consistent combined loss function across all training phases.

\begin{align} \label{eq:ddpg_loss}
    \mathcal{L}_{DDPG} (\theta_Q,\theta_\pi) =& \lambda_{Q_1}\mathcal{L}_{Q_1}(\theta_Q) + \lambda_{A}\mathcal{L}_{A}(\theta_\pi)\nonumber \\ 
    &+ \lambda_{L2Q}\mathcal{L}_{L2}(\theta_Q)+ \lambda_{L2\pi}\mathcal{L}_{L2}(\theta_\pi).
\end{align}

\subsubsection{Effects of Human Experience Replay Sampling}
\begin{figure}[!t]%
    \centering
    \includegraphics[width=0.7\linewidth]{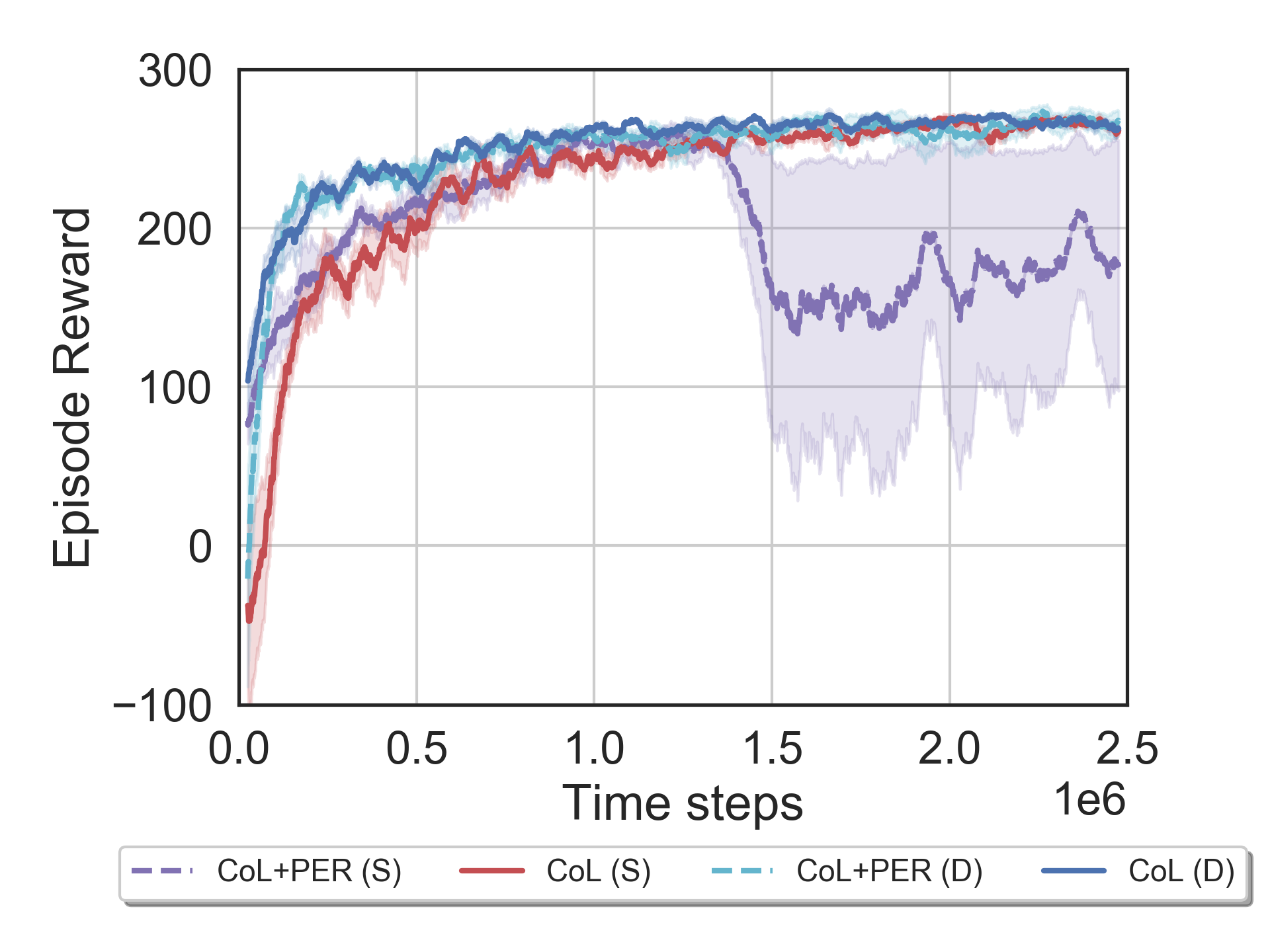}
    \caption{Effects of human experience replay sampling in the Cycle-of-Learning. Results for 3 random seeds (bold line representing the mean and shaded area the standard error) showing ablation study in Lunar\-Lander\-Continuous-v2 environment, dense (D) and sparse (S) reward cases, comparing complete Cycle-of-Learning (CoL) trained with fixed ratio of expert and agent samples and complete Cycle-of-Learning using Prioritized Experience Replay (CoL+PER) with a variable ratio of expert and agent samples ranked based on their temporal difference error. Reprinted from \cite{goecks2019integrating}.}%
    \label{fig:ablation_results_per}%
\end{figure}

To determine the effects of the experience replay buffer on performance we compare the standard CoL, which utilizes a fixed ratio buffer of samples comprising 25\% expert data and 75\% agent data, against an implementation with Prioritized Experience Replay (PER) \cite{Schaul2015}, with a data buffer prioritized by the magnitude of each transition's temporal difference (TD) error, denoted as \emph{CoL+PER}.
The comparative performance of these implementations, for both the dense- (D) and sparse-reward (S) cases of the Lunar\-Lander\-Continuous-v2 scenario, are shown in Figure \ref{fig:ablation_results_per}.
For the dense-reward condition, there is no significant difference in the learning performance between the fixed ratio and PER buffers.
However, for the sparse-reward case of the \emph{CoL+PER} implementation, the learning breaks down after approximately 1.3 million training steps, resulting in a significantly decreased performance thereafter.
This result illustrates that the fixed sampling ratio for the replay buffer in the standard CoL is a more robust mechanism of incorporating experience data, particularly in sparse-reward environments, likely because it grounds performance to demonstrated human behavior throughout training.

\section{Summary}
In this work, we present a novel method for combining behavior cloning with reinforcement learning using an actor-critic architecture that implements a combined loss function and a demonstration-based pre-training phase.
We compare our approach against state-of-the-art baselines, including BC, DDPG, and DAPG, and demonstrate the superiority of our method in terms of learning speed, stability, and performance with respect to these baselines.
This is shown in the OpenAI Gym Lunar\-Lander\-Continuous-v2 and the high-fidelity Microsoft AirSim quadrotor simulation environments in both dense and sparse reward settings.
This result is especially noticeable in the AirSim landing task (Figure \ref{fig:col_results_compact_c}), an environment designed to exhibit a high degree of stochasticity. 
The BC and DDPG baselines fail to converge to an effective and stable policy after five million training steps on the Lunar\-Lander\-Continuous-v2 environment with dense reward and the modified version with a sparse reward signal.
DAPG, although successful in both Lunar\-Lander\-Continuous-v2 environments and the custom AirSim landing task, converges at a slower rate when compared to the proposed method and starts the training at a lower performance value after pre-training with demonstration data.
Conversely, our method, CoL, is able to quickly achieve high performance without degradation, surpassing both behavior cloning and reinforcement learning algorithms alone, in both dense and sparse reward cases.
Additionally, we demonstrate through separate analyses of several components of our architecture that pre-training, the use of a combined loss function, and a fixed ratio of human-generated experience are critical to the performance improvements.
This component analysis also indicated that simply sequencing standard behavior cloning and reinforcement learning algorithms does not produce these gains and highlighted the importance of grounding the training to the demonstrated data by using a fixed ratio of expert and agent trajectories in the experience replay buffer.

Future work will investigate how to effectively integrate multiple forms of human feedback into an efficient human-in-the-loop RL system capable of rapidly adapting autonomous systems in dynamically changing environments. 
Actor-critic methods, such as the CoL method proposed in this paper, provide an interesting opportunity to integrate different human feedback modalities as additional learning signals at different stages of policy learning \cite{Waytowich2018}. 
For example, existing works have shown the utility of leveraging human interventions \cite{goecks2018efficiently,Saunders2017}, and specifically learning a predictive model of what actions to ignore at every time step \cite{Zahavy2018}, which could be used to improve the quality of the actor's policy. 
Deep reinforcement learning with human evaluative feedback has also been shown to quickly train policies across a variety of domains \cite{Warnell2018,Arumugam2017} and can be a particularly useful approach when the human is unable to provide a demonstration of desired behavior but can articulate when desired behavior is achieved. 
Feedback of this type can be interpreted as a critique of an agent's current behavior relative to the human's expectation of desired behavior \cite{Arumugam2017}, thus making it conceptually similar to an advantage function which can be used to improve the quality of the critic. 
Further, the capability our approach provides, to transition from a limited number of human demonstrations to a baseline behavior cloning agent and subsequent improvement through reinforcement learning without significant losses in performance, is largely motivated by the goal of human-in-the-loop learning on physical systems.
Thus our aim is to integrate this method onto such systems and demonstrate rapid, safe, and stable learning from limited human interaction.

%
%
%
%

\chapter{CONCLUSIONS AND RECOMMENDATIONS} \label{ch:conclusions}



The following conclusions are made based on the results presented in this dissertation:

\begin{enumerate}
    \item The Cycle-of-Learning is able to quickly achieve high performance without degradation, surpassing both behavior cloning and reinforcement learning algorithms alone, despite receiving only sparse rewards. 
    Compared CoL to BC, DDPG, and DAPG, showing that CoL converges to better policy performance on the continuous environments studied. On LunarLanderContinuous-v2, this translates to 636\% better when compared to BC, 71\% better compared to DDPG, and 104\% better than DAPG.
    Not only converges to better performance, but it starts with average performance higher than the baselines and, more important, behavior cloning. On the Microsoft AirSim environment, this translates to 161\% initial higher performance when compared to DAPG, and 297\% when compared to BC.

    \item Component Analysis showed that the pre-training phase, each loss component, and the different forms of combining expert and agent samples in the experience replay are critical to the performance improvements.
    This component analysis also indicated that simply sequencing standard behavior cloning and reinforcement learning algorithms does not produce these gains.
    
    \item Learning from demonstrations in combination with learning from interventions yields a more proficient policy based on less data, when compared to either approach in isolation.
    Averaging the results over all presented conditions and datasets, the task completion increased by 12.8\% ($\pm$ 3.6\% std. error) using 32.1\% ($\pm$ 3.2\% std. error) less human samples, which results in a CoL agent that overall has a task completion rate per sample 1.84 times higher than its counterparts.
    This supports the notion that the Cycle-of-Learning is a more data efficient approach to training via human inputs and further supports the notion that a combinatorial learning strategy inherently samples more data rich inputs from the human observer.
    
    
    \item Human interaction modalities, or any expert prior knowledge or policy, should be exploited in all aspects of the Reinforcement Learning cycle for better results, and not only on the reward function.
    
    \item Further, the case studies presented the application of inverse reinforcement learning and generative models to learn from human demonstrations, showing that an agent could learn how to perform a UAS landing task and surpass human mean performance after only 100 learning iterations.
    
\end{enumerate}

Several recommendations are made based on the research in this dissertation:

\begin{enumerate}
    \item Investigate the trade-offs between learning in a simulated environment and transferring to hardware compared to directly learning in hardware. High-fidelity simulation environments take great effort and time to be implemented and there are no theoretical guarantees that the policy will be transferred successfully.
    \begin{itemize}
        \item How: As shown in this research with the Cycle-of-Learning and similar algorithms, as DAPG, if expert demonstrations are available, it is possible to integrate them to the reinforcement learning loop and learn directly in hardware. The time allocated to the development of the simulated environment can be shifted to collecting expert demonstrations.
    \end{itemize}
    \item Investigate the impact of neural network architecture size, mainly the number of hidden layers and neurons, on task performance when using machine learning algorithms, focusing on learning direct on hardware.
    \begin{itemize}
        \item How: Advances in the fields of Neural Architecture Search (NAS) and Neural Network Compression can be leverage to find smaller architectures with similar accuracy that can increase inference speed and allow complex models to be deployed on hardware.
    \end{itemize}
    \item Develop new algorithms focusing on realistic, and consequently more complex, scenarios, even at the cost of reduced performance. Although simple environments are important to develop and benchmark new algorithms, there is no theoretical guarantees that the performance will hold when they are applied to more realistic scenarios.
    \begin{itemize}
        \item How: When working with simulated environments, consider adding sensor noise, time synchronization, and high-fidelity dynamic models. Machine learning models are generally not robust to changes in the inputs distribution.
    \end{itemize}
    \item Consider the dynamic nature of human policies. This work assumed that the human policy remained fixed while training the learning agent and the human demonstrators were given practice time before performing the task. In reality, human performance can increase during training while the human learns to become better at performing the task or can degrade due to psychological factors.
    \begin{itemize}
        \item How: Add a confidence metric to the behavior cloning loss based on expertise of the human expert or current psychological states.
    \end{itemize}
    \item Investigate how to extend Cycle-of-Learning to teach a learning agent multiple tasks, as opposed to a single task as presented in this dissertation. This is essential to create learning agents that are able to fulfil additional roles in real-time tasks.
    \begin{itemize}
        \item How: Currently being developed by another graduate student in VSCL, Ritwik Bera, is a learning model that segments given demonstrations into differents options, which represent sub-tasks, and learns a option-conditioned policy.
    \end{itemize}

\end{enumerate}


\let\oldbibitem\bibitem
\renewcommand{\bibitem}{\setlength{\itemsep}{0pt}\oldbibitem}
\bibliographystyle{unsrtnat}

\phantomsection
\addcontentsline{toc}{chapter}{REFERENCES}

\renewcommand{\bibname}{{\normalsize\rm REFERENCES}}

\bibliography{data/myReference}


\end{document}